\documentclass[fleqn,10pt]{wlscirep}
\usepackage[utf8]{inputenc}
\usepackage[T1]{fontenc}
\usepackage{multirow}
\usepackage{adjustbox}
\usepackage{subcaption}
\usepackage{pifont}
\usepackage{xr}
\usepackage{float}
\usepackage{nameref}
\usepackage{xcolor}
\usepackage{longtable}
\usepackage{hyperref}  
\usepackage{marvosym}
\usepackage[nameinlink,noabbrev]{cleveref}  

\crefname{figure}{Fig.}{Figs.}         
\crefname{table}{Table}{Tables}
\crefname{section}{Section}{Sections}
\crefname{appendix}{Supplementary}{Supplementary}

\title{3D Foundation Model for Generalizable Disease Detection in Head Computed Tomography}

\author[1†]{Weicheng Zhu}
\author[1†\Letter]{Haoxu Huang}
\author[1]{Huanze Tang}
\author[2]{Rushabh Musthyala}
\author[1]{Boyang Yu}
\author[1]{Long Chen}
\author[3]{Emilio Vega}
\author[3,8]{Thomas O'Donnell}
\author[4]{Reya Hayek}
\author[4]{Lindsey Kuohn}
\author[3]{Seena Dehkharghani}
\author[4]{Jennifer A. Frontera}
\author[4,5,6]{Arjun V. Masurkar}
\author[4]{Kara Melmed}
\author[3,7\Letter]{Narges Razavian}
\affil[1]{New York University, Center for Data Science, New York, NY, 10001, USA}
\affil[2]{New York University, Courant Institute of Mathematical Sciences, New York, NY, 10001, USA}
\affil[3]{NYU Grossman School of Medicine, Department of Radiology, New York, NY, 10016, USA}
\affil[4]{NYU Grossman School of Medicine, Department of Neurology, New York, NY, 10016, USA}
\affil[5]{NYU Grossman School of Medicine, Department of Neuroscience and Physiology, New York, NY, 10016, USA}
\affil[6]{NYU Grossman School of Medicine, Neuroscience Institute, New York, NY, 10016, USA}
\affil[7]{NYU Grossman School of Medicine, Department of Population Health, New York, NY, 10016, USA}
\affil[8]{Siemens Healthineers, Malvern, PA, 19355, USA}

\affil[\Letter]{\textbf{Corresponding Authors:}{Narges.Razavian@nyulangone.org, hh2740@nyu.edu}}

\affil[$\dagger$]{These authors contributed equally to this work}
 
\begin{abstract}
Head computed tomography (CT) imaging is a widely-used imaging modality with multitudes of medical indications, particularly in assessing pathology of the brain, skull, and cerebrovascular system. It is commonly the first-line imaging in neurologic emergencies given its rapidity of image acquisition, safety, cost, and ubiquity. Deep learning models may facilitate detection of a wide range of diseases. However, the scarcity of high-quality labels and annotations, particularly among less common conditions, significantly hinders the development of powerful models. To address this challenge, we introduce \textbf{FM-HCT}: a \textbf{F}oundation \textbf{M}odel for \textbf{H}ead \textbf{CT} for generalizable disease detection, trained using self-supervised learning. Our approach pre-trains a deep learning model on a large, diverse dataset of 361,663 non-contrast 3D head CT scans without the need for manual annotations, enabling the model to learn robust, generalizable features. To investigate the potential of self-supervised learning in head CT, we employed both discrimination with self-distillation and masked image modeling, and we construct our model in 3D rather than at the slice level (2D) to exploit the structure of head CT scans more comprehensively and efficiently. The pre-training phase is followed by fine-tuning on smaller, annotated downstream datasets, thereby optimizing the model for specific diagnostic tasks, such as detecting hemorrhages, tumors, and other abnormalities. The model's downstream classification performance is evaluated using internal and three external datasets, encompassing both in-distribution (ID) and out-of-distribution (OOD) data. Our results demonstrate that the self-supervised foundation model significantly improves performance on downstream diagnostic tasks compared to models trained from scratch and previous 3D CT foundation models on scarce annotated datasets. Furthermore, the model maintains strong generalization across different datasets, indicating its potential for broad clinical applicability. This work highlights the effectiveness of self-supervised learning in medical imaging and sets a new benchmark for head CT image analysis in 3D, enabling broader use of artificial intelligence for head CT-based diagnosis.
\end{abstract} 
\begin{document}

\flushbottom
\maketitle
\thispagestyle{empty}

\section*{Introduction}
Head computed tomography (CT) is often the first step in diagnosing a wide range of neurological disorders, including head trauma, hemorrhages, hydrocephalus, and malignancies. Head CT scans are faster, more accessible, and generally less expensive than magnetic resonance imaging (MRIs), making them ideal for emergencies like traumatic brain injury (TBI) or suspected stroke. They are also effective in detecting bone fractures, or neurovascular pathologies such as arterial venous malformations. Despite its widespread use, CT lacks the contrast resolution and hence the sensitivity for many disorders dependent upon diagnosis by MRI, thus MRI is the imaging modality of choice for many neurologic diseases. MRI, however, is more costly, risks potential heating or displacement of indwelling implants, and suffers generally slower acquisition times, increasing patient discomfort and risking non-diagnostic examinations due to its greater sensitivity to motion-related artifacts. It is also more expensive than CT and is contraindicated in specific patients. Access to MRI is a major challenge in resource-limited countries. The timely and arduous determination of certain pathologies can delay appropriate medical and surgical treatment for patients.

There is significant potential to harness artificial intelligence (AI) algorithms to enhance the diagnostic and early detection capabilities of head CT, providing critical support in clinical decision-making and improving patient outcomes. Early and accurate diagnosis can potentially lead to more effective treatments, reduce complications, and improve patient survival.

Current research on AI-driven diagnosis using head CT is limited due to both lack of data availability and the complexity of model architectures. Although datasets such as RSNA~\cite{flanders_construction_2020} and CQ500~\cite{CQ500} provide publicly available head CT data, they remain small (RSNA includes approximately 10K samples and CQ500 approximately 1K), and their primary focus is on hemorrhage detection, which restricts broader applicability as a credible path to clinical decision support. Moreover, many existing models are designed with highly task-specific architectures that may not generalize well to diverse clinical applications. These models typically apply 2D convolution neural networks (CNN) to sequentially process 3D volumes slice-by-slice under the supervision of slice-level labels~\cite{wang_deep_2021,CQ500,yun_deep_2023}. 
Slice-level labels are often expensive to acquire, and models trained on 2D slices often struggle to generalize to conditions like neurodegenerative diseases, where slice-level labels are not easily defined. Developing models that can harness the information embedded within the 3D structure of CT images while requiring minimal slice-level labeled data can thus expand the impact of such approaches. To address these challenges, we have developed FM-HCT: a Foundation Model for Head CT, and demonstrated robust performance across multiple tasks and datasets, which highlights our model's potential for broad clinical applicability.

Recent advancements in AI \textit{foundation models}, deep learning models pre-trained on extensive datasets in a self-supervised manner, have enabled rapid adaptation and robust performance across a wide range of tasks~\cite{pmlr-v139-radford21a,zhou2021ibot,oquab2024dinov, rishi24foundation}. Multiple studies have shown that foundation models trained on large-scale medical data can enhance model performance in various medical imaging tasks, including chest X-rays~\cite{yao2024evaxfoundationmodelgeneral}, histopathology~\cite{wang_pathology_2024,huang_visuallanguage_2023,chen_towards_2024, Vorontsov2024},  retina imaging~\cite{zhou2023foundation}, fMRI~\cite{dong2024brainjepa} and more. Additionally, several generalist vision-language models show promise for multimodal medical applications~\cite{codella2024medimageinsight,yang2024advancingmultimodalmedicalcapabilities,zhang2024generalist}. Although some research has focused on CT scans~\cite{Tang_2022_CVPR,blankemeier2024merlinvisionlanguagefoundation,codella2024medimageinsight}, these studies remain limited to abdominal CTs and cannot generalize to other part of the body. While Google CT Foundation model has explored report generation for head CTs~\cite{yang2024advancingmultimodalmedicalcapabilities}, it uses default video encoders to interpret 3D head CTs and has only been subjectively evaluated on fewer than 100 samples, for which the generated reports were worse than original ones. Given these limitations, developing a dedicated vision foundation model for head CTs is essential to advance AI-driven diagnosis and facilitate early detection of cranial and neurological conditions.

In this work, we introduce FM-HCT, a 3D foundation model for head CT scans, developed using self-supervised learning (SSL). While SSL has shown success in natural images~\cite{chen2020simple, he2020momentum, caron2020unsupervised, caron2021emerging, bao2022beit, He2021MaskedAA, zhou2021ibot, oquab2024dinov, zbontar2021barlow, bardes2022vicreg} and in medical imaging~\cite{Liu_2023_CVPR, zhu2022interpretablepredictionlungsquamous, chen_towards_2024, zhou2023foundation, Huang2023, azizi21big, Vorontsov2024, huang2023radiology, huang21GLoRIA, chen23masked, Azizi2023}, training a robust 3D CT volume encoder presents distinct challenges, such as selecting appropriate pretext tasks, managing spatial normalization, and addressing high computational demands. To overcome these challenges, we developed a standardized pipeline that normalizes head CT scans from various protocols, producing consistent input for our foundation model. For pre-training, we adapted two SSL frameworks—self-distillation, inspired by DINO~\cite{caron2021emerging}, and masked prediction, inspired by MAE~\cite{He2021MaskedAA}. These methods were tailored to train a volumetric encoder based on a customized vision Transformer (ViT)\cite{dosovitskiy2020vit}. The full details of our design and choices are described in Section \hyperref[sec:methods]{``Method''}. pre-training was conducted on a large-scale dataset comprising 361,663 head CT scans from a major clinical institution.

\begin{figure}[htbp]
    \centering
    \includegraphics[width=0.90\textwidth]{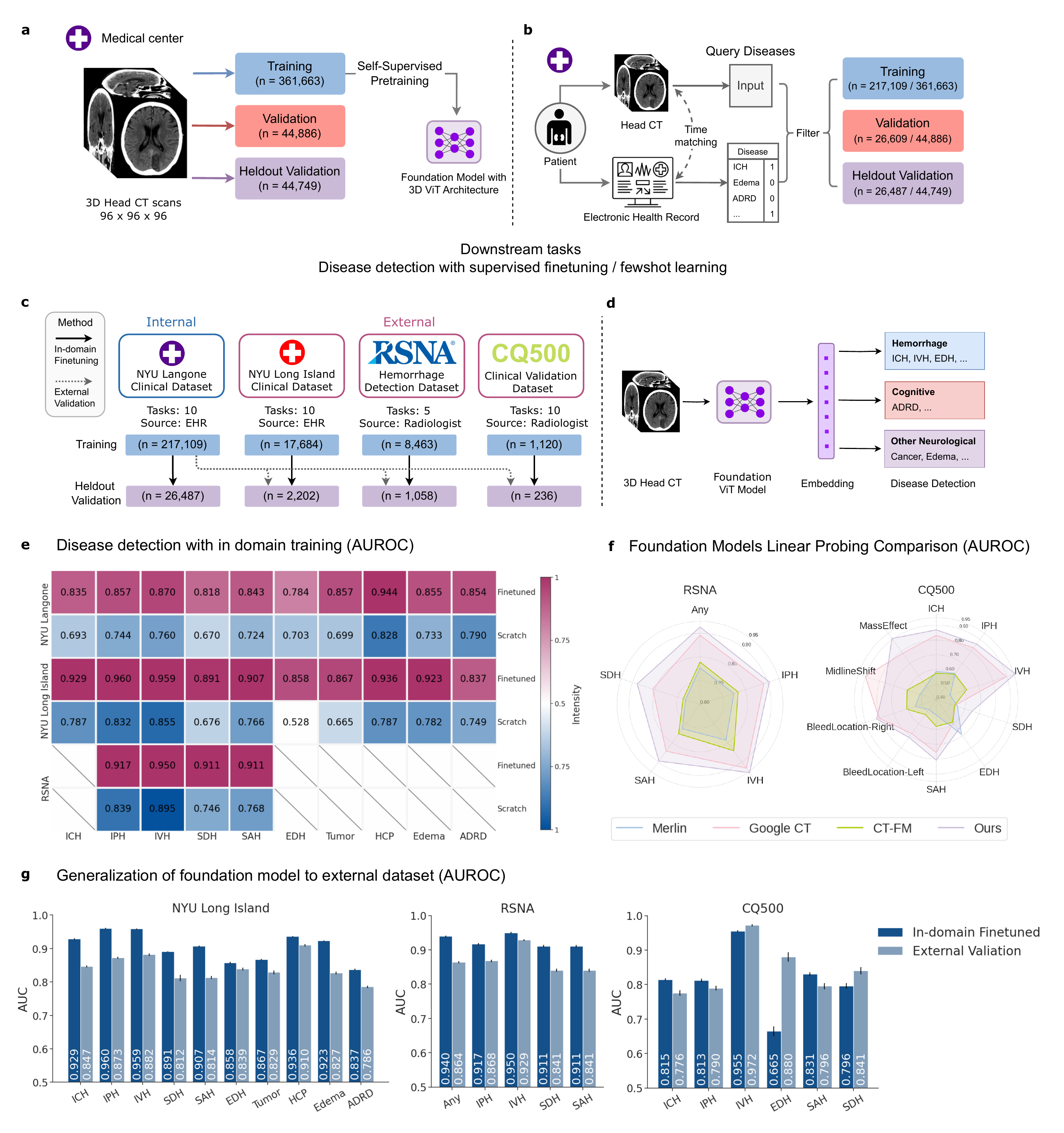}
    \caption{\textbf{Overview of the study} - the approach to developing a foundation model for head CT and its performance in disease detection tasks. $n$ in the Figure refers to number of samples for each dataset. \textbf{a,} Collection of training data and pretraining of the foundation model. \textbf{b,} Query disease labels associated with head CT scans for downstream tasks. \textbf{c,} Evaluation design of the foundation model using both internal and external datasets. \textbf{d,} Application of the foundation model to various downstream disease detection tasks. \textbf{e-g,} The performance comparison among different scenarios: (e) training with and without foundation model, (f) comparison of our CT foundation model vs. other CT foundation models, (g) in-domain finetuning and external validation in transfer learning.}
    \label{fig:overview}
\end{figure}

To evaluate the foundation model, we systematically assessed its performance and generalizability across 10 downstream disease detection tasks using diverse internal and external datasets, as illustrated in \Cref{fig:overview}. Beyond commonly studied hemorrhages, our evaluation includes crucial yet less-explored tasks in head CT, such as identifying brain tumors, Alzheimer’s disease and related dementia (ADRD), edema, and hydrocephalus (HCP). For each downstream task, the foundation model was fine-tuned using task-specific labels. Given the scarcity of expert-annotated public datasets for these conditions, we leverage electronic health records (EHR) to acquire labels of each task. While EHRs may include missing data and suffer potential label-noise, they remain a valuable and practical source for large-scale patient status labeling that can be used to evaluate the performance of the foundation model. In order to assess the label quality, we provide label sensitivity analysis by comparing 1000 manually labeled cases by one radiologist in Supplementary \Cref{tab:label_noise}.

Our results reveal substantial performance improvements enabled by our foundation model. Downstream models initialized with the pre-trained weights of foundation model achieve a $16.07\%$ improvement in macro-AUC over models trained from scratch with random initialization on internal NYU Langone data, and $20.86\%$ and $12.01\%$ improvements on external datasets from NYU Long Island (previously a separate hospital) and RSNA, respectively ($P<0.001$ for all comparisons). These findings underscore the potential of our foundation model to advance AI-based interpretation of head CT scans, supporting more accurate diagnosis and early disease detection. Furthermore, as described in Section~\hyperref[sec:results]{``Results''}, we demonstrate the model’s capabilities in out-of-distribution generalization (\Cref{fig:overview}), few-shot learning (\Cref{fig:fewshot}), and scalability (\Cref{fig:scaling_law}), highlighting the method’s potential in scenarios with limited annotated fine-tuning data, or scenarios such as federated learning which provide access to orders of magnitude more data. Overall, the experimental results on multiple datasets and tasks underscore the generalizability, adaptability and effectiveness of the model, and pave the way for significant impact in real-world clinical applications.

\section*{Results}
\label{sec:results}
\subsection*{Foundation model for disease detection with 3D head CT scans}

The key aim of the foundation model is to develop a single model that improves performance on a wide range of downstream tasks of detecting recognizable abnormalities from head CT scans. To evaluate the capability of the foundation model, we train classification models for multiple disease detection tasks by fine-tuning the foundation model separately per disease, and assessing the fine-tuned model's performance on held-out validation and external data sets. The selected downstream tasks include detecting various types of hemorrhages (intraparenchymal hemorrhage (IPH), intraventricular hemorrhage (IVH), subdural hemorrhage (SDH), epidural hemorrhage (EDH), subarachnoid hemorrhage (SAH), and intracranial hemorrhage (ICH)), brain tumors, hydrocephalus (HCP), edema, and Alzheimer's diseases and related dementia (ADRD). \Cref{fig:overview}a,b,c show the overview of our pre-training framework and included data, EHR-matching, and datasets used in pre-training, in-domain fine-tuning, and external validation. Overall N=361,663 scans were used during pre-training, and four distinct datasets from different sources were used for various forms of validation (NYU Langone N=26,487; NYU Long Island N=2,202; RSNA N=1,058; and CQ500 N=236). NYU Langone is a hospital system comprised of multiple geographically distinct hospitals including two Level 1 Trauma Centers and three Comprehensive Stroke Centers. NYU Long Island, a Level 1 Trauma Center/Comprehensive Stroke Center, is treated as an external dataset for the purposes of this study. 

The first two rows of \Cref{fig:overview}e report the task-specific AUCs for Vision Transformer (ViT) classifiers trained from scratch with random initialization, namely \textit{scratch}, versus those fine-tuned from the foundation model, namely \textit{fine-tuned} on NYU Langone data. The fine-tuned models consistently outperform the scratch model across all 10 disease detection tasks, achieving a macro-AUC of 0.852 --- a $16.07\%$ relative increase over the scratch model’s 0.734 ($P<0.001$). Additionally, in \Cref{fig:overview}f and Supplementary \Cref{fig:radar-comparison-merlin} we compared the foundation model with three other foundation model for 3D CT scans --- Merlin~\cite{blankemeier2024merlinvisionlanguagefoundation}, Google's CT Foundation~\cite{yang2024advancingmultimodalmedicalcapabilities} model and CT-FM~\cite{pai2025visionfoundationmodelscomputed}. Across four compared datasets illustrated in Supplementary \Cref{fig:radar-comparison-merlin}, Merlin outperforms the scratch model with a macro-AUC relative improvement of $8.07\%$ and $27.50\%$ on Average Precision (AP) while falling short compared to our foundation model with $13.05\%$ relative lower macro-AUC ($P<0.001$) and $27.50\%$ on AP ($P<0.001$). Although Merlin is not directly comparable to our foundation model as it was pre-trained on abdominal CT, it still provides a valuable baseline. To complement the shortcoming that Merlin is a baseline not pre-trained on Head CT, we compare our model with CT-FM \cite{pai2025visionfoundationmodelscomputed}, a foundation model pretrained on 148,000 diverse CT scans, including head CT. Our model demonstrates an macro-AUC relative improvement of $9.56\%$ ($P<0.001$) and $44.60\%$ on AP ($P<0.001$) as shown in Supplementary \Cref{fig:radar-comparison-merlin}, demonstrating the effectiveness and scalability of our approach. Additionally, we compare our model to Google CT Foundation model with linear probing, because trainable weights for end-to-end fine-tuning are not provided for this model. We consistently observe improved model performance across the board (in \Cref{fig:overview}f and Supplementary \Cref{fig:probing-comparison-gemini}).  
These findings demonstrate that despite the progress in general domain multimodal models, specialized foundation model pre-trained on head CT data still significantly enhance the understanding of brain diseases. 

To assess our foundation model's generalization to out-of-distribution data, we compiled three external datasets from multiple institutions and sources: NYU Long Island, RSNA~\cite{flanders_construction_2020}, and CQ500~\cite{CQ500}, as shown in \Cref{fig:overview}c (NYU Langone and NYU Long Island are geographically separate and distinct institutions within the broader health system). The data in these external datasets has a different distribution than the data used for pre-training. We evaluate the generalization on external datasets via two common practices to utilize the foundation model: (1) in-domain fine-tuning on separated datasets and tasks, and (2) fully external validation of the disease detection models without any site-specific fine-tuning.

\begin{figure}[htbp]
    \centering
    NYU Langone \\
    \includegraphics[trim={0 0 140pt 0},clip,height=0.3\textwidth]{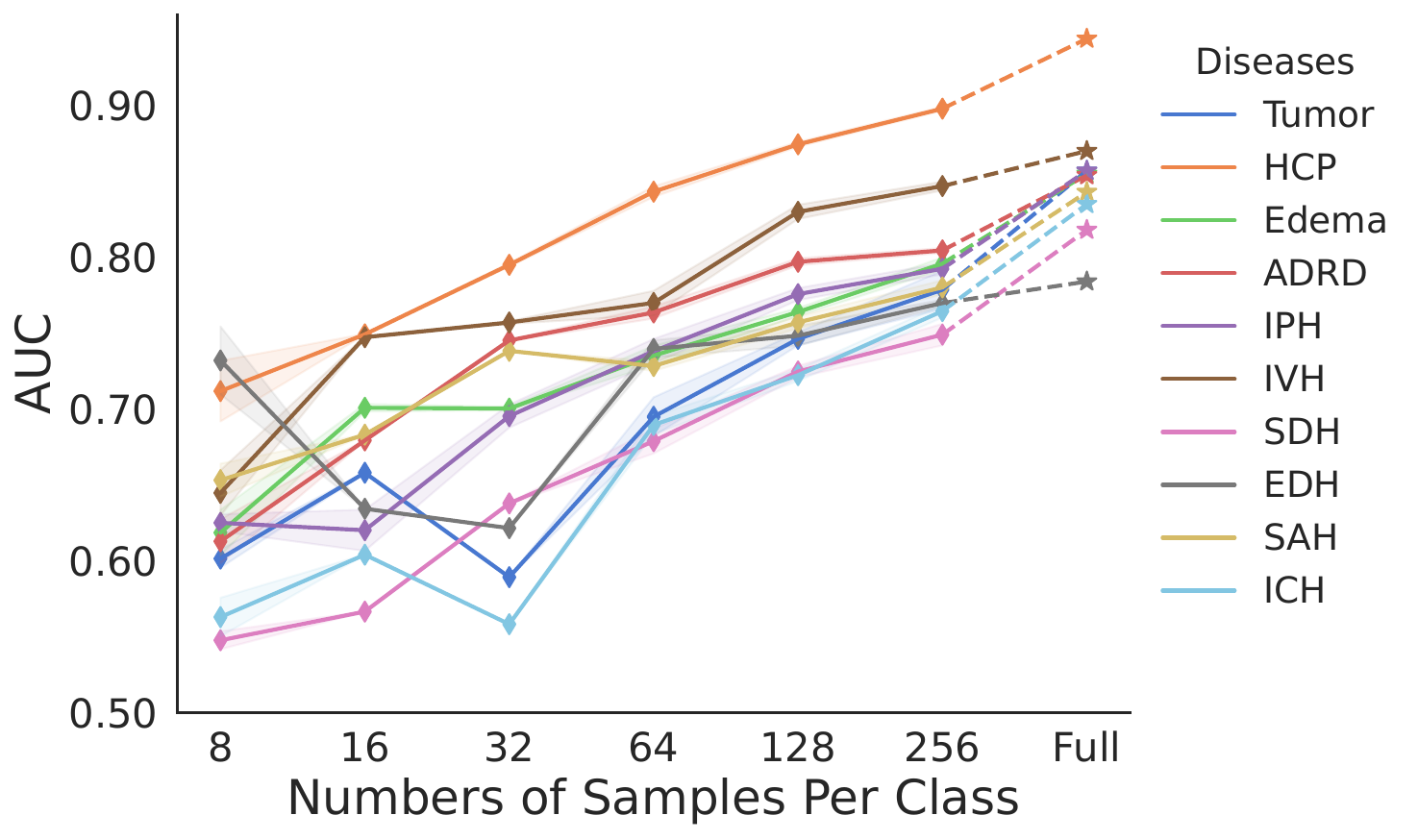}
    \includegraphics[height=0.3\textwidth]{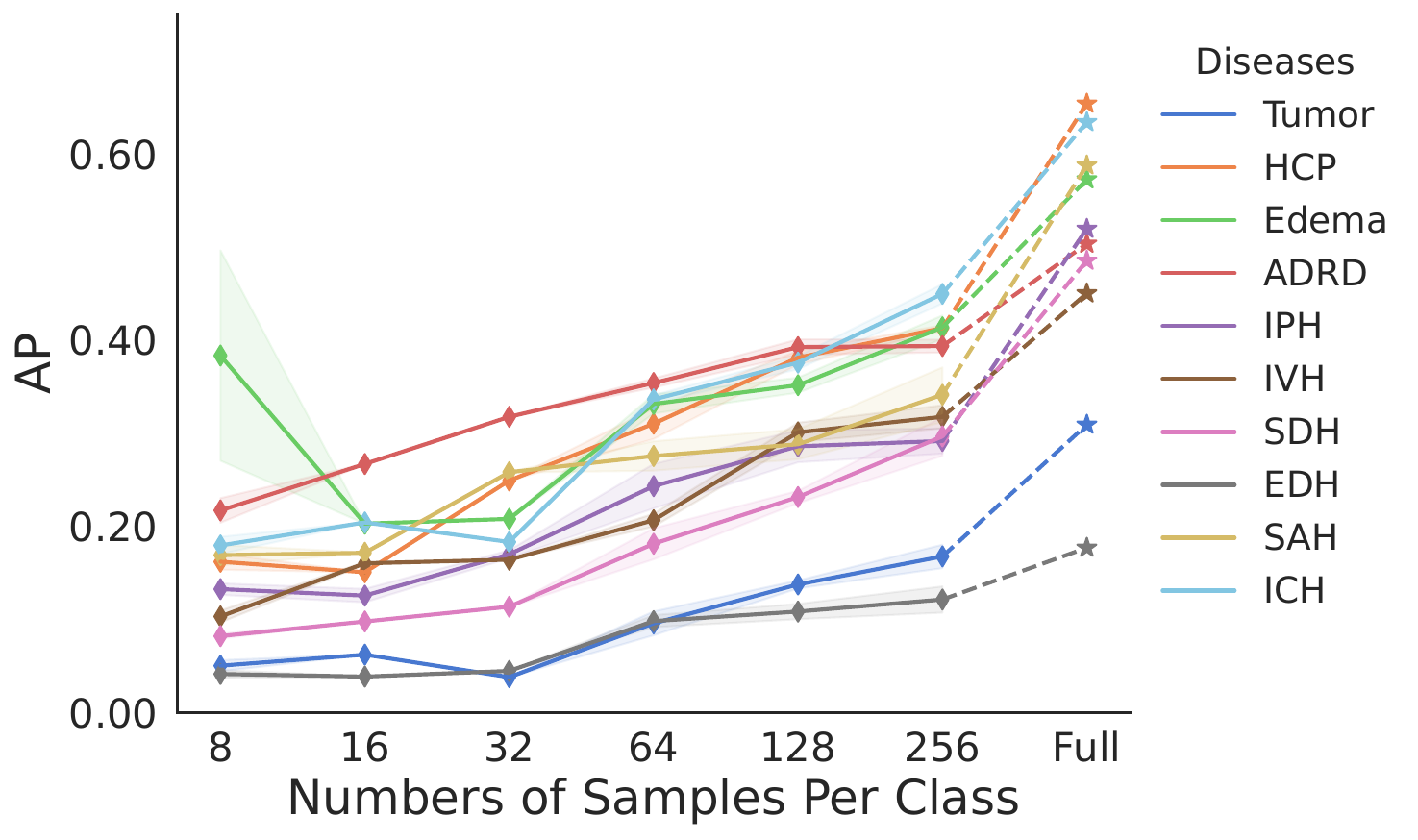} \\
    NYU Long Island \\
    \includegraphics[trim={0 0 140pt 0},clip,height=0.3\textwidth]{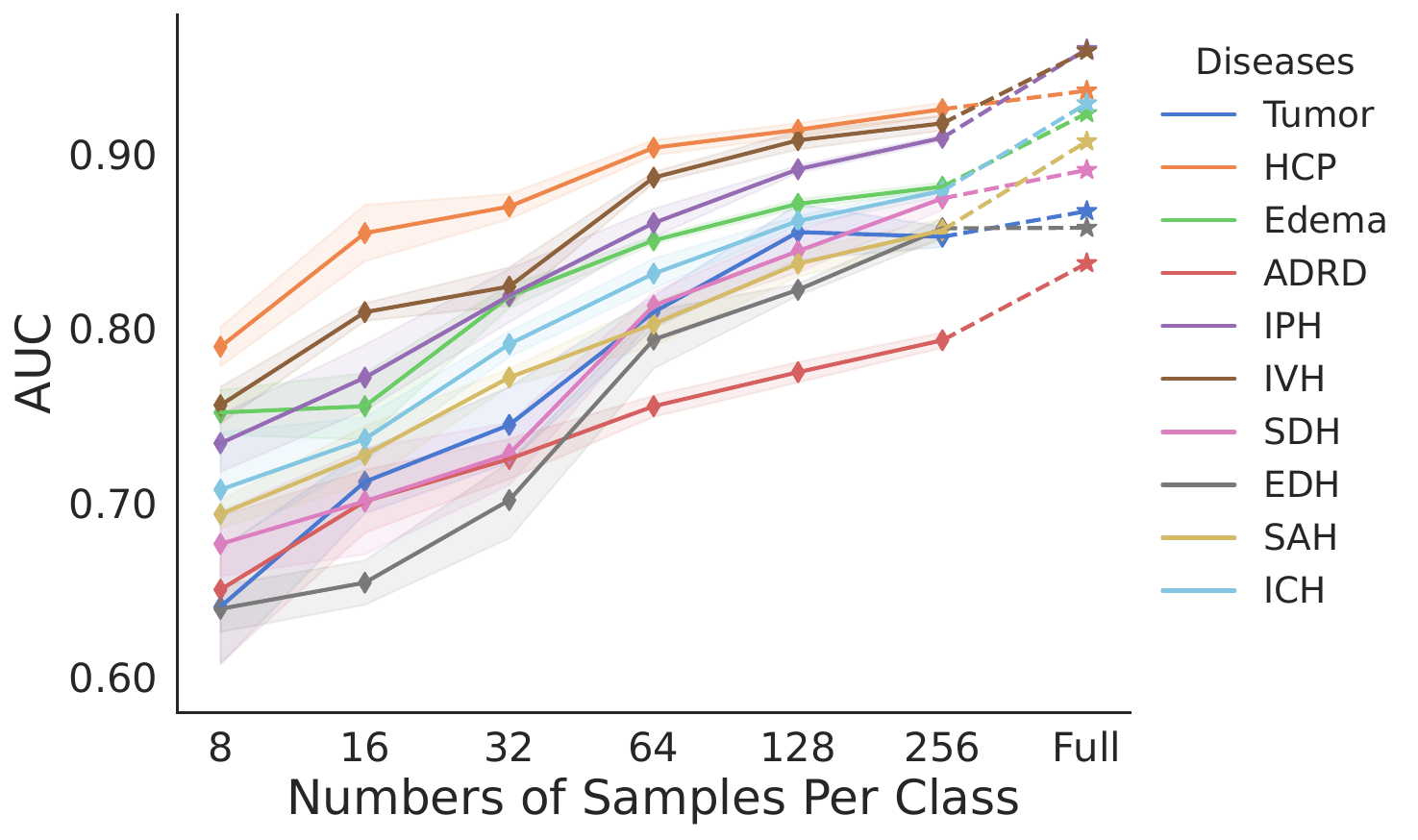}
    \includegraphics[height=0.3\textwidth]{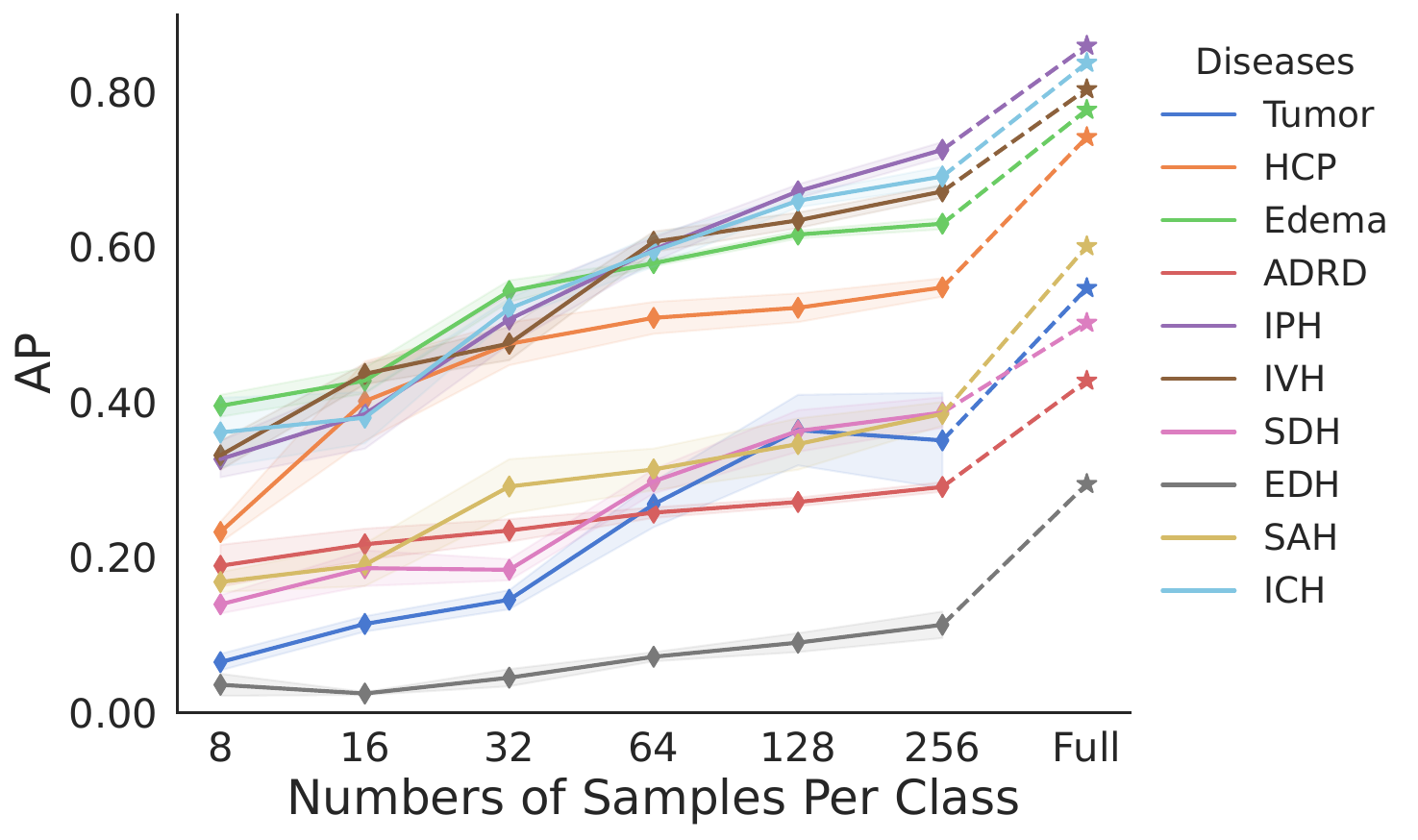} \\
    RSNA \\
    \includegraphics[trim={0 0 125pt 0},clip,height=0.295\textwidth]{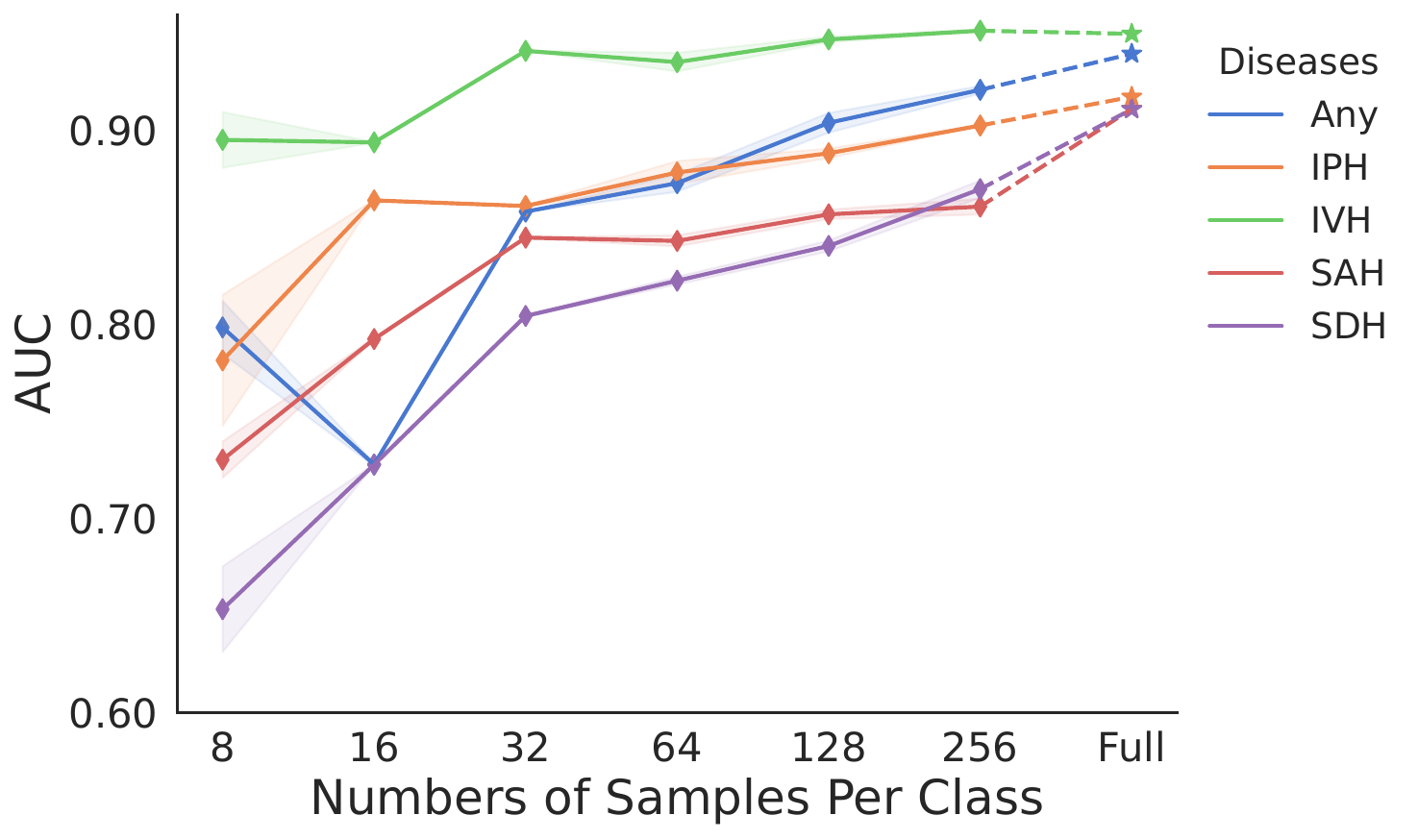}
    \includegraphics[height=0.295\textwidth]{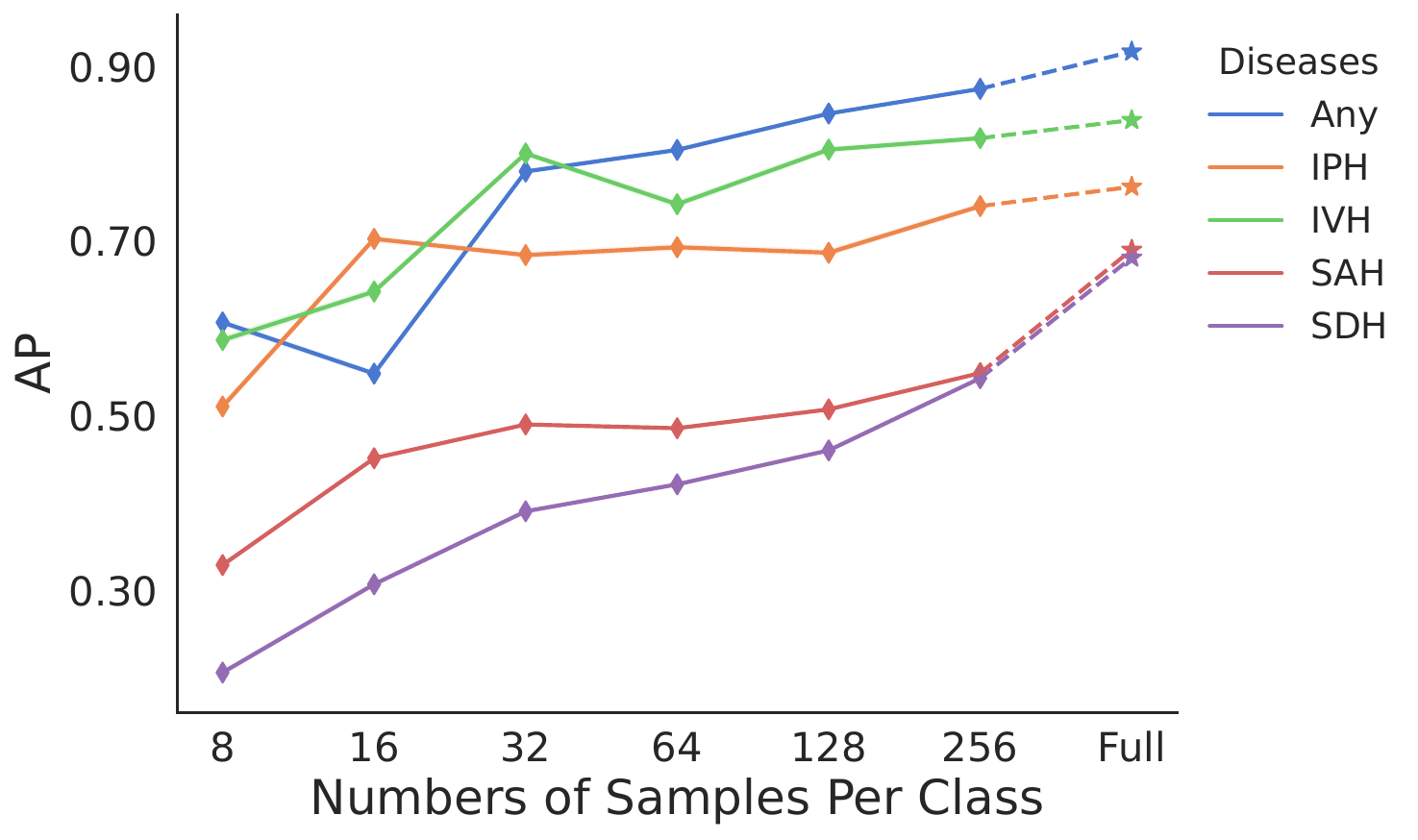}
    \caption{\textbf{Few-shot performance of the foundation model.}  The plots display the per-pathology AUC and average precision (AP) of the disease detection model under a few-shot learning setting, evaluated with varying numbers of training samples from the NYU Langone, NYU Long Island, and RSNA datasets. CQ500 is excluded since its small dataset size gives no enough positive samples for many diseases. Few-shot learning performance is compared to supervised finetuning with all training data (denoted by stars), demonstrating the strong generalization ability of the foundation model with limited training data. The confidence intervals are computed by 5 repeated experiments on resampling the training data and retraining the model.}
    \label{fig:fewshot}
\end{figure}

For in-domain fine-tuning, the foundation model is fine-tuned on each external dataset’s training set and validated on held-out sets from the same source. The bottom four rows in \Cref{fig:overview}e report the tasks-level performances on NYU Long Island and RSNA datasets. The fine-tuned model yields a macro-AUC of 0.904 across the 10 tasks on NYU Long Island dataset and a macro-AUC of 0.923 for five types of hemorrhages on the RSNA dataset. In comparison, the scratch model results in macro-AUC scores of 0.748 and 0.824, respectively.  Moreover, the foundation model also significantly outperforms Merlin, as shown in Supplementary \Cref{fig:radar-comparison-merlin}. The superior performances on external datasets indicate the generalizability of the foundation model. Note that the limited data size of CQ500 forbids training a effective deep learning model from scratch, reinforcing the importance of the foundation model in label efficiency, which is further studied in Section \hyperref[sec:label_efficiency]{"Label efficiency"}. Interestingly, when comparing performances across different datasets, \Cref{fig:overview}e demonstrates that the AUCs of the in-domain fine-tuned model on the external dataset even exceed the AUCs achieved on the internal dataset. For instance, the fine-tuned models consistently obtained AUCs greater than 0.90 in all the hemorrhage detection tasks on RSNA dataset, surpassing the AUCs on NYU Langone data. This may be attributed to the higher label quality in radiologist-reviewed datasets, for which label noise may be better controlled by comparison to EHR-derived labels.

In the full external validation without any site-specific fine-tuning (illustrated in \Cref{fig:overview}c), we evaluated classification models fine-tuned on the NYU Langone training set, as-is, on the held-out validation sets from each external dataset. \Cref{fig:overview}f compares performance between external validation and in-domain fine-tuning. Results show that, for the NYU Long Island and RSNA datasets —where the training set used for fine-tuning includes a sufficient number of high-quality labeled samples— in-domain fine-tuning does enhance the model performance. However, on the CQ500 dataset, with only 1,120 training samples, the in-domain fine-tuned model performs worse than the model transferred from NYU Langone, especially for EDH and SDH, which have a greater class imbalance. These comparisons highlight two typical use cases for foundation models depending on the availability of labeled data for fine-tuning. Additionally, comparing the first row of \Cref{fig:overview}e and external validation in \Cref{fig:overview}f, the fine-tuned model on NYU Langone achieves similar AUC values on both internal and external datasets, indicating robust generalizability to external data.

In addition to in-domain fine-tuning model performance comparison present in Supplementary \Cref{fig:radar-comparison-merlin}. Out-of-domain generalizability of our model is assessed in \Cref{fig:overview}g, where our model is first fine-tuned on in-domain finetuning dataset and then evaluated on external datasets (NYU Longisland, RSNA and CQ500). The result shows our model can demonstrate comparable performance on external dataset.

\subsection*{Volume to volume hemorrhage sub-types retrieval performance}
\label{sec:volume-retrieval}
\begin{figure}
\centering
\makebox[\textwidth][l]{%
    \hspace{0.25\textwidth}\textbf{RSNA}\hspace{0.33\textwidth}\textbf{CQ500}
} \\[0.2cm]
\includegraphics[trim={0 0 0mm 0},clip,height=0.27\textwidth]{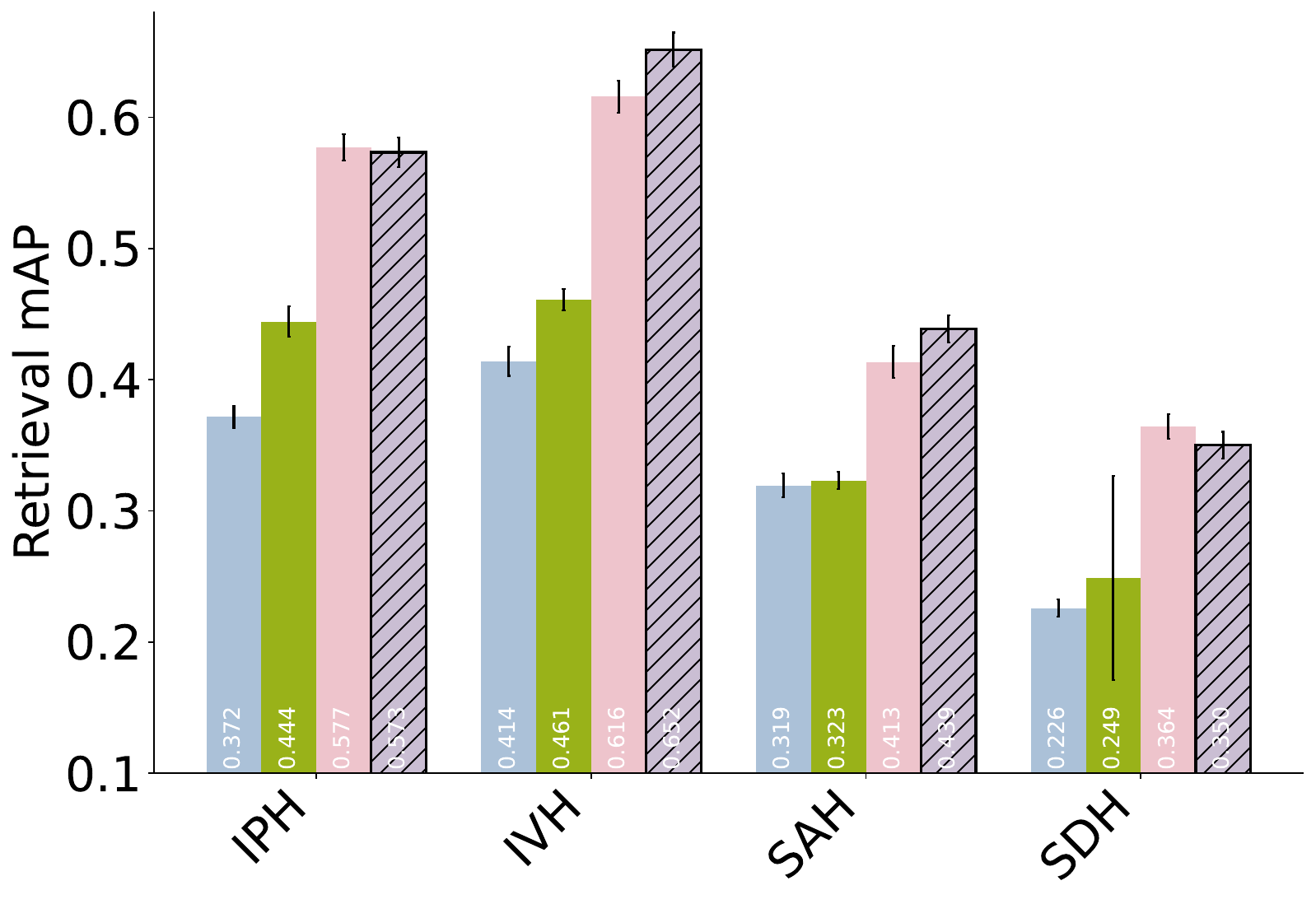}
\includegraphics[trim={0 0 5mm 0},clip,height=0.27\textwidth]{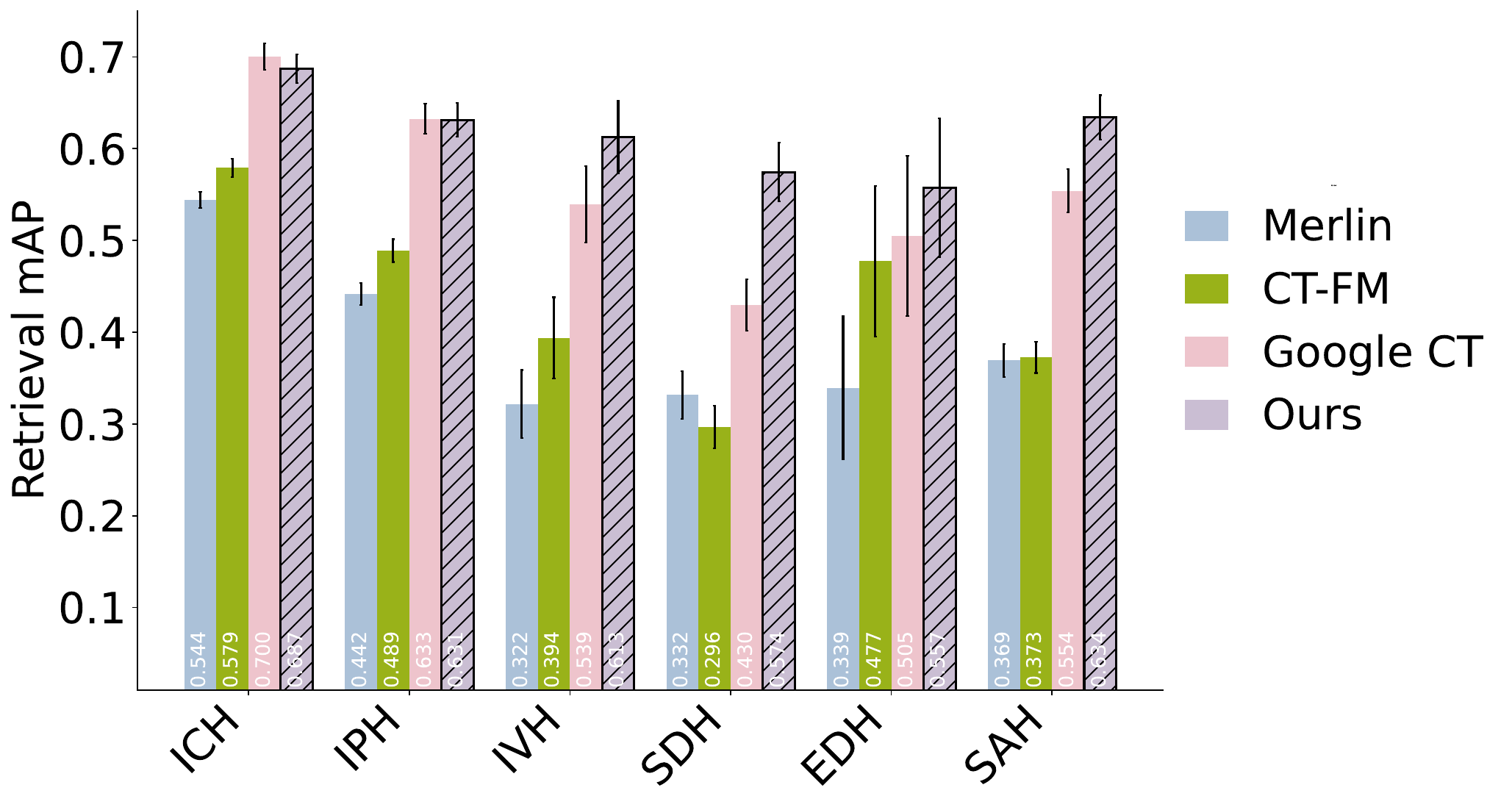}

\caption{\textbf{Volume-to-Volume Retrieval Performance Comparison} The plot presents mean Average-Precision (Retrieval mAP) for volume-to-volume retrieval with hemorrhage sub-types retrieval on RSNA and CQ500. All vs. all image retrieval is performed in this study, where every image in the dataset is used as a query once, and the gallery (the search space) is the entire dataset itself. Additional methodological details are provided in \hyperref[sec:methods]{``Method’’} Section. Additional evaluation on Precision$@K$ is present in Supplementary \Cref{fig:precision_retrieval}. In the plot, we show that our model shows better retrieval performance compared alternative models in majority of cases.}
\label{fig:mAP-retrieval}
\end{figure}

To further rigorously evaluate the representation quality of our pre-trained foundation model, we conducted a volume-to-volume hemorrhage subtyping retrieval study on RSNA and CQ500, comparing against Google CT, CT-FM, and Merlin (\Cref{fig:mAP-retrieval}). 

Retrieval performance was assessed using mean Average Precision (Retrieval mAP), which is conceptually distinct from the Average Precision (AP) metric used earlier in the paper for classification tasks. In the retrieval setting, mAP is computed as the mean of all individual query-level AP scores, where each AP score is defined as the average precision at the ranks where relevant items are retrieved, i.e., at positions $K=\{k_1,k_2,...,k_R\}$. In our study, we report AP at ranks $K=\{1,5,10\}$. A more detailed formulation of retrieval mAP can be found in the \hyperref[sec:methods]{“Method”} Section.

Our model demonstrates substantial improvements over CT-FM and Merlin, and achieves notable relative gains of $9.99\%$ on CQ500 and $2.21\%$ on RSNA compared with Google CT, averaged across hemorrhage subtypes. Additional retrieval methodological details are provided in \hyperref[sec:methods]{``Method’’} Section, and a comprehensive evaluation with Precision$@K$ is reported in the Supplementary \Cref{fig:precision_retrieval}.

\subsection*{Label efficiency of few-shot classification performance}
\label{sec:label_efficiency}

Another key advantage of the foundation model is its ability to facilitate transfer learning and fine-tuning tasks with minimal labeled data. For example, as shown in \Cref{fig:overview}c, the CQ500 dataset contains only 1,585 scans. Despite the small dataset size, fine-tuning our foundation model on CQ500 achieves promising results, with an AUC of 0.863. 

To systematically evaluate the label efficiency of our foundation model, we also assess the generalization capabilities of models on new tasks given a limited number of examples within the paradigm of few-shot learning, where only $K$ positive and negative samples each are used for training in each task. Since the quality of few-shot learning is largely determined by the sampled $K$-shots training data, we re-sampled and re-trained the model 5 times for calculating means and confidence intervals. As expected, \Cref{fig:fewshot} shows that performance improves as more data is used for training, with narrower confidence intervals. Surprisingly, even with a small number of examples (e.g., 512 total, with $K=256$), the model achieves performance comparable to training with the full dataset, which contains over at least 16 times more training examples in the RSNA. Notably, for tasks like detecting IVH in the RSNA dataset, the 8-shots model achieves an AUC above 0.90, a result that rivals full-data training. These findings suggest that our foundation model has learned diverse and expressive features/representations during SSL pre-training, making it highly effective for new tasks even when trained on small labeled datasets.

We additionally show few-shot model performance on CT-FM in Supplementary \Cref{fig:moedl_comparison_fewshot,fig:ct-fm_fewshot}, where it shows lower few-shot capability in comparison to our model.

\begin{figure}[t]
    \centering
    \includegraphics[width=0.24\textwidth]{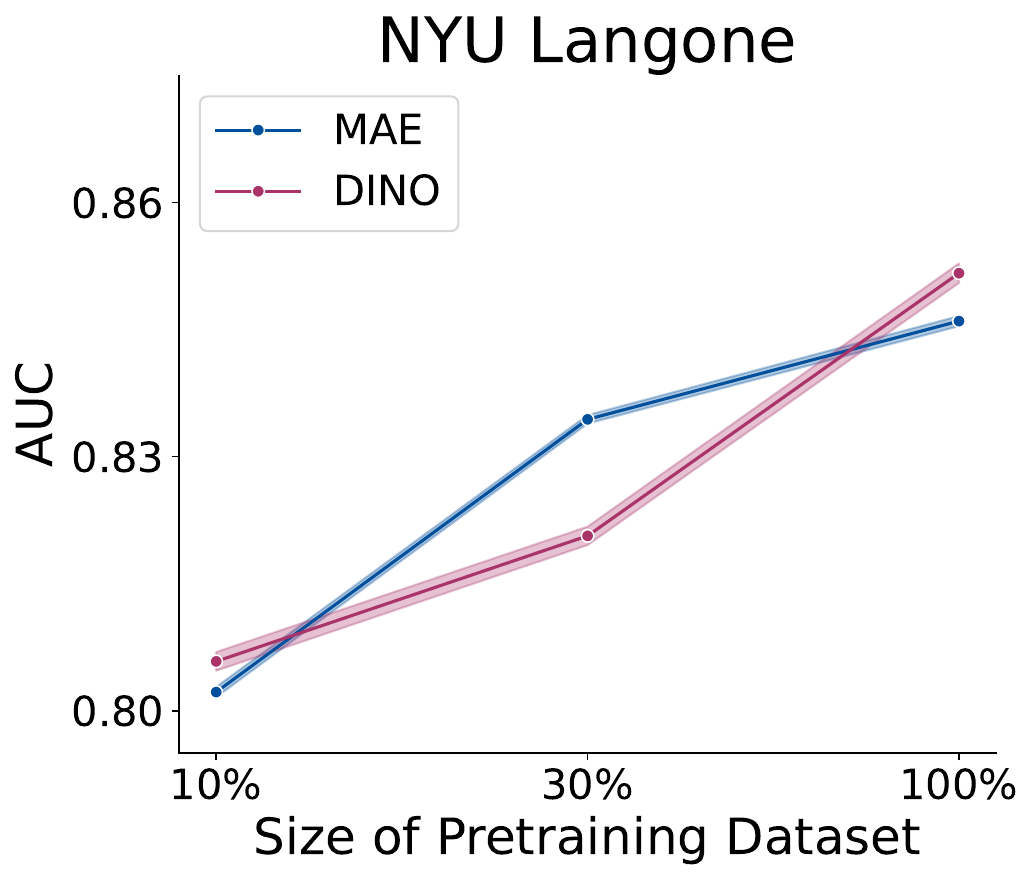} 
    \includegraphics[width=0.24\textwidth]{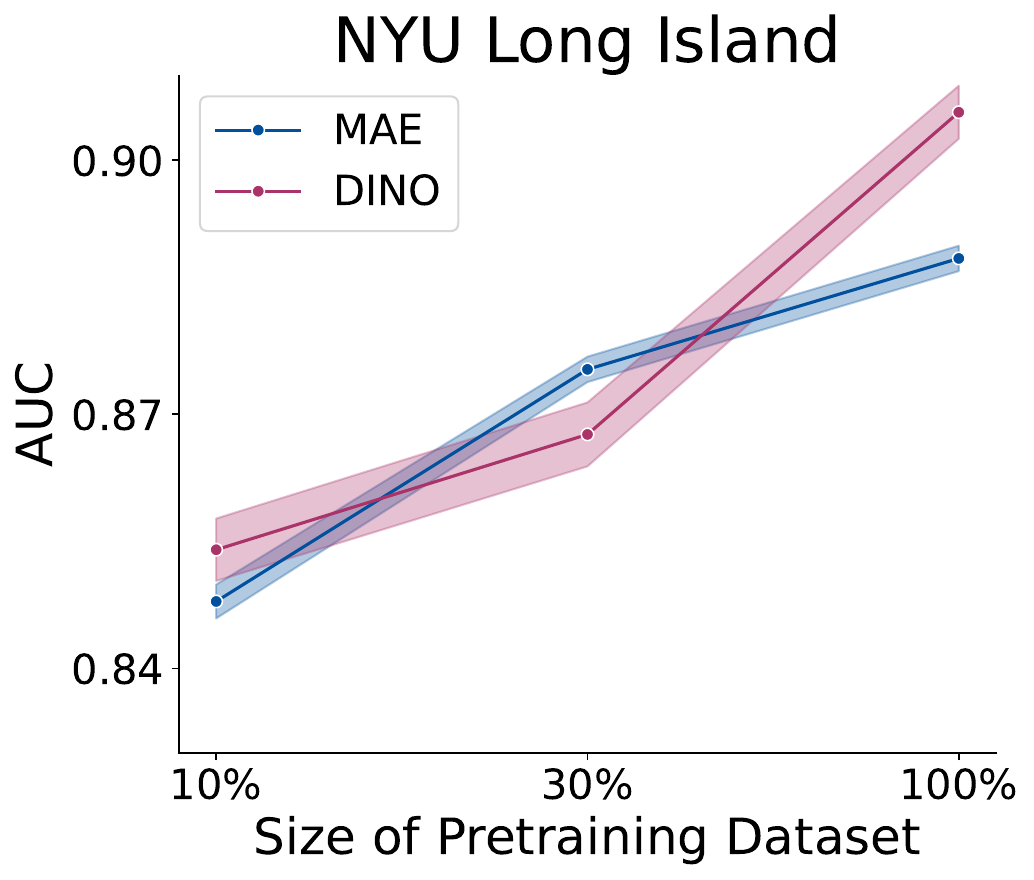}
    \includegraphics[width=0.24\textwidth]{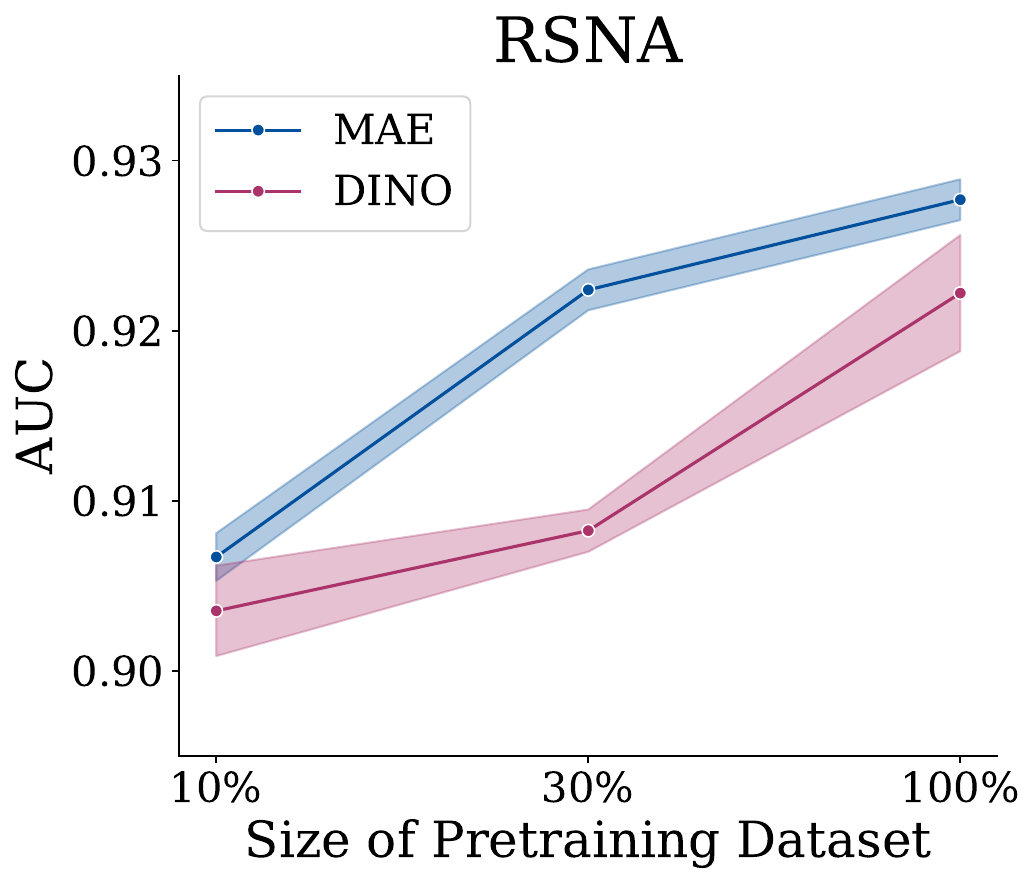}
    \includegraphics[width=0.24\textwidth]{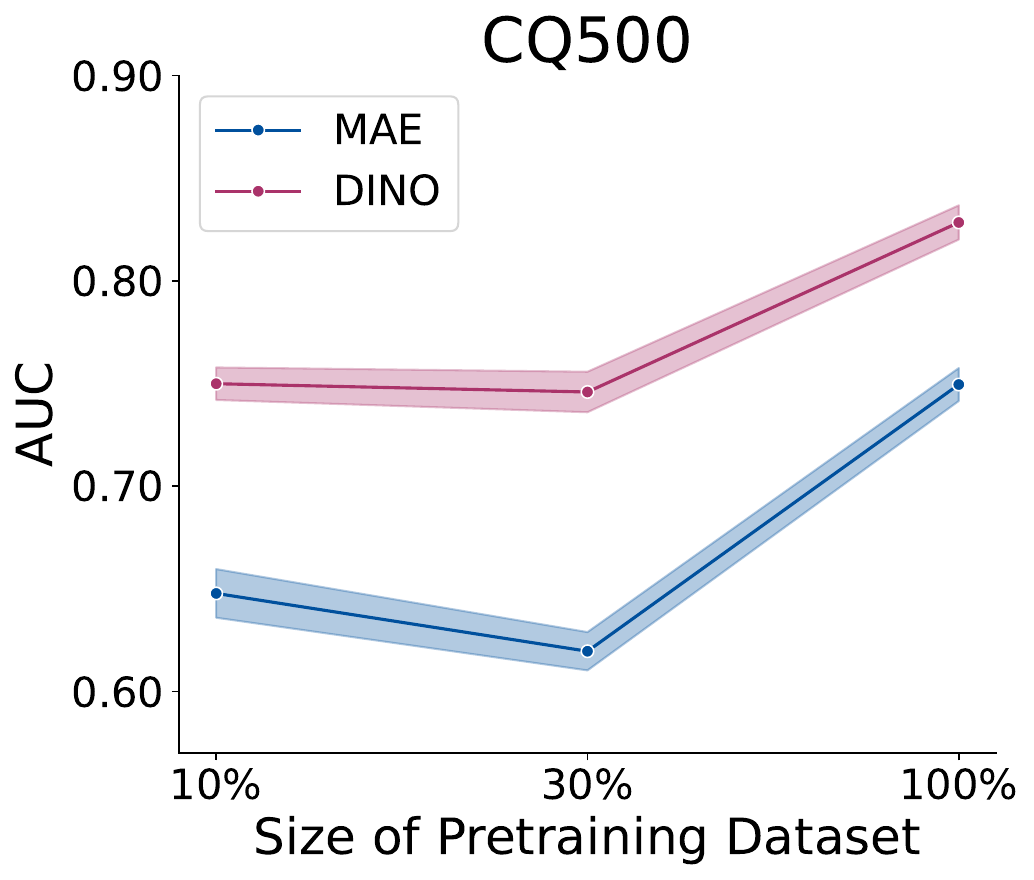}
    \includegraphics[width=0.24\textwidth]{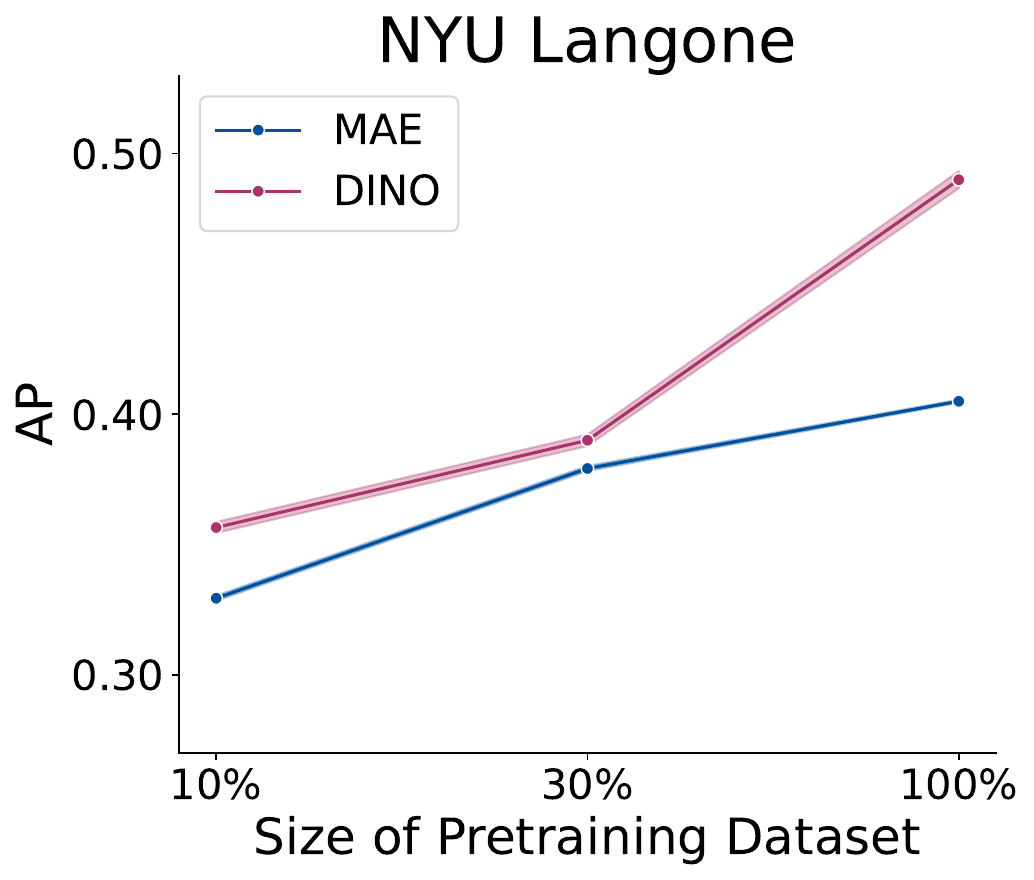}
    \includegraphics[width=0.24\textwidth]{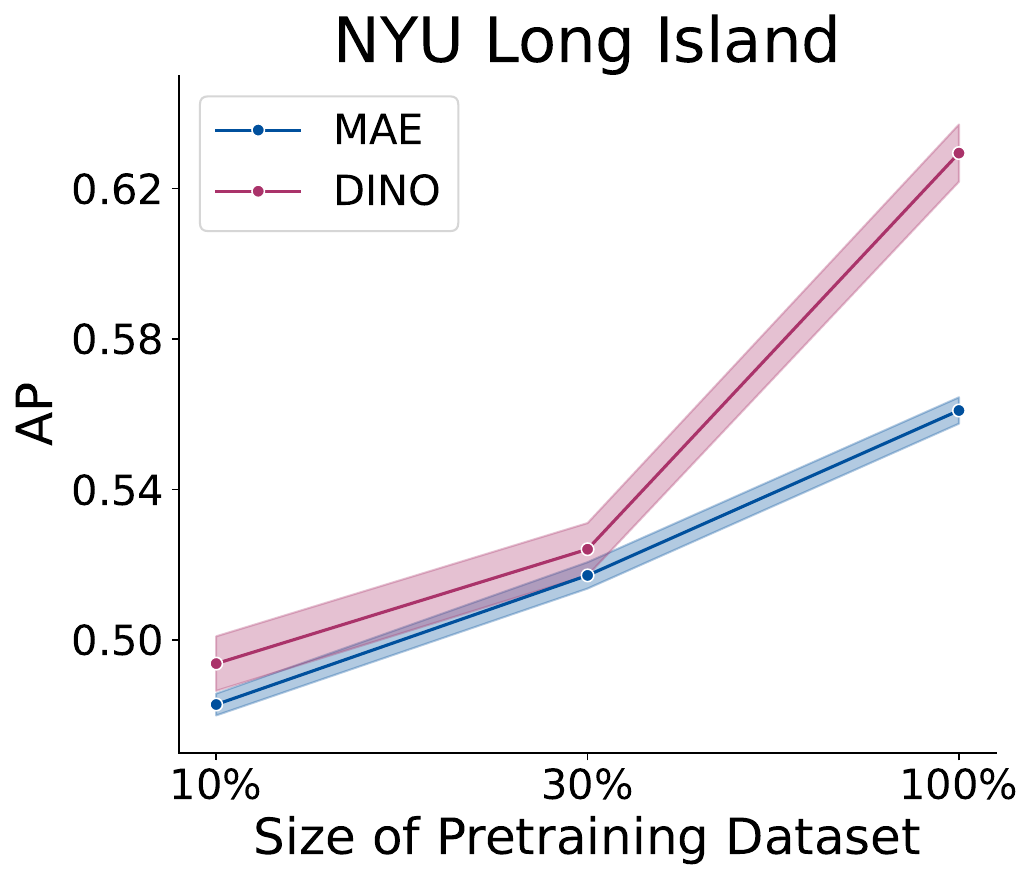}
    \includegraphics[width=0.24\textwidth]{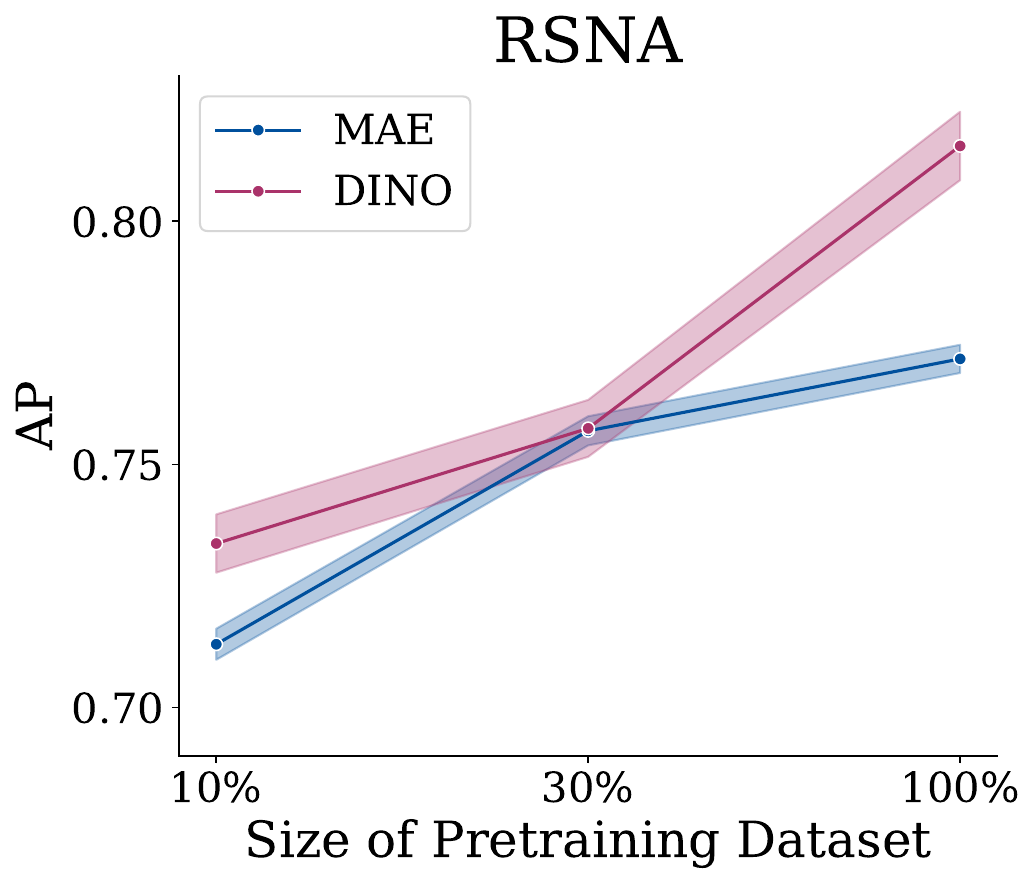}
    \includegraphics[width=0.24\textwidth]{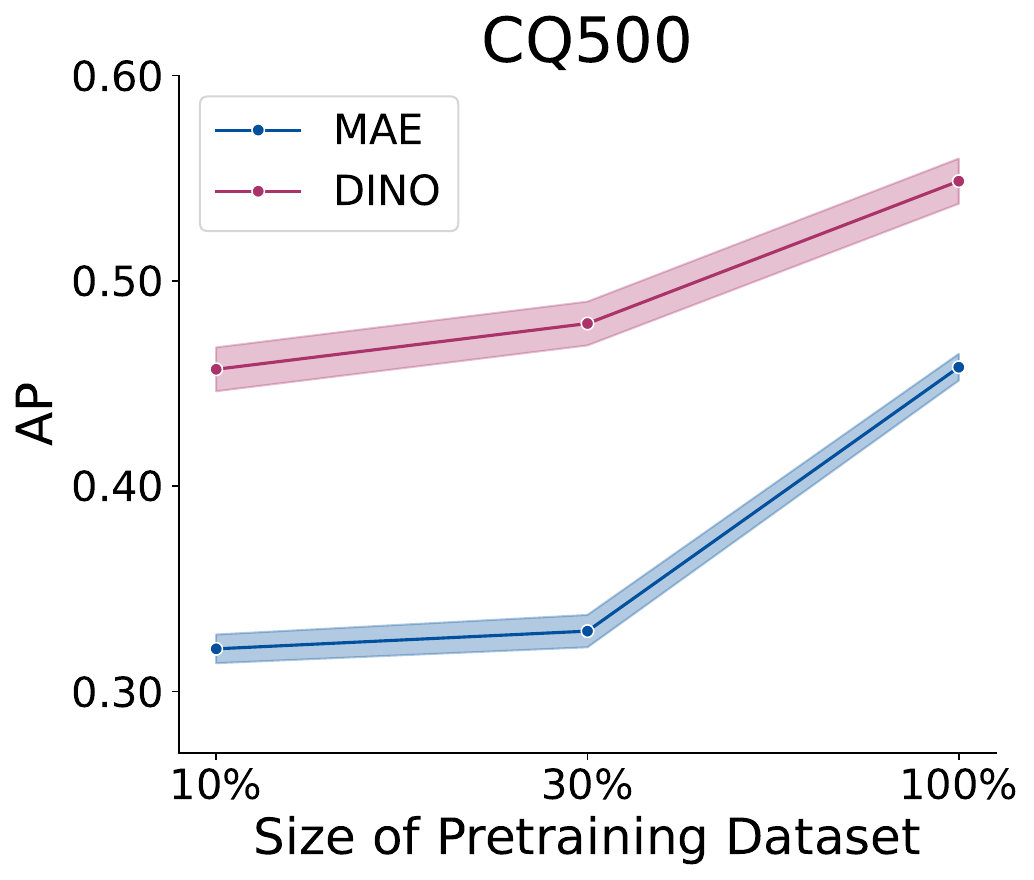}
    \caption{\textbf{Performance for Different Percentage of Pre-training Samples (Mean):} we compare the label efficiency in terms of different percentage of pre-training data for MAE vs. DINO. The $95\%$ CI are plotted in colour bands and the centre points of the bands indicate the mean value. We show that although DINO present higher label efficiency plot, both MAE and DINO efficiently scale up on downstream performance as more pre-training data is incorporated.}
    \label{fig:scaling_law}
\end{figure}

\subsection*{Comparison to alternative modeling choices}
\label{sec:model_efficiency}
In order to further verify the effectiveness of our proposed method on modeling 3D CT scans, we additionally compare to two alternative modeling choices --- 1) multiple instance learning (ABMIL~\cite{pmlr-v80-ilse18a}) with a state-of-the-art 2D foundation model DINOv3~\cite{siméoni2025dinov3}, 2) mean pooling with DINOv3, 3) modeling 3D CT Scans as video with a state-of-the-art video foundation model VJEPA2~\cite{assran2025vjepa2selfsupervisedvideo}. The detailed model fine-tuning comparison across four datasets is provided in Supplementary \Cref{fig:video_model_comparison}. Since both DINOv3 and VJEPA2 support dynamic resolutions, we keep the input size to be the same as our model for fair comparison ($96\times 96\times 96$). The result shows that our model perform best for majority of tasks with $2.57\%$ relative improvement on AUROC and $15.54\%$ on AP against DINOv3 with ABMIL, $9,26\%$ relative improvement on AUROC and $46.57\%$ on AP against DINOv3 with mean pooling, and $6.26\%$ relative improvement on AUROC and $26.25\%$ on AP against VJEPA2. Furthermore, from model efficiency analysis in Supplementary \Cref{fig:model_efficiency}, we show that our modeling method has significant advantage over other benchmarked architectures in term of both model throughput and memory cost.

\subsection*{Scaling up pre-training data}

Scaling laws have proven effective in enhancing the performance of foundation models by increasing the size of the training dataset~\cite{kaplan2020scalinglawsneurallanguage}. This phenomenon is not only observed in natural language and image domains~\cite{zhai22scalingvit, pmlr-v202-dehghani23a}, but also extends to medical imaging~\cite{zhou2023foundation, li2024well}. As shown in \Cref{fig:scaling_law}, scaling up the foundation model by incorporating more data during self-supervised pre-training significantly improves downstream tasks performances. We compared models pre-trained with varying proportions of the available data --- $10\%$, $30\%$, and $100\%$ (full dataset), observing that larger pre-training datasets consistently led to better downstream task performance. These findings highlight the potential of leveraging more data to achieve superior results, further suggesting the value of multi-institutional collaboration and federated approaches to aggregating larger datasets to enhance model quality. Noticeably, the performance for CQ500 does not change a lot from $10\%$ to $30\%$, but $100\%$ gives a sudden performance improvement, this indicates that for smaller datasets like CQ500, scaling up the data size is crucial for learning meaningful representations.

\subsection*{Visual Interpretation}
\begin{figure}[t]
    \centering
    \includegraphics[height=0.6\linewidth]{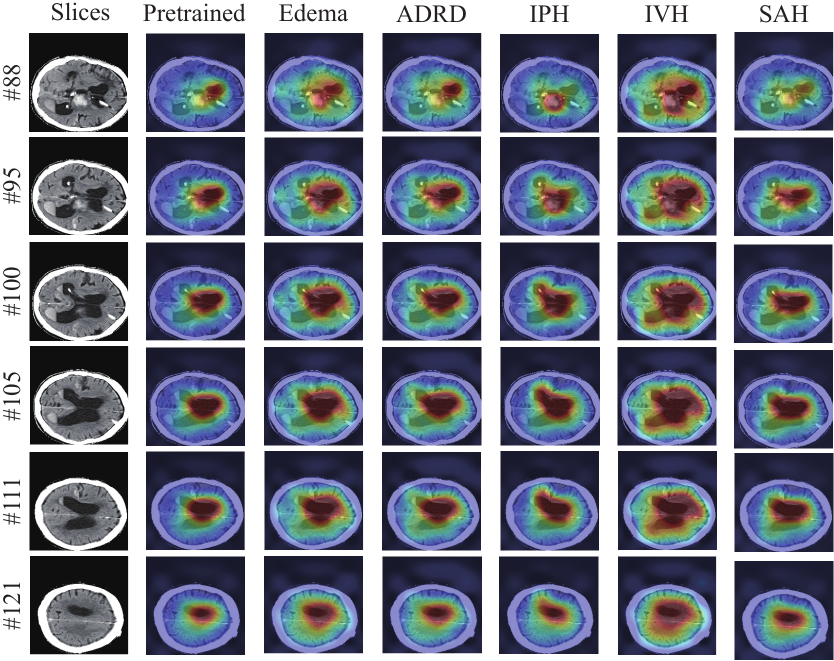}
    \caption{Visualization of ViT attentions on the scan.}
    \label{fig:attention_interpretation}
\end{figure}

To gain insight into the features learned through self-supervised pre-training and supervised fine-tuning of the foundation model, we visualize the attention maps within the Vision Transformer (ViT), as shown in \Cref{fig:attention_interpretation}. These heatmaps highlight the regions where the ViT model focuses most strongly. In the second column, we see that the pre-trained foundation model captures generic brain features, with dark red indicating attention on abnormal ventricular shapes and green marking areas of hemorrhage. After fine-tuning on specific tasks, the ViT’s attention becomes more focused on patterns relevant to each disease. For instance, in the edema task (third column), the heatmap extends across most of the brain, reflecting generalized swelling. For ADRD (fourth column), the model emphasizes regions of ventricular enlargement and cerebral atrophy. Multiple hemorrhages are also present in this sample, with attention covering both the IPH in the dense central region (fifth column) and extending toward the left end of the ventricle where IVH appears (sixth column). In the case of SAH (seventh column), the attention map is less prominent due to the small, peripheral area of the SAH in the lower part of the slice, although the model still predicts it accurately.

The comparison between the pre-trained and fine-tuned ViT explains the performance difference between linear probing and fine-tuning (shown in Supplementary \Cref{fig:probing-comparison-perpath}, as end-to-end fine-tuning allows the model to learn task-specific features more effectively. Details on the computation of the visualized attention maps are provided in Section \hyperref[sec:methods]{“Methods”}.

\section*{Discussion}
Despite advances in disease detection using 3D head CT scans, current solutions are limited by the availability of annotated data and the complex, task-specific design requirements of network architectures. These constraints hinder the broader application of machine learning in clinical disease detection. To address this, we developed a foundation model, trained on a large unlabeled dataset, to enable fine-tuning for multiple tasks with minimal labeled data under a unified network architecture. 

Highly accurate detection of intracerebral hemorrhages without delay is a critical clinical issue for the diagnostic decision making and treatment in an emergency room \cite{Hemphill2015-yw,Qureshi2009-ve}. Our results indicate that 3D Head CT scans can also be used to help identify hemorrhage subtypes and, more interestingly, etiology. High performances and generalizability observed by our model in detecting intracerebral hemorrhage have a potential to greatly assist in pre-hospital and early hospital management of blood pressure. This is particularly important given that early blood pressure control is a key factor in preventing hematoma expansion and improving patient outcomes~\cite{Macellari2014-zj,Morotti2022-lc}.

This approach is also particularly valuable for extending detection capabilities to new diseases in CT imaging. For example, early detection of ADRD with deep learning has traditionally relied on MRI scans \cite{Li2019-jx, pmlr-v116-liu20a, Xue2024}. However, access to MRI machines is costly and often restricted by patients’ geographic location and socioeconomic status~\cite{https://doi.org/10.1002/neo2.10}. Head CT, in contrast, is fast, accessible, and is the first-line imaging test in emergency and diagnostic settings. Our foundation model enables more accessible ADRD detection using head CT scans. This advancement holds the potential for expanding early ADRD detection in common public health settings for the older population~\cite{lin_dementia_2020, kim2021racial}, such as emergency rooms, as well as in underserved communities nationally and internationally in which CT is more available than MRI. Similarly, our model could facilitate the development of detection tools for other conditions, such as cancers and neuroinfectious diseases, thus supporting population health on a broader scale.

Our study demonstrates that this pre-trained foundation model significantly outperforms models trained from scratch and other CT foundation models on the same labeled data. Moreover, it exhibits strong performance even with limited data, as shown in few-shot learning experiments, and suggests promising potential for scaling up with larger datasets. In clinical practice, head CT scans are typically acquired using heterogeneous protocols, including variations in slice thickness and scanner modalities. A robust foundation model for CT should generalize effectively across these diverse protocols. In this study, we utilized scans with slice thicknesses ranging from 0.5 mm to 5 mm and data from two major manufacturers (Siemens and Toshiba) to develop and assess the generalization capabilities of our foundation model. As illustrated in Supplementary \Cref{fig:batch_effect}, the embeddings produced by the foundation model show separability based on scanner manufacturer and slice thickness, likely reflecting variations in protocol distribution. However, by comparing the distribution of “All” patients to that of positive cases for each condition, we observe that the embeddings do not tend to collapse or bias towards a certain protocol. Supplementary \Cref{fig:thickness-ablation} further demonstrates that fine-tuned models achieve comparable performance across scanner protocols. Detailed per-task performance results are provided in Supplementary \Cref{fig:slice_thickness_per_pathology,fig:manufacturer_per_pathology}.  Additionally, in comparisons of Toshiba and Siemens scanners, we noted a systemically higher prevalence of positive cases across all tasks in Toshiba scans, leading to a modestly higher AUC in specific instances. Despite these variations, our foundation model demonstrates robust generalization capabilities across diverse CT protocols, highlighting its potential for broad clinical application.

However, our evaluation is limited by label noise in real-world datasets. Labels derived from electronic health records (EHRs) can suffer from missing or incomplete information. This issue is evidenced by the model’s lower performance on NYU Langone data compared to RSNA data, where labels were rigorously reviewed by radiologists. Another limitation is that, due to constraints on training samples and computational resources, our model does not yet fully explore the potential of scaling laws. The 361K scans used for pre-training represent the entirety of CT scans available from a single large clinical institution, highlighting the need for multi-institutional collaborations to enhance the dataset diversity and volume. With greater computational resources, we could also scale up the model’s size, resolution of image patches, and number of tokens used in the ViT architecture, potentially improving performance for detecting conditions with small spatial manifestations, such as subarachnoid hemorrhage (SAH).

While our current results primarily focused on disease detection, our foundation model holds significant potential for advancing disease prognosis analysis. For instance, the prediction of decompensation, particularly hemorrhagic expansion, is an important potential use of the foundation model and may lead to the development of novel hyperacute treatment strategies~\cite{hematoma_expansion}. Additionally, critical applications in acute ischemic stroke, such as predicting hemorrhagic transformation and the development of malignant edema can benefit from the foundation model. Beyond acute conditions, the foundation model can potentially also be used to predict the development of ADRD~\cite{Zhu2024-zd}.

\section*{Methods}
\label{sec:methods}
\subsection*{Datasets}
\subsubsection*{Dataset for pre-training foundation model}
We utilized a large-scale head CT scan dataset from NYU Langone, consisting of 499,084 scans across 203,665 patients, collected between 2009 and 2023. These scans were acquired using Siemens and Toshiba machines. We included all the non-contrast head CT scans with ranging from 0.5mm to 5mm, kVp values between 70 and 150, and convolution kernels Hr/Qr/J with sharpness levels of 35-45. We filtered out corrupted scan series with missing DICOM files and those containing less than 10 slices, resulting in 451,298 scans. We partitioned these scans by the patient IDs into training, validation, and held-out validation sets in an 8:1:1 ratio to avoid the leakage of scans from the sample patient. As illustrated in \Cref{fig:overview}a, this led to training, validation and held-out validation set with 361,663, 44,886 and 44,749 scans, respectively. The scans in the training set were used to train the foundation model.

\subsubsection*{Datasets for downstream tasks}
We evaluated our model using four datasets: one \emph{in-domain (ID)} dataset from NYU Langone and three \emph{out-of-domain (OOD)} datasets from NYU Long Island, the RSNA Challenge, and the public CQ500 dataset. Each dataset includes multiple head CT disease detection classes, with some classes abbreviated as follows: Hydrocephalus (HCP), Dementia (ADRD), Intraparenchymal Hemorrhage (IPH), Intraventricular Hemorrhage (IVH), Subdural Hemorrhage (SDH), Epidural Hemorrhage (EDH), Subarachnoid Hemorrhage (SAH), and Intracerebral Hemorrhage (ICH). These classes can have co-occur in the same Head CT scan. The characteristics of the patients are shown in \Cref{tab:characteristic}. We split all datasets by patients to avoid information leakage, Further dataset details of our dataset are provided below.

\paragraph{NYU Langone - 10 detection tasks} The NYU Langone main campus dataset serves as the internal ID dataset for downstream evaluation. As depicted in \Cref{fig:overview}b, patient health status was derived from Electronic Health Records (EHR) within a 3-month window centered around the scan date, with conditions defined by ICD-10 diagnostic codes and medications, outlined in Supplementary \Cref{tab:disease_definition}. This cohort includes 270,205 scans from 66,801 patients with valid EHRs, covering 10 classes: Tumor, HCP, Edema, ADRD, IPH, IVH, SDH, EDH, SAH, and ICH. This cohort was partitioned following the same split used used during pre-training: matched patients within the training, validation, and held-out subsets of the SSL pre-training phase were assigned to the corresponding sets of the supervised fine-tuning phase. This resulted in 217,109, 26,609, and 26,487 scans in the training, validation and test set, respectively.

\paragraph{NYU Long Island - 10 detection tasks} NYU Long Island data was acquired in Long Island hospital which used to be a separate hospital, severing as an OOD dataset. This dataset includes 22,158 samples with 10 classes, labeled similarly to the NYU Langone dataset using ICD-10 codes. It is partitioned into an 8:1:1 train-validation-test split.

\paragraph{RSNA - 5 detection tasks} The RSNA Head CT Challenge dataset~\cite{flanders_construction_2020} serves as a public external OOD dataset, collected from Stanford University, Thomas Jefferson University, Unity Health Toronto, and Universidade Federal de São Paulo (UNIFESP). The dataset, initially provided as 2D slices, was reorganized by subject ID, retaining subjects with complete slice data. After preprocessing, the dataset consists of 10,579 samples across five classes: Any (any hemorrhage type), IPH, IVH, SAH, and SDH. Dataset labels were assigned by 60 volunteers from the American Society of Neuroradiology (ASNR). We partitioned this cohort into an 8:1:1 train-validation-test split.

\paragraph{CQ500 - 10 detection tasks} The CQ500 Head CT dataset~\cite{CQ500} serves as another public external OOD dataset, collected from multiple centers in India. This dataset includes 1,585 samples including varying slice thickness across 10 selected classes: ICH, IPH, IVH, SDH, EDH, SAH, BleedLocation-Left, BleedLocation-Right, MidlineShift, MassEffect. Each scan was labeled by three senior radiologists, and the cohort was split into an 8:1:1 train-validation-test ratio.

\paragraph{RSNA — 4 retrieval tasks; CQ500 — 6 retrieval tasks}
We evaluate volume-to-volume hemorrhage subtype retrieval on both RSNA and CQ500. For each positive case of a given subtype (IPH, IVH, SAH, and SDH in RSNA; ICH, IPH, IVH, SDH, EDH, and SAH in CQ500), the objective is to retrieve other cases with the same subtype. Retrieval is conducted in an all-vs-all setting, where every positive sample in the dataset is used once as a query against the remaining cases in the gallery (rest of samples in the dataset).

\subsubsection*{Label acquisition from electronic health records}

As illustrated in \Cref{fig:overview}b, we labeled head CT scans from NYU Langone and Long Island Hospital using electronic health records (EHR). For each head CT, we retrieved an EHR snippet for the corresponding patient based on their Medical Record Number (MRN), starting from the time of the scan and covering a 90-day period. We then checked for the presence of any diagnosis codes (ICD-10 codes) and medication records, within this EHR snippet that matched the predefined definitions for each disease, allowing us to create binary labels for each condition. The complete list of ICD-10 codes and the medications used for disease definitions is provided in Supplementary \Cref{tab:disease_definition}.


\subsubsection*{Data preprocessing}
For the NYU Langone and Long Island datasets, we converted the DICOM files into NIfTI format using MRIcroGL dcm2nii~\cite{li_first_2016}, standardizing the file format with those from the RSNA and CQ500 datasets. Given the variability in scan protocols, which can result in differing orientation, resolutions and slice thicknesses, we applied spatial normalization to transform the volume orientation to right-anterior-superior (RAS) angle and resample with bicubic interpolation to the isotropic resolution ratio of $(1.0, 1.0, 1.0)$ in the world coordinate system. This ensures uniform pixel spacing across all scans and axes. 

Head CT scans use Hounsfield Units (HU) to represent various tissue types, which span a broad range of values. To better capture tissue characteristics, we applied three windowing ranges, each emphasizing specific tissue types: (40, 80) for soft tissue, (80, 200) for contrast-enhanced tissues and blood vessels, and (600, 2800) for bone. We then stacked the values from each window, producing a 3-channel 3D volume that enhances the representation of these key tissues. Similar strategy has been applied in Chilamkurthy \textit{et al.}~\cite{CQ500}.

To ensure compatibility with model input requirements, we transformed each volume into the desired size. We first padded or cropped each volume to a size of (224, 224, 224), preserving the whole brain across all axes. Then for training, we applied data augmentations detailed in Supplementary Section \hyperref[sec:dataaug_details]{``Data Augmentation details’’}; for evaluation, we center-cropped the volumes to (192, 192, 192). Finally, we resized each volume to (96, 96, 96) as the input size for the model.

\subsection*{Model architecture}
\label{sec:model_architecture}
Numerous studies have demonstrated that ViT can effectively learn high-quality representations for 2D medical images at scale~\cite{chen_towards_2024, zhou2023foundation, MedSAM, Vorontsov2024, Azizi2023}. Our study extends this by exploring whether representations of 3D medical images (specifically head CT scans) can also be effectively learned at scale through the direct compression of 3D patches as model input. We employ the Vision Transformer (ViT)\cite{dosovitskiy2020vit} as the volume encoder for our foundation model, as well as for baseline comparisons in all experiments. Our model uses a ViT-Base architecture with an embedding dimension of 768, 12 self-attention layers, 12 heads, and feed-forward layers with a hidden size of 3072. We apply sine-cosine absolute positional encoding\cite{NIPS2017_3f5ee243} across all pre-training and fine-tuning stages.

For the 3D input volume, instead of creating 196 patches of size $16 \times 16$ from a $224 \times 224$ image as in standard 2D ViT, we segment $96 \times 96 \times 96$ 3D volumes into 512 patches of size $12 \times 12 \times 12$ for ViT input. This customized patch design considers the trade-off between performance and computational cost. As shown in Supplementary \Cref{fig:patches-ablation}, our model outperforms a version using 216 patches of size $16 \times 16 \times 16$, indicating that smaller, more numerous patches enhance model performance. This supports the importance of capturing fine-grained features in 3D medical imaging, consistent with prior findings~\cite{Tang_2022_CVPR, li2024well}. However, computational costs increase significantly with respect to $s$ ($s$ defined as patch size reducing factor), at a rate of $O(s^{6})$, due to the cubic growth of patch numbers in 3D and the quadratic growth in self-attention computation (\Cref{apd:self_attention_rate}). To balance performance with computational efficiency, we adopt 512 patches of $12 \times 12 \times 12$ as the optimal input size for ViT in our foundation model.

\subsection*{Self-supervised pretaining}
Self-Supervised Learning recently has been widely adopted as learning framework for building medical foundation models~\cite{chen_towards_2024, zhou2023foundation, Huang2023, azizi21big, Vorontsov2024}. While previous works mainly focus on directly applying existing self-supervised learning algorithms on 2D medical images, we explore how to effectively leverage these algorithms with 3D medical images. Specifically, we explore two main branches of self-supervised learning framework for building our 3D foundation model --- discriminative with self-distillation (DINO) and masked image modeling (MAE).

\paragraph{Self-Distillation Modelling (DINO)}
DINO\cite{caron2021emerging, oquab2024dinov} is a self-supervised learning method shown promising and robust downstream evaluation performance in previous studies on different areas~\cite{chen_towards_2024, Vorontsov2024}. DINO uses a student-teacher framework for learning meaningful representations. Both student and teacher networks share the same model architecture, while the teacher’s parameters are updated using an exponential moving average of the student’s parameters. Each input image is augmented multiple times to create different views as student and teacher networks input. Specifically, we applied random global and local crops, random flips, shifts in intensity and contrasts, and Gaussian blurs for augmented views. Then the student’s output is trained to match the teacher’s output using a distillation loss, ensuring similar representations for different views of the same image. We pre-trained the ViT in the DINO framework for $1000$ epochs with batch size at $64$ per GPU and an AdamW~\cite{loshchilov2018decoupled} optimizer ($\beta_1=0.9, \beta_2=0.95$, $0.05$ weight decay). A base learning rate $3\times10^{-4}$ was applied combined with cosine scheduling and a linear warmup on the first $5$ epochs. During pre-training, two global augmentations and three local augmentations were applied to enable ViT to learn both global and local features of the head CT. Because small region of brain is likely to be dissimilar, we observed cropping too small brain regions would cause unstable model training by making the learning task to be too challenging. Therefore, we first resample the input images to $224\times224\times224$. Subsequently, we perform multi-scale cropping by extracting both global and local crops regions, ranging from $112\times112\times112$ to $224\times224\times224$ for global crops and from $64\times64\times64$ to $112\times112\times112$ for local crops. After the cropping, all cropped regions are resampled to $96\times96\times96$. For training on $100\%$ data, convergence on the performance for downstream tasks is observed at around $300$ epochs, which took around one week on four 80GB NVIDIA A100 GPUs. 

\paragraph{Masked Image Modeling (MAE)}
MAE~\cite{He2021MaskedAA} is another self-supervised learning method for vision tasks, inspired by masked language modeling in Natural Language Processing (NLP). MAE is trained to reconstruct randomly-masked patches via an encoder-decoder architecture, where the encoder processes visible patches of an image, while the decoder reconstructs the image from encoded patches and mask tokens. Specifically, we randomly masked the patches from each volume with a probability of 0.75. Mean squared error (MSE) loss is optimized to minimize the difference between the reconstructed volume and the original volume. We pre-trained the ViT in MAE framework for $400$ epochs with batch size at 64 per GPU and an AdamW~\cite{loshchilov2018decoupled} optimizer ($\beta_1=0.9, \beta_2=0.95$, $0.05$ weight decay). A base learning rate $1.5\times10^{-3}$ was applied combined with cosine scheduling and a linear warmup on the first $5\%$ steps,  For training on $100\%$ data, convergence is observed at around $250$ epochs, which took around 4 days on four 80GB NVIDIA A100 GPUs for MAE. Similar to DINO, MAE has shown success in learning robust representations in many previous works~\cite{ravi2024sam2, tong2022videomae, gupta2023siamese, zhou23self, huang2022masked, cong2022satmae, chen23masked}, including the studies on both 2D and 3D data.

We compared the performance on downstream tasks between two versions of foundation models pre-trained using DINO and MAE, as shown in \Cref{fig:scaling_law} and Supplementary \Cref{fig:probing_comparison,fig:probing-comparison-perpath,fig:probing-comparison-perpath-dino}. The results indicate that DINO consistently outperforms MAE across all datasets. Based on this finding, we selected the DINO-pre-trained model as our final foundation model.

\subsection*{Evaluation setting}
\paragraph{Baseline comparisons}
Since no prior foundation models have been specifically trained on 3D Head CT for direct comparison, we benchmark our model against Merlin~\cite{blankemeier2024merlinvisionlanguagefoundation} and Google CT Foundation model~\cite{yang2024advancingmultimodalmedicalcapabilities} to highlight the advantages of our domain-specific foundation model. Merlin is a 3D Abdomen CT foundation model pre-trained on vision-language pairs with contrastive learning~\cite{pmlr-v139-radford21a} and ICD code prediction task, where 6+ million images from 15,331 CTs, 1.8+ million diagnostic ICD codes from EHR, and 6+ million tokens from radiology reports are used. Different from our model architecture, Merlin used ResNet-152 ($\sim60.4$M Parameters) as vision model with reshaped image size of $224\times224\times160$. The performance comparison between our model and Merlin is shown in Supplementary \Cref{fig:radar-comparison-merlin}, where our model shows substantial improvement across most datasets and diseases. Google CT Foundation model is trained on a comprehensive private dataset comprising 527,078 CT studies with associated radiology reports from 430,772 patients. The model is first trained by Contrastive Captioning with CoCa~\cite{yu2022coca} on 2D medical images and then adapting to CT by training on series of CT slices with VideoCoCa~\cite{yan2023videococavideotextmodelingzeroshot}. The performance comparison between our model and Google CT Foundation model is shown in Supplementary \Cref{fig:probing-comparison-gemini}, where our model shows a consistent improvement across the board. We additionally show comparison of our model against model trained from scratch in \Cref{fig:overview} and Supplementary \Cref{fig:radar-comparison-merlin}, where the overall significantly improved performance shows the effectiveness of our pre-training strategies on 3D Head CT images.

\paragraph{Fine-tuning and Probing classification evaluation}
We assessed pre-trained model performance through full fine-tuning (updating all weights) and various probing methods (updating only the classification layers). For both approaches, images were normalized to isotropic spacing, transformed to three HU interval channels, and reshaped to $3\times96\times96\times96$. The entire transformed 3D image was then input into the ViT model for feature extraction, followed by an additional classification layer for downstream tasks. Probing utilized two strategies: linear probing, which adds a linear layer atop the ViT backbone, and attentive probing, which incorporates an attention layer. Attentive probing is chosen since MAE does not use \texttt{[CLS]} token as the learning objective. Linear probing only relies on \texttt{[CLS]} token to perform classification and attentive probing explores the interaction among all tokens~\cite{Chen2024}. Given the imbalances of downstream task labels, we randomly sampled a balanced subset from the training set per epoch, consisting of 5,000 samples (when fine-tuning on the NYU Langone, NYU Long Island, and RSNA datasets), and 500 samples when fine-tuning on CQ500. We trained all methods using the AdamW~\cite{loshchilov2018decoupled} optimizer with a cosine learning rate scheduler, a learning rate of $1\times10^{-5}$ for backbone and $1\times10^{-3}$ for classification layers, cross-entropy loss, and a maximum of 10 epochs. The main evaluation result with linear probing is shown in \Cref{fig:overview} with fine-tuning and probing comparison shown in Supplementary \Cref{fig:probing_comparison} for average performance across all diseases and Supplementary \Cref{fig:probing-comparison-perpath} for per disease performance. The result indicates that probing achieves performance levels close to full fine-tuning, underscoring the high quality of learned representations in our model.

For fine-tuning model from scratch, as we observe more unstable model performance from different hyper-parameters across different datasets, we perform hyper-parameters sweep across different setting and report the best performance model. The sweeping hyper-parameters are lr=\{1e-3, 1e-4, 1e-5\}, weight decay=\{0.01, 0.05, 0.0001, 0.00001\}, epochs=\{10, 15, 30, 50\}, optimizer=\{SGD, Adam, AdamW\}.

\paragraph{Few-shots classification evaluation}
In order to evaluate the effectiveness of our model under scare label conditions, we applied few-shots learning where each class is only sampled $K$-times. Specifically, we chose $K=8, 16, 32, 64, 128, 256$, where the data is sampled such that positive and negative samples equal to $K$ for each disease. Few-shot training was performed using full fine-tuning with the same hyper-parameter settings. While we also attempted some other commonly used few-shots classification methods such as k-nearest neighbors (KNN), Simple Shots~\cite{want19simpleshot} and Prototypical Networks~\cite{jake17proto}, we did not observe performance improvement on our datasets over full fine-tuning. The main evaluation for few-shots classification is present in \Cref{fig:fewshot}, where we observed our model can already reach performance close to full fine-tuning with only $K=256$ samples. This demonstrates the effectiveness of our model under scare data training regime.

\paragraph{Volume-to-volume retrieval evaluation}
We evaluate volume-to-volume hemorrhage subtype retrieval in an all vs. all setting. Each CT volume is represented by a fixed-length embedding stored along with multi-hot subtype labels. At evaluation time, embeddings are first $\ell2$-normalized and a nearest-neighbor index is built on the gallery embeddings with inner-product search (equivalent to cosine similarity). For each subtype $c$, every positive case for $c$ is used once as a query and is retrieved against the entire gallery minus the query itself (self-match excluded). Relevance is defined as "same subtype $c$". The ranked hit list per query yields a boolean relevance vector for which we compute mean Average Precision (mAP) and Precision$@K$ (for $K=\{1,5,10\}$). Queries and gallery include all available positives (all-vs-all) for the evaluated datasets.

if a query has R relevant items in the galaeery, and the ranked list retrieves them at position $K=\{k_1, k_2, ..., k_R\}$ ($K=\{1,3,5\}$ in our study), average precision for retreival is defined as $Ap=\frac{1}{R}\sum^{R}_{i=1}P(k_i)$, where $P(k_i)$ is the precision at rank $k_i$. Mean Average Precision is then defined as $mAP=\frac{1}{Q}\sum^{Q}_{q=1}AP(q)$ for query q and total number of queries Q.

\subsection*{Visual Interpretation}
\label{sec:visual_interpretation}
Self-attention enables the Vision Transformer (ViT) to integrate information across the entire volume, even in its lowest layers. To analyze the relationships among different patches within the CT volumes, we calculate the average spatial distance over which information is integrated, using the attention weights. 

Let $\mathbf{A}^{(l,h)} \in \mathbb{R}^{N \times N}$  represent the attention weight matrix for the $h$th attention head in the $l$th layer of ViT and $N$ is the number of patches in a CT volume. $d(i, j)$ denotes the spatial distance between patch $i$ and patch $j$ within the 3D volume. The attention distance for each patch $i$ is computed as a weighted average distance to other patches, based on the attention weights:
\begin{equation}
    \centering
    D_i^{(l, h)} = \sum_{j=1}^{N} A^{(l, h)}_{ij}  d(i, j)
\end{equation}

We visualize the average attention distances across all heads and layers for every patch in the volume in \Cref{fig:attention_interpretation}. This “attention distance” serves as an estimate of the ViT’s receptive field within the CT volumes, indicating the regions of the brain that the model focuses on. This visualization helps illustrate how the model integrates information across spatial areas to capture meaningful patterns within the volume.

\subsection*{Statistical analysis}

In each experiment, we report the mean and confidence interval, calculated by bootstrapping the held-out validation set 100 times. For few-shot learning, where model variance is also influenced by the specific training data samples, we repeated the training and evaluation process five times with randomly sampled training data, reporting the mean and confidence interval of the resulting metrics. For all statistical significance (p-values) reported in this study, we used a two-sided paired permutation test with $1,000$ permutations to assess the performance difference of two compared models.

\subsection*{Computing Hardware Software}
All experiments are performed under Python (v3.8.11), PyTorch (v2.4.1), CUDA (12.1) and MONAI (v1.2.0). We extend ViT, MAE, DINO implementation from original their corresponding repositories (\url{https://github.com/facebookresearch/mae}, \url{https://github.com/facebookresearch/dino}) to match our need for 3D CT image encoding. For comparison with Merlin~\cite{blankemeier2024merlinvisionlanguagefoundation}, we integrated their original model weight checkpoints and model backbone code (\url{https://github.com/louisblankemeier/merlin}) to our downstream fine-tuning code base. ResNet50-3D~\cite{hara3dcnns} from (\url{https://github.com/kenshohara/3D-ResNets-PyTorch/tree/master}) is integrated to our code base for evaluation. Nearest-neighbor indexing for volume-to-volume retrieval was done with faiss (v.1.12.0) (\url{https://github.com/facebookresearch/faiss}). All plots and figures were created by Matplotlib (v0.1.6) and Seaborn (v0.13.2). All downstream experiments were conducted on single 80 GB NVIDIA A100 GPU (graphics processing unit). All pre-training experiments were conducted on four 80 GB NVIDIA A100 GPUs.

\section*{Data availability}
The internal clinical data involved in the study is unavailable due to privacy concerns and institutional policy. Public dataset RSNA is available from \url{https://www.kaggle.com/competitions/rsna-intracranial-hemorrhage-detection}. Public dataset CQ500 is available from \url{https://www.kaggle.com/datasets/crawford/qureai-headct}.  The original data is provided as DICOM files. We converted each scan from DICOM to NIfTI files and removed the scans with missing slices for creating 3D imaging datasets in our evaluation. We use all slice thickness scan protocols in each scan (e.g. thin, plain thin, and plain scan) for CQ500, hence providing a more exhaustive evaluation on our model adaptability on different slice thickness for scan.

\section*{Code availability}
The code for pre-training, fine-tuning and evaluation of the foundation model is available on \url{https://github.com/NYUMedML/headCT_foundation}. Due to the possibility of inferring patient face from headCT data, the model weights are only available upon request after signing institutional agreement. Requests for model weight should be sent to the corresponding author and the NYU Langone Data Sharing Strategy Board (DSSB) Committee (DataSharing@nyulangone.org).

\section*{Acknowledgements}
W.Z., H.H., L.C., A.V.M. and N.R. were supported by the National Institute On Aging of the National Institutes of Health under Award R01AG085617. W.Z. H.H., B.Y. and L.C. received partial support from NSF Award 1922658. N.R. and A.V.M. were also partially supported by the National Institute On Aging of the National Institutes of Health under Awards R01AG079175 and P30AG066512.

\newpage
\appendix

\renewcommand{\figurename}{Supplementary Figure}
\renewcommand{\tablename}{Supplementary Table}
\setcounter{figure}{0}
\setcounter{table}{0}

    

\section{Details of datasets}
This section provides additional details about the dataset used to evaluate the downstream tasks. \Cref{tab:disease_definition} lists the ICD-10 codes and medications used to identify the diagnoses for each disease. \Cref{tab:characteristic} presents the distribution of patient characteristics for each disease. 

\begin{figure}[!ht]
    \centering
    \includegraphics[width=0.8\textwidth]{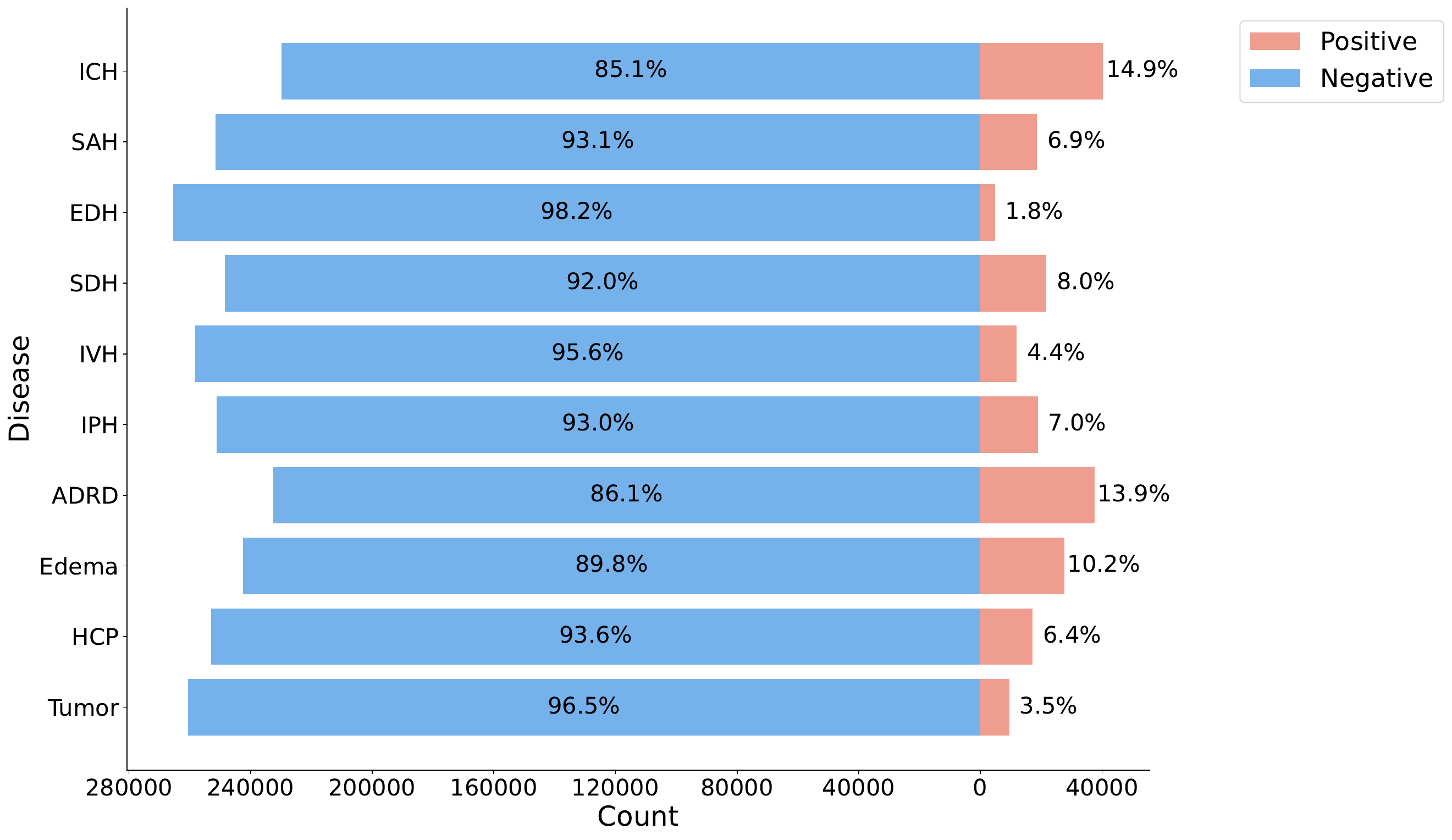}
    \caption{Disease prevalence of NYU Langone }
    \label{fig:nyu_langone_prevalence}
\end{figure}

\begin{figure}[!h]
    \centering
    \includegraphics[width=0.8\textwidth]{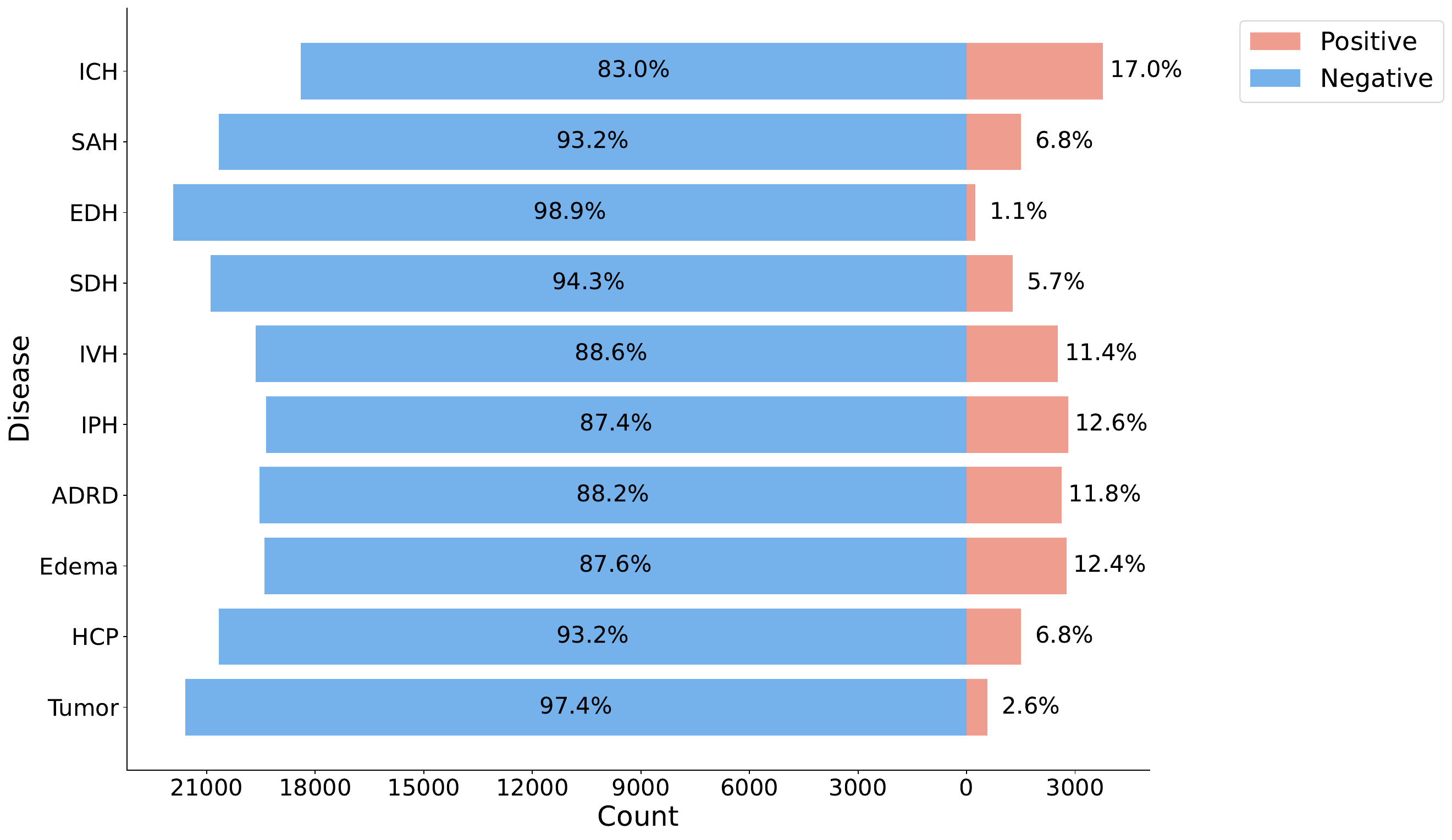}
    \caption{Disease prevalence of NYU Longisland dataset}
    \label{fig:nyu_longisland_prevalence}
\end{figure}

\Cref{fig:nyu_langone_prevalence,fig:nyu_longisland_prevalence} illustrates the prevalence of each disease in the downstream tasks for the NYU Langone and NYU Long Island datasets, highlighting the imbalances present in these tasks.

\Cref{tab:label_noise} presents label sensitivity analysis based on ground truth labels from one radiologist on 1000 samples (891 samples in total after filtering duplicates) in comparison to ICD code mapped labels. The result is reported with Cohen's Kappa $\kappa$, sensitivity and specificity. While majority of diseases show reasonably consistent agreement ($\kappa>0.3$ for $7$ out of $10$ diseases), specificity ($>0.7$ for $10$ out of $10$ diseases) and sensitivity ($>0.7$ for $7$ out of $10$ diseases). Epidural hemorrhage (EDH), edema and hydrocephalus show low agreement ($<0.3$) and low sensitivity ($<0.5$). Although sesativity of these three classes are low, when available at research grade labeling (i.e EDH available in CQ500), the direct external evaluation of the model trained on EHR based labeled data, without any fine-tuning on CQ500, reflects very high performance (See \Cref{fig:overview})g. Since research-grade data is unavailable for Edema and Hydrocephalus we are not able to further verify this but we anticipate that the model performance even trained on low sensitivity labels (which are high in specificity) leads to sufficiently strong models.


\begin{table}[!htpb]
    \centering
    \caption{The definition of diseases in EHR by diagnosis codes and medications.}
    \begin{tabular}{lr}
    \toprule
         Disease &  Definition in EHR \\
    \midrule
       IPH  &  I61.0, I61.1, I61.2, I61.3, I61.4, I61.8, I61.9 \\
       IVH  &  I61.5, P52.1, P52.2, P52.3  \\
       ICH  &  IPH + IVH + I61.6, I62.9, P10.9, P52.4, P52.9 \\
       SDH  &  S06.5, I62.0 \\
       EDH  &  S06.4, I62.1 \\
       SAH  &  I60.*, S06.6, P52.5, P10.3  \\
       Tumor  &  C71.*, C79.3, D33.0, D33.1, D33.2, D33.3, D33.7, D33.9  \\
       Hydrocephalus  &  G91.* \\
       Edema  &  G93.1, G93.5, G93.6, G93.82, S06.1 \\
       \multirow{2}{*}{ADRD}  &  G23.1, G30.*, G31.01, G31.09, G31.83, G31.85, G31.9, F01.*, F02.*, F03.*, G31.84, G31.1, \\ 
       & \textbf{Medication:} DONEPEZIL, RIVASTIGMINE, GALANTAMINE, MEMANTINE, TACRINE \\ 
    \bottomrule
    \end{tabular}
    \label{tab:disease_definition}
\end{table}
\begin{table}[!htbp]
\centering
\caption{Demographic characteristics of patients associated with scans from the NYU Langone dataset, matched with electronic health records (EHR) and utilized in downstream tasks.}
\label{tab:characteristic}

 The characteristic table on NYU Langone dataset matched with EHR.
\begin{tabular}{ll|rr|r}
\toprule
                       \textbf{Cohort} &  &           \textbf{Male (\%)} &          \textbf{Female (\%)} &     \textbf{Age (std)} \\
\midrule
 --- & All (n=270,205) & 128,113 (47.41\%) & 142,092 (52.59\%) & 63.64 (19.68) \\
\midrule
       Tumor & Neg (n=260,704) & 123,338 (47.31\%) & 137,366 (52.69\%) & 63.85 (19.72) \\
             & Pos (n=9,501) &   4,775 (50.26\%) &   4,726 (49.74\%) & 57.80 (17.67) \\
\midrule
HCP & Neg (n=253,000) & 118,881 (46.99\%) & 134,119 (53.01\%) & 63.67 (19.72) \\
              & Pos (n=17,205) &   9,232 (53.66\%) &   7,973 (46.34\%) & 63.18 (19.11) \\
\midrule
Edema & Neg (n=242,576) & 112,987 (46.58\%) & 129,589 (53.42\%) & 63.96 (19.84) \\
      & Pos (n=27,629) &  15,126 (54.75\%) &  12,503 (45.25\%) & 60.81 (17.97) \\
\midrule
ADRD  & Neg (n=232,667) & 111,159 (47.78\%) & 121,508 (52.22\%) & 61.31 (19.55) \\
      & Pos (n=37,538) &  16,954 (45.16\%) &  20,584 (54.84\%) & 78.09 (13.30) \\
\midrule
          IPH & Neg (n=251,308) & 117,692 (46.83\%) & 133,616 (53.17\%) & 63.58 (19.82) \\
              & Pos (n=18,897) &  10,421 (55.15\%) &   8,476 (44.85\%) & 64.39 (17.69) \\
\midrule
          IVH & Neg (n=258,232) & 121,686 (47.12\%) & 136,546 (52.88\%) & 63.65 (19.79) \\
              & Pos (n=11,973) &   6,427 (53.68\%) &   5,546 (46.32\%) & 63.45 (17.19) \\
\midrule
          SDH & Neg (n=248,468) & 114,869 (46.23\%) & 133,599 (53.77\%) & 63.44 (19.78) \\
              & Pos (n=21,737) &  13,244 (60.93\%) &   8,493 (39.07\%) & 65.95 (18.33) \\
\midrule
          EDH & Neg (n=265,431) & 125,113 (47.14\%) & 140,318 (52.86\%) & 63.77 (19.64) \\
              & Pos (n=4,774) &   3,000 (62.84\%) &   1,774 (37.16\%) & 56.53 (20.75) \\
\midrule
          SAH & Neg (n=251,594) & 118,424 (47.07\%) & 133,170 (52.93\%) & 63.79 (19.76) \\
              & Pos (n=18,611) &   9,689 (52.06\%) &   8,922 (47.94\%) & 61.59 (18.49) \\
\midrule
          ICH & Neg (n=229,851) & 105,498 (45.90\%) & 124,353 (54.10\%) & 63.41 (19.93) \\
              & Pos (n=40,354) &  22,615 (56.04\%) &  17,739 (43.96\%) & 64.93 (18.14) \\
\bottomrule
\end{tabular}
\end{table}


\begin{table}[!h]
    \centering
    \caption*{\textbf{Supplementary \Cref{tab:characteristic} Continued.} Demographic characteristics of patients associated with scans from the NYU Long Island dataset, matched with electronic health records (EHR) and utilized in downstream tasks.}
\begin{tabular}{ll|rr|r}
\toprule
                       \textbf{Cohort} &  &           \textbf{Male (\%)} &          \textbf{Female (\%)} &     \textbf{Age (std)} \\
\midrule
--- & All (n=22,158) & 9,580 (43.23\%) & 12,578 (56.77\%) & 68.33 (18.14) \\
\midrule
Tumor & Neg (n=21,578) & 9,275 (42.98\%) & 12,303 (57.02\%) & 68.59 (18.08) \\
      & Pos (n=580) &   305 (52.59\%) &    275 (47.41\%) & 58.78 (17.79) \\
\midrule
HCP   & Neg (n=20,653) & 8,718 (42.21\%) & 11,935 (57.79\%) & 69.05 (17.90) \\
      & Pos (n=1,505) &   862 (57.28\%) &    643 (42.72\%) & 58.52 (18.48) \\
\midrule
Edema & Neg (n=19,402) & 8,068 (41.58\%) & 11,334 (58.42\%) & 68.89 (18.27) \\
      & Pos (n=2,756) & 1,512 (54.86\%) &  1,244 (45.14\%) & 64.36 (16.66) \\
\midrule
ADRD  & Neg (n=19,537) & 8,391 (42.95\%) & 11,146 (57.05\%) & 66.78 (18.28) \\
      & Pos (n=2,621) & 1,189 (45.36\%) &  1,432 (54.64\%) & 79.90 (11.77) \\
\midrule
IPH   & Neg (n=19,357) & 7,974 (41.19\%) & 11,383 (58.81\%) & 68.97 (18.27) \\
      & Pos (n=2,801) & 1,606 (57.34\%) &  1,195 (42.66\%) & 63.89 (16.48) \\
\midrule
IVH   & Neg (n=19,636) & 8,164 (41.58\%) & 11,472 (58.42\%) & 68.96 (18.22) \\
      & Pos (n=2,522) & 1,416 (56.15\%) &  1,106 (43.85\%) & 63.43 (16.66) \\
\midrule
SDH   & Neg (n=20,885) & 8,870 (42.47\%) & 12,015 (57.53\%) & 68.33 (18.21) \\
      & Pos (n=1,273) &   710 (55.77\%) &    563 (44.23\%) & 68.37 (16.83) \\
\midrule
EDH   & Neg (n=21,912) & 9,443 (43.10\%) & 12,469 (56.90\%) & 68.33 (18.16) \\
      & Pos (n=246) &   137 (55.69\%) &    109 (44.31\%) & 68.19 (15.59) \\
\midrule
SAH   & Neg (n=20,652) & 8,824 (42.73\%) & 11,828 (57.27\%) & 68.68 (18.12) \\
      & Pos (n=1,506) &   756 (50.20\%) &    750 (49.80\%) & 63.58 (17.65) \\
\midrule
ICH   & Neg (n=18,388) & 7,456 (40.55\%) & 10,932 (59.45\%) & 68.92 (18.35) \\
      & Pos (n=3,770) & 2,124 (56.34\%) &  1,646 (43.66\%) & 65.48 (16.77) \\
\bottomrule
\end{tabular}
\end{table}

\begin{table}[htbp]
\centering
\caption{Per-disease label sensitivity analysis}
\begin{tabular}{lccccc}
\toprule
disease & prevalence & Cohen's $\kappa$ & sensitivity & specificity & accuracy \\
\midrule
EDH           & 0.0078 & 0.0134 & 0.1429 & 0.9265 & 0.9203 \\
ICH           & 0.1111 & 0.3761 & 0.9394 & 0.7525 & 0.7733 \\
IPH           & 0.1111 & 0.5413 & 0.7677 & 0.9028 & 0.8878 \\
IVH           & 0.0606 & 0.4513 & 0.7963 & 0.9092 & 0.9024 \\
SAH           & 0.0955 & 0.6035 & 0.9529 & 0.8983 & 0.9035 \\
SDH           & 0.0976 & 0.5645 & 0.8161 & 0.9104 & 0.9012 \\
dementia      & 0.1470 & 0.7283 & 0.8855 & 0.9303 & 0.9237 \\
edema         & 0.1279 & 0.0756 & 0.2719 & 0.8198 & 0.7497 \\
hydrocephalus & 0.0988 & 0.1118 & 0.2273 & 0.8941 & 0.8283 \\
tumor         & 0.0539 & 0.5966 & 0.7917 & 0.9573 & 0.9484 \\
\bottomrule
\label{tab:label_noise}
\end{tabular}
\end{table}

\newpage
\section{Data augmentation details}
\label{sec:dataaug_details}
We applied Random Flipping across all three dimensions, Random Shift Intensity with offset $0.1$ for both pre-training and fine-tuning. For DINO training. random Gaussian Smoothing with sigma=$(0.5-1.0)$ is applied across all dimensions, Random Gamma Adjust is applied with gamma=$(0.2-1.0)$.

\section{Clarification on model evaluation}
While previous literature achieved promising performance on Head CT disease detection tasks, they either used large scale private dataset (Qure25k - around 21k samples) for fine-tuning~\cite{Chilamkurthy2018} or applied slice-level labels for subject classification~\cite{WANG2021102785}. Both cases are not practical for real work clinical application. We want to clarify that our approach, along with recent efforts on building end-to-end 3D foundation models \cite{blankemeier2024merlinvisionlanguagefoundation,yang2024advancingmultimodalmedicalcapabilities,pai2025visionfoundationmodelscomputed}, is completely different from either of these two methods to make any direct comparisons. In our case, each scan only has subject-level labels, which constitute a more difficult and clinically realistic task. To be specific, models trained with slice-level labels benefit from substantially more granular supervision—each slice of an imaging volume is explicitly labeled and directly contributes to the training objective. In contrast, our setting only provides a single subject-level label for an entire 3D volume, which is inherently a weaker form of supervision. Our study specifically focuses on the subject-level label setting, which better reflects many real-world clinical scenarios where fine-grained annotations are impractical or unavailable. We therefore emphasize that our contribution lies not in outperforming slice-level supervised methods, but in demonstrating strong performance under the much weaker subject-level supervision paradigm with entire 3D volumetric scan as model input.

\section{Additional experiment results}
\begin{figure}[!htpb]
    \centering
    \makebox[\textwidth][l]{%
        \hspace{0.3\textwidth}\textbf{NYU Langone}
    } \\[0.2cm]
    \includegraphics[trim={0 0 0 0},clip,height=0.3\textwidth, width=0.3\textwidth]{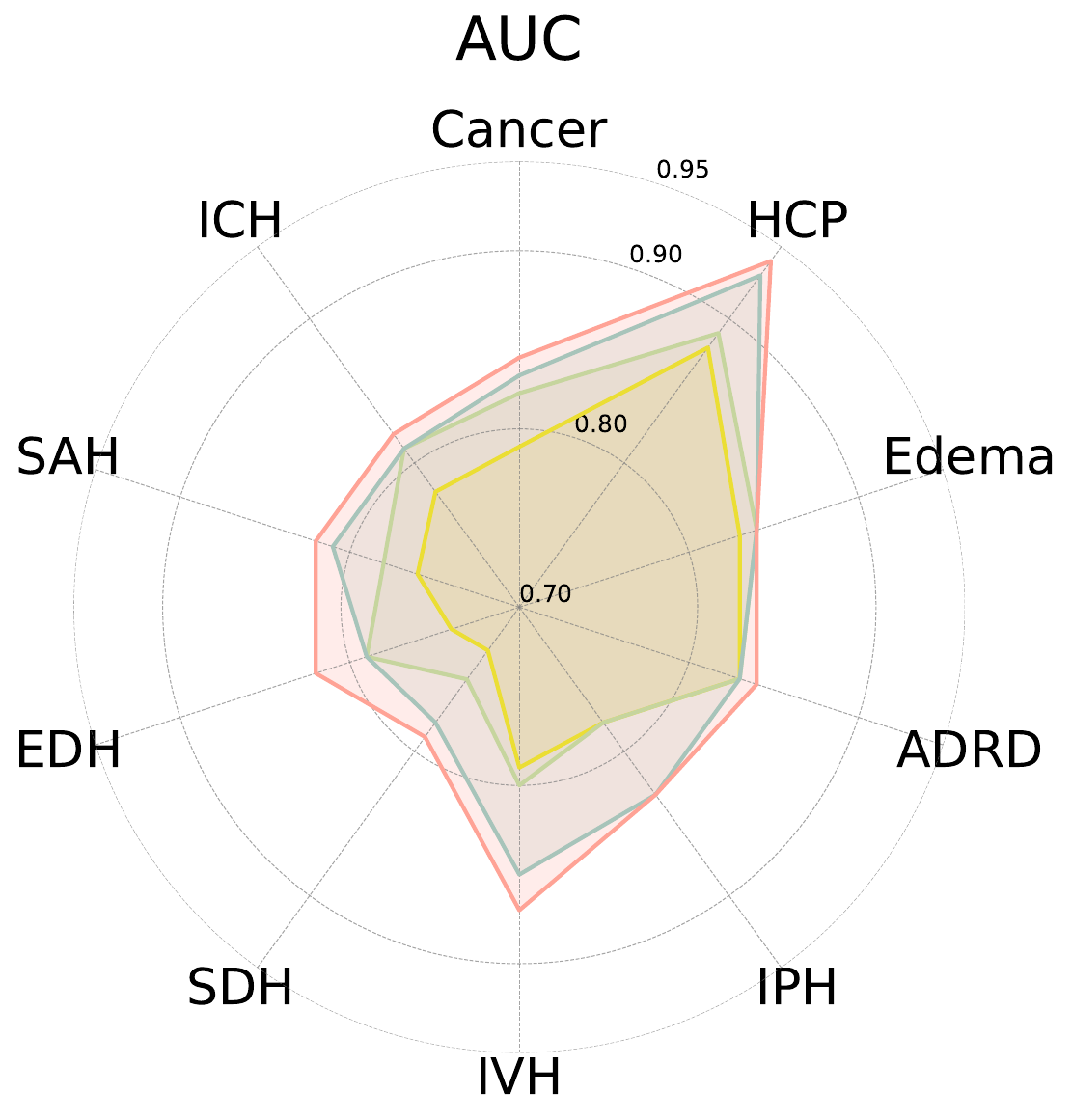}
    \includegraphics[trim={0 0 0 0},clip,height=0.3\textwidth, width=0.55\textwidth]{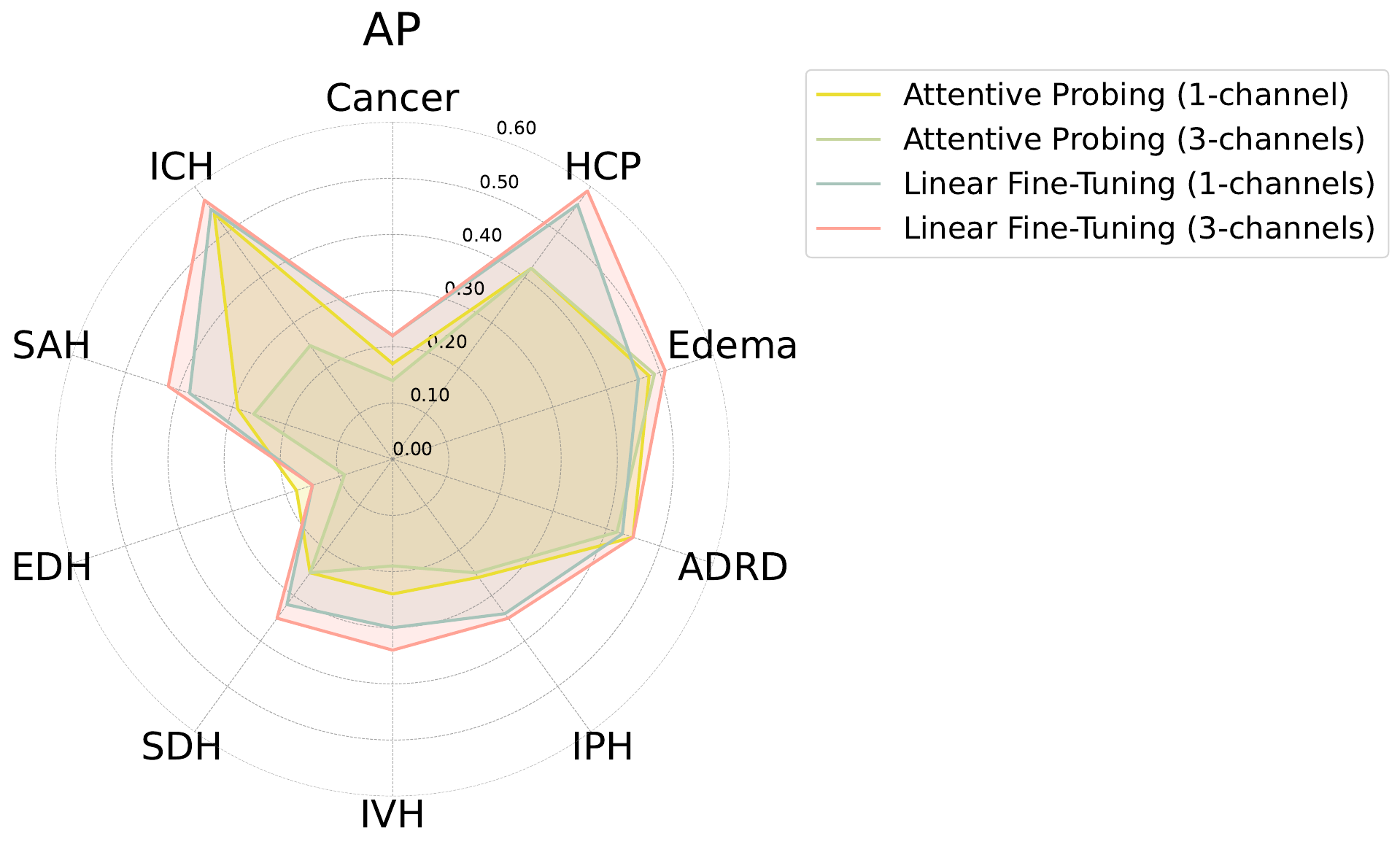}\\
    \makebox[\textwidth][l]{
        \hspace{0.34\textwidth}\textbf{RSNA}
    } \\[0.2cm]
    \includegraphics[trim={0 0 0 0},clip,height=0.3\textwidth, width=0.3\textwidth]{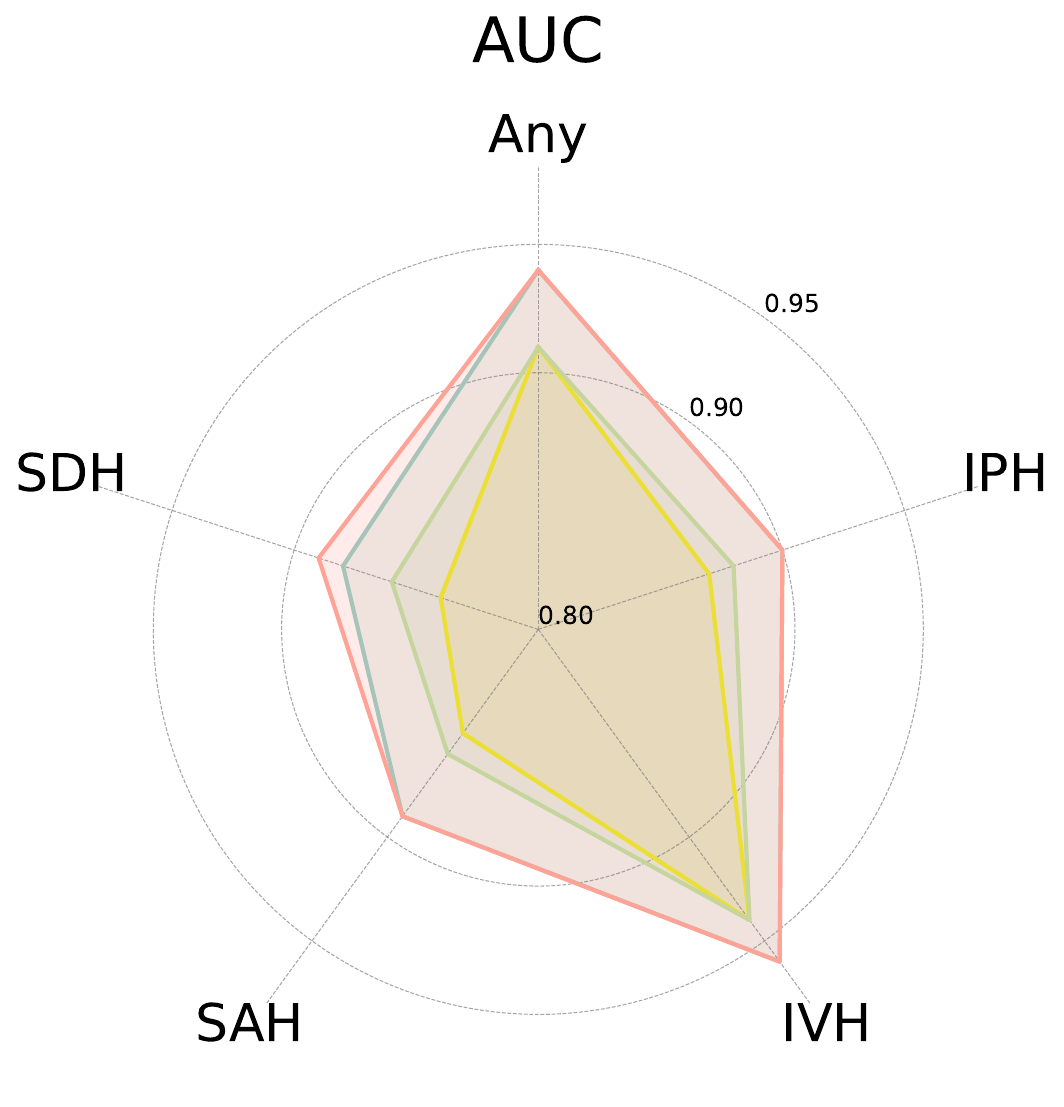}
    \includegraphics[height=0.3\textwidth, width=0.55\textwidth]{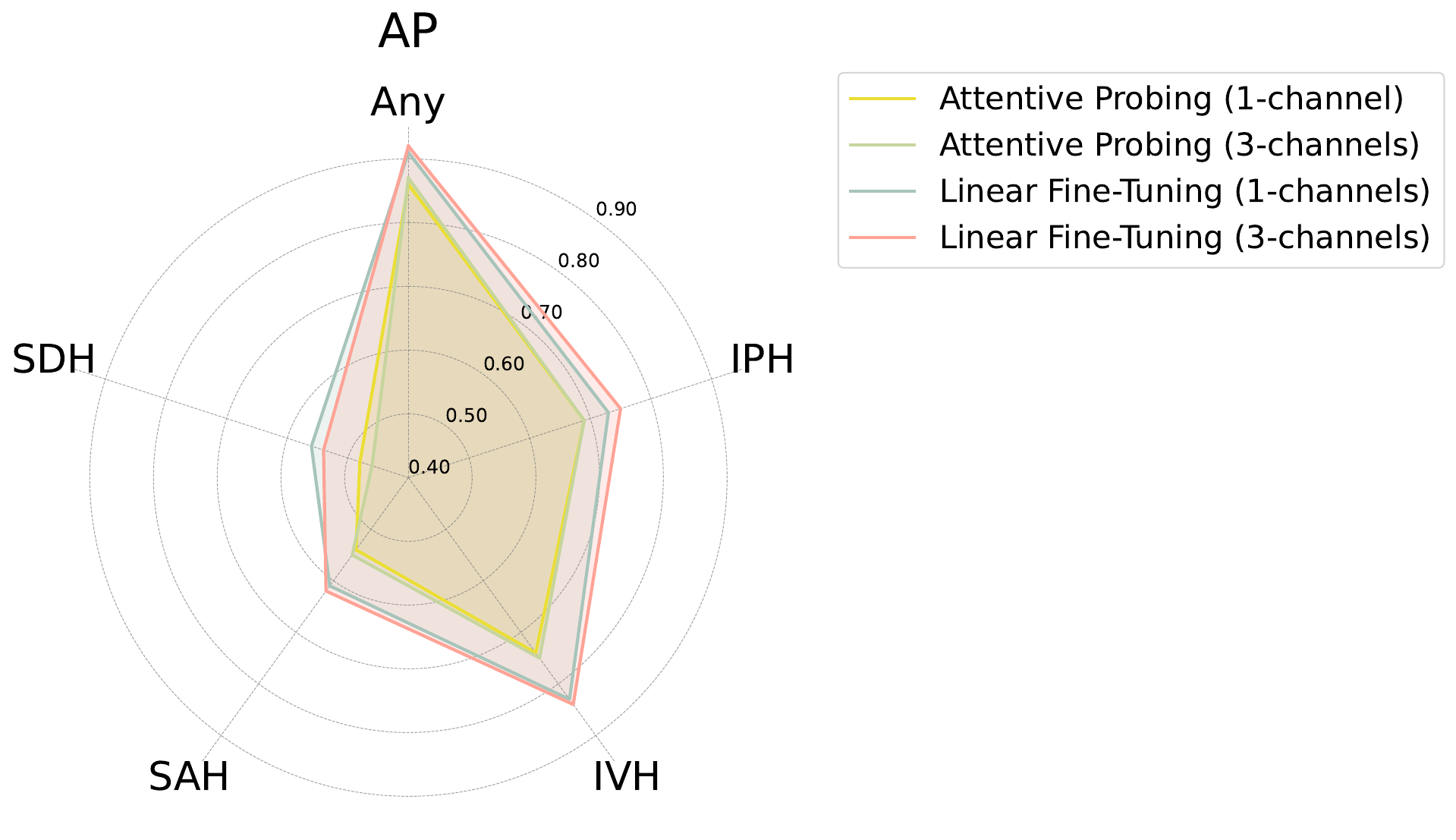} 
    \caption{\textbf{Comparison of Different Channels Performance.} This plot compares the performance of models trained using different numbers of channels (channels from multiple HU intervals vs. a single HU interval). Across two datasets, models using three channels from different HU intervals consistently outperform those using a single channel with a fixed HU interval. All models were pre-trained on $100\%$ of the pretraining data with MAE.}
    \label{fig:channels-ablation}
\end{figure}

\begin{figure}[!htpb]
    \centering
    \makebox[\textwidth][l]{%
        \hspace{0.3\textwidth}\textbf{NYU Langone}
    } \\[0.2cm]
    \includegraphics[trim={0 0 0 0},clip,height=0.3\textwidth, width=0.3\textwidth]{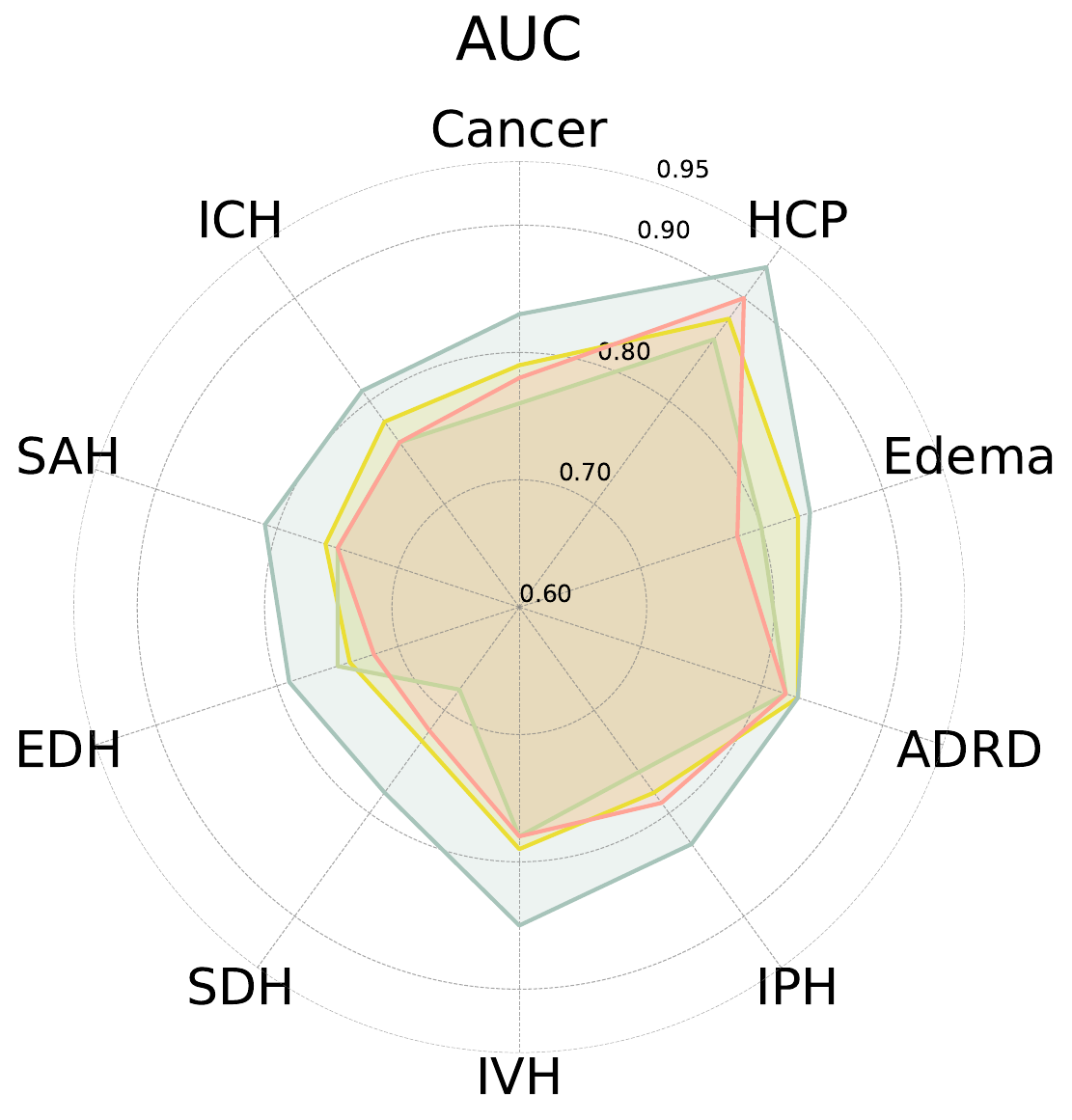}
    \includegraphics[trim={0 0 0 0},clip,height=0.3\textwidth, width=0.55\textwidth]{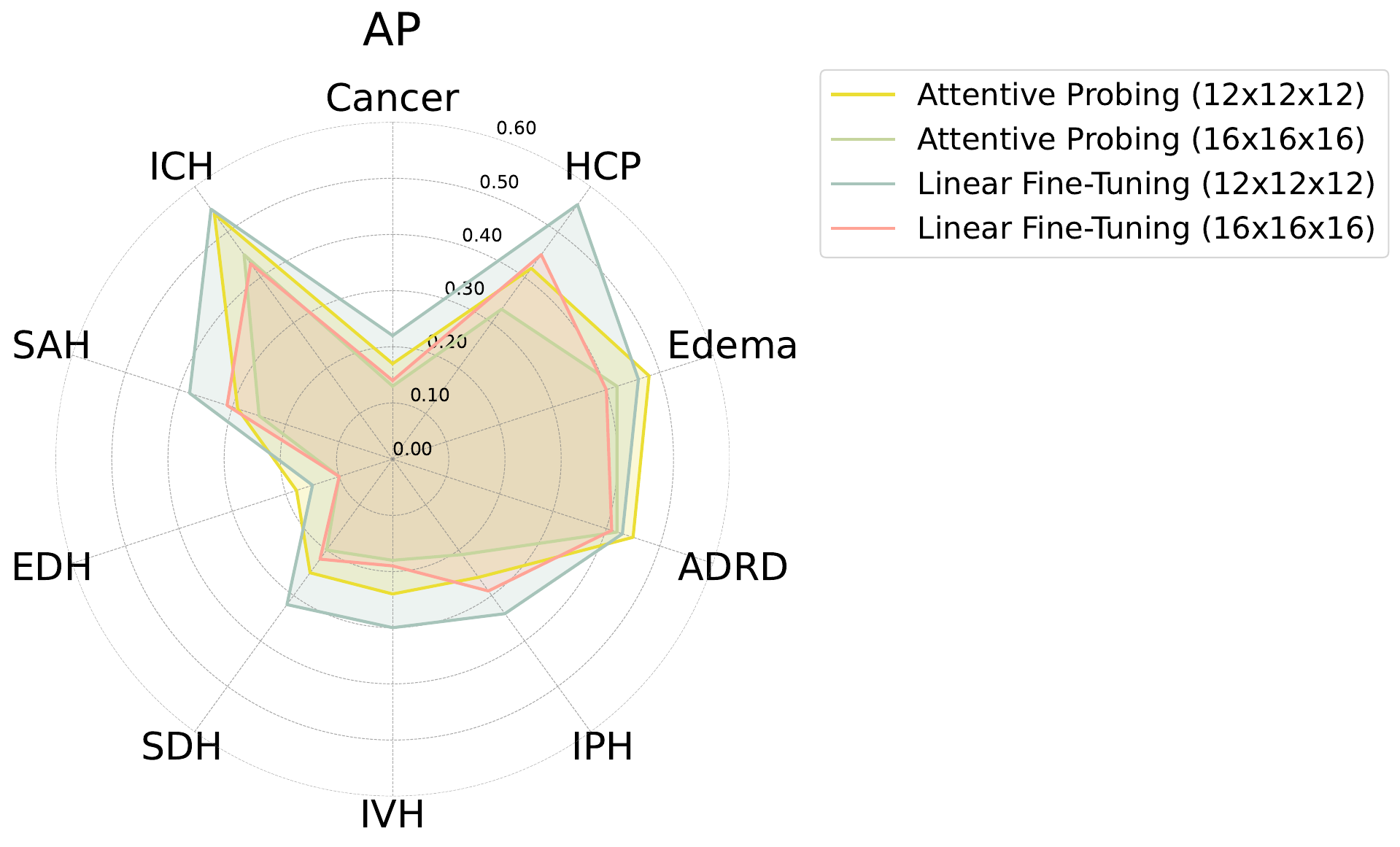}\\
    \makebox[\textwidth][l]{
        \hspace{0.34\textwidth}\textbf{RSNA}
    } \\[0.2cm]
    \includegraphics[trim={0 0 0 0},clip,height=0.3\textwidth, width=0.3\textwidth]{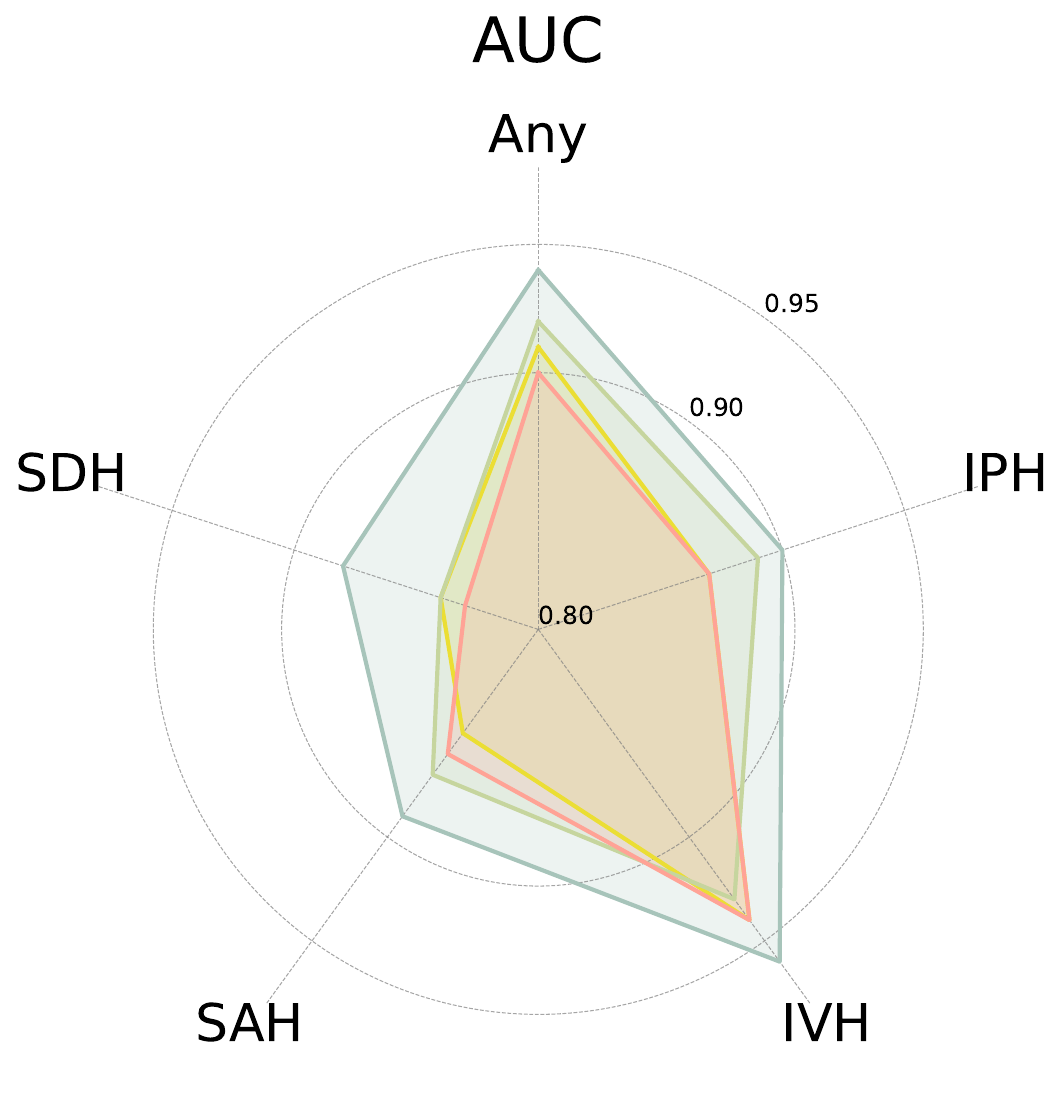}
    \includegraphics[height=0.3\textwidth, width=0.55\textwidth]{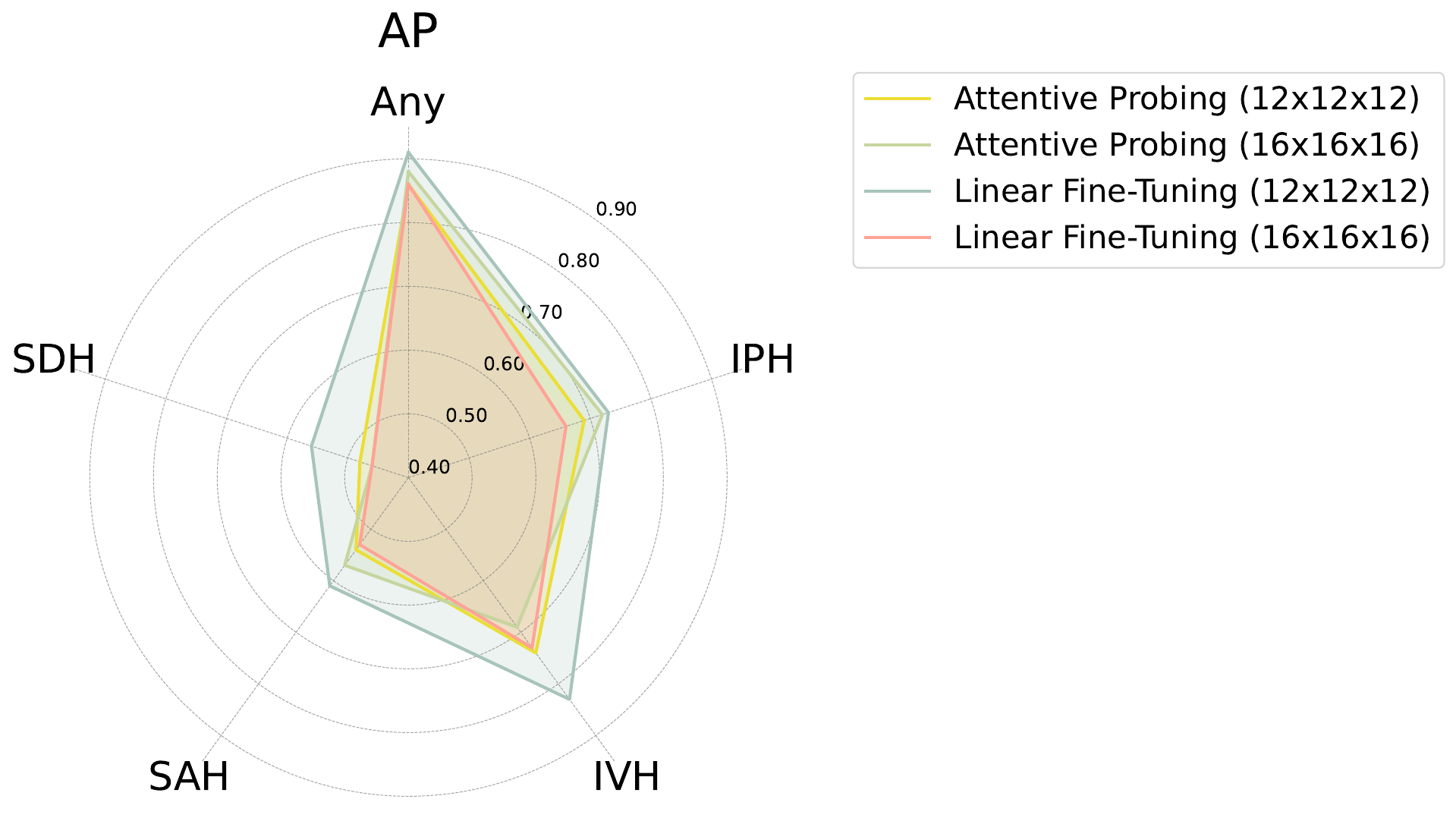} 
    \caption{\textbf{Comparison of Different Patches Performance.} This plot compares the performance of models trained with different patch sizes (12 vs. 16). The results demonstrate that smaller patch sizes consistently achieve better performance. All models were pre-trained on $100\%$ of the pretraining data with MAE.}
    \label{fig:patches-ablation}
\end{figure}

This section provides additional experimental results with more details.
Supplementary \Cref{fig:channels-ablation,fig:patches-ablation} compares the performance of the foundation model using different numbers of channels and patch sizes, demonstrating that the architecture design of our foundation model is optimal. 

\begin{figure*}
    \centering
    \makebox[\textwidth][l]{%
        \hspace{0.06\textwidth}
        \textbf{NYU Langone} \hspace{0.06\textwidth} \textbf{NYU Long Island} \hspace{0.09\textwidth} \textbf{RSNA} \hspace{0.18\textwidth} \textbf{CQ500}
    } \\[0.2cm]
    \includegraphics[trim={0 0 0 0},clip,height=0.21\textwidth, width=0.21\textwidth]{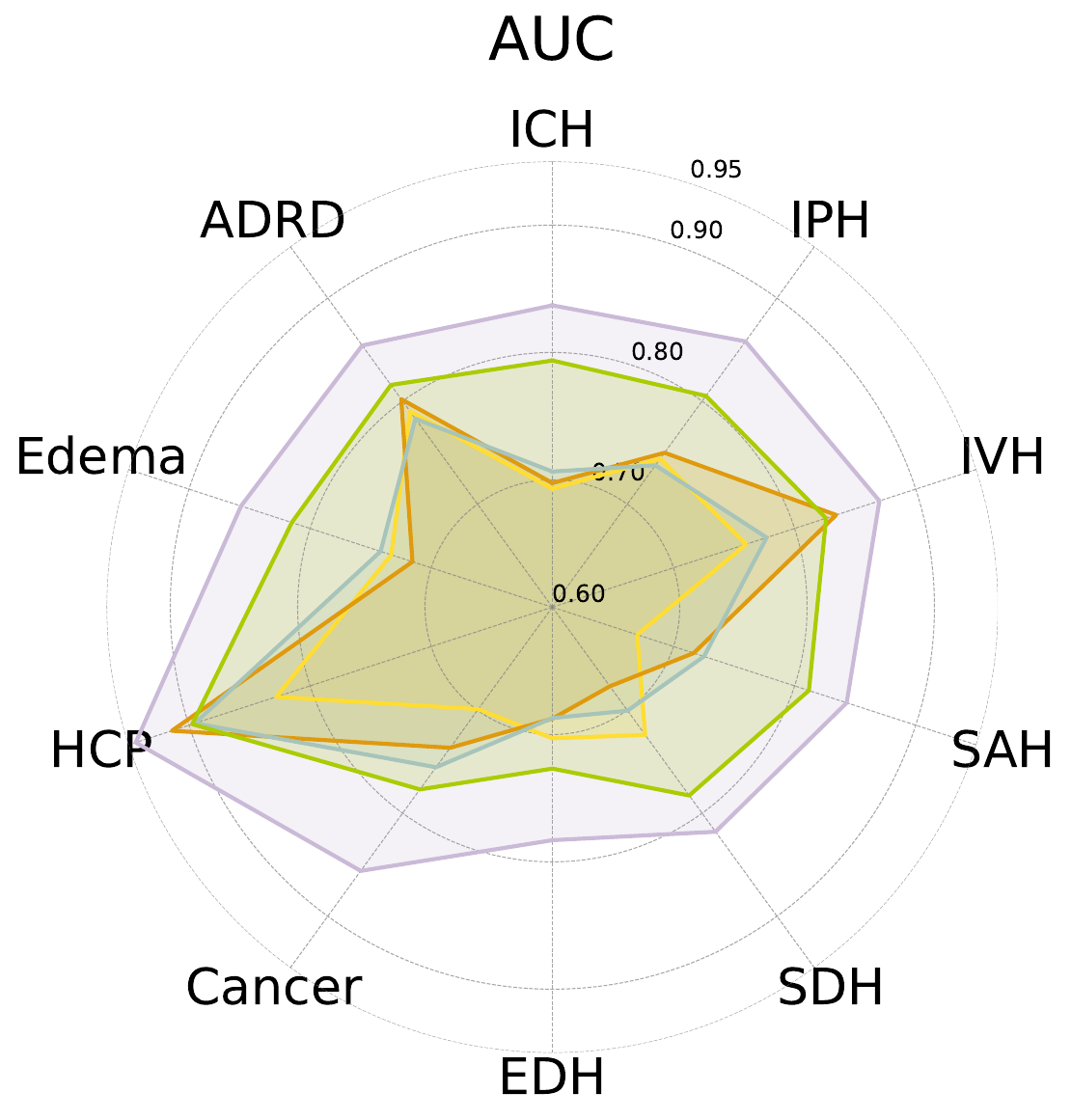}
    \includegraphics[trim={0 0 0 0},clip,height=0.21\textwidth, width=0.21\textwidth]{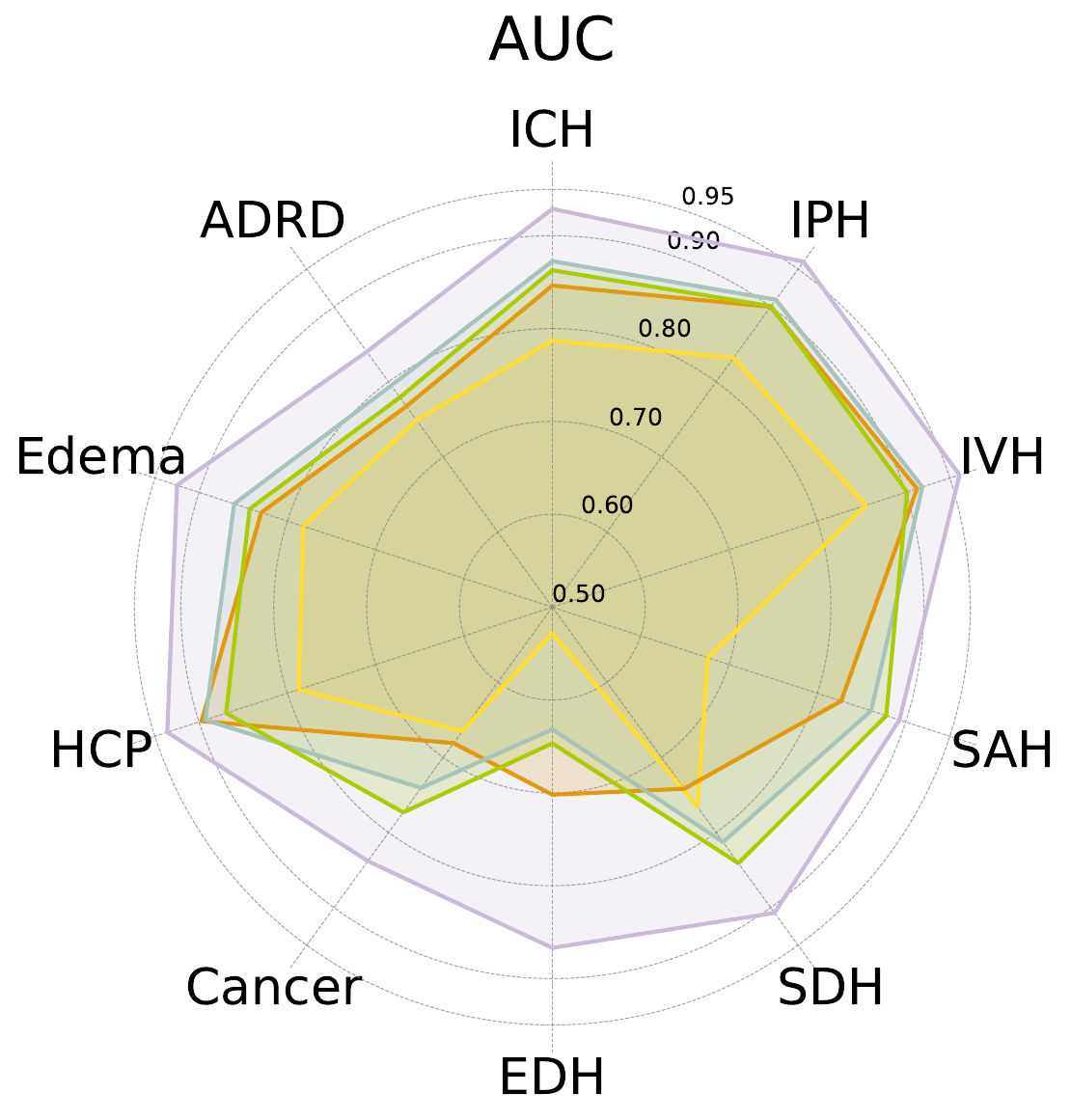}
    \includegraphics[trim={0 0 0 0},clip,height=0.21\textwidth, width=0.21\textwidth]{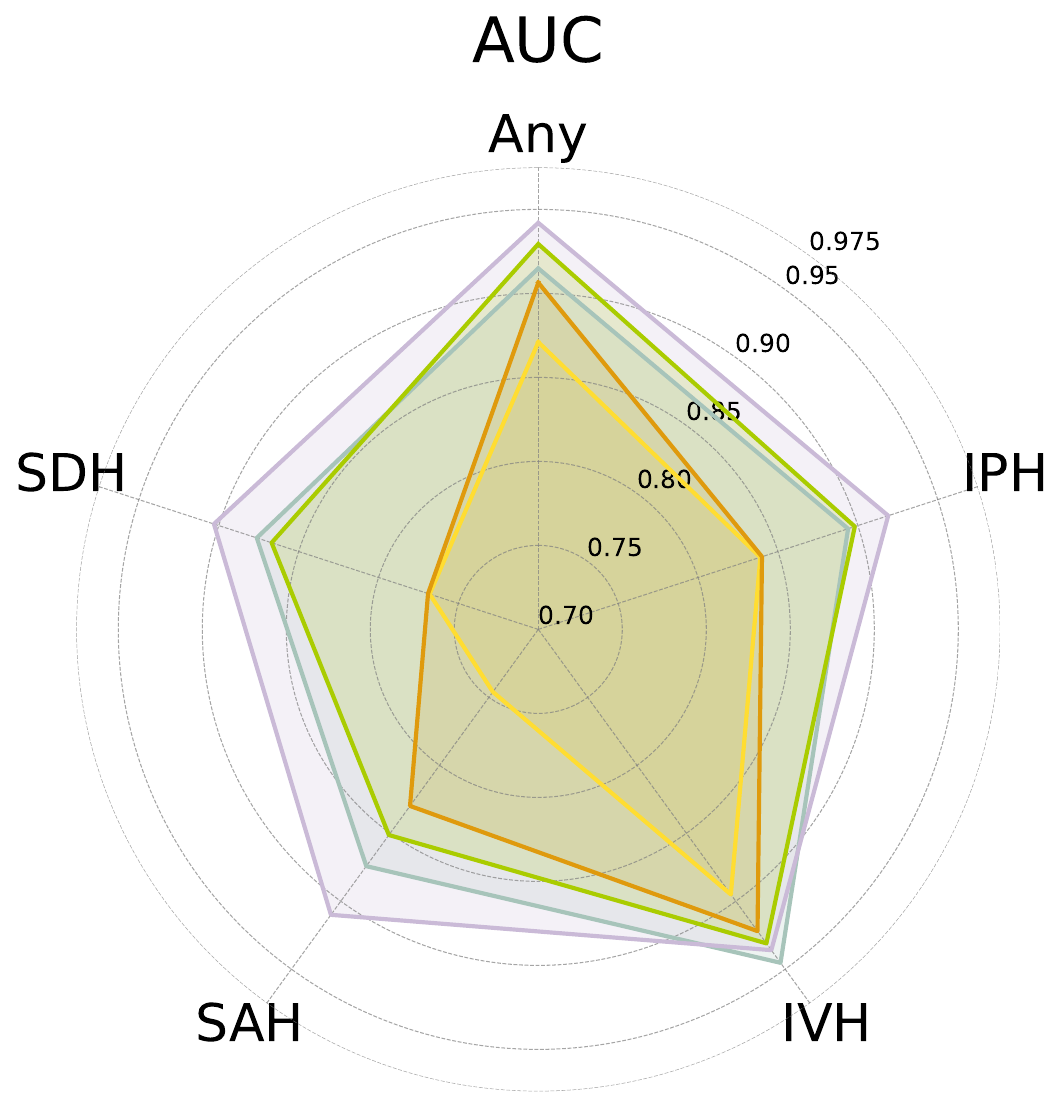}
    \includegraphics[trim={0 0 0 0},clip,height=0.21\textwidth, width=0.35\textwidth]{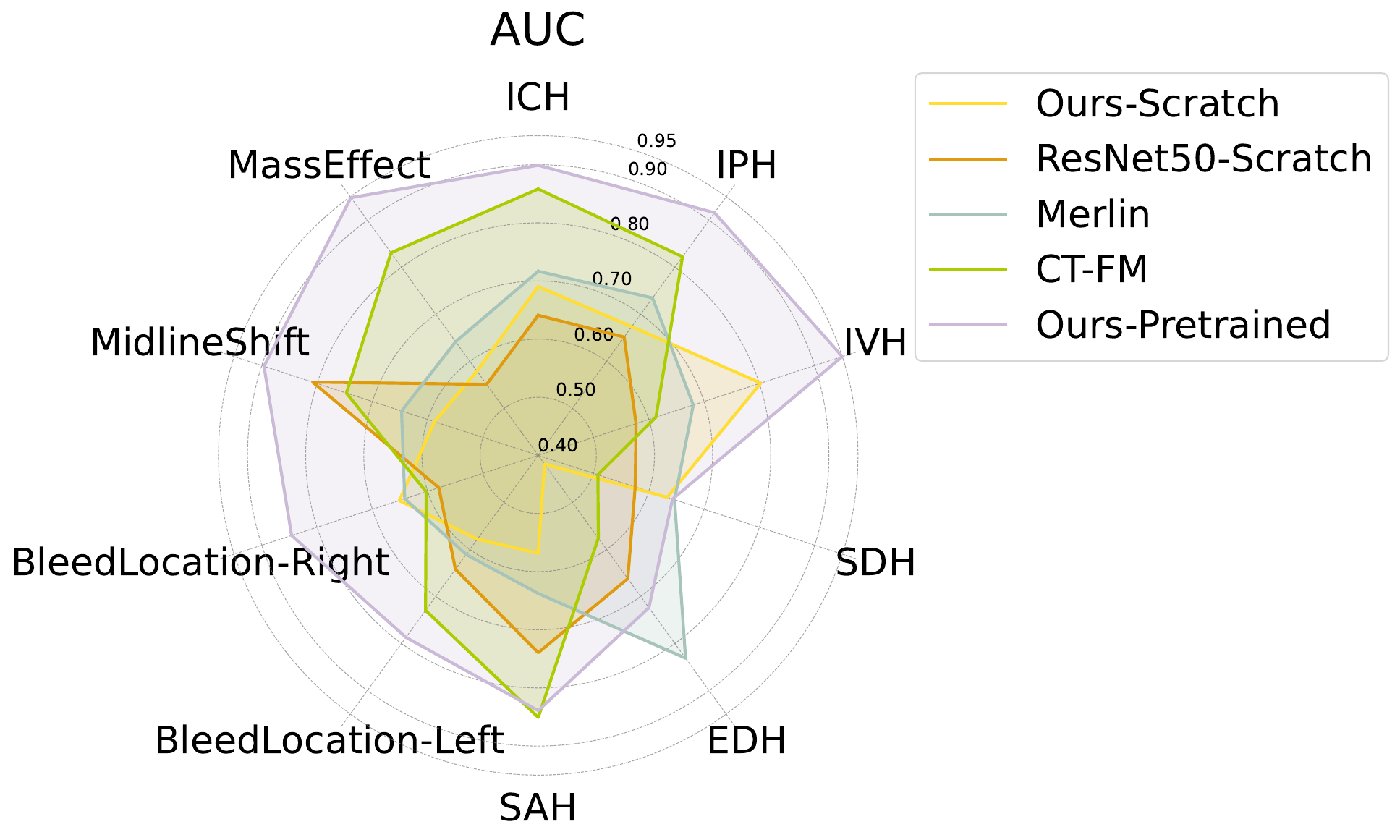}\\[0.2cm]
    \includegraphics[height=0.21\textwidth, width=0.21\textwidth]{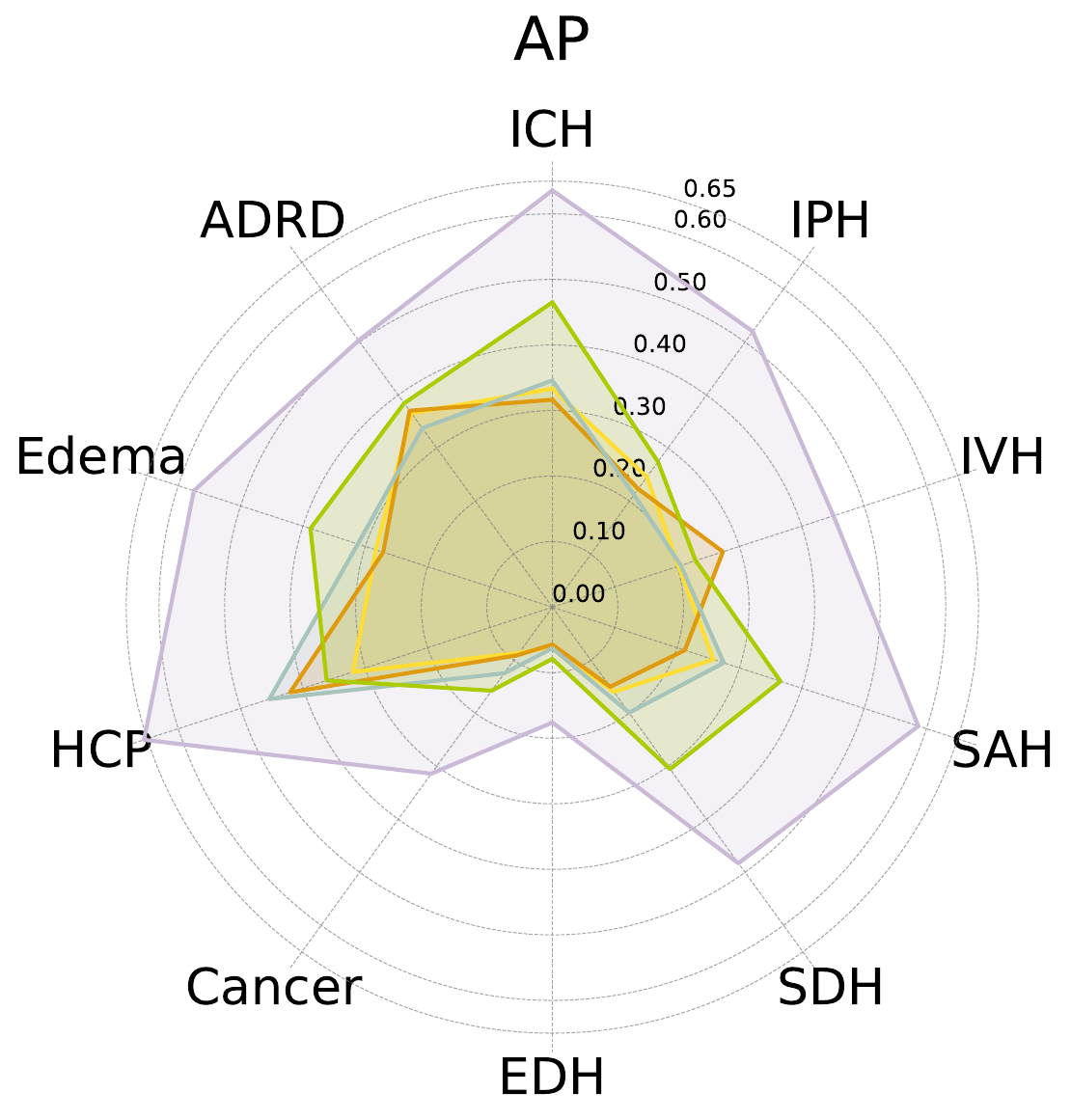} 
    \includegraphics[height=0.21\textwidth, width=0.21\textwidth]{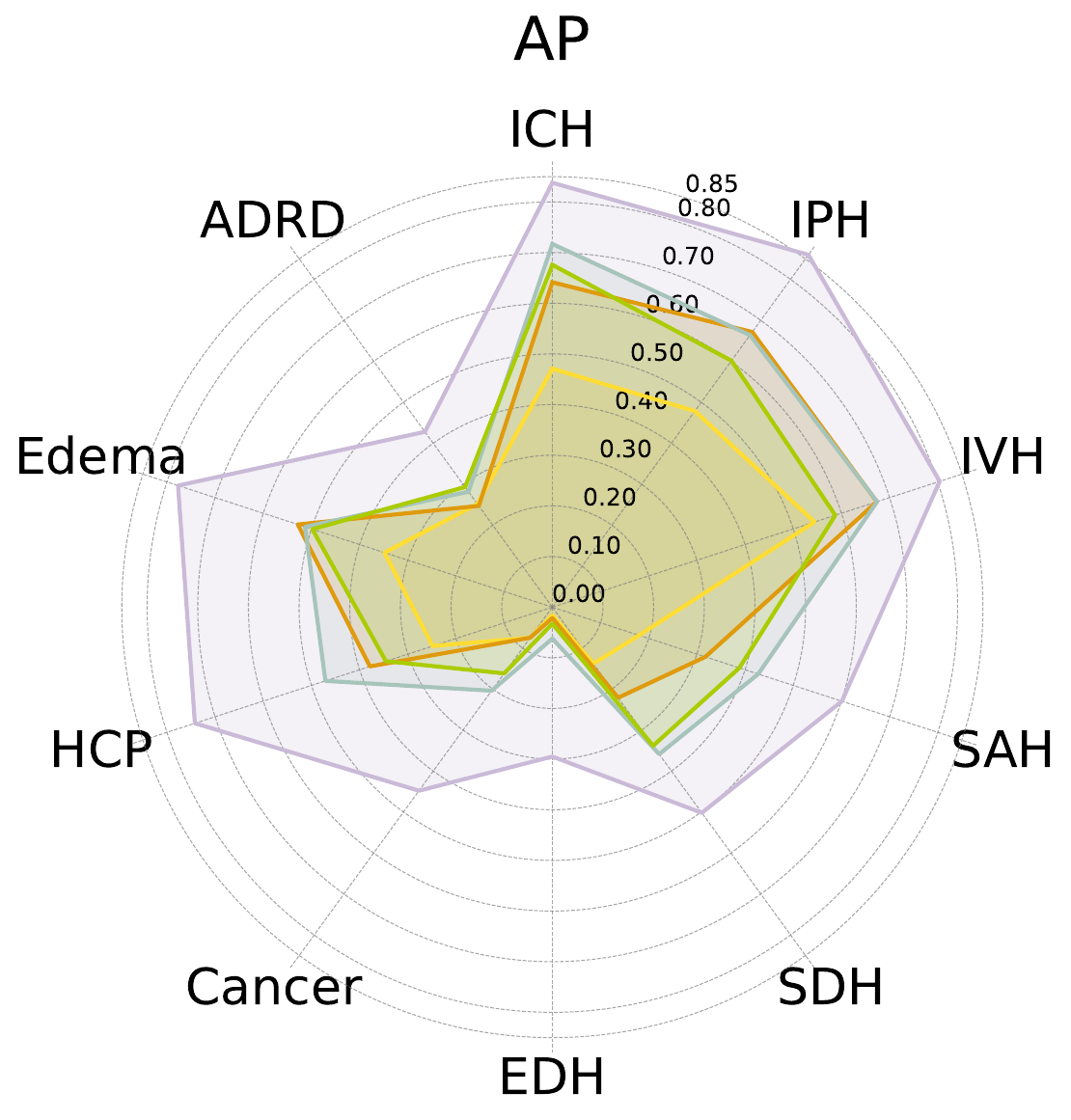} 
    \includegraphics[height=0.21\textwidth, width=0.21\textwidth]{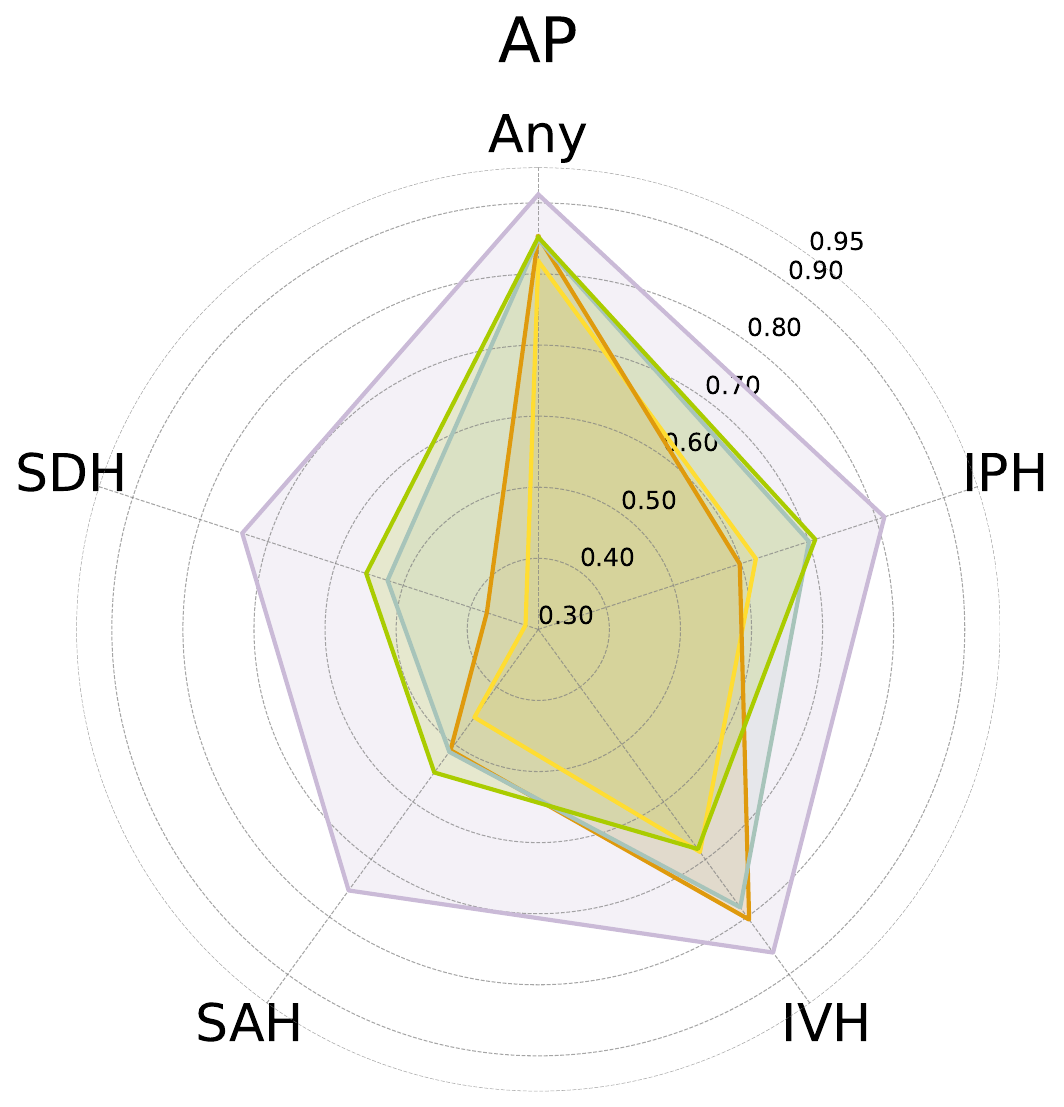}
    \includegraphics[height=0.21\textwidth, width=0.35\textwidth]{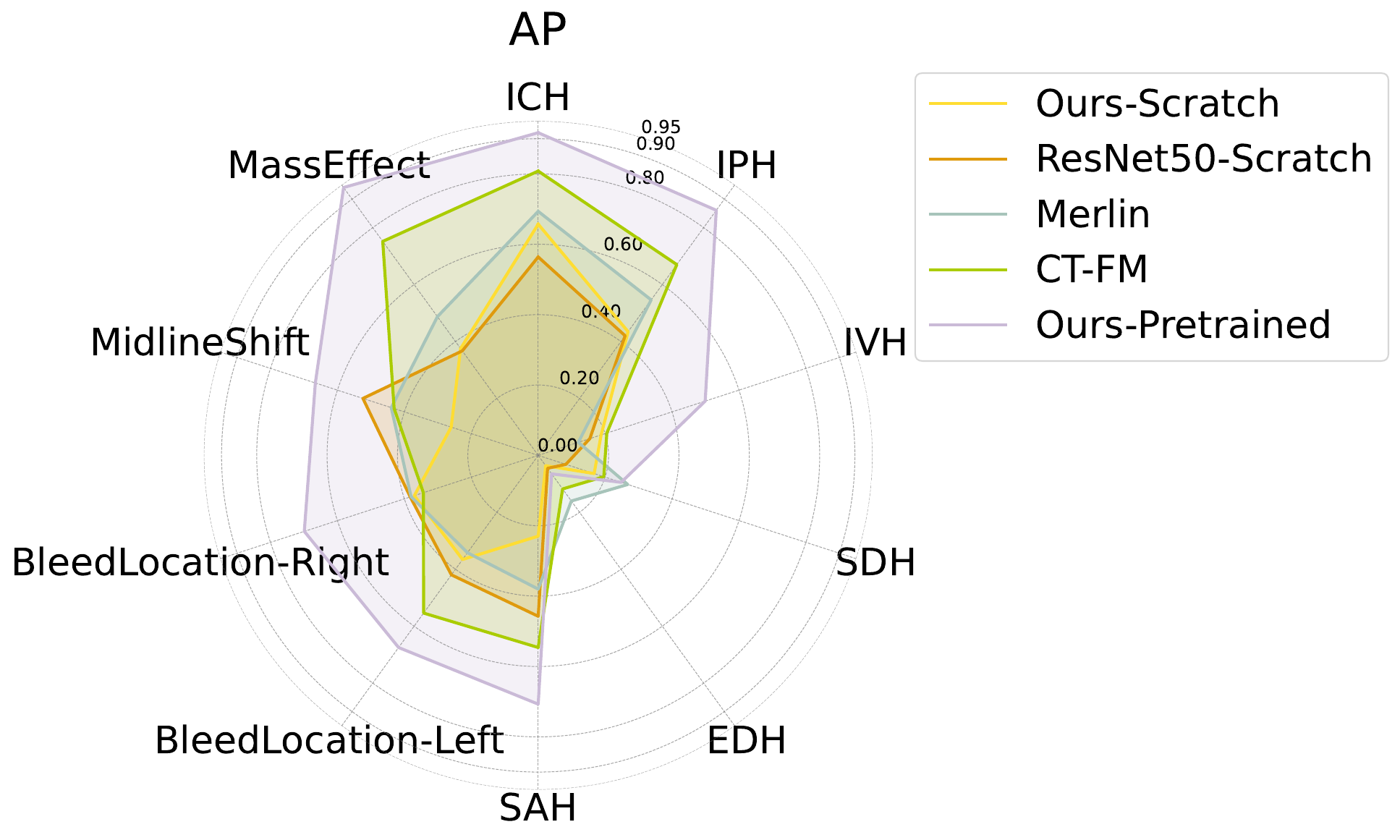}
    \caption{\textbf{Comparison to previous 3D Foundation Model with Full Finetuning.} This plot compares the performance of our model with Merlin~\cite{blankemeier2024merlinvisionlanguagefoundation}, CT-FM~\cite{pai2025visionfoundationmodelscomputed}, ResNet3D, our train from scratch and pre-trained model across four datasets. All experiments are done with fine-tuning following the original data pre-processing pipeline of respective models. Our DINO trained model is used in this comparison. Our model demonstrates consistently superior performance across majority of diseases, with the exception of epidural hemorrhage (EDH) in the CQ500 dataset.}
    \label{fig:radar-comparison-merlin}
\end{figure*}

Supplementary \Cref{fig:radar-comparison-merlin} compares our foundation model with a foundation CT model from previous studies under full fine-tuning, Merlin\cite{blankemeier2024merlinvisionlanguagefoundation}, which was pre-trained on abdomen CT scans with corresponding radiology report pairs, CT-FM~\cite{pai2025visionfoundationmodelscomputed}, which was pre-trained on diverse CT scans from different anatomy including head CT. Our model demonstrates superior performance on head CT scans over both models.

\begin{figure}
\centering
\makebox[\textwidth][l]{%
    \hspace{0.39\textwidth}\textbf{RSNA}
} \\[0.2cm]
\includegraphics[trim={0 0 0mm 0},clip,height=0.27\textwidth]{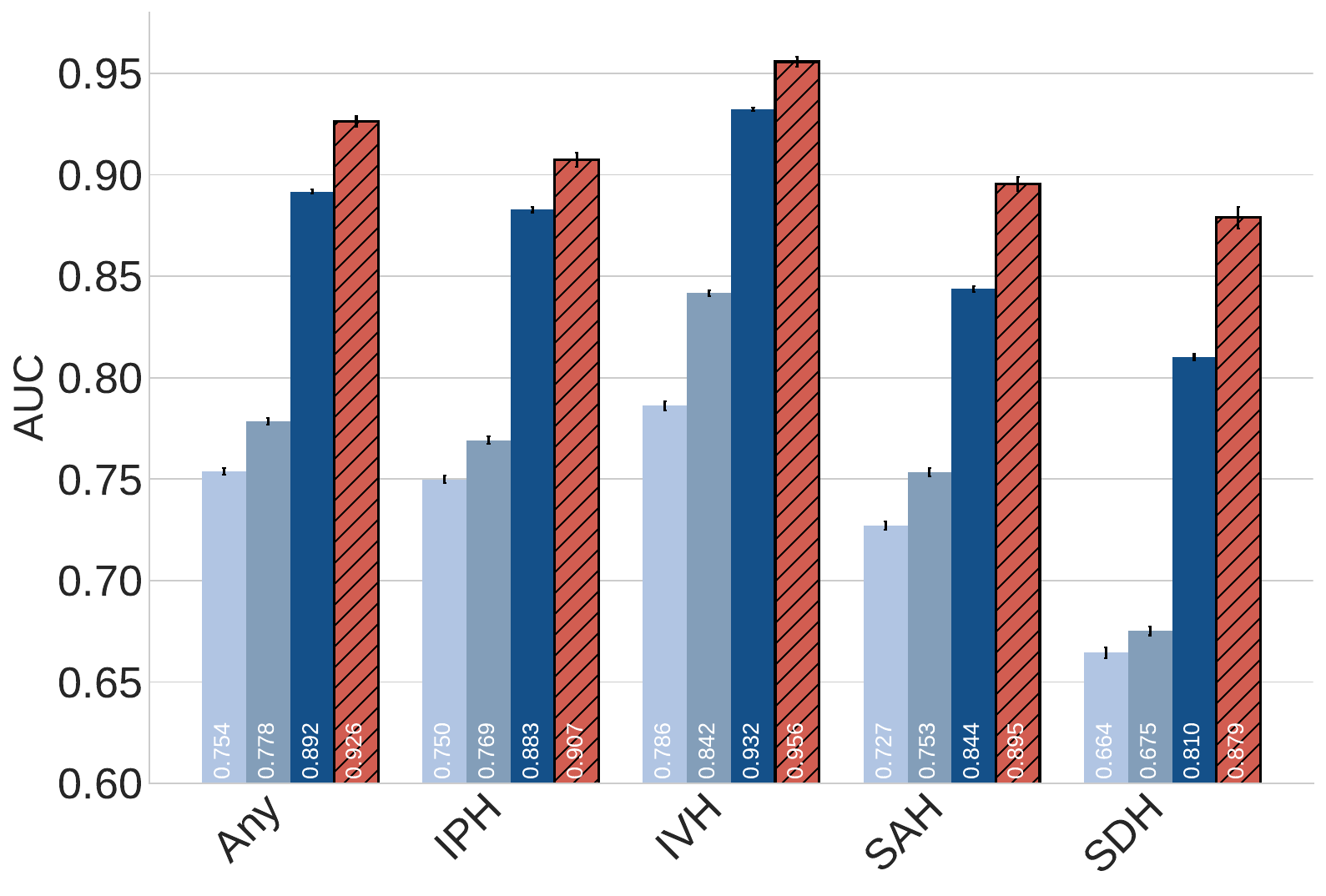}
\includegraphics[trim={0 0 5mm 0},clip,height=0.27\textwidth]{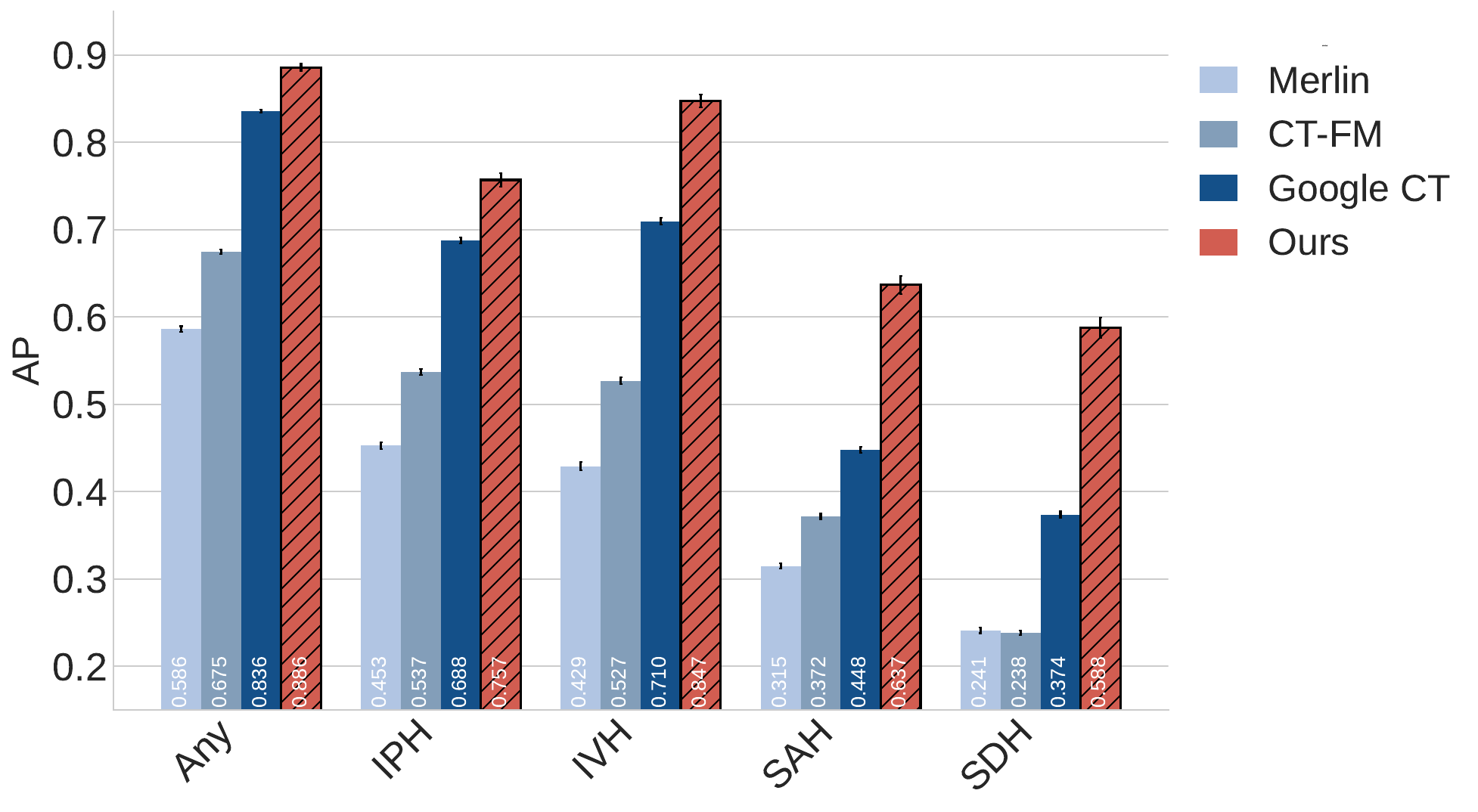}

\makebox[\textwidth][l]{%
    \hspace{0.38\textwidth}\textbf{CQ500}
} \\[0.2cm]
\includegraphics[trim={0 0 10mm 0},clip,height=0.345\textwidth]{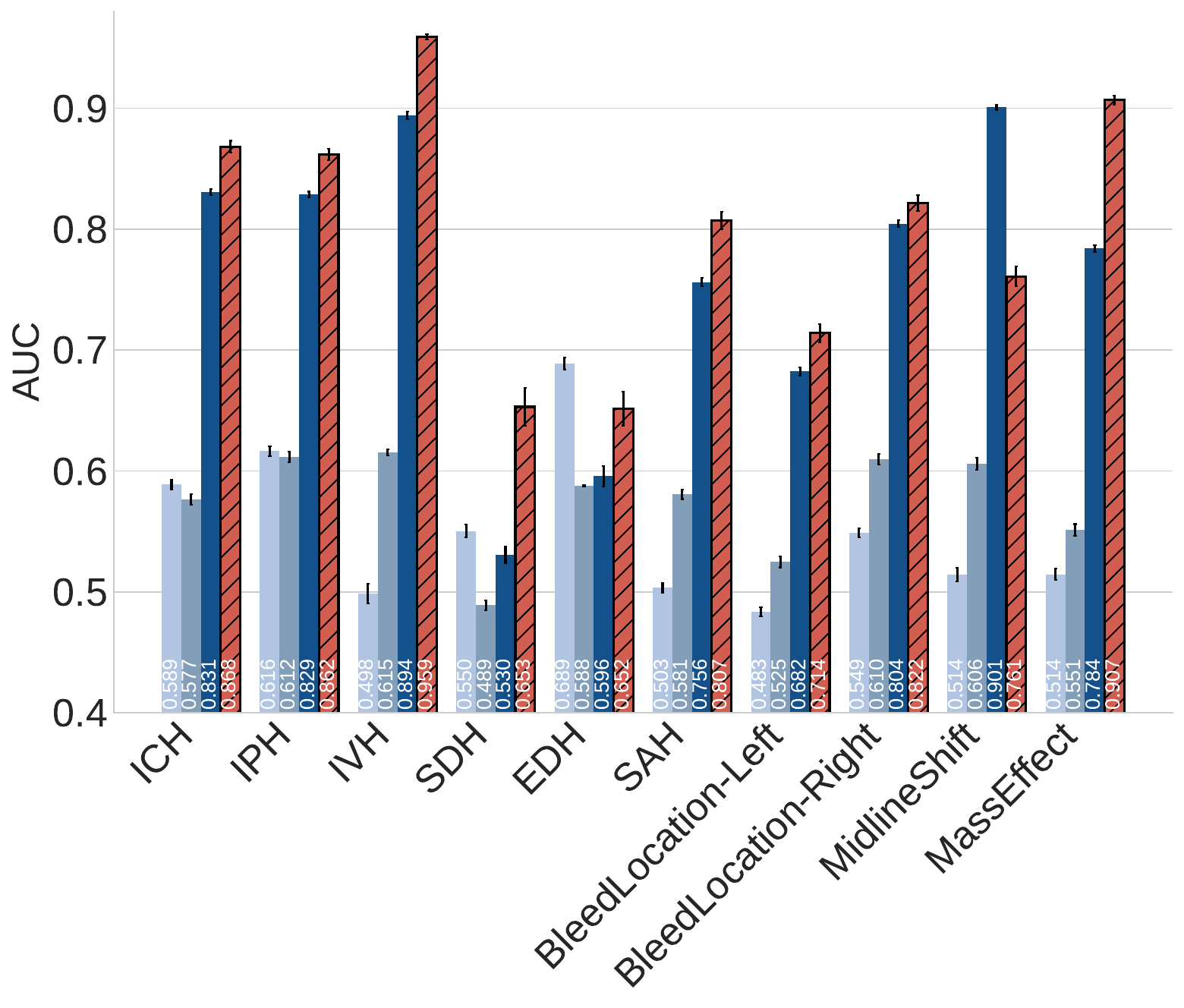}
\includegraphics[trim={0 0 5mm 0},clip,height=0.345\textwidth]{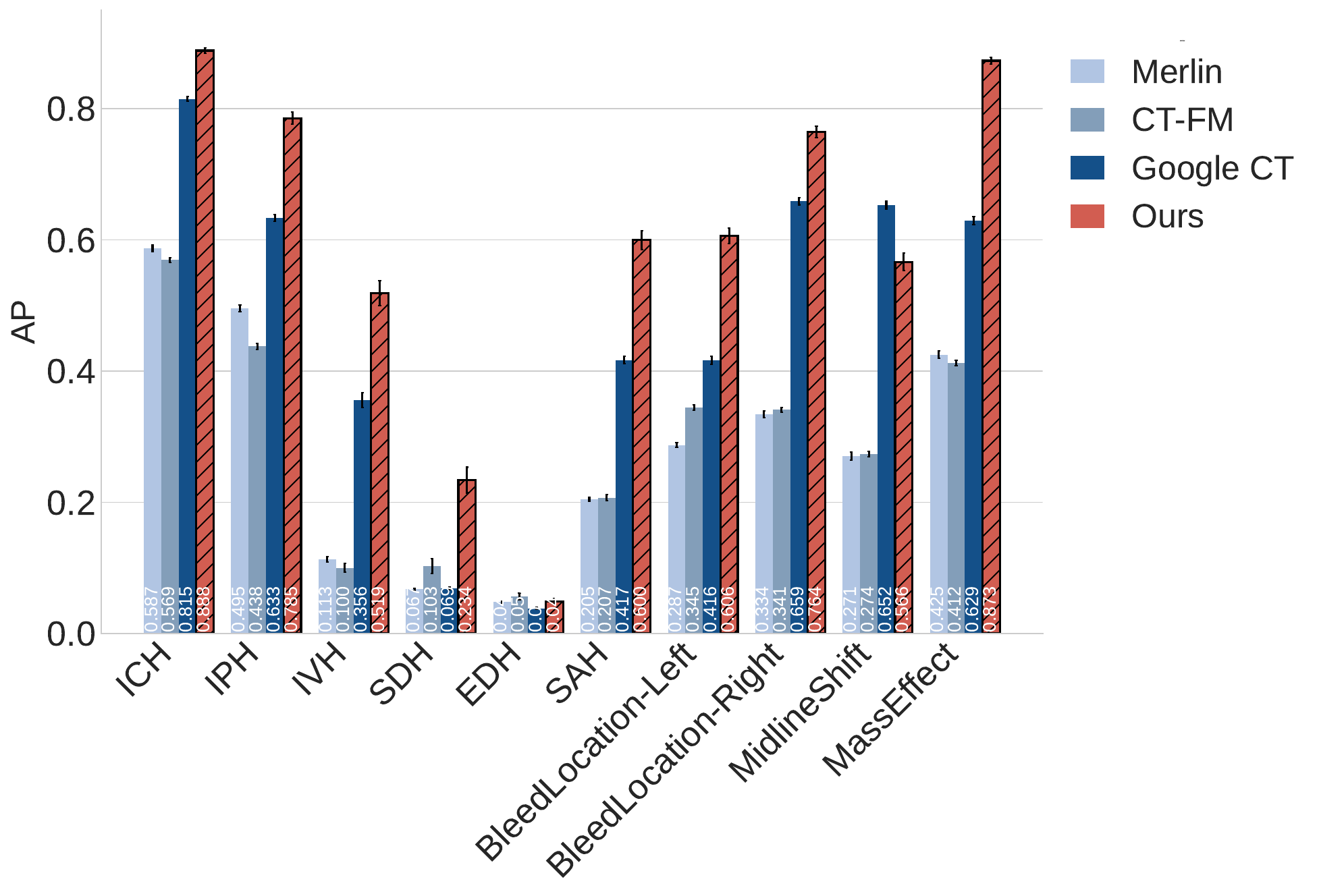}

\caption{\textbf{Performance comparison for linear probing} This plot compares our model performance vs. Google CT Foundation model\cite{yang2024advancingmultimodalmedicalcapabilities}, Merlin \cite{blankemeier2024merlinvisionlanguagefoundation} and CT-FM \cite{pai2025visionfoundationmodelscomputed} across all diseases on RSNA and CQ500. Since Google CT Foundation model requires uploading data to Google Cloud (not allowed on our private data) for requesting model embeddings with model weights inaccessible, only public dataset comparison is provided in this study. Similar to other evaluations, we observed that our model outperforms Google CT Foundation model across the board with the only exception on Midline Shift for Google CT Foundation model and EDH for Merlin. Note that CT-FM original data pre-process used width-wise concatenation for different CT windows different from its pre-training input shape, which causes sub-optimal performance when linear probing is performed (not able to adjust backbone parameters for adapting to this specific data input).}
\label{fig:probing-comparison-gemini}
\end{figure}

Supplementary \Cref{fig:probing-comparison-gemini} compares our foundation model with Google CT Foundation model~\cite{yang2024advancingmultimodalmedicalcapabilities}, CT-FM~\cite{pai2025visionfoundationmodelscomputed} and Merlin~\cite{blankemeier2024merlinvisionlanguagefoundation} under linear probing. Google CT Foundation were pre-trained with large-scale CT scans from different anatomy. Our model consistently shows improved performance across the board even though our model was pre-trained with less samples than Google CT Foundation.

\begin{figure}[htbp]
    \centering

    \begin{subfigure}[t]{0.32\textwidth}
        \centering
        \includegraphics[width=\linewidth]{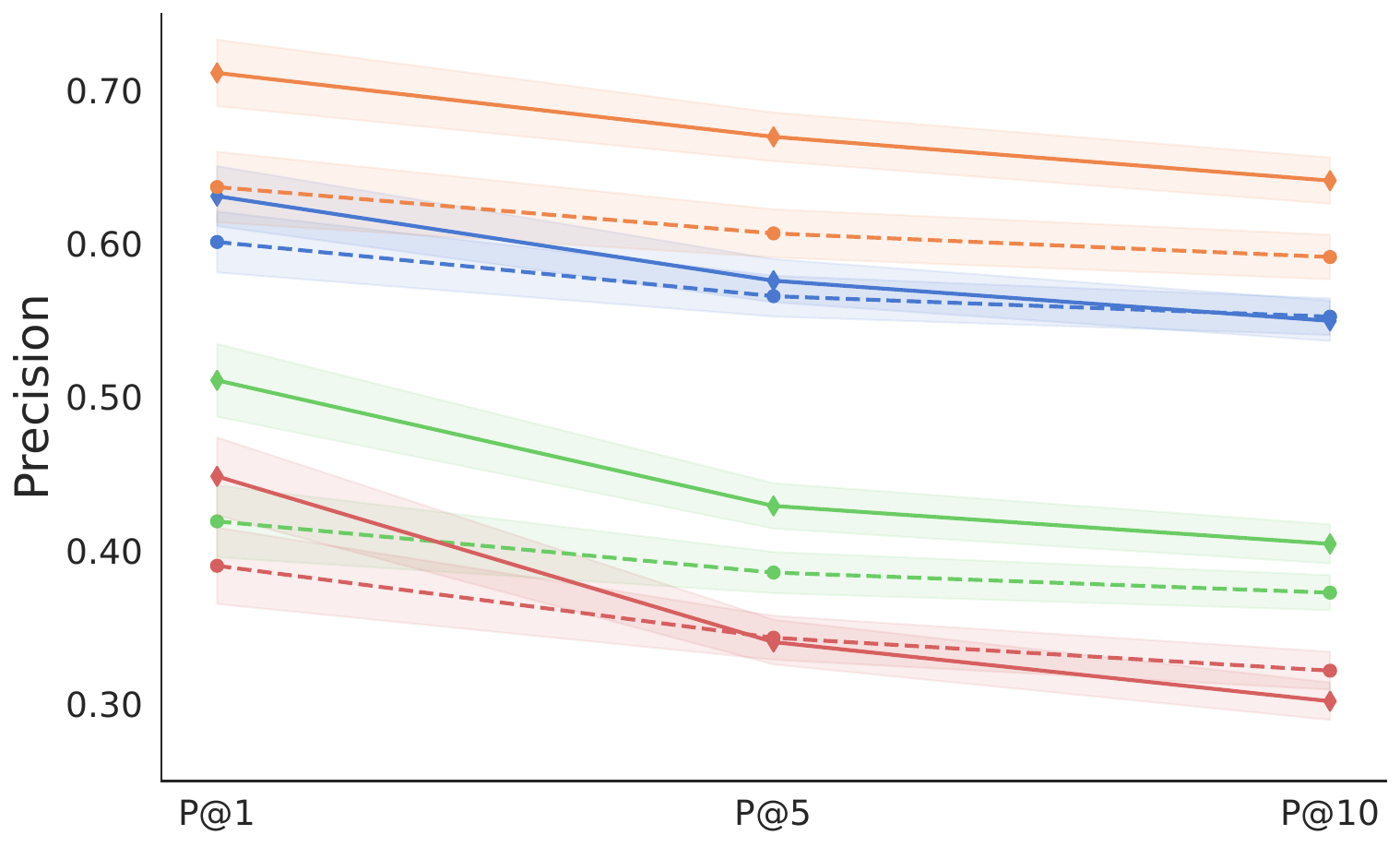}
        \caption{RSNA (vs. Google CT)}
    \end{subfigure}
    \hfill
    \begin{subfigure}[t]{0.32\textwidth}
        \centering
        \includegraphics[width=\linewidth]{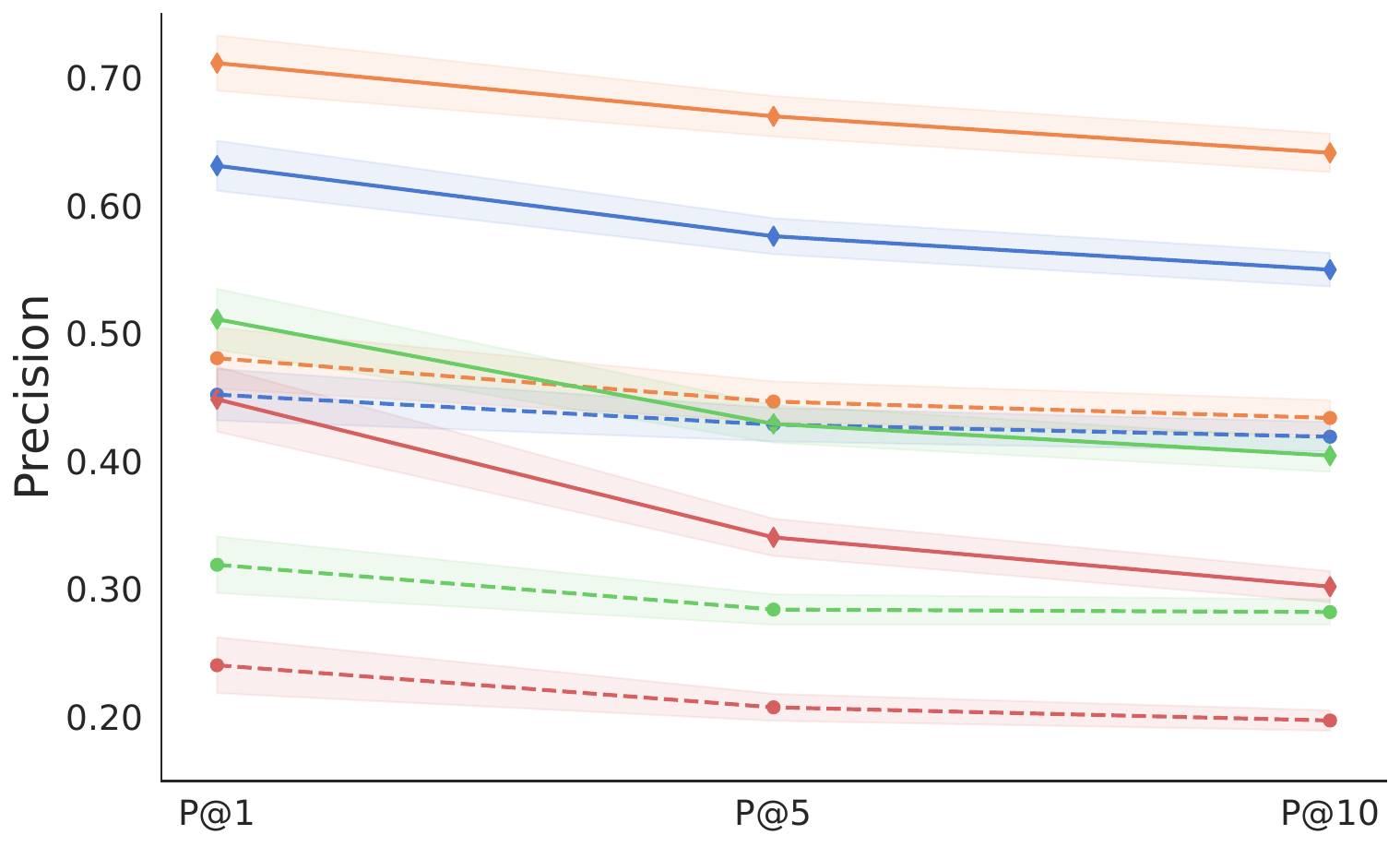}
        \caption{RSNA (vs. CT-FM)}
    \end{subfigure}
    \hfill
    \begin{subfigure}[t]{0.32\textwidth}
        \centering
        \includegraphics[width=\linewidth]{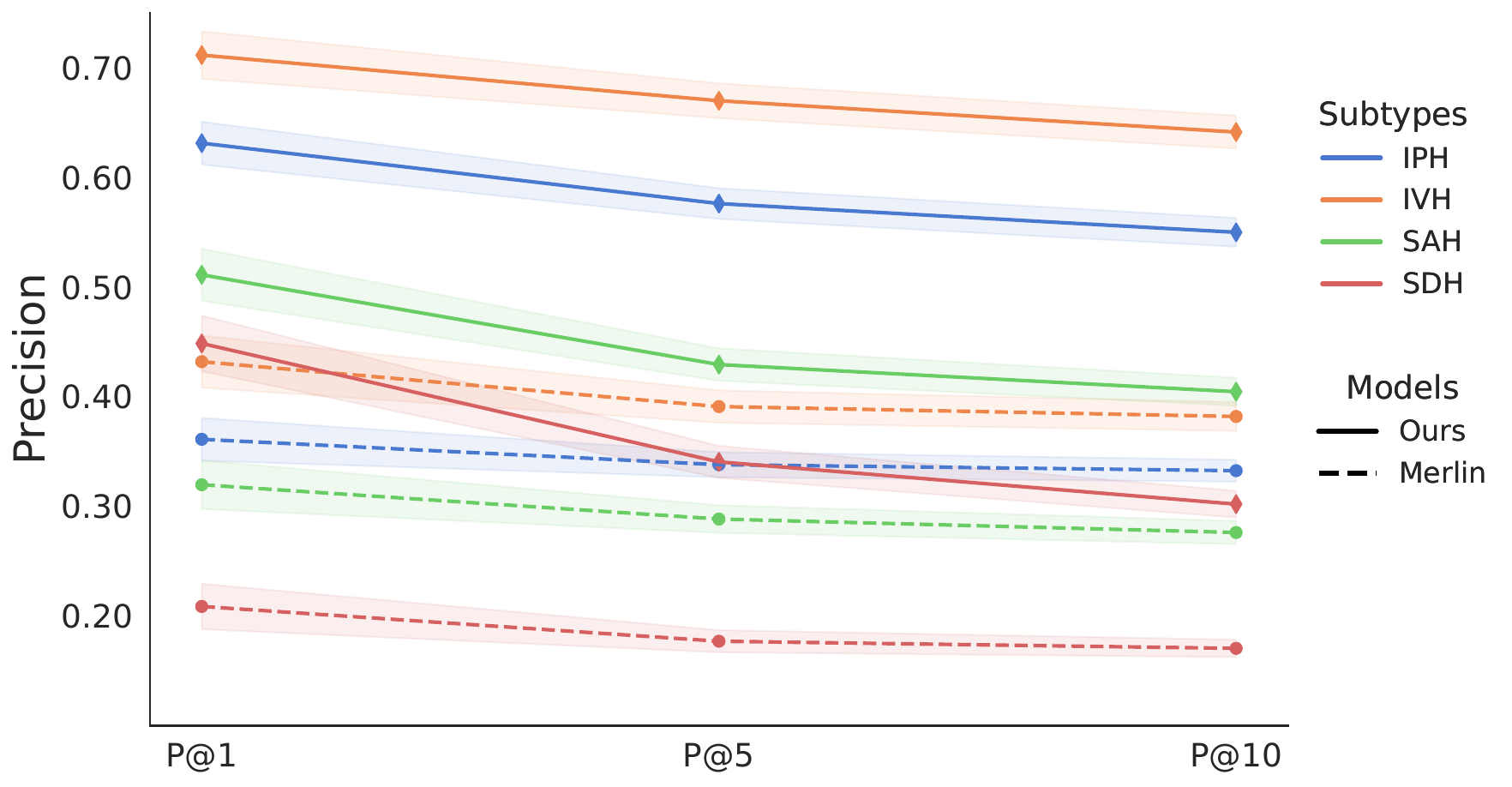}
        \caption{RSNA (vs. Merlin)}
    \end{subfigure}

    \vspace{0.4cm} 

    \begin{subfigure}[t]{0.32\textwidth}
        \centering
        \includegraphics[width=\linewidth]{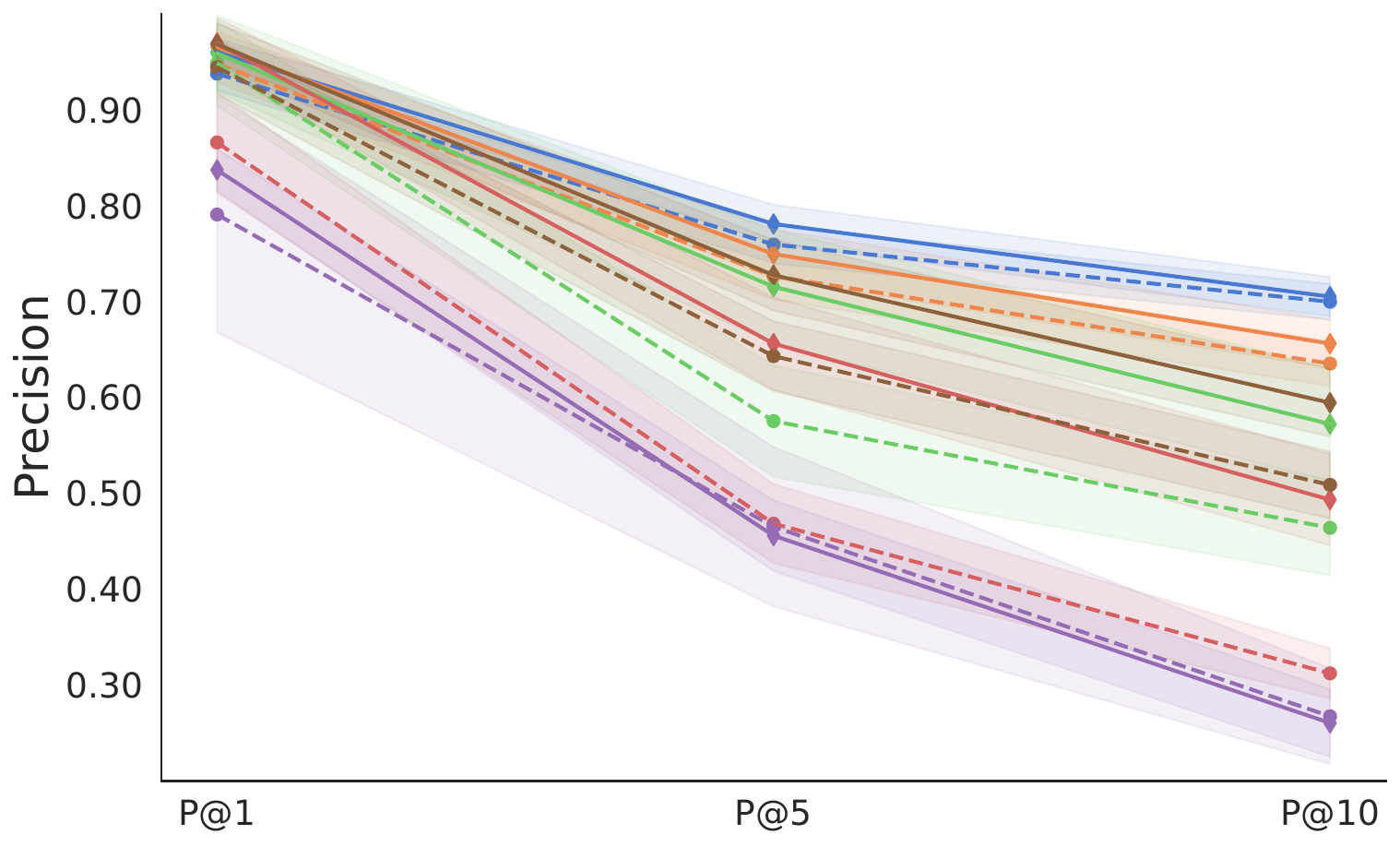}
        \caption{CQ500 (vs. Google CT)}
    \end{subfigure}
    \hfill
    \begin{subfigure}[t]{0.32\textwidth}
        \centering
        \includegraphics[width=\linewidth]{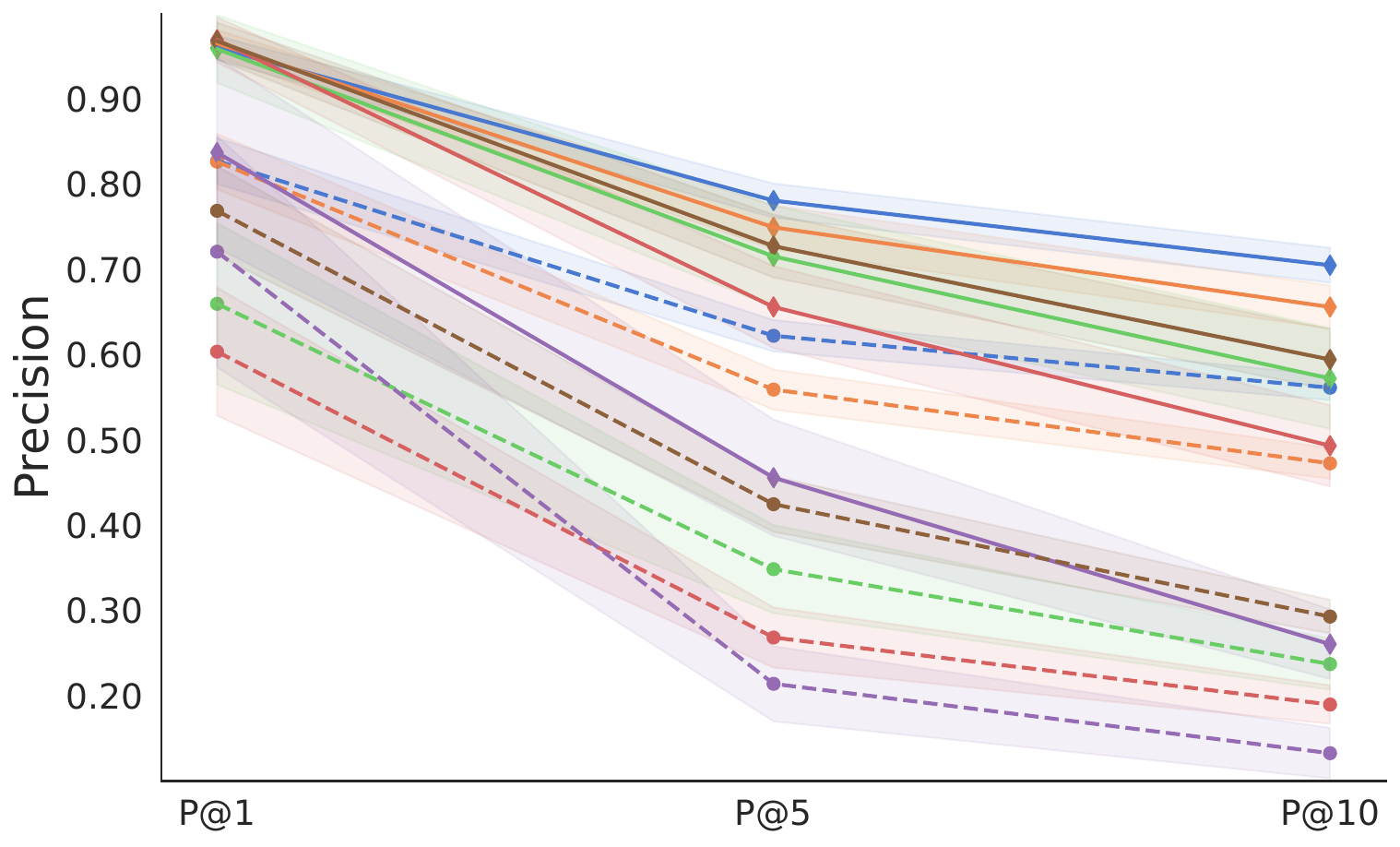}
        \caption{CQ500 (vs. CT-FM)}
    \end{subfigure}
    \hfill
    \begin{subfigure}[t]{0.32\textwidth}
        \centering
        \includegraphics[width=\linewidth]{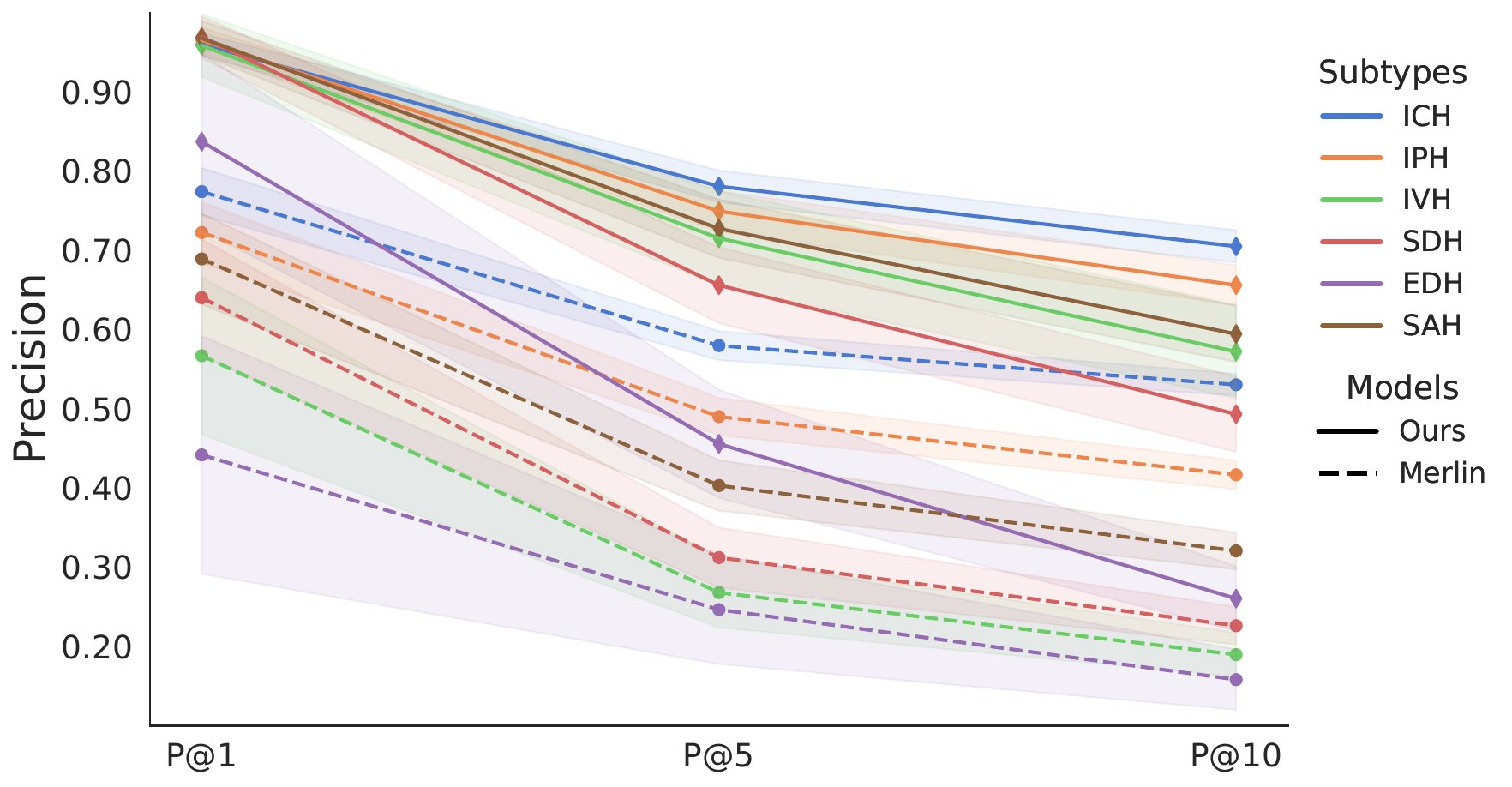}
        \caption{CQ500 (vs. Merlin)}
    \end{subfigure}

    \caption{\textbf{Precision$@K$ performance comparison for volume-to-volume retrieval.} 
    Solid lines represent our model, dotted lines the compared model, and different colors represent subtypes. 
    Our model maintains competitive performance across all tasks.}
    \label{fig:precision_retrieval}
\end{figure}

Supplementary \Cref{fig:precision_retrieval} presents the volume-to-volume subtype retrieval performance evaluated with Precision$@K$. Our model consistently outperforms CT-FM and Merlin across subtypes. Compared to Google CT, the improvements are more modest, with certain subtypes reaching comparable performance.

\begin{figure}[htbp]
    \centering
    NYU Langone \\
    \includegraphics[trim={0 0 140pt 0},clip,height=0.3\textwidth]{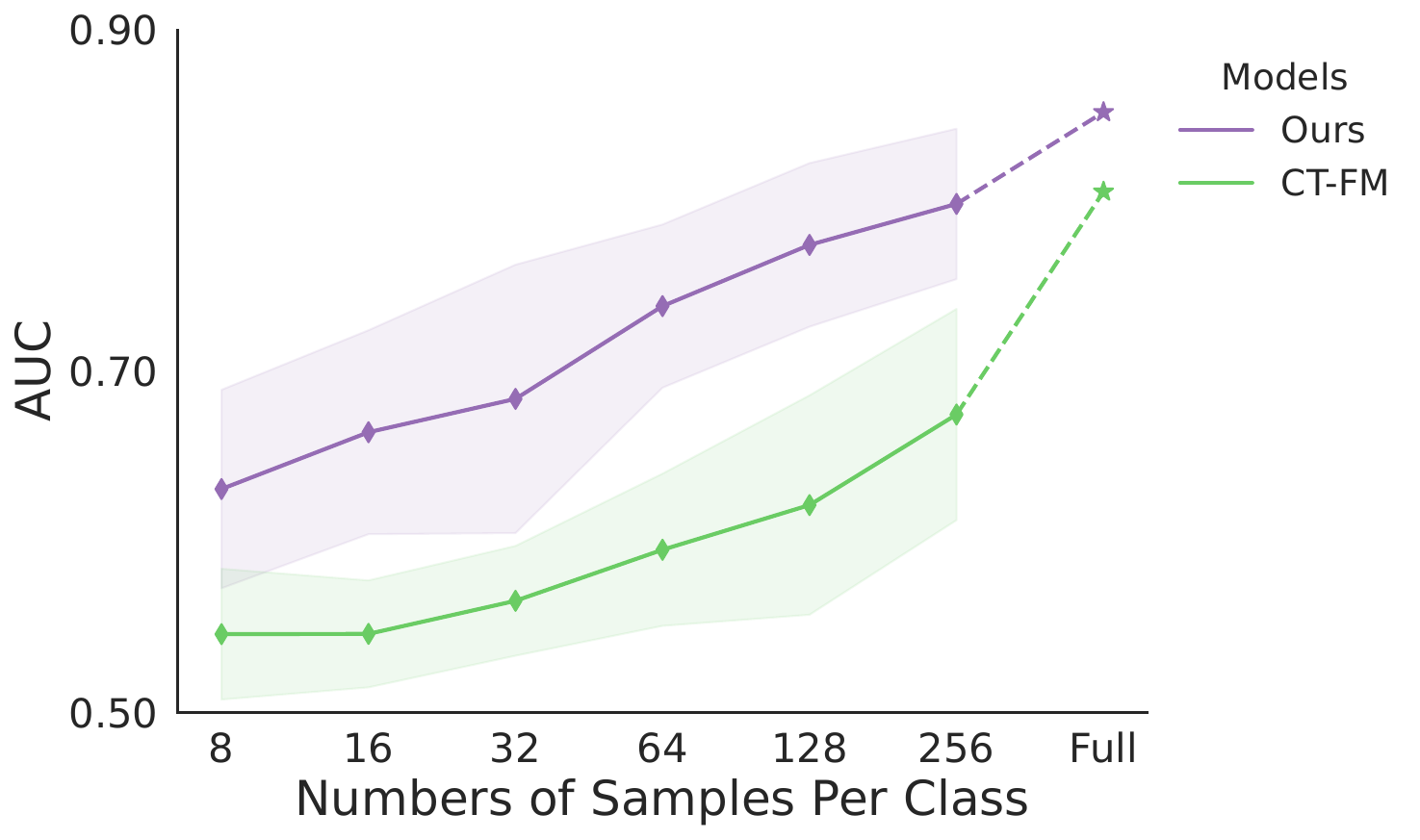}
    \includegraphics[height=0.3\textwidth]{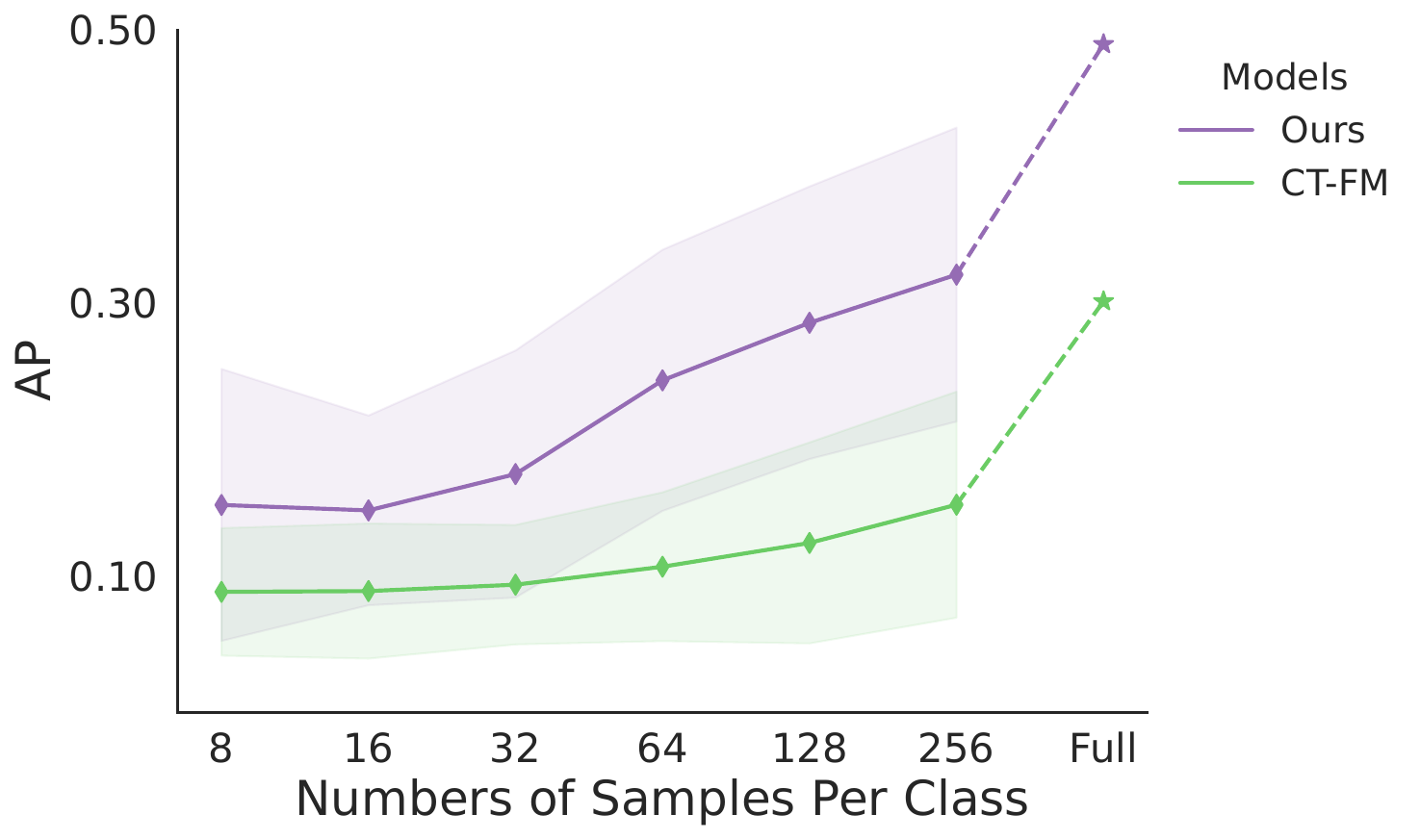} \\
    NYU Long Island \\
    \includegraphics[trim={0 0 140pt 0},clip,height=0.3\textwidth]{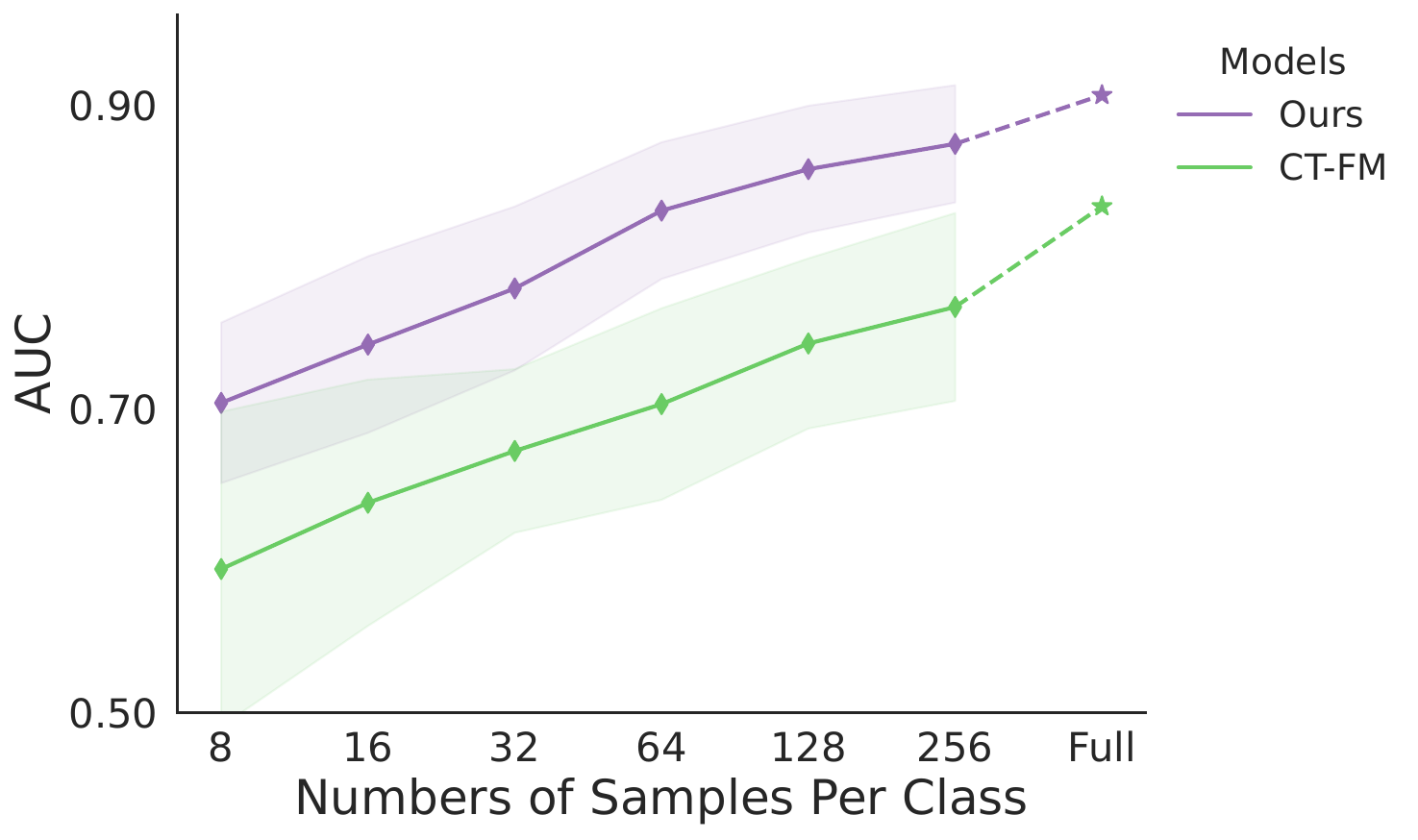}
    \includegraphics[height=0.3\textwidth]{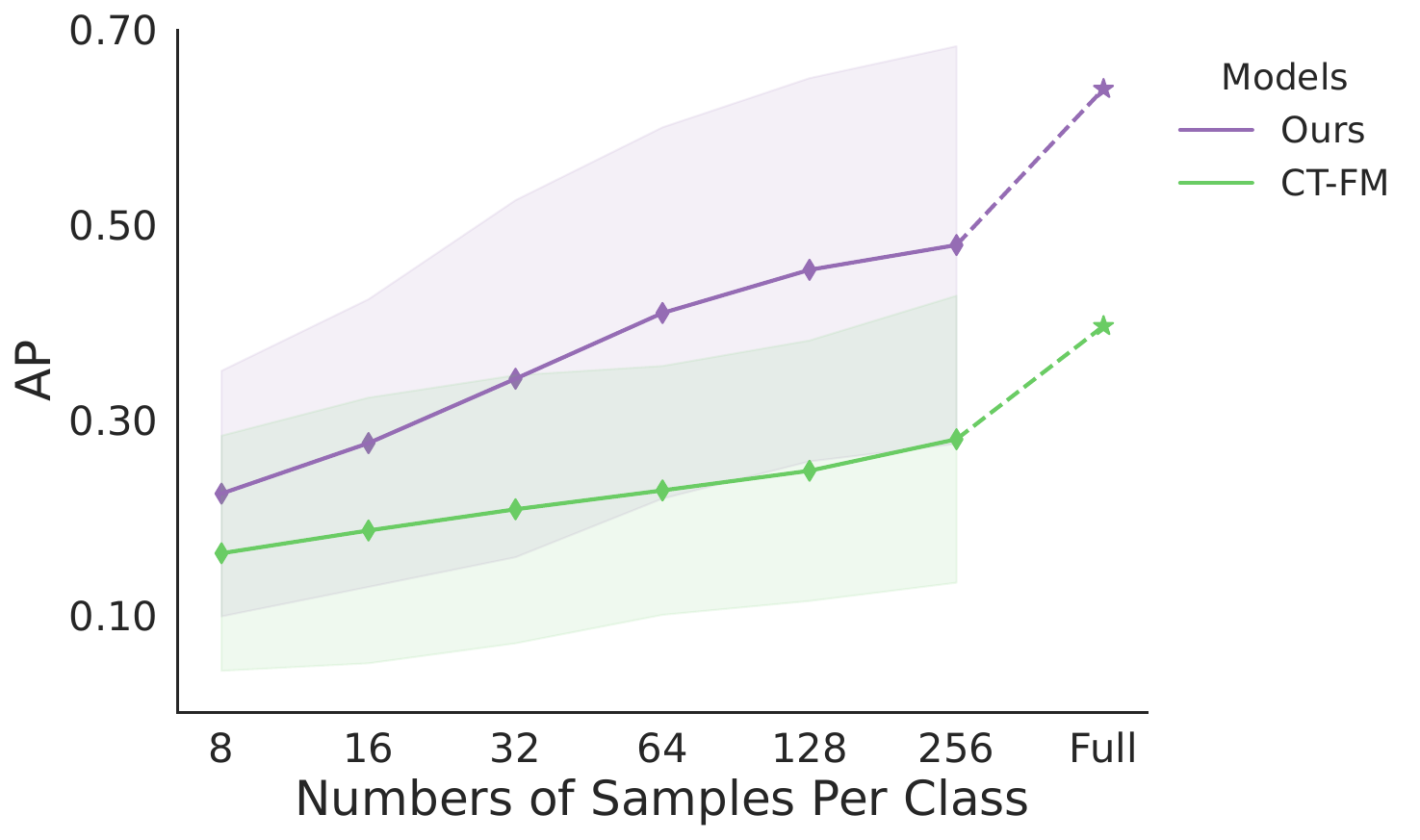} \\
    RSNA \\
    \includegraphics[trim={0 0 125pt 0},clip,height=0.295\textwidth]{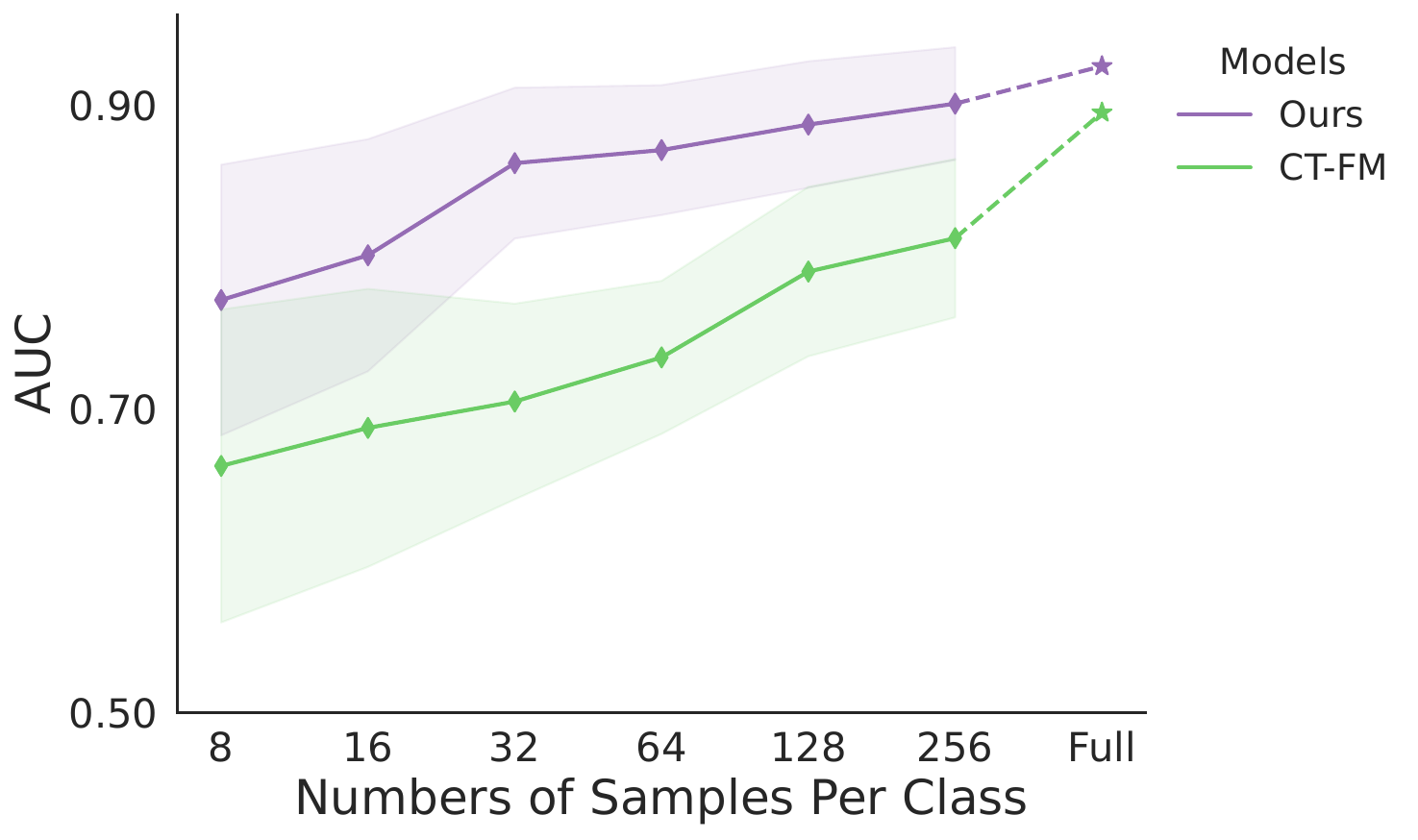}
    \includegraphics[height=0.295\textwidth]{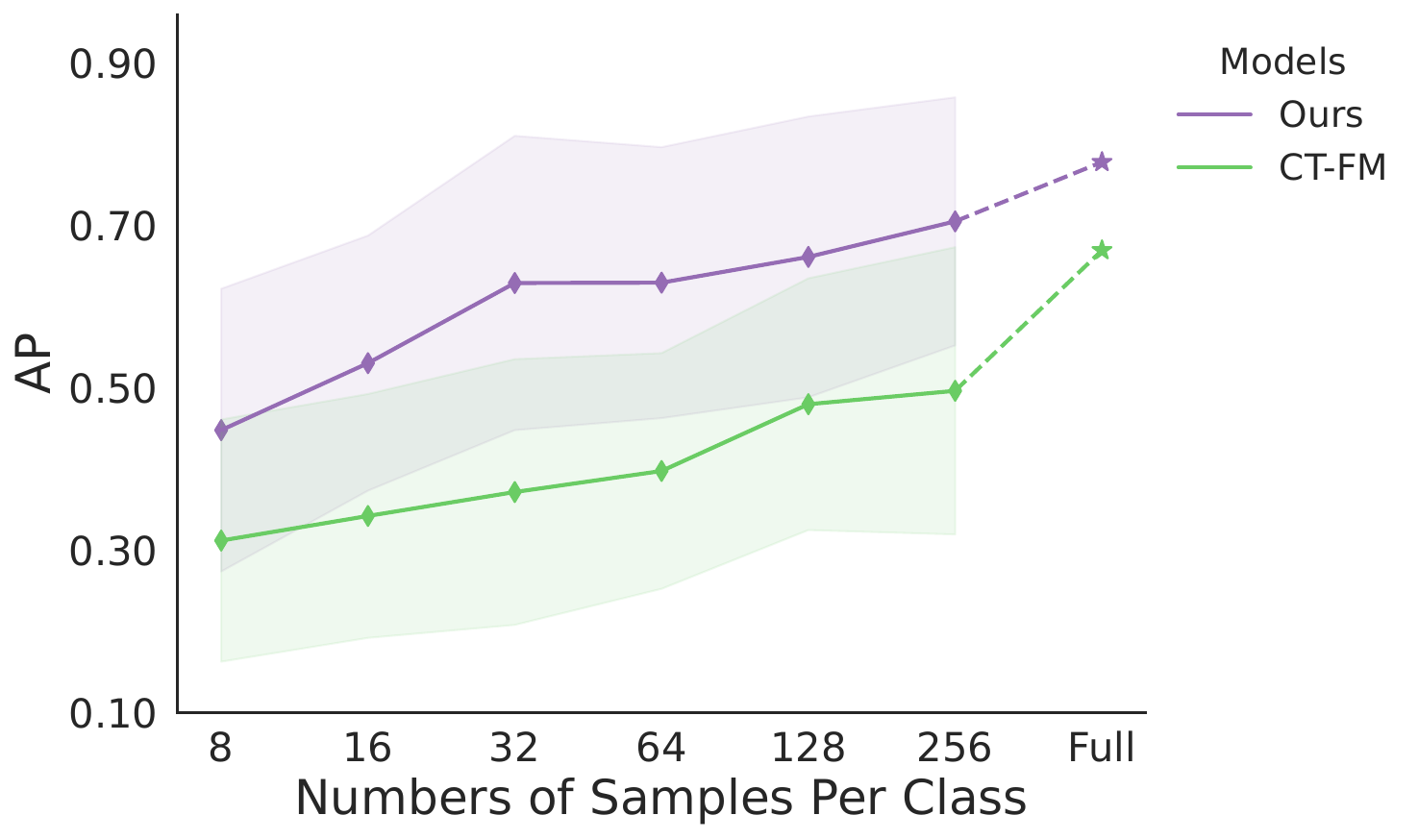}
    \caption{\textbf{Few-shot performance of different foundation models.} We compare different few-shot models performance averaged across diseases for different datasets. The experiment setup follows \Cref{fig:fewshot} in the main article (evaluated on 8, 16, 32, 64, 128, 256 balanced positive/negative samples). We show our model consistently outperform CT-FM under few-shot setting.}
    \label{fig:moedl_comparison_fewshot}
\end{figure}

\begin{figure}[htbp]
    \centering
    NYU Langone \\
    \includegraphics[trim={0 0 140pt 0},clip,height=0.3\textwidth]{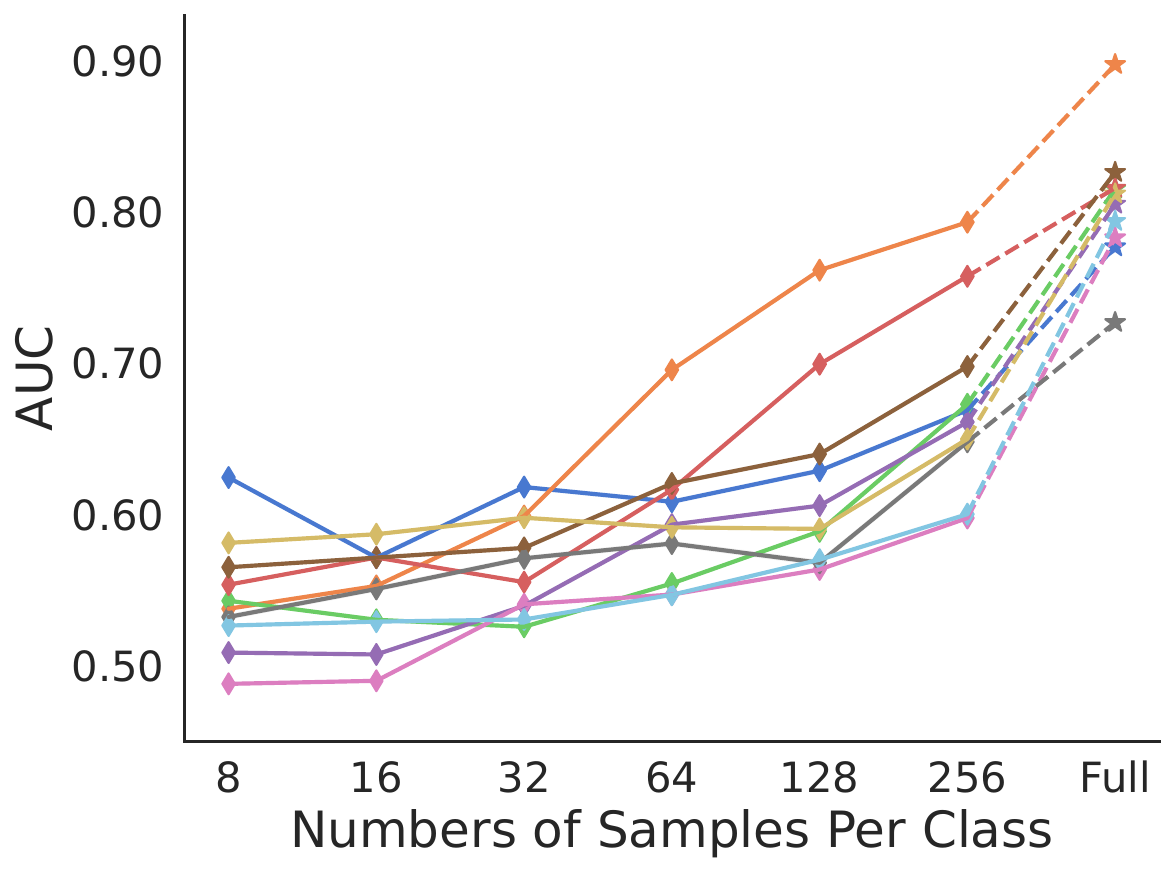}
    \includegraphics[height=0.3\textwidth]{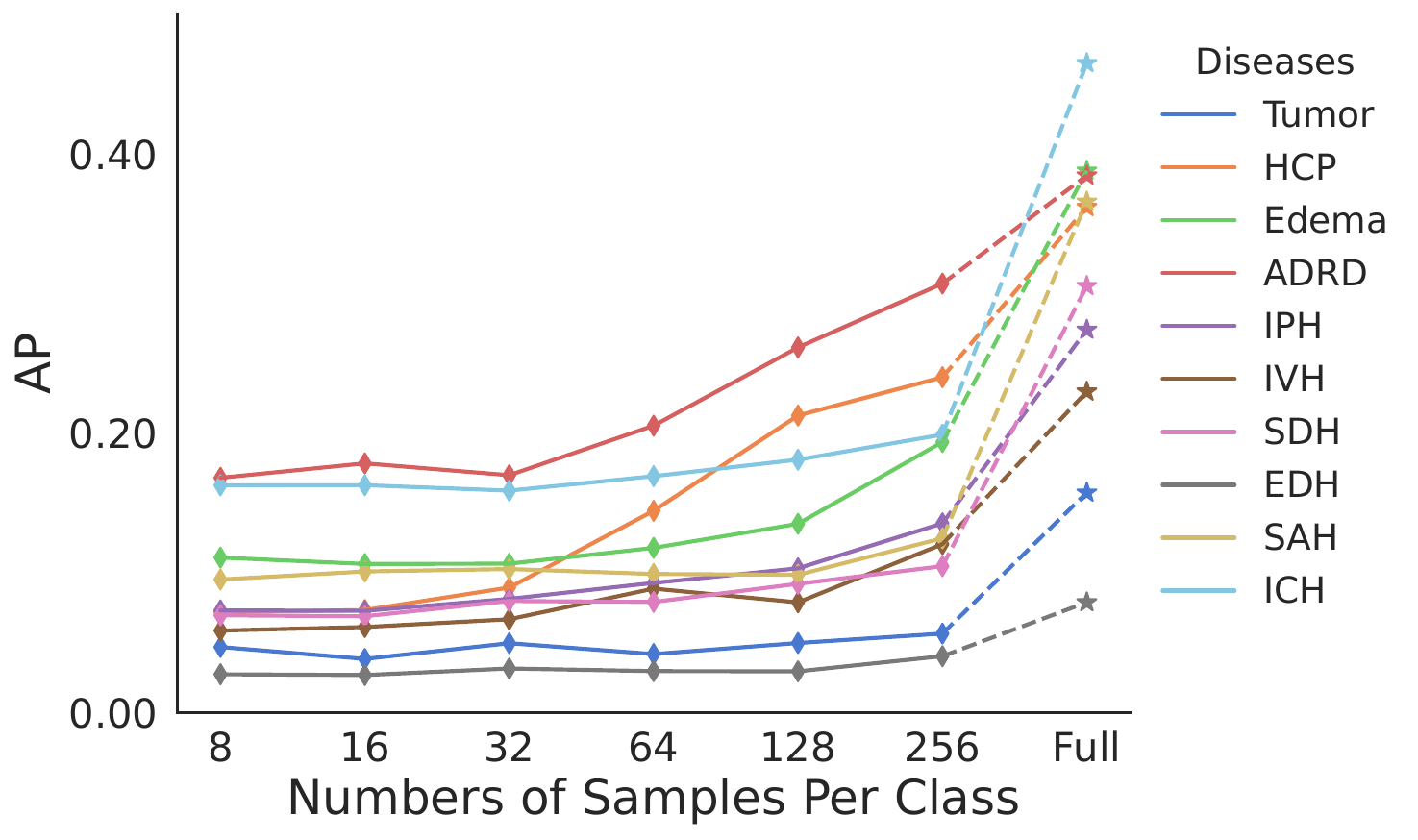} \\
    NYU Long Island \\
    \includegraphics[trim={0 0 140pt 0},clip,height=0.3\textwidth]{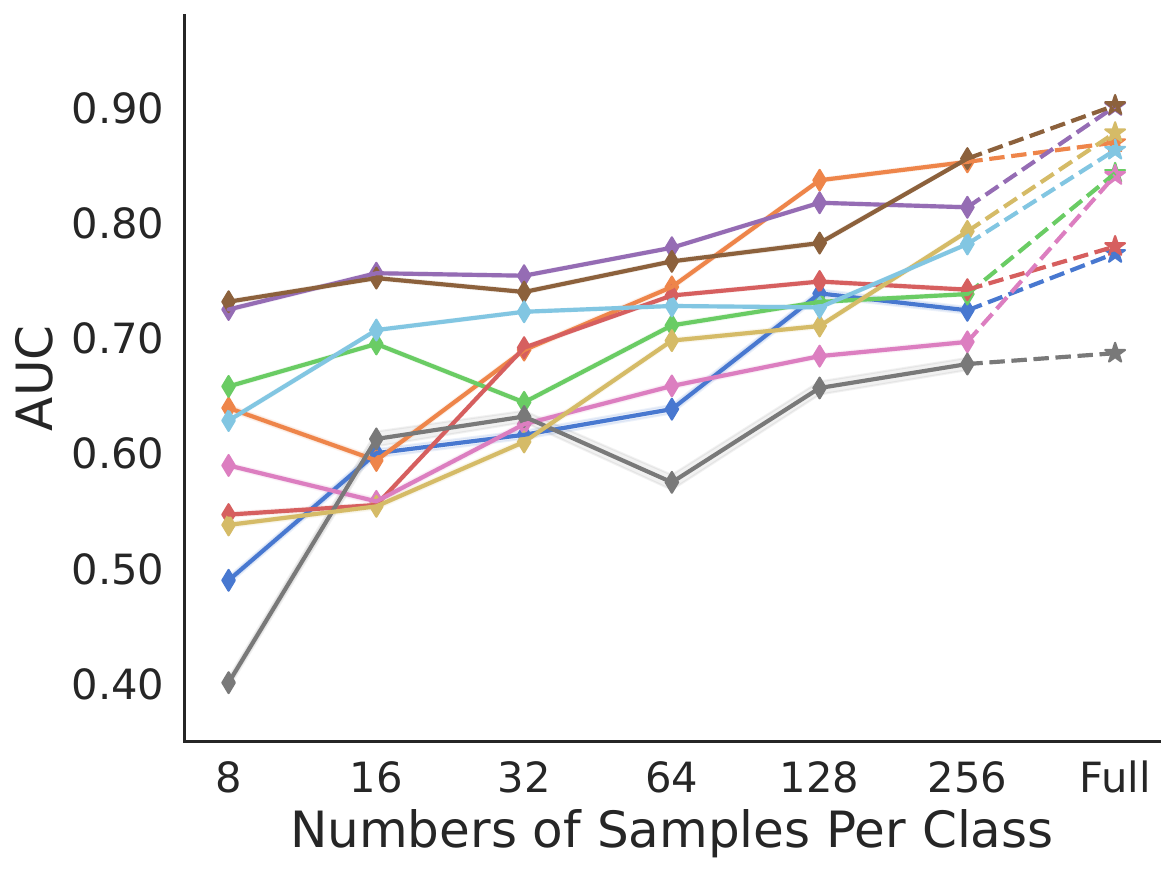}
    \includegraphics[height=0.3\textwidth]{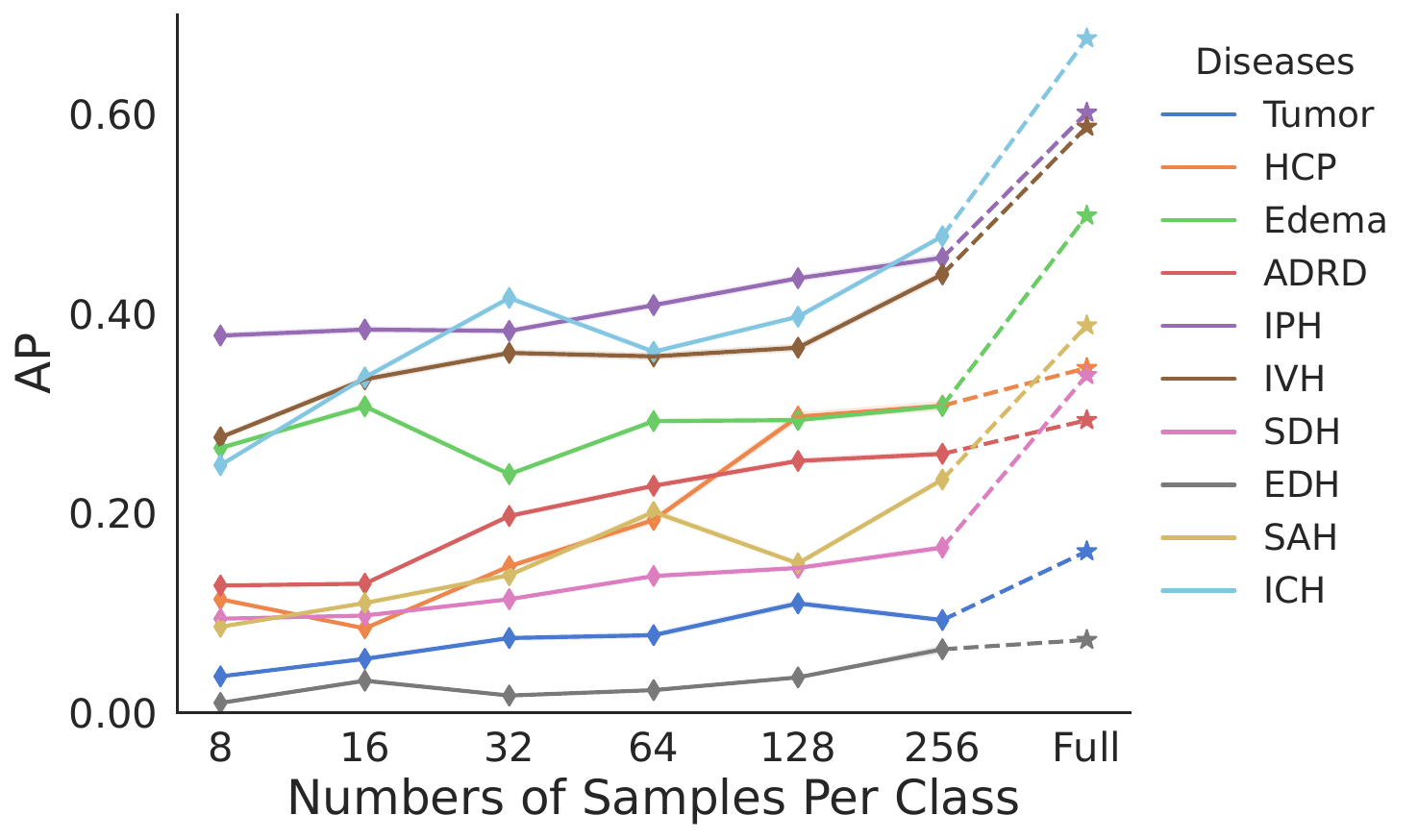} \\
    RSNA \\
    \includegraphics[trim={0 0 125pt 0},clip,height=0.295\textwidth]{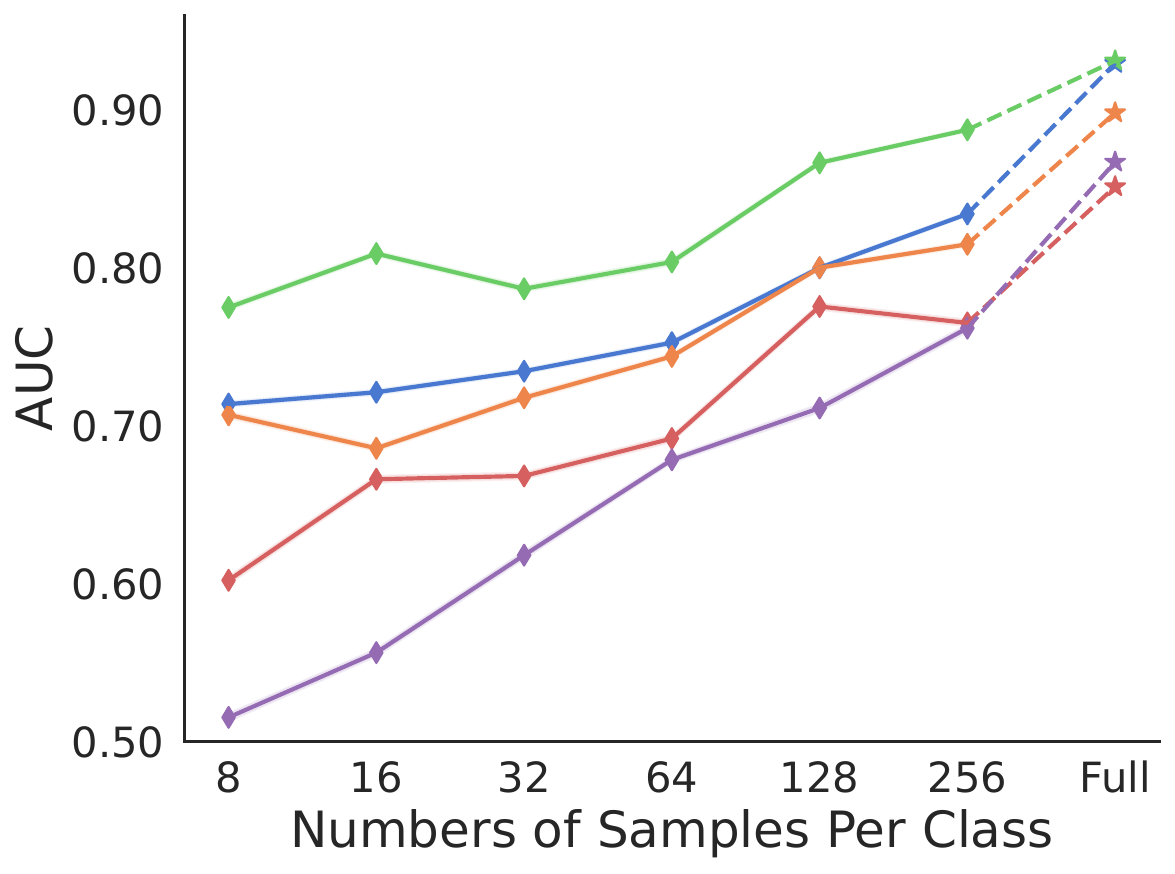}
    \includegraphics[height=0.295\textwidth]{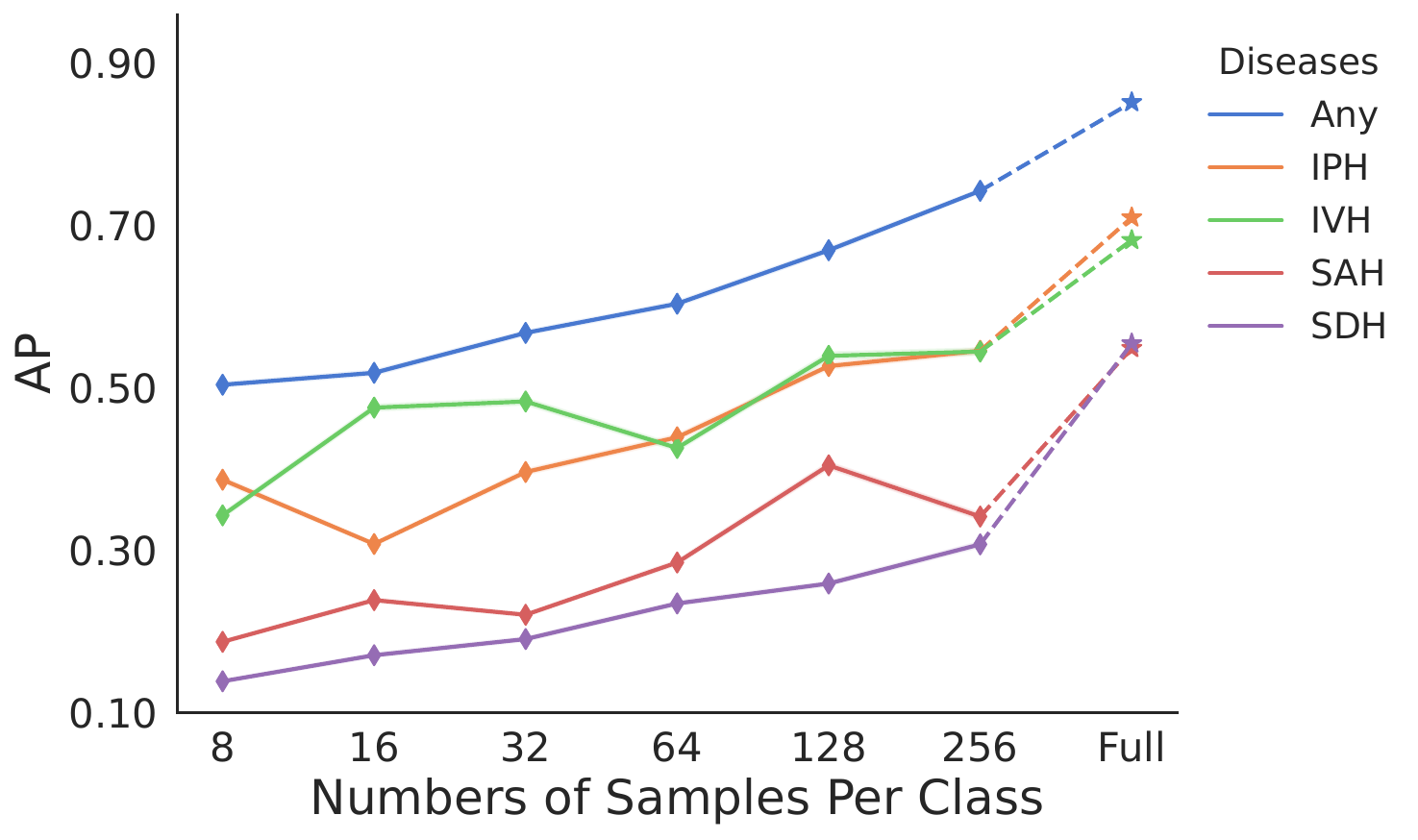}
    \caption{\textbf{Few-shot performance of CT-FM.}  The plots display the per-pathology AUROC and average precision (AP) of the disease detection model under a few-shot learning setting, following evaluation method of \Cref{fig:fewshot}}
    \label{fig:ct-fm_fewshot}
\end{figure}

Supplementary \Cref{fig:moedl_comparison_fewshot,fig:ct-fm_fewshot} compares our model with CT-FM under few-shot setting. The result shows our model also presents superior few-shot performance across the board over CT-FM, demonstrating the label efficiency advantage of our model.

\begin{figure*}
    \centering
    \makebox[\textwidth][l]{%
        \hspace{0.10\textwidth}
        \textbf{NYU Langone} \hspace{0.08\textwidth} \textbf{NYU Long Island} \hspace{0.1\textwidth} \textbf{RSNA} \hspace{0.15\textwidth} \textbf{CQ500}
    } \\[0.2cm]
    \includegraphics[trim={0 0 0 0},clip, width=0.22\textwidth]{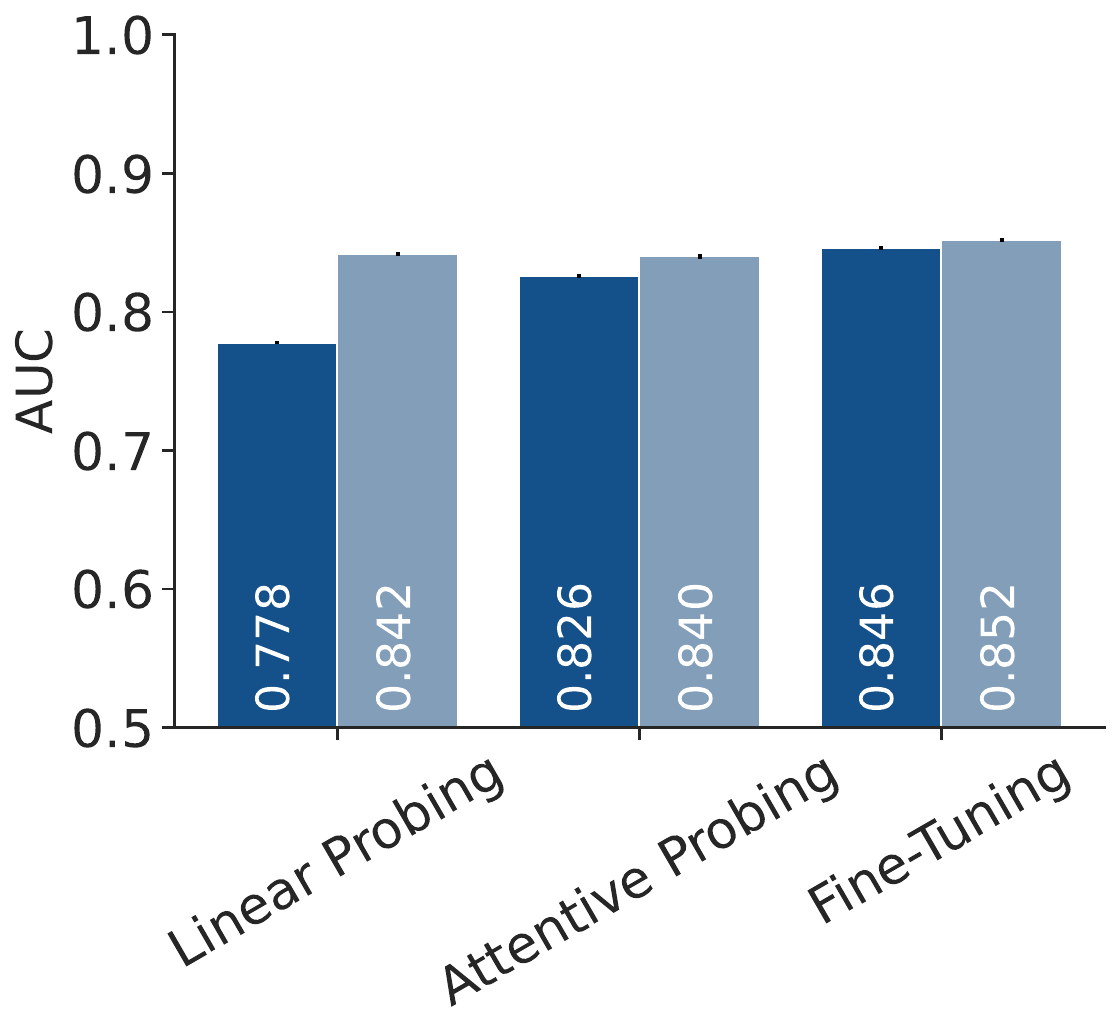}
    \includegraphics[trim={0 0 0 0},clip, width=0.22\textwidth]{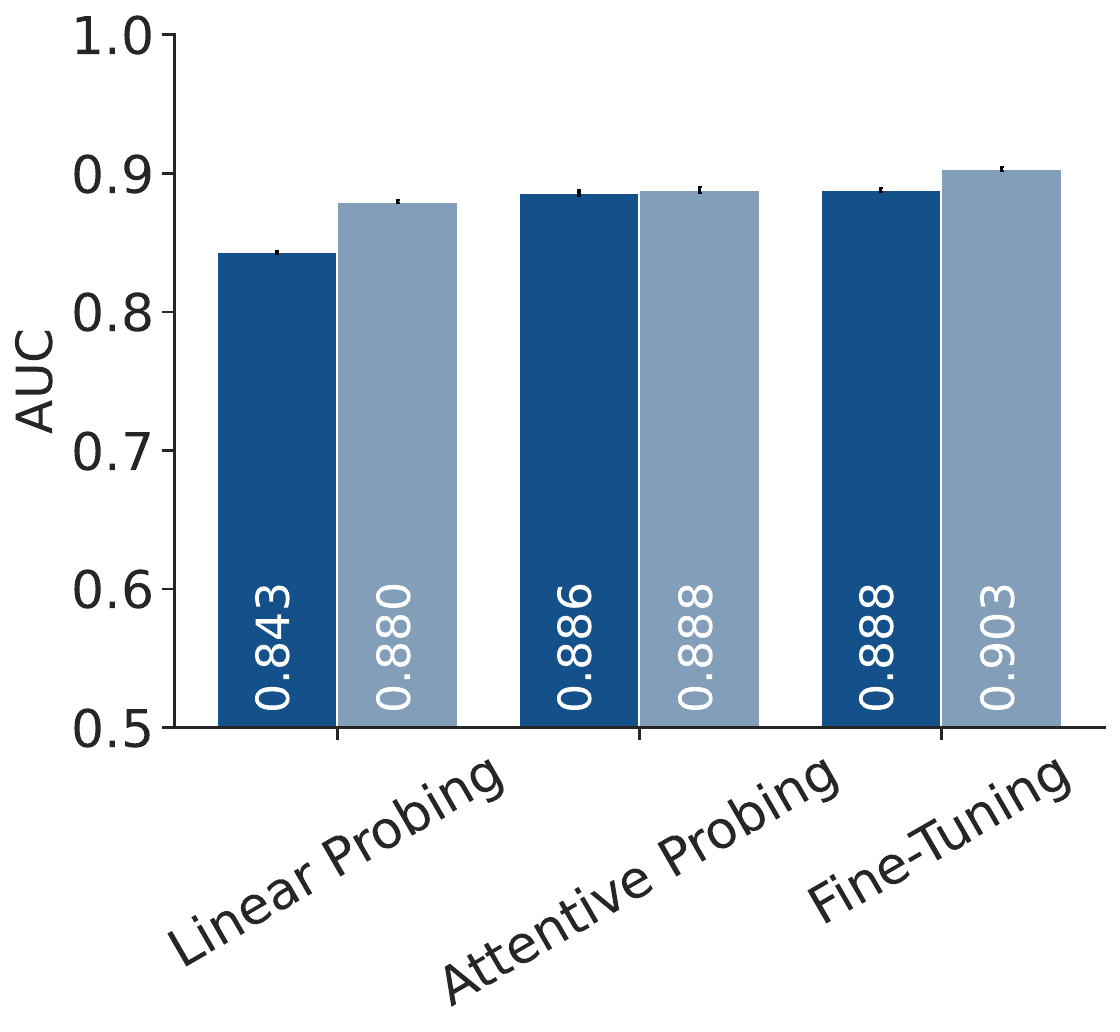}
    \includegraphics[trim={0 0 0 0},clip, width=0.22\textwidth]{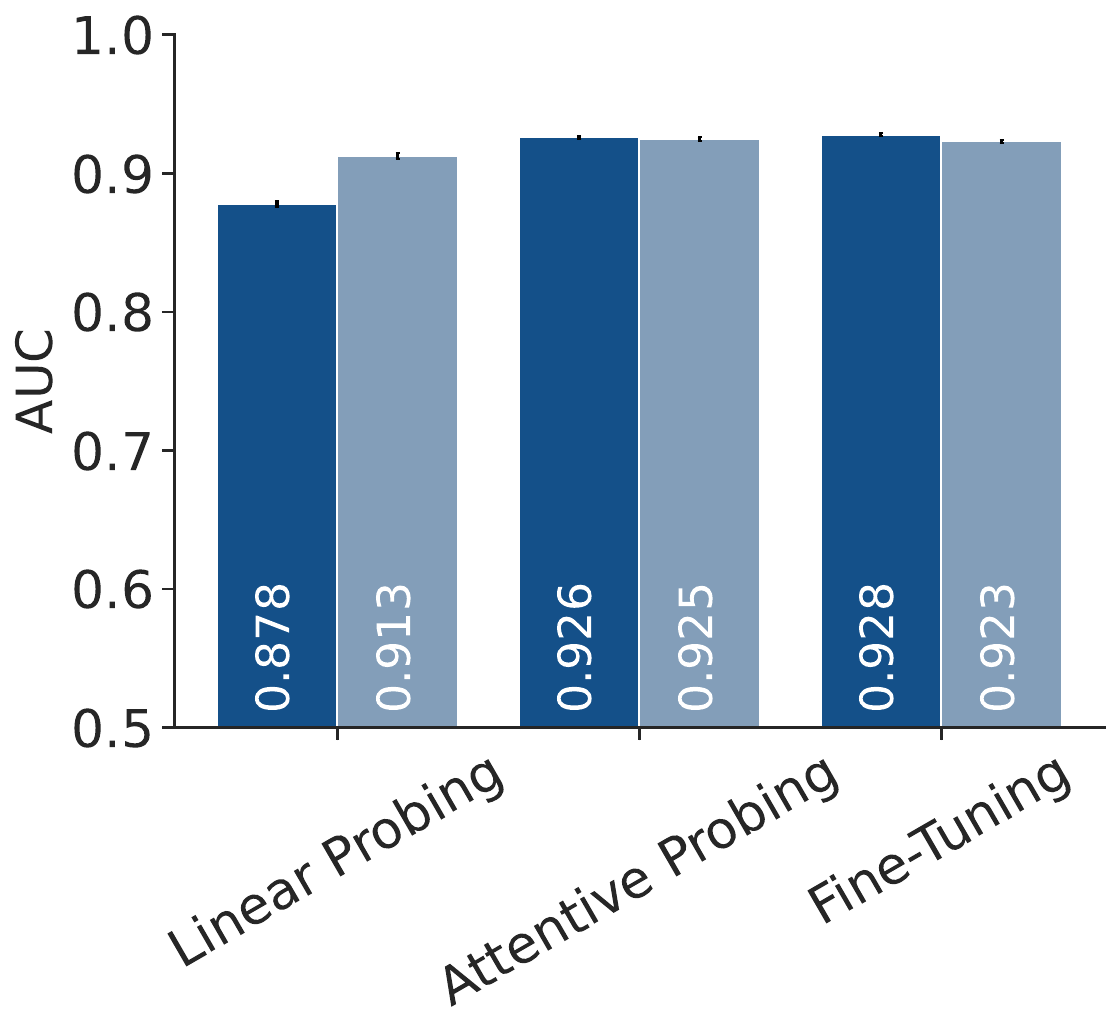}
    \includegraphics[trim={0 0 0 0},clip, width=0.28\textwidth]{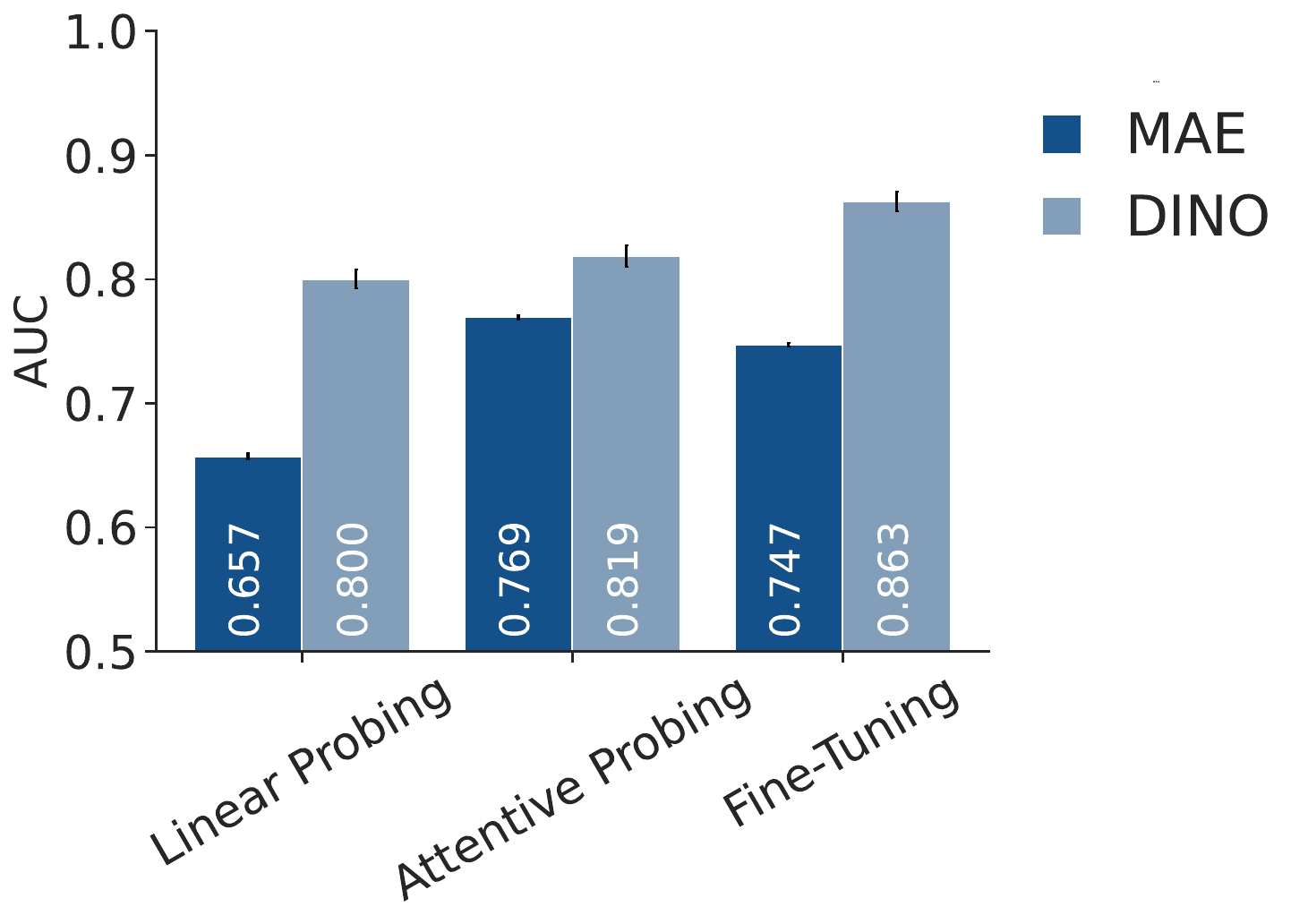}
    \\[0.2cm]
    \includegraphics[width=0.22\textwidth]{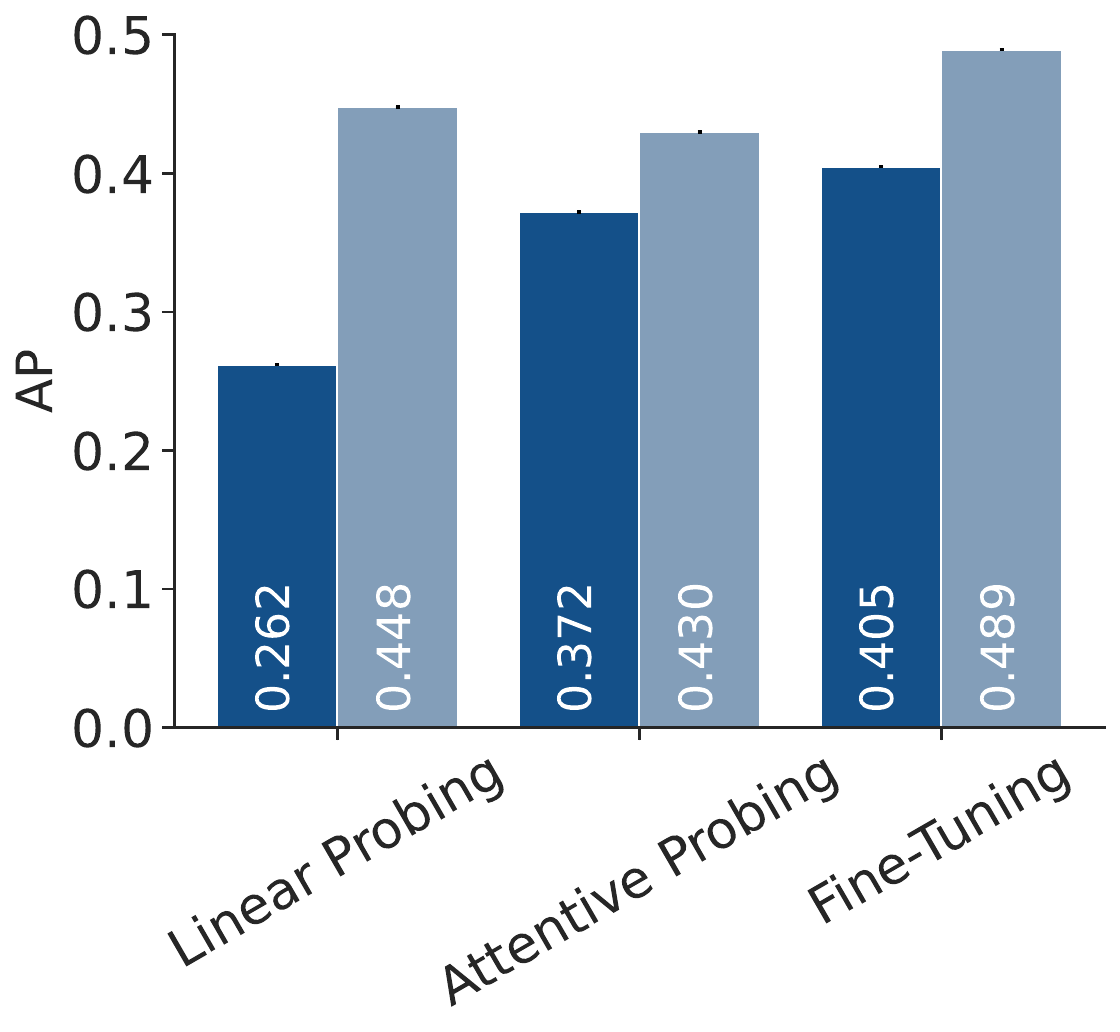} 
    \includegraphics[width=0.22\textwidth]{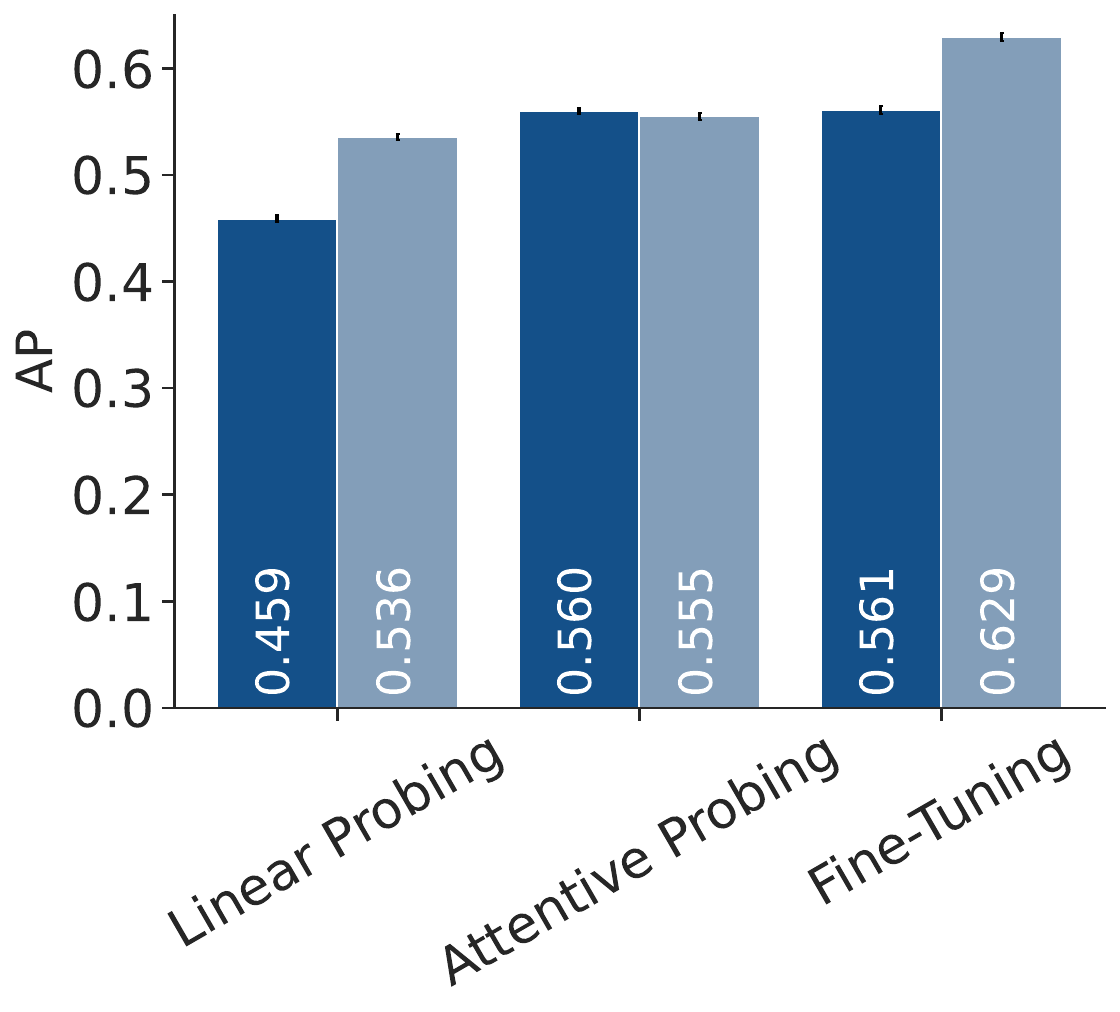} 
    \includegraphics[width=0.22\textwidth]{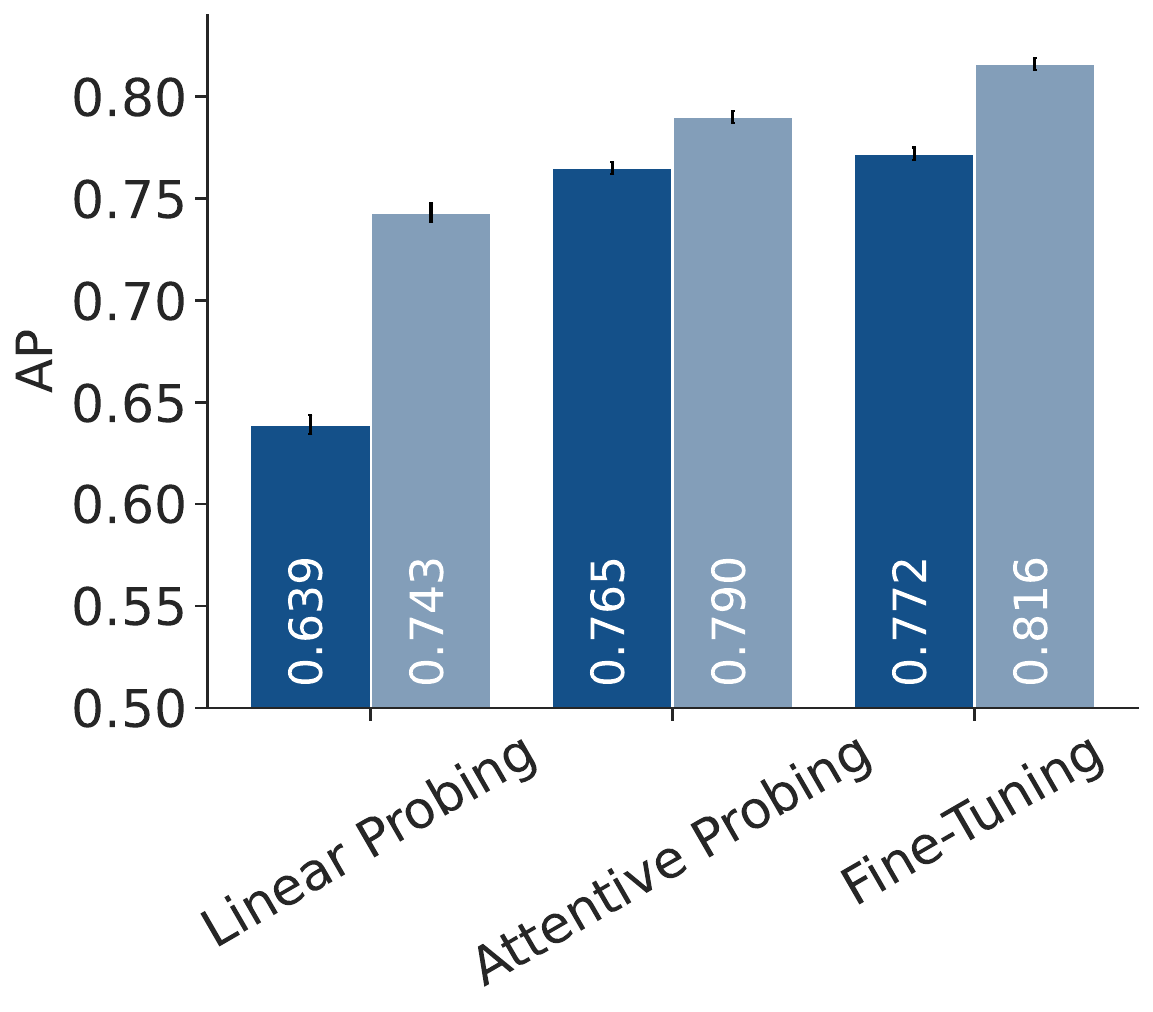}
    \includegraphics[width=0.28\textwidth]{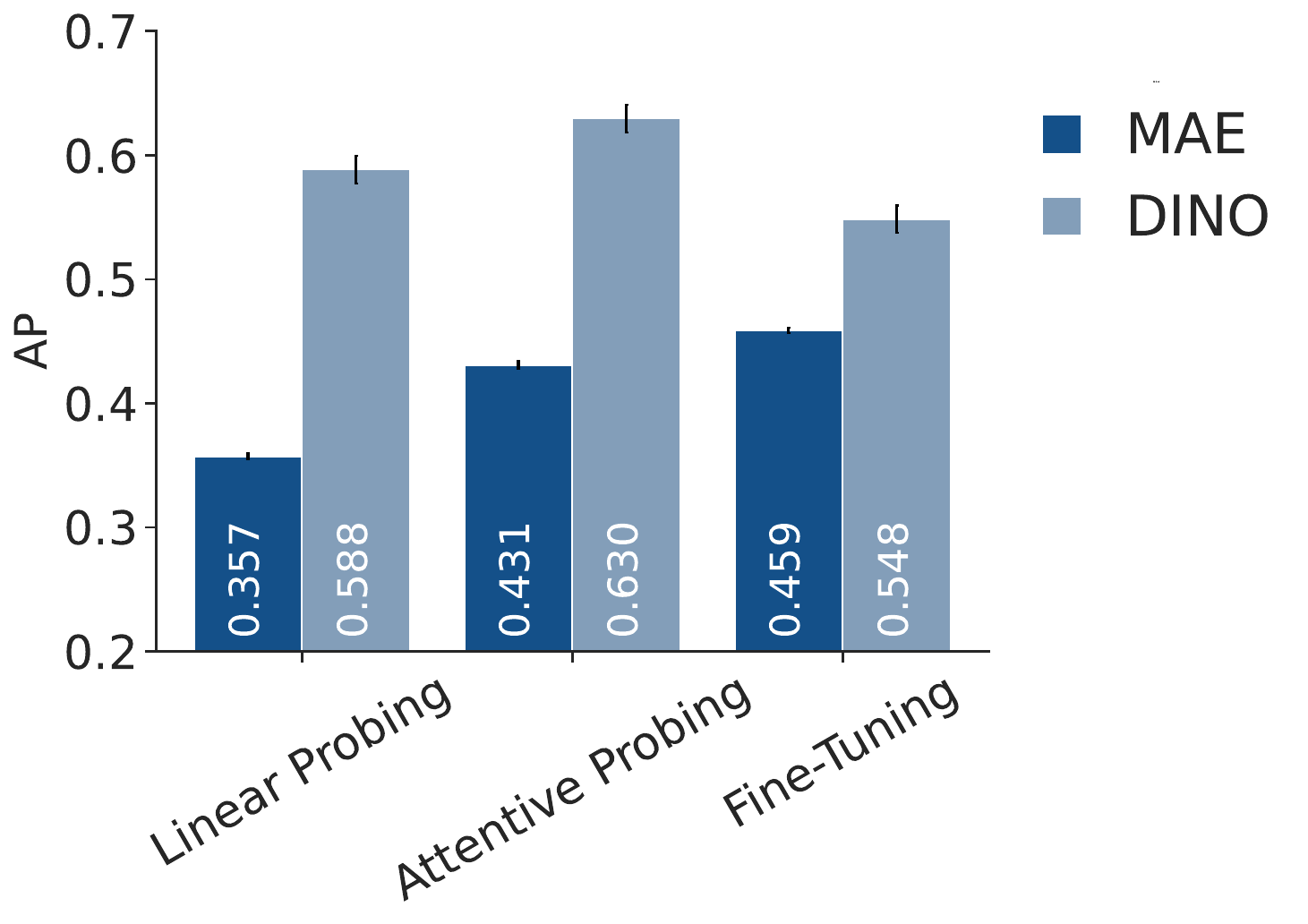}
    \caption{\textbf{Comparison of Different Downstream Training Methods.} This plot illustrates the downstream performance of models evaluated using fine-tuning and various probing methods across four datasets. In most cases, the DINO pre-trained model outperforms the MAE pre-trained model. All models were pre-trained on $100\%$ of the available pretraining data.}
    \label{fig:probing_comparison}
\end{figure*}

\begin{figure}
    \centering
    \textbf{NYU Langone} \\
    \includegraphics[trim={0 0 0 0},clip,height=0.24\textwidth, width=0.38\textwidth]{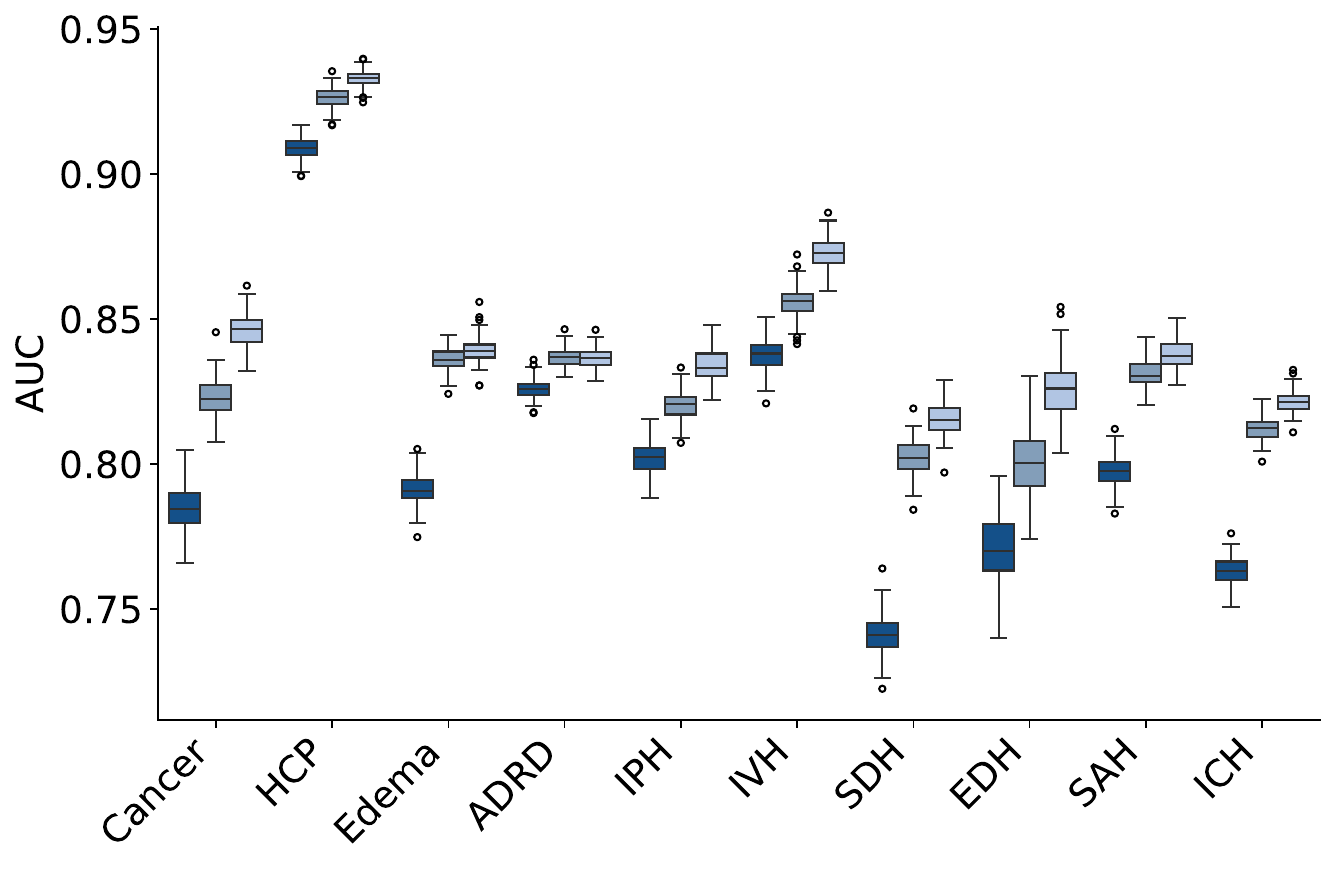}
    \includegraphics[height=0.24\textwidth, width=0.45\textwidth]{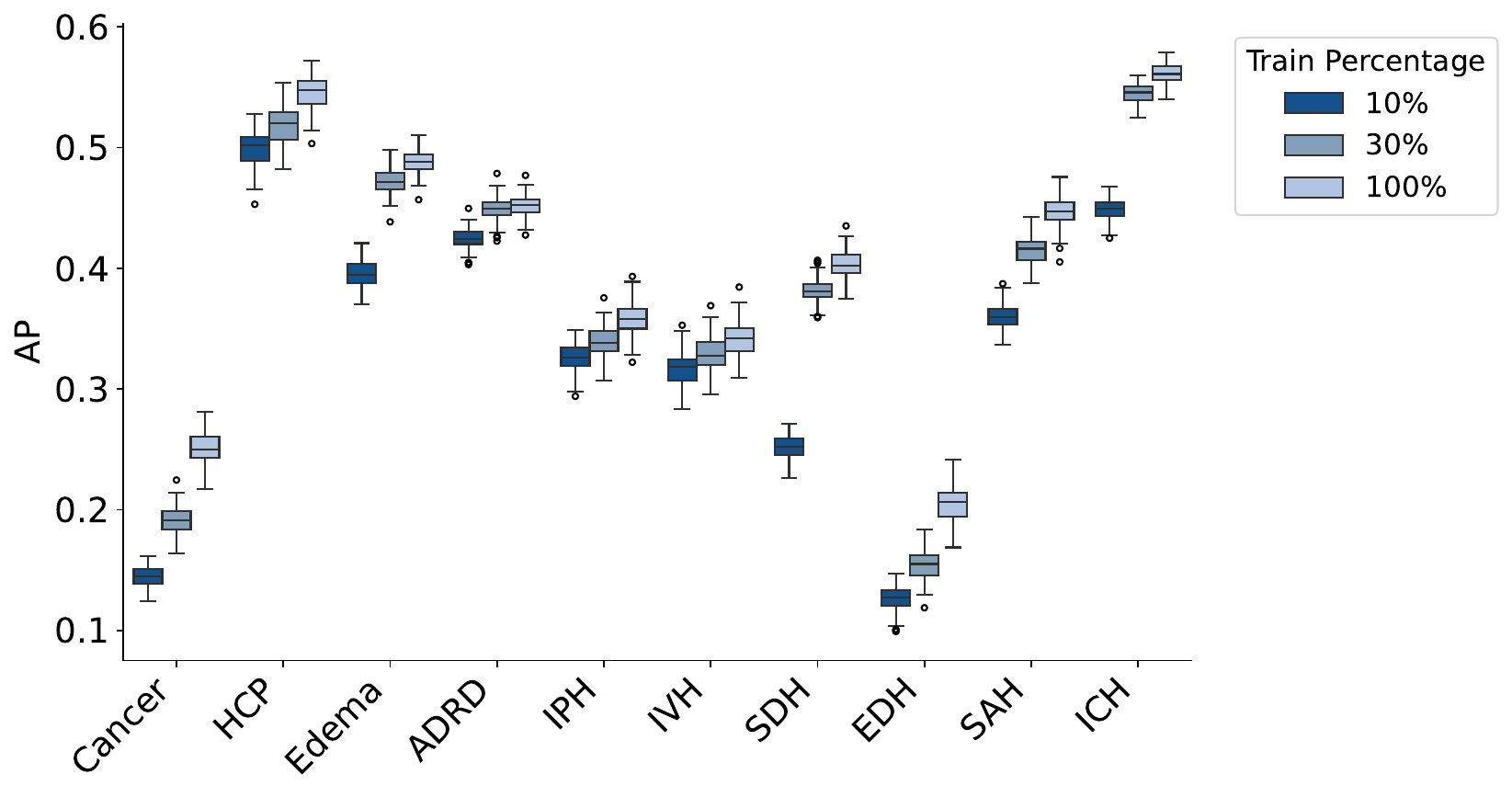} \\
    \textbf{NYU Long Island} \\
    \includegraphics[trim={0 0 0 0},clip,height=0.24\textwidth, width=0.38\textwidth]{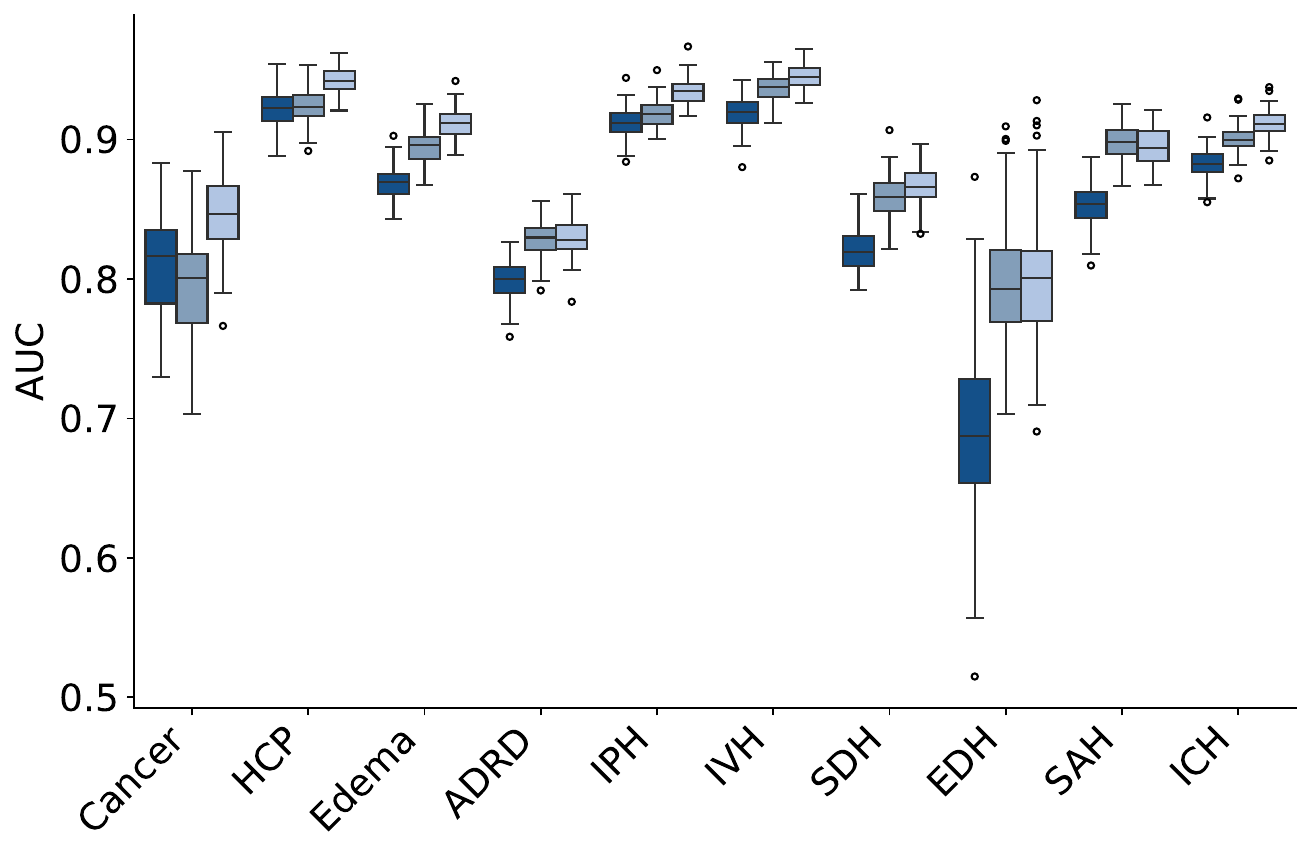}
    \includegraphics[height=0.24\textwidth, width=0.45\textwidth]{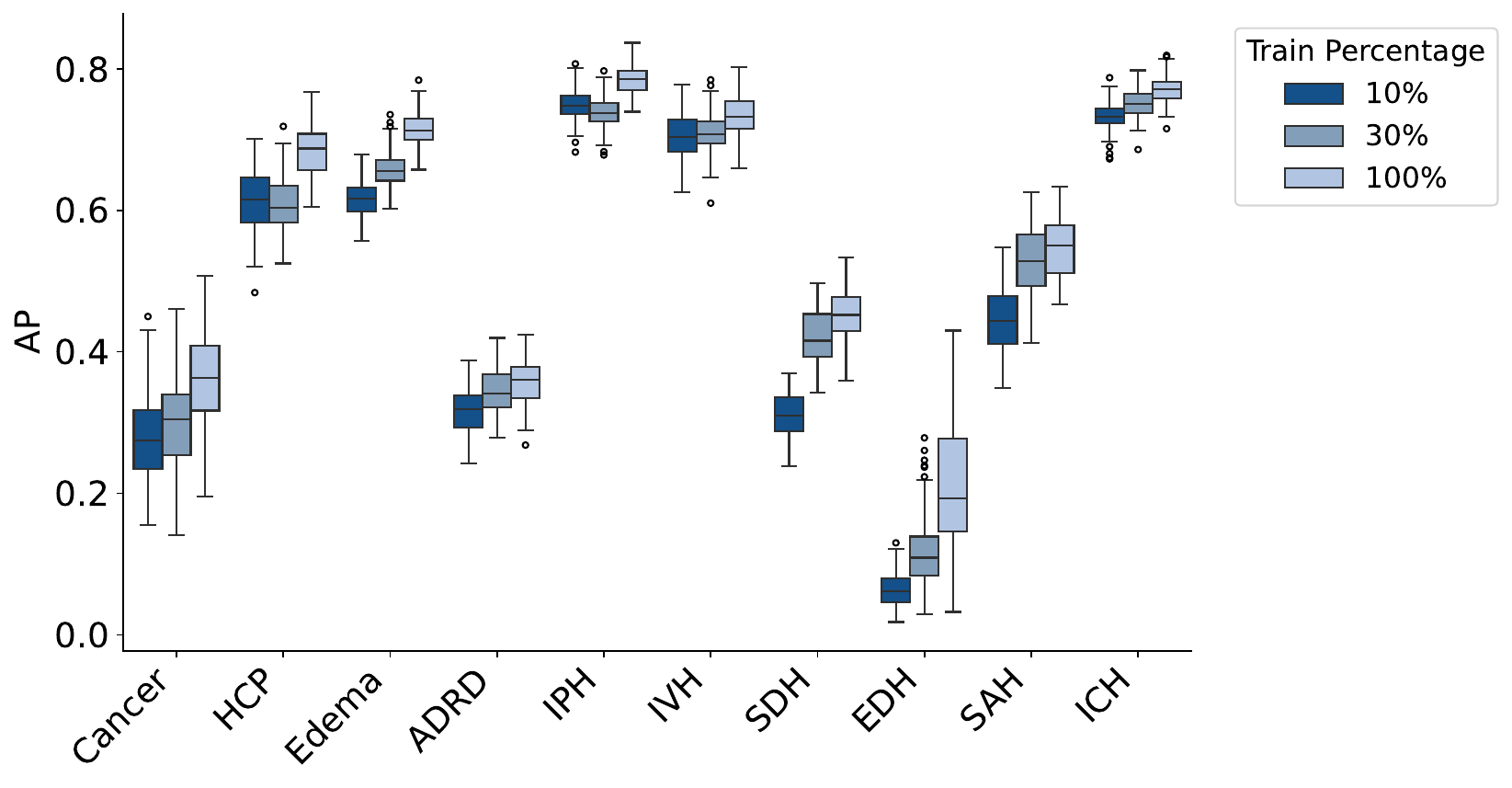} \\
    \textbf{RSNA} \\
    \includegraphics[trim={0 0 0 0},clip,height=0.24\textwidth, width=0.38\textwidth]{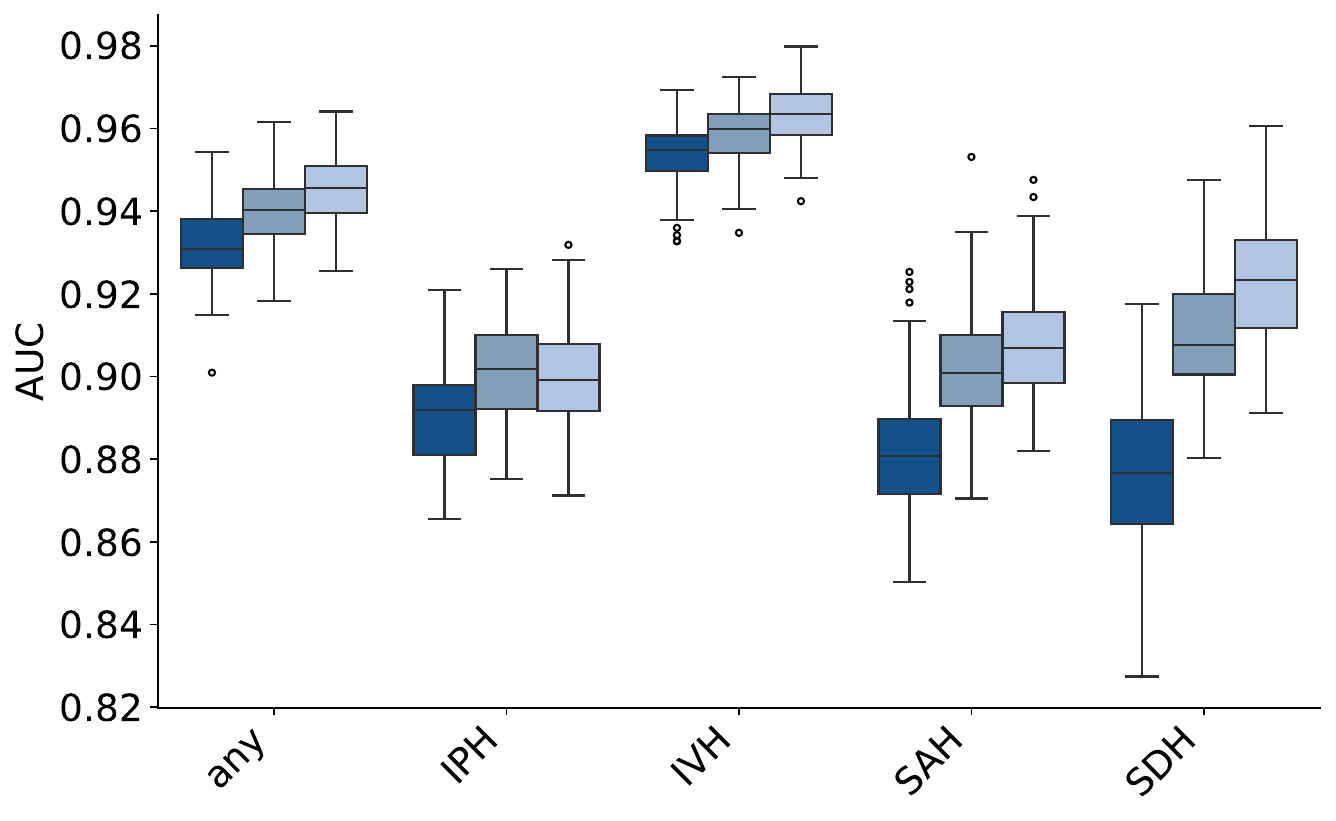}
    \includegraphics[height=0.24\textwidth, width=0.45\textwidth]{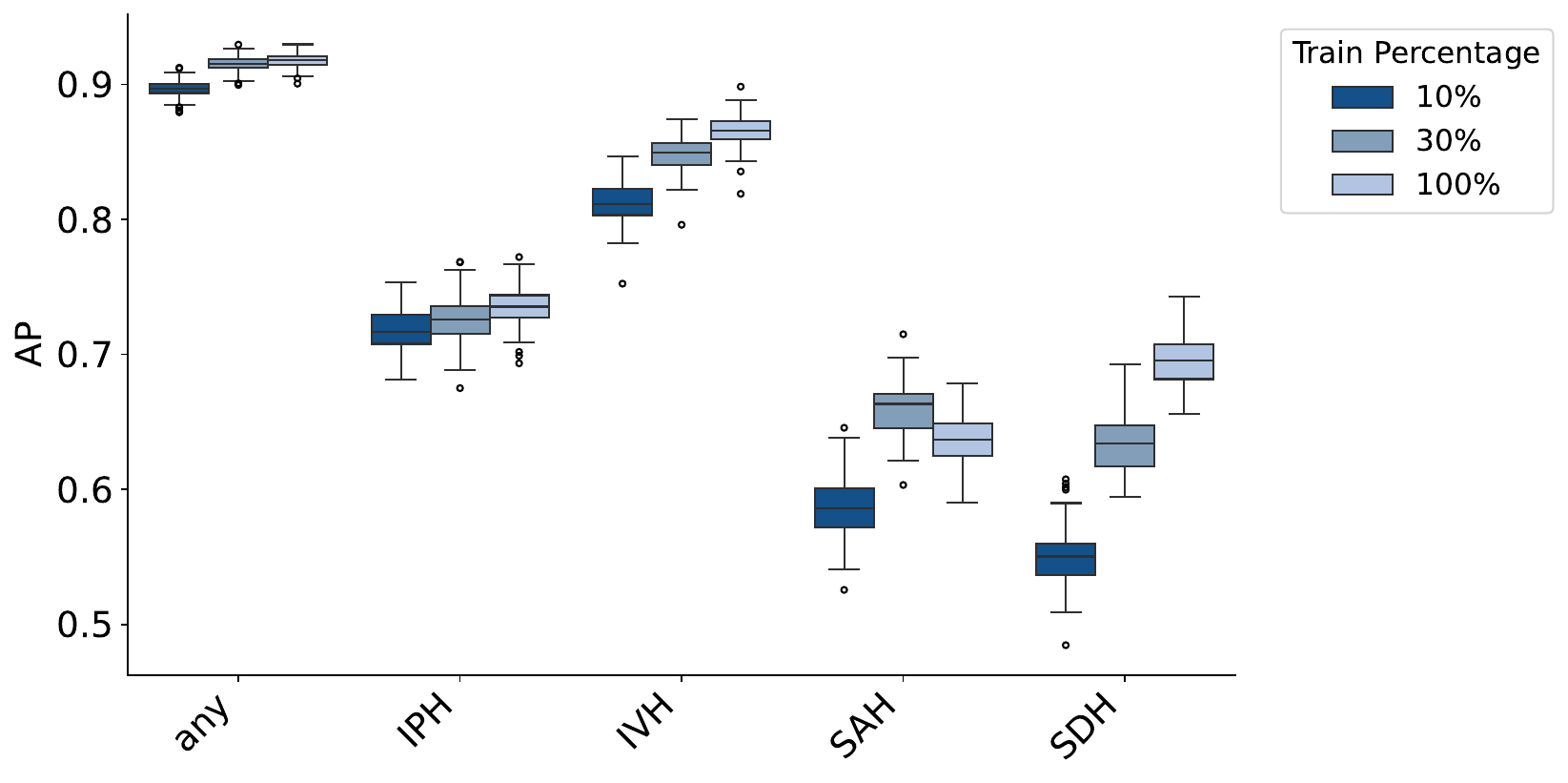}\\
    \textbf{CQ500} \\
    \includegraphics[trim={0 0 0 0},clip,height=0.24\textwidth, width=0.38\textwidth]{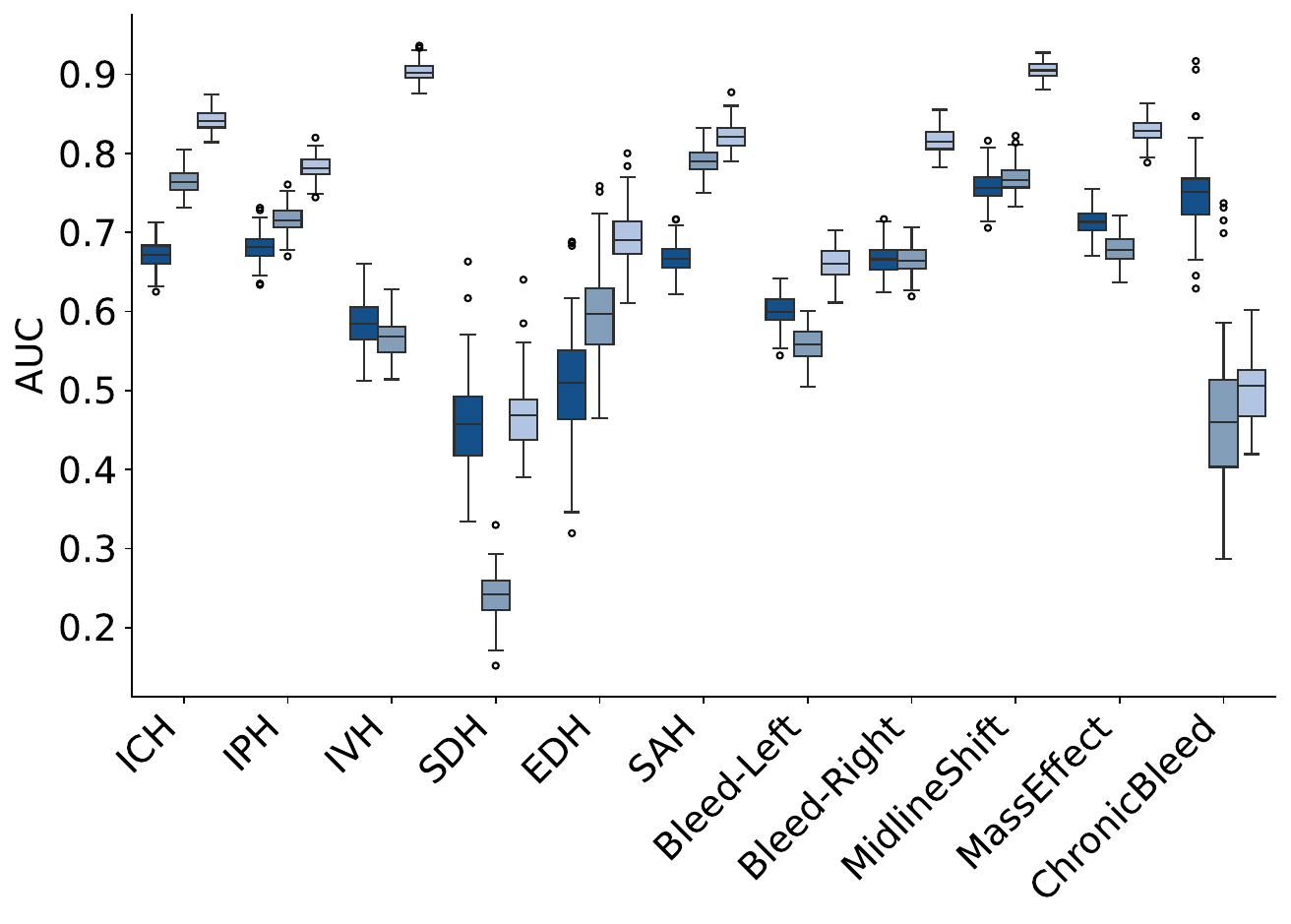}
    \includegraphics[height=0.24\textwidth, width=0.45\textwidth]{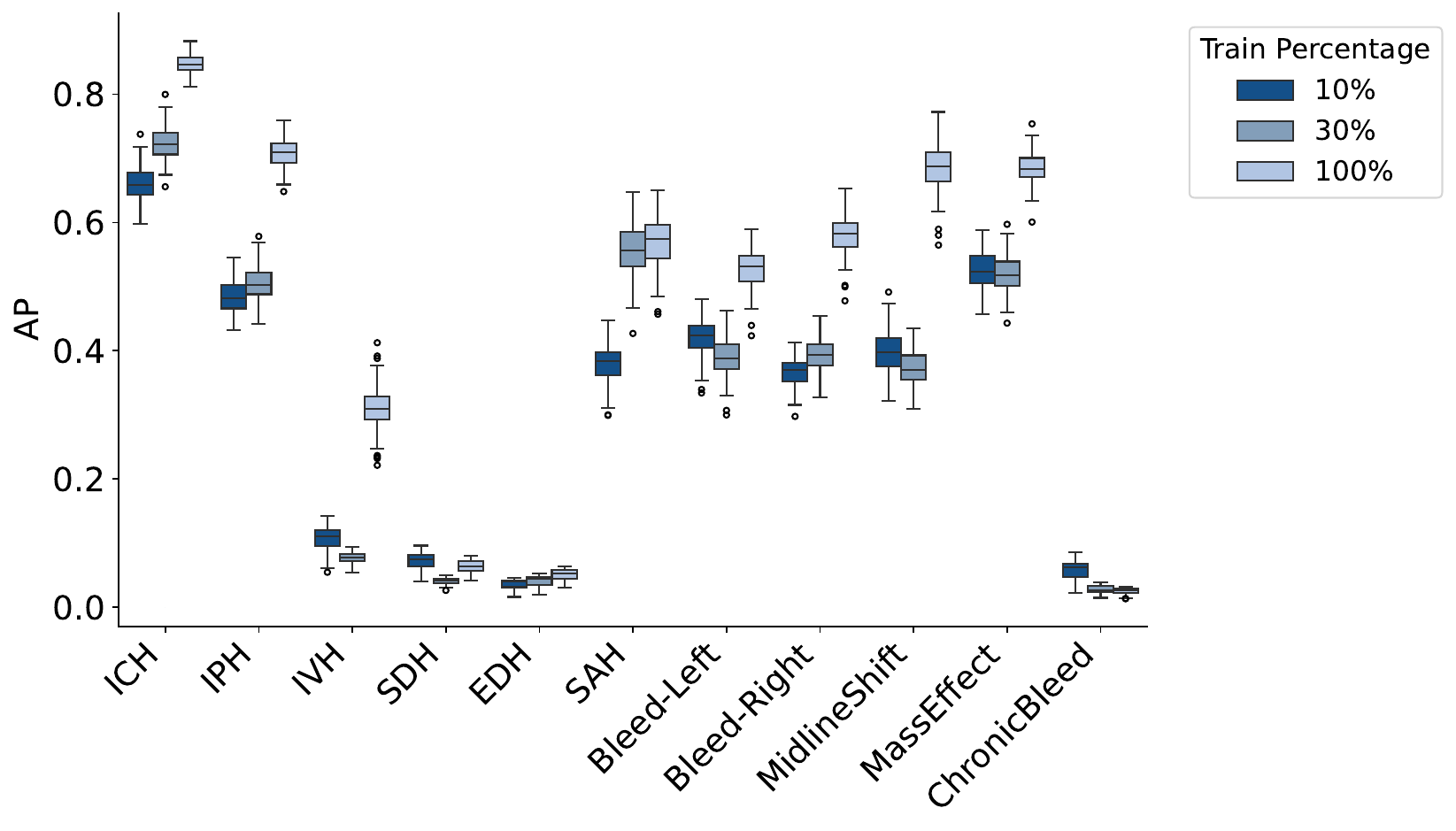}
    \caption{\textbf{Performance for Different Percentage of Pre-training Samples (Per-Pathology).} This plot illustrates label efficiency for individual pathologies using Tukey plots, alongside the average performance across all diseases shown in Figure 1 from main paper. The results indicate that the majority of pathologies show improved downstream performance as the amount of pretraining data increases.}
    \label{fig:boxplot_scaling}
\end{figure}

Supplementary \Cref{fig:probing_comparison} compares the performance on downstream tasks with various supervised tuning methods applied to foundation models pretrained with the MAE and DINO frameworks. Per-pathology comparisons are shown in Supplementary \Cref{fig:probing-comparison-perpath,fig:probing-comparison-perpath-dino}. Meanwhile, supplementary \Cref{fig:boxplot_scaling} complements \Cref{fig:scaling_law} in the main article, illustrating the per-pathology performances of foundation models pretrained with different scales of training data.

\begin{figure}[ht]
    \centering
    \includegraphics[width=\textwidth]{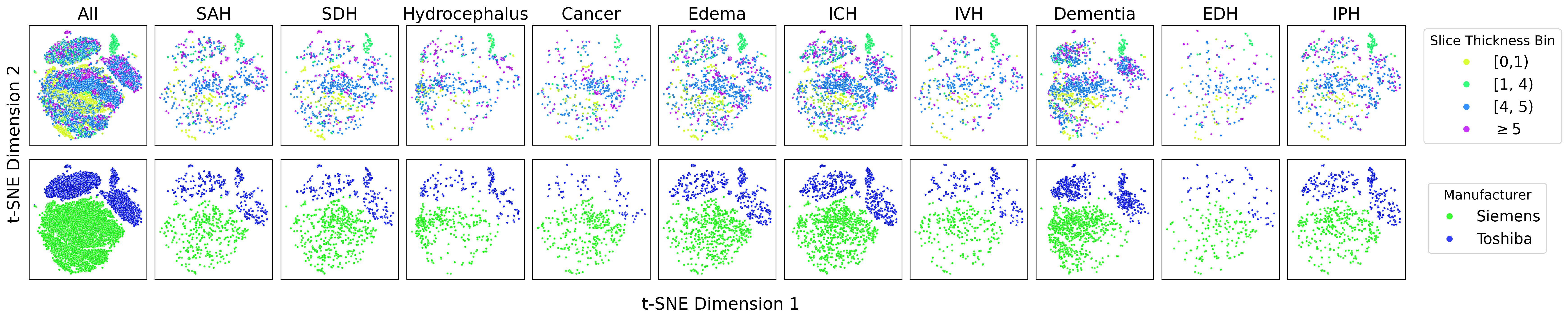}
    \caption{The 2D projection with t-SNE of CT volume representation extracted from the foundation model. Interestingly, certain subgroups still exhibited slightly better AUCs. For instance, scans with slice thicknesses between 1–4 mm (represented by light green points in the upper panel of \Cref{fig:batch_effect}) align with a specialized protocol for CT angiography (CTA), which uses contrast dye to improve diagnosis on particular diseases.}
    \label{fig:batch_effect}
\end{figure}

\begin{figure*}[ht]
    \centering
    \begin{subfigure}[b]{0.33\textwidth}
        \centering
        \includegraphics[width=\textwidth]{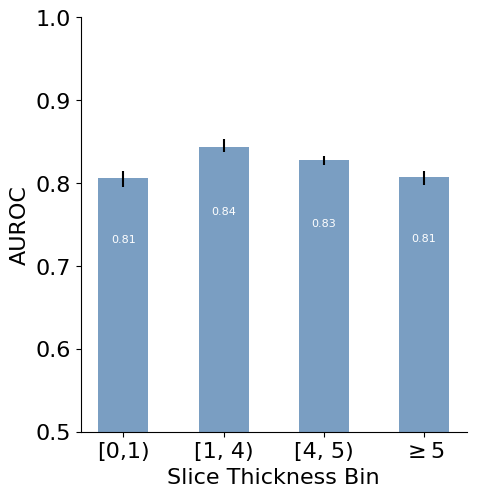}
        \caption{AUROC Performance}
    \end{subfigure}
    \hfill
    \begin{subfigure}[b]{0.33\textwidth}
        \centering
        \includegraphics[width=\textwidth]{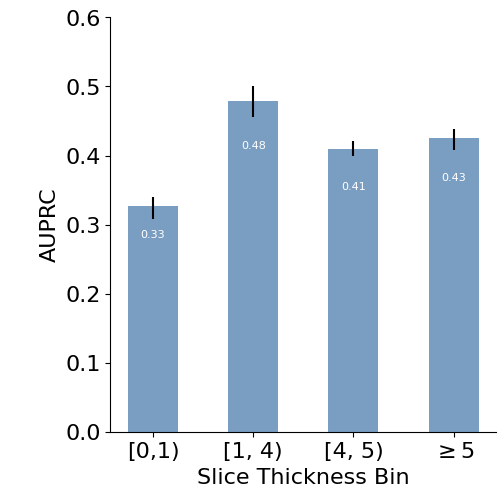}
        \caption{AUPRC Performance}
    \end{subfigure}
    \hfill
    \begin{subfigure}[b]{0.33\textwidth}
        \centering
        \includegraphics[width=\textwidth]{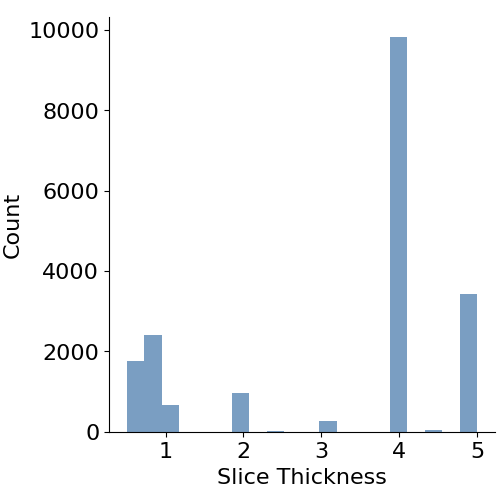}
        \caption{Histogram of slice thickness distribution}
    \end{subfigure}
    \caption{The downstream task performances on various ranges of slice thickness.}
    \label{fig:thickness-ablation}
\end{figure*}

Supplementary \Cref{fig:batch_effect,fig:thickness-ablation} studies the impact of batch effect caused by different CT scan protocols of slice thickness and machine manufacturer. Detailed per-pathology performances are shown in Supplementary \Cref{fig:slice_thickness_per_pathology,fig:manufacturer_per_pathology}.

\vspace{50pt}

\begin{figure}
    \centering
    \makebox[\textwidth][l]{%
        \hspace{0.35\textwidth}\textbf{NYU Langone}
    } \\[0.2cm]
    \includegraphics[trim={0 0 0 0},clip,height=0.25\textwidth]{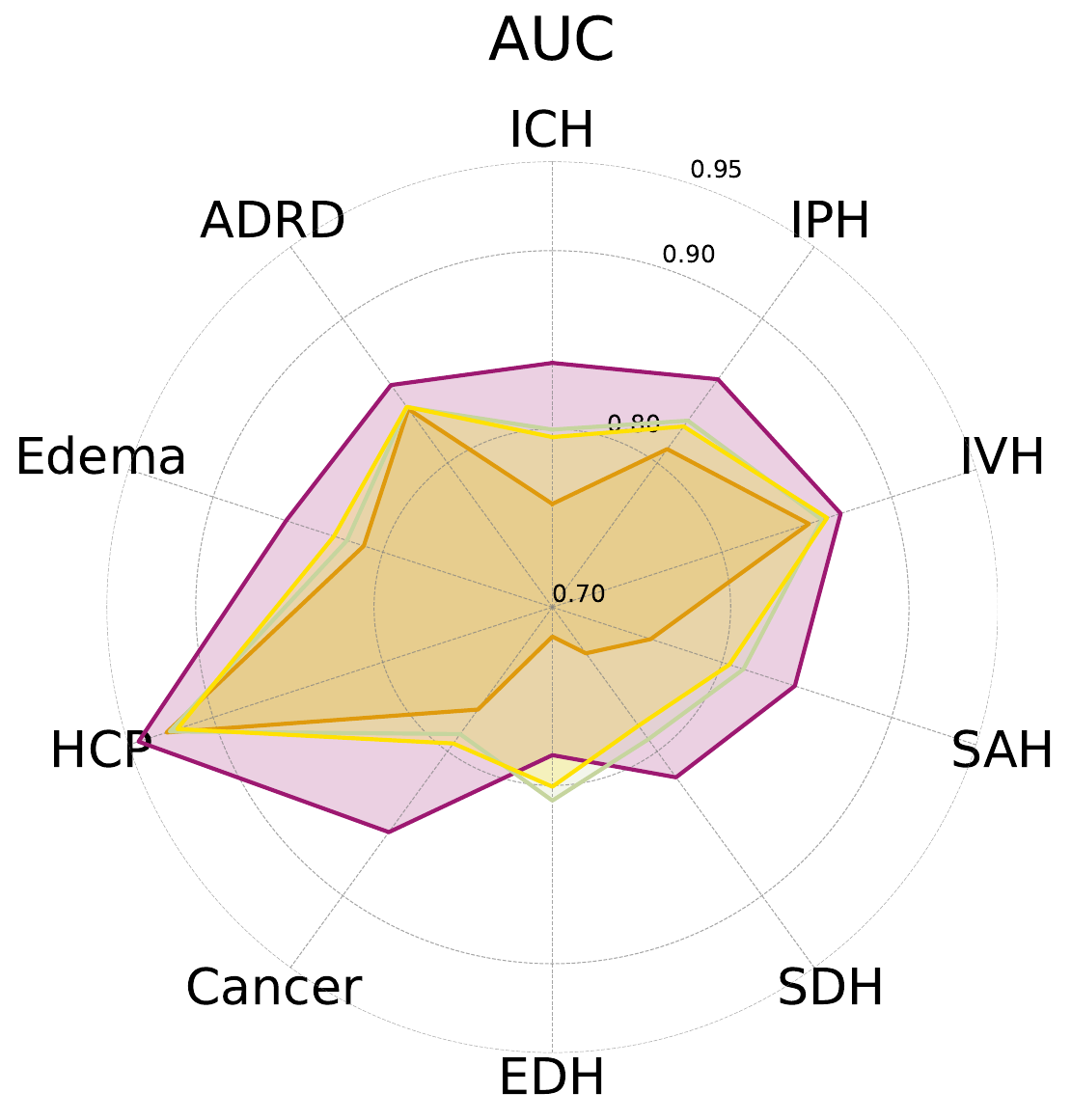}
    \includegraphics[trim={0 0 0 0},clip,height=0.25\textwidth]{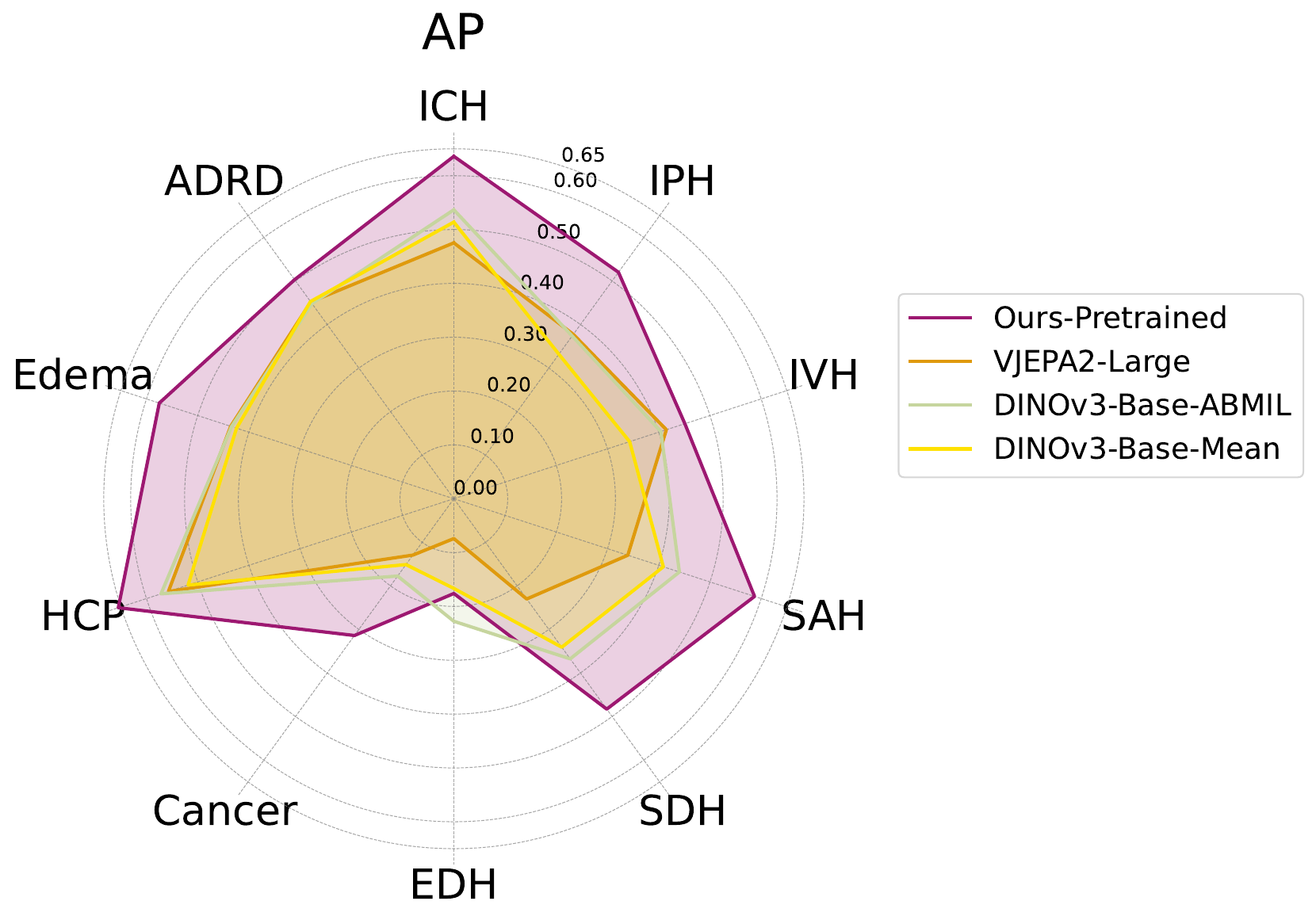} \\
    \makebox[\textwidth][l]{
        \hspace{0.35\textwidth}\textbf{NYU Long Island}
    } \\[0.2cm]
    \includegraphics[trim={0 0 0 0},clip,height=0.25\textwidth]{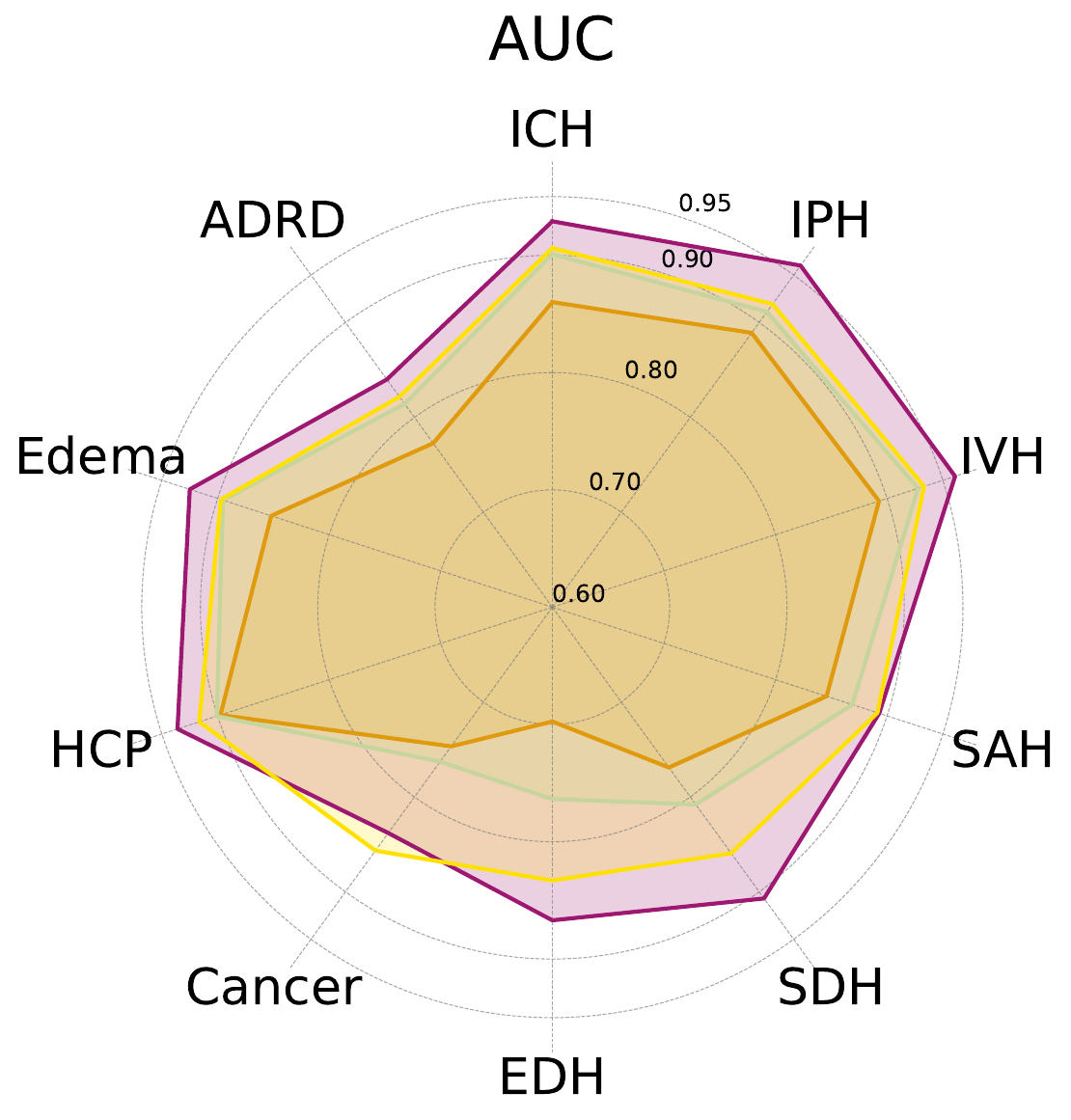}
    \includegraphics[trim={0 0 0 0},clip,height=0.25\textwidth]{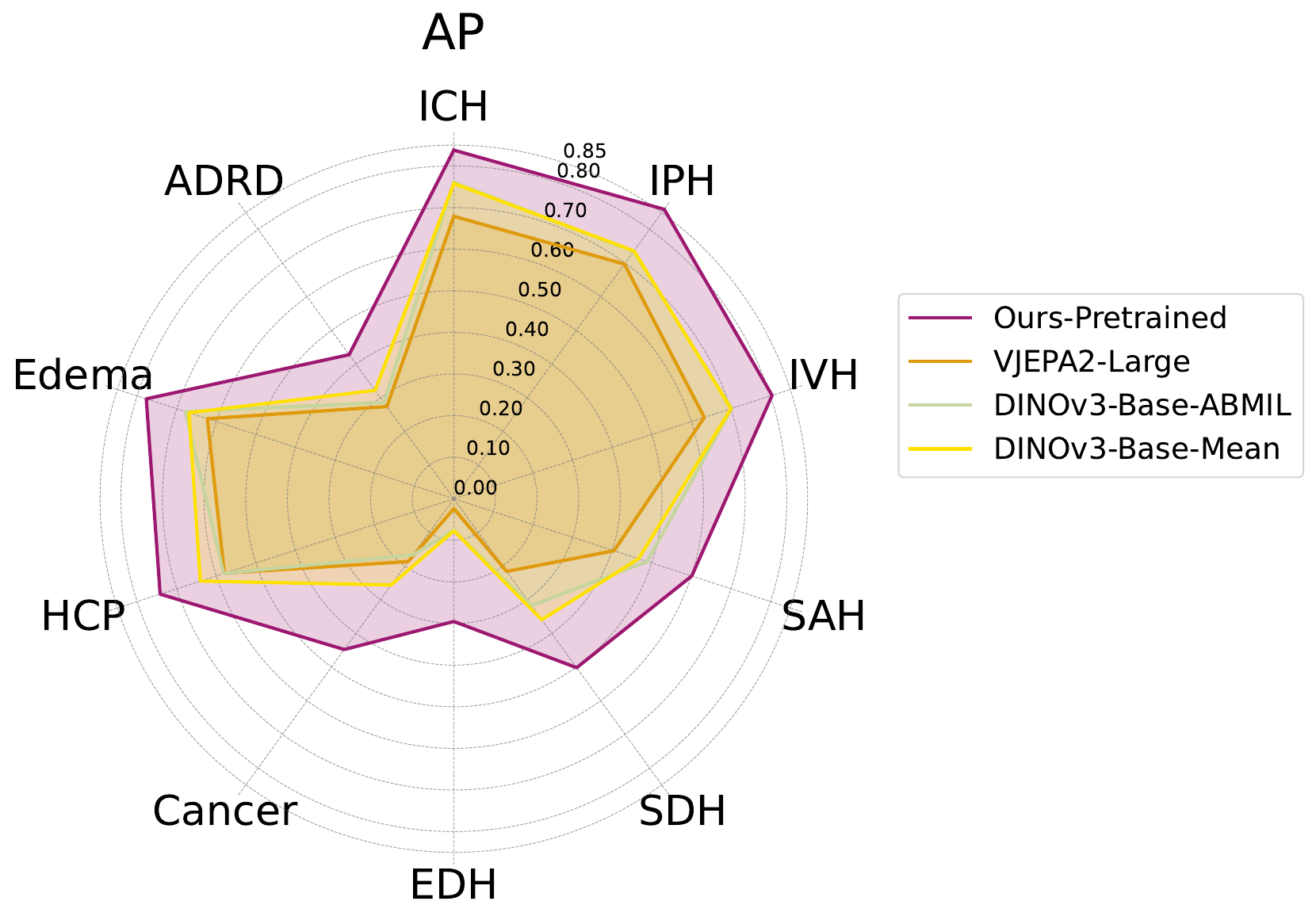} 
    \makebox[\textwidth][l]{
        \hspace{0.4\textwidth}\textbf{RSNA}
    } \\[0.2cm]
    \includegraphics[trim={0 0 0 0},clip,height=0.25\textwidth]{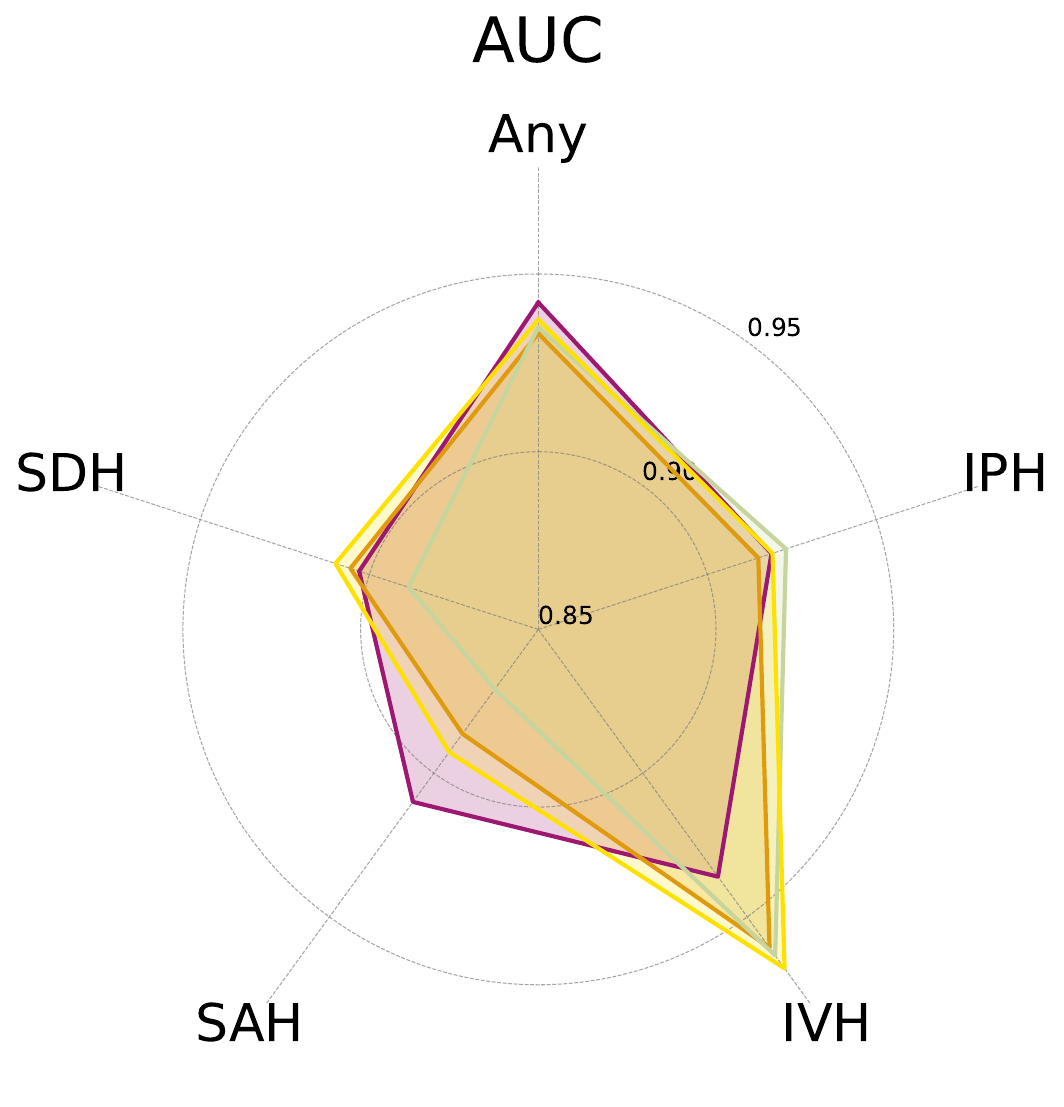}
    \hspace{5mm}
    \includegraphics[trim={0 0 0 0},clip,height=0.25\textwidth]{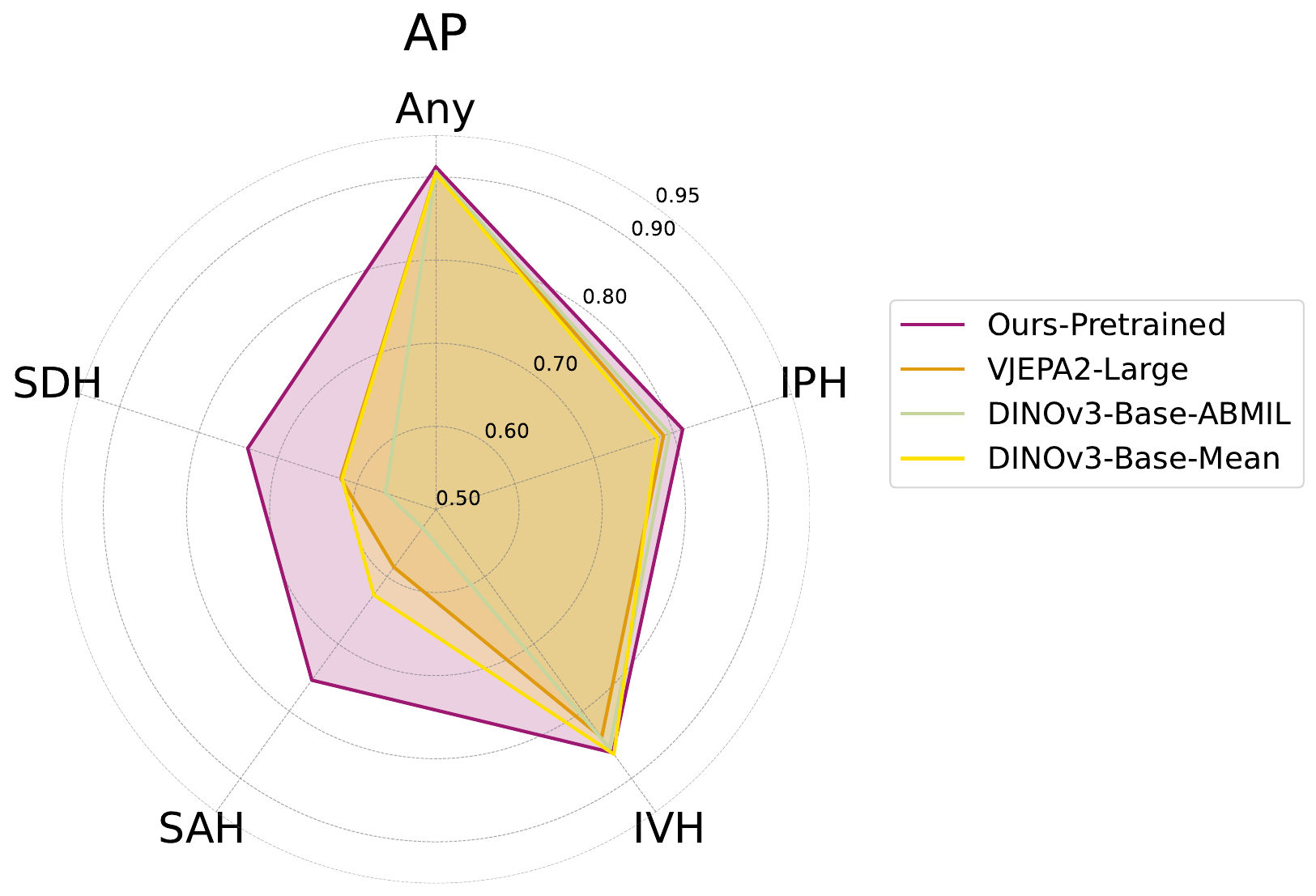} 
    \makebox[\textwidth][l]{
        \hspace{0.4\textwidth}\textbf{CQ500}
    } \\[0.2cm]
    \includegraphics[trim={0 0 0 0},clip,height=0.25\textwidth]{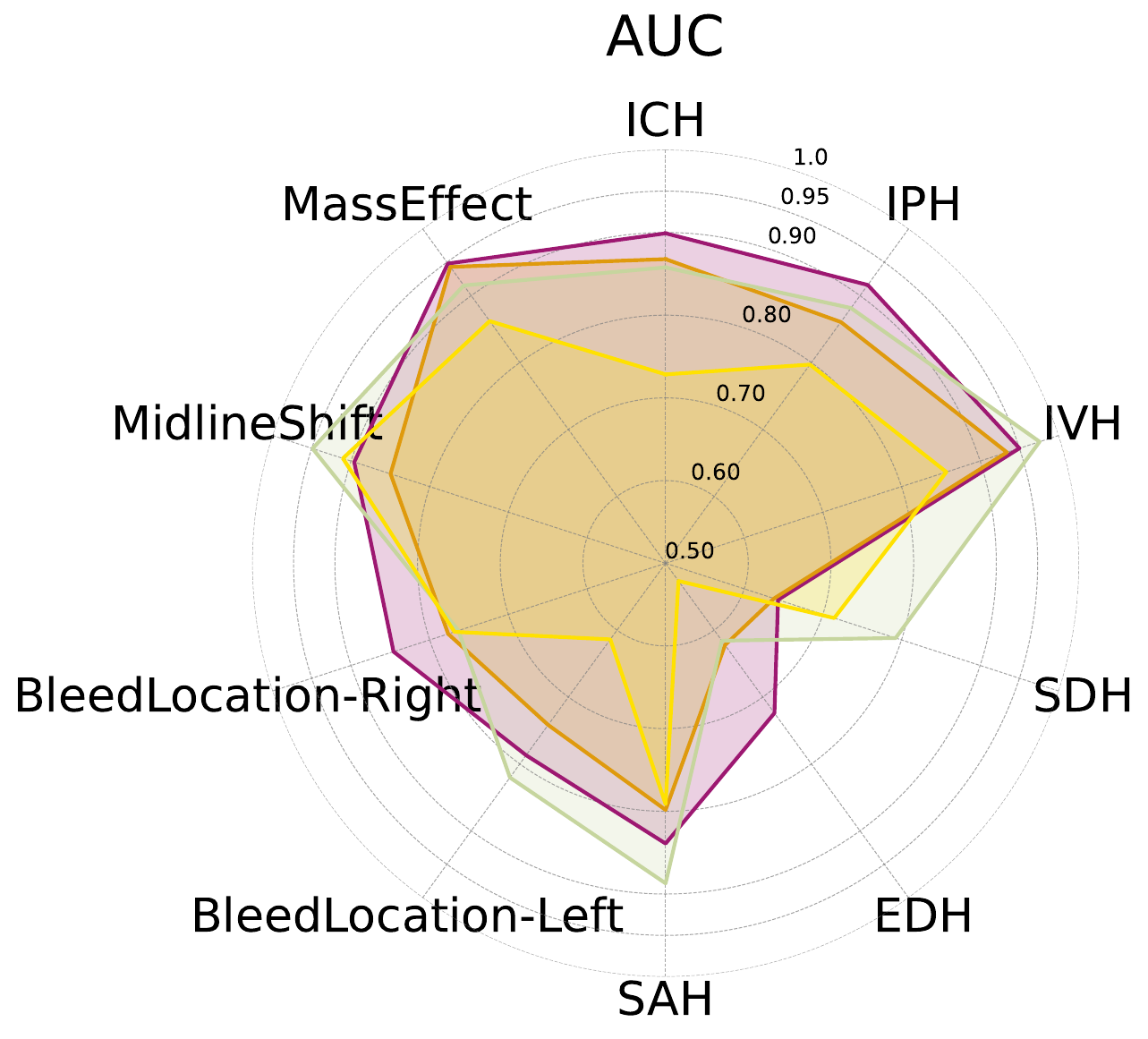} \hspace{5mm}
    \includegraphics[trim={0 0 0 0},clip,height=0.25\textwidth]{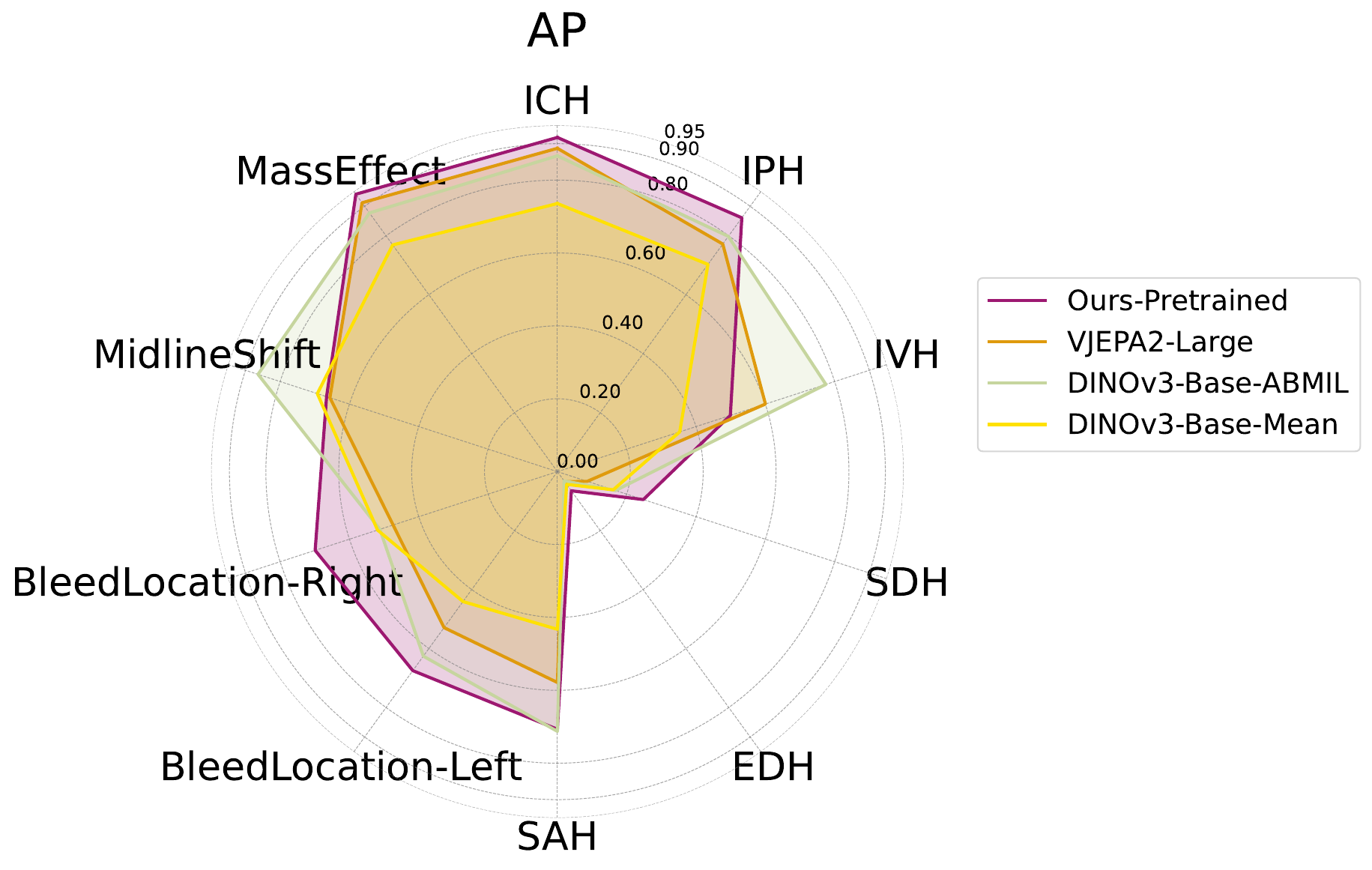} 
    \caption{\textbf{Performance comparison with 2D and video foundation model} This plot shows model performance comparison on our model against MIL using 2D foundation model DINOv3~\cite{siméoni2025dinov3} with attention based multiple-instance learning \cite{pmlr-v80-ilse18a} (simplied as MIL in the plot) and video foundation VJEPA2~\cite{assran2025vjepa2selfsupervisedvideo}. All comparisons are conducted on same image resolution (96$\times$96$\times$96) as DINOv3 and VJEPA2 both support dynamic resolutions. Comparing to these two methods, our model shows consistently improvement on AUROC and AP across majority of tasks with significant advantages on inference speed and memory cost (as shown in Supplementary \Cref{fig:model_efficiency})}
    \label{fig:video_model_comparison}
\end{figure}

\begin{figure}[ht]
    \centering
    \includegraphics[width=0.7\textwidth]{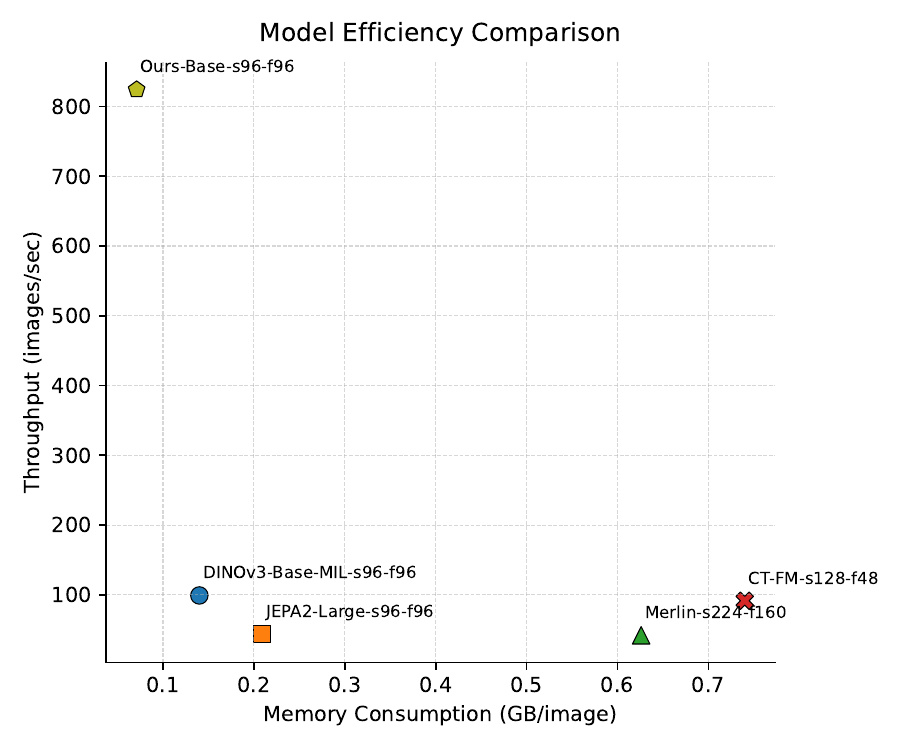}
    \caption{\textbf{Throughput vs. Memory Cost Analysis:} in this plot, we perform model efficiency analysis on throughput and memory cost for each of benchmarked models in our study (all models code are taken from official code repository). We use video model terms to standardize input shape naming of each model in this plot where $s$ means height and weight and $f$ means depth (frames for video). The analysis shows that our model with vanilla 3D ViT demonstrates significant advantage in terms of both throughput and memory cost.}
    \label{fig:model_efficiency}
\end{figure}

\begin{figure}
    \centering
    \makebox[\textwidth][l]{%
        \hspace{0.35\textwidth}\textbf{NYU Langone}
    } \\[0.2cm]
    \includegraphics[trim={0 0 120mm 0},clip,height=0.255\textwidth]{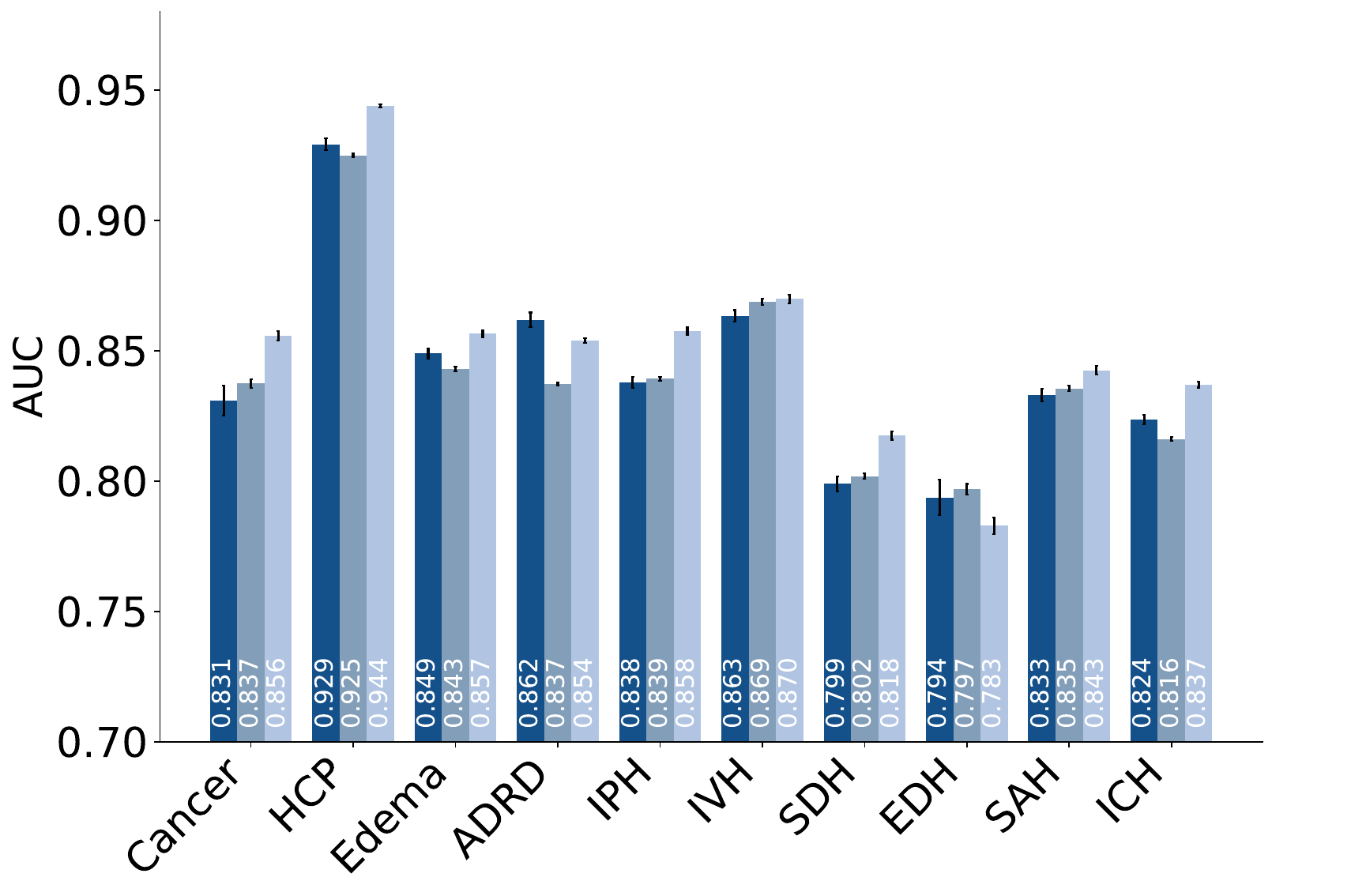}
    \includegraphics[trim={0 0 0 0},clip,height=0.255\textwidth]{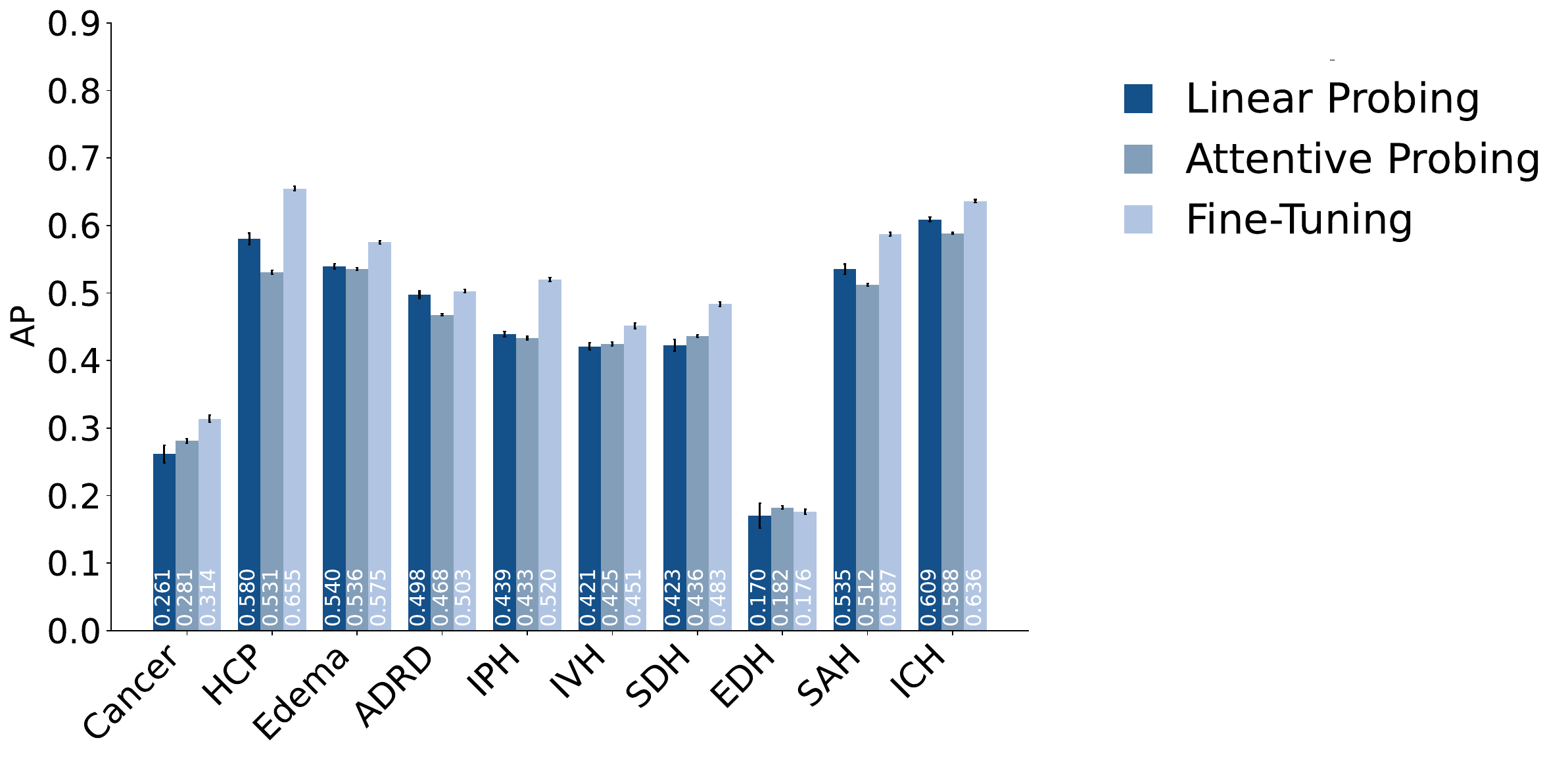} \\
    \makebox[\textwidth][l]{
        \hspace{0.35\textwidth}\textbf{NYU Long Island}
    } \\[0.2cm]
    \includegraphics[trim={0 0 120mm 0},clip,height=0.255\textwidth]{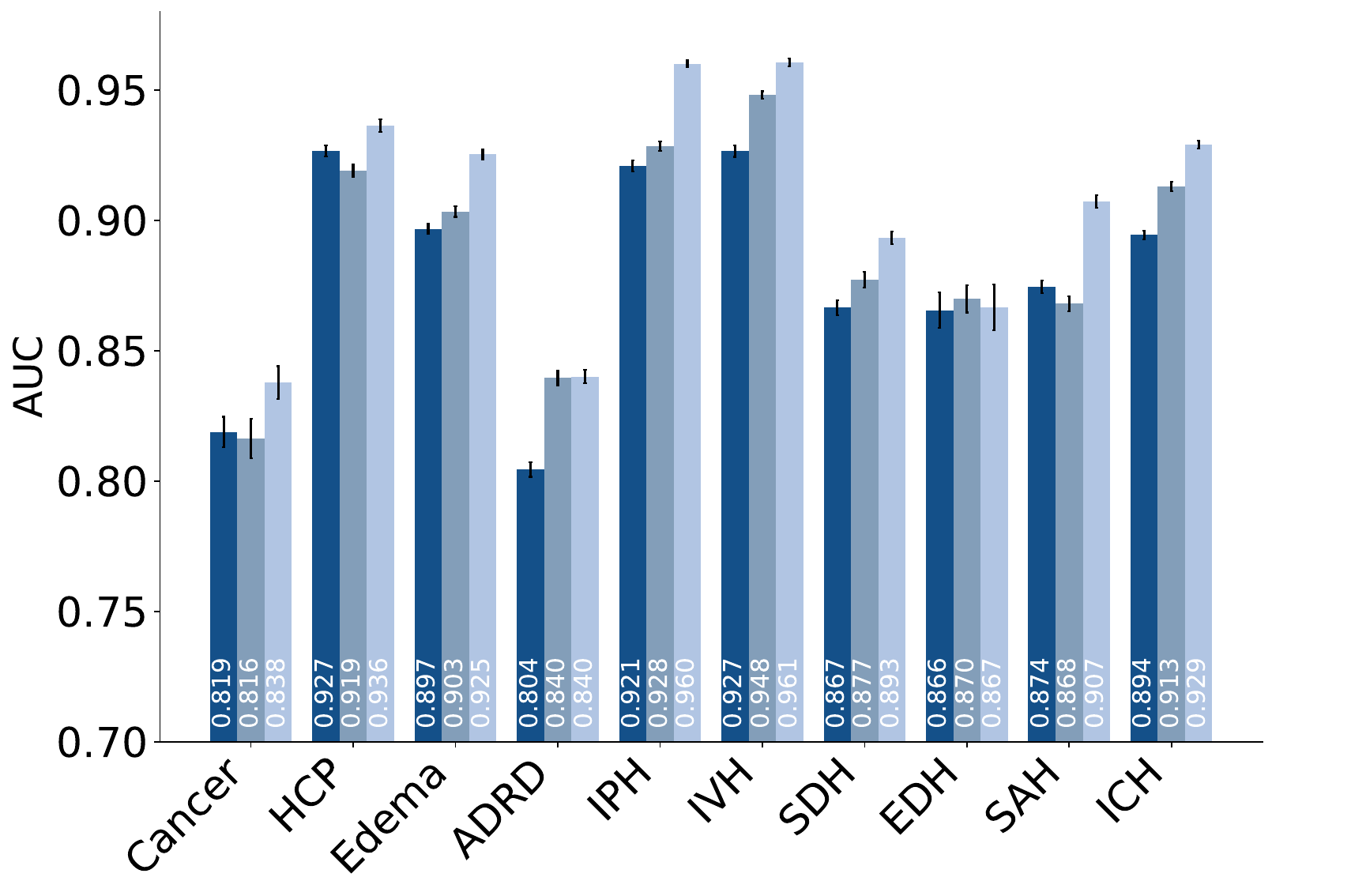}
    \includegraphics[trim={0 0 0 0},clip,height=0.255\textwidth]{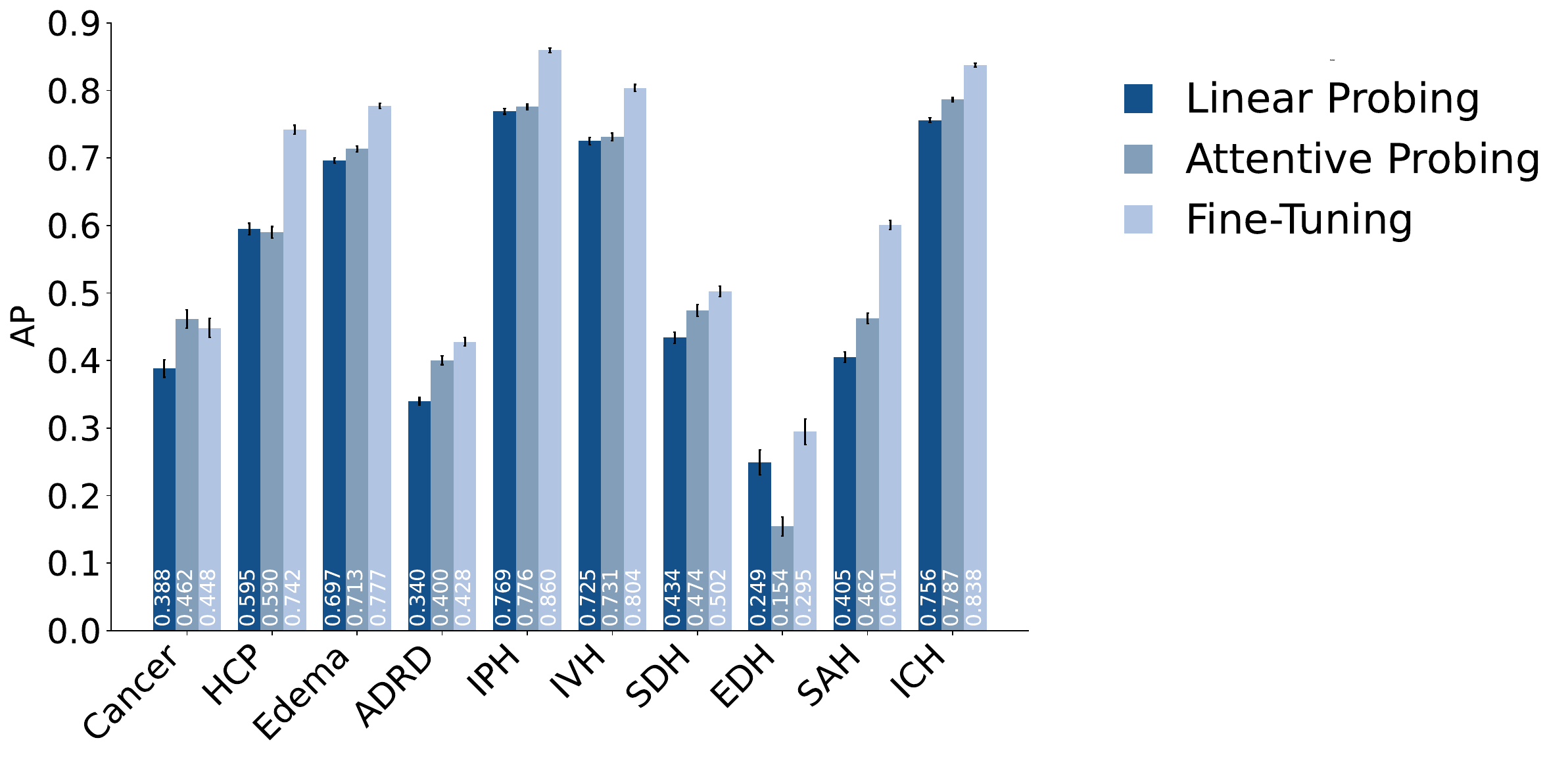} 
    \makebox[\textwidth][l]{
        \hspace{0.4\textwidth}\textbf{RSNA}
    } \\[0.2cm]
    \includegraphics[trim={0 0 120mm 0},clip,height=0.24\textwidth]{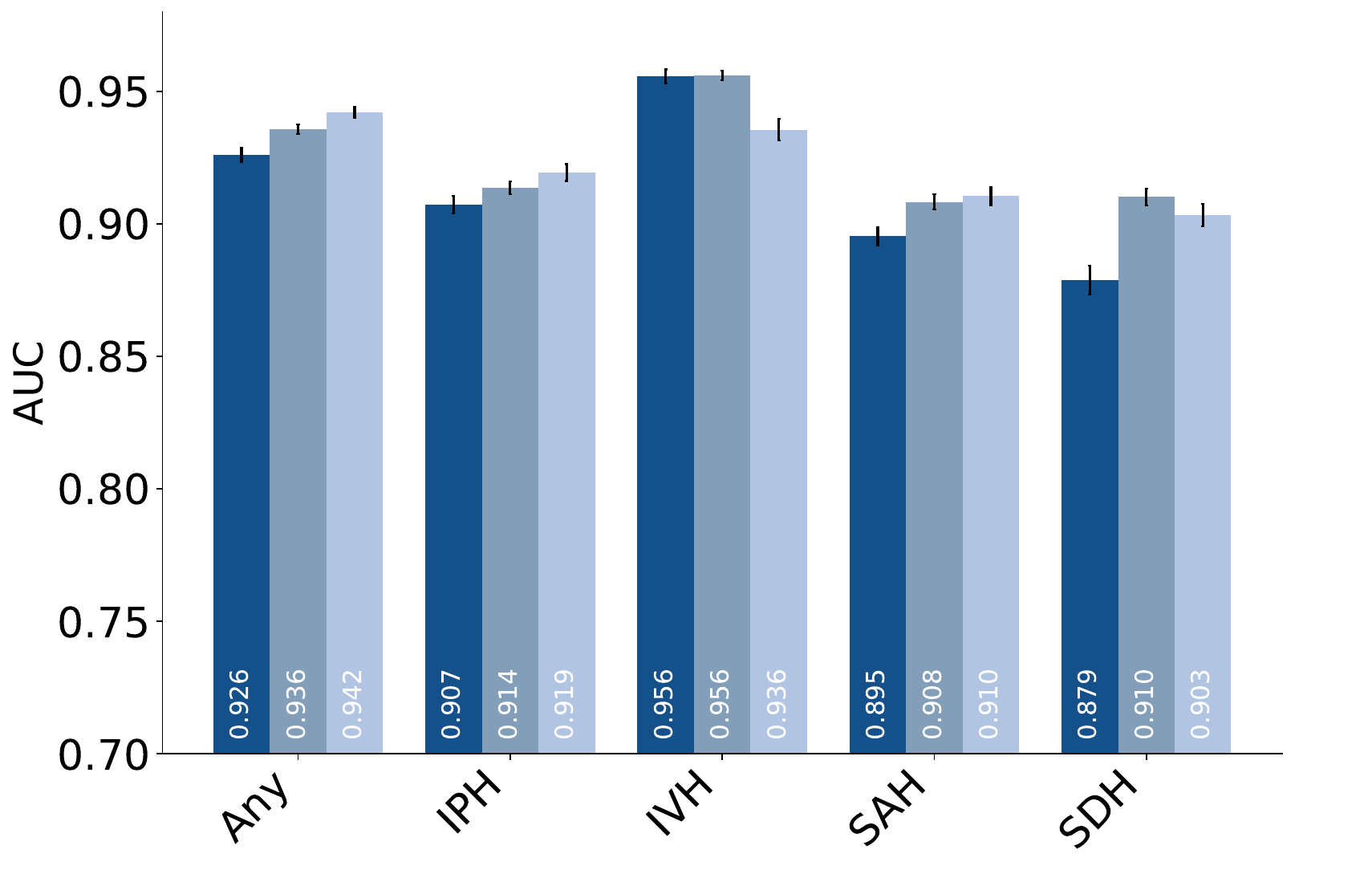}
    \hspace{5mm}
    \includegraphics[trim={0 0 0 0},clip,height=0.24\textwidth]{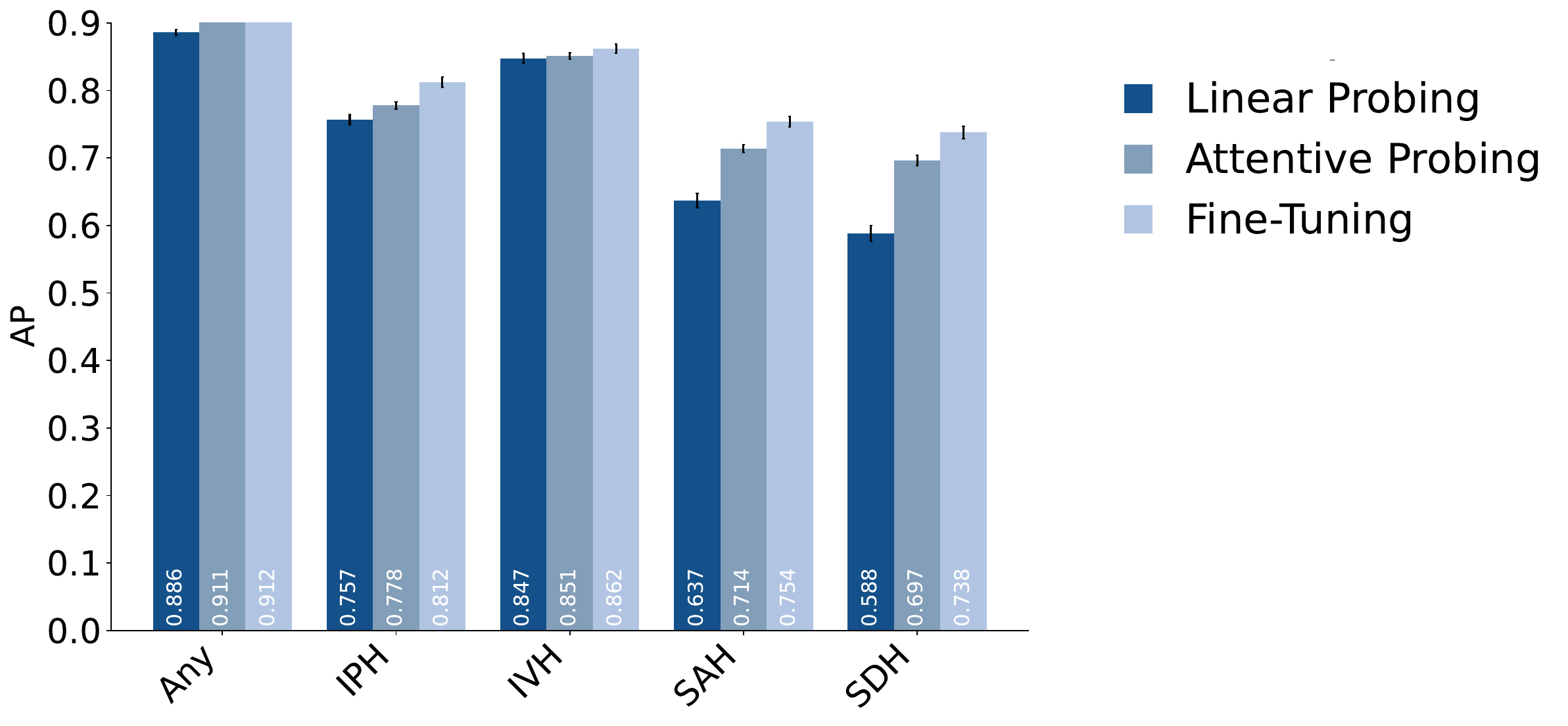} 
    \makebox[\textwidth][l]{
        \hspace{0.4\textwidth}\textbf{CQ500}
    } \\[0.2cm]
    \includegraphics[trim={0 0 120mm 0},clip,height=0.30\textwidth]{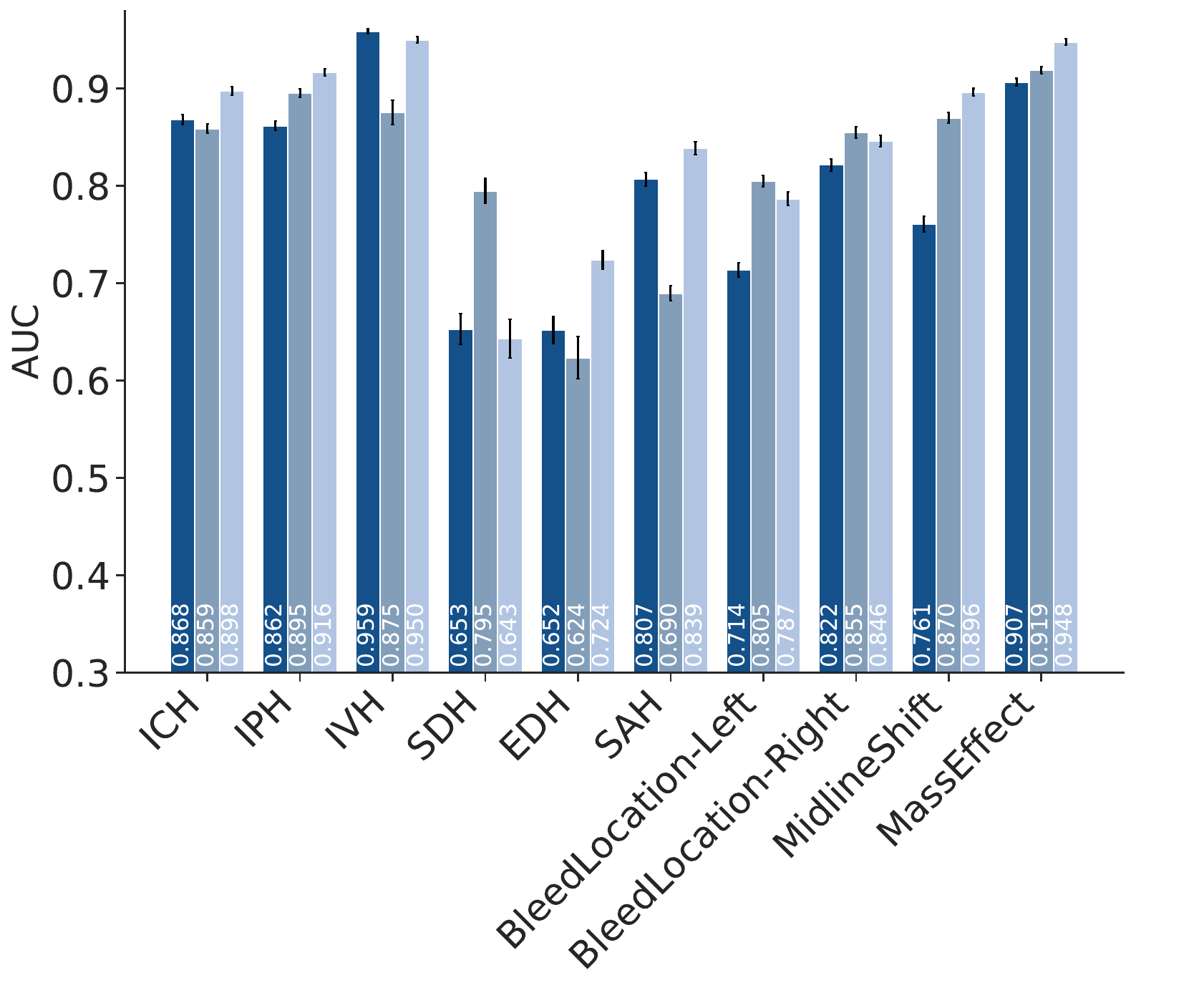} \hspace{5mm}
    \includegraphics[trim={0 0 0 0},clip,height=0.30\textwidth]{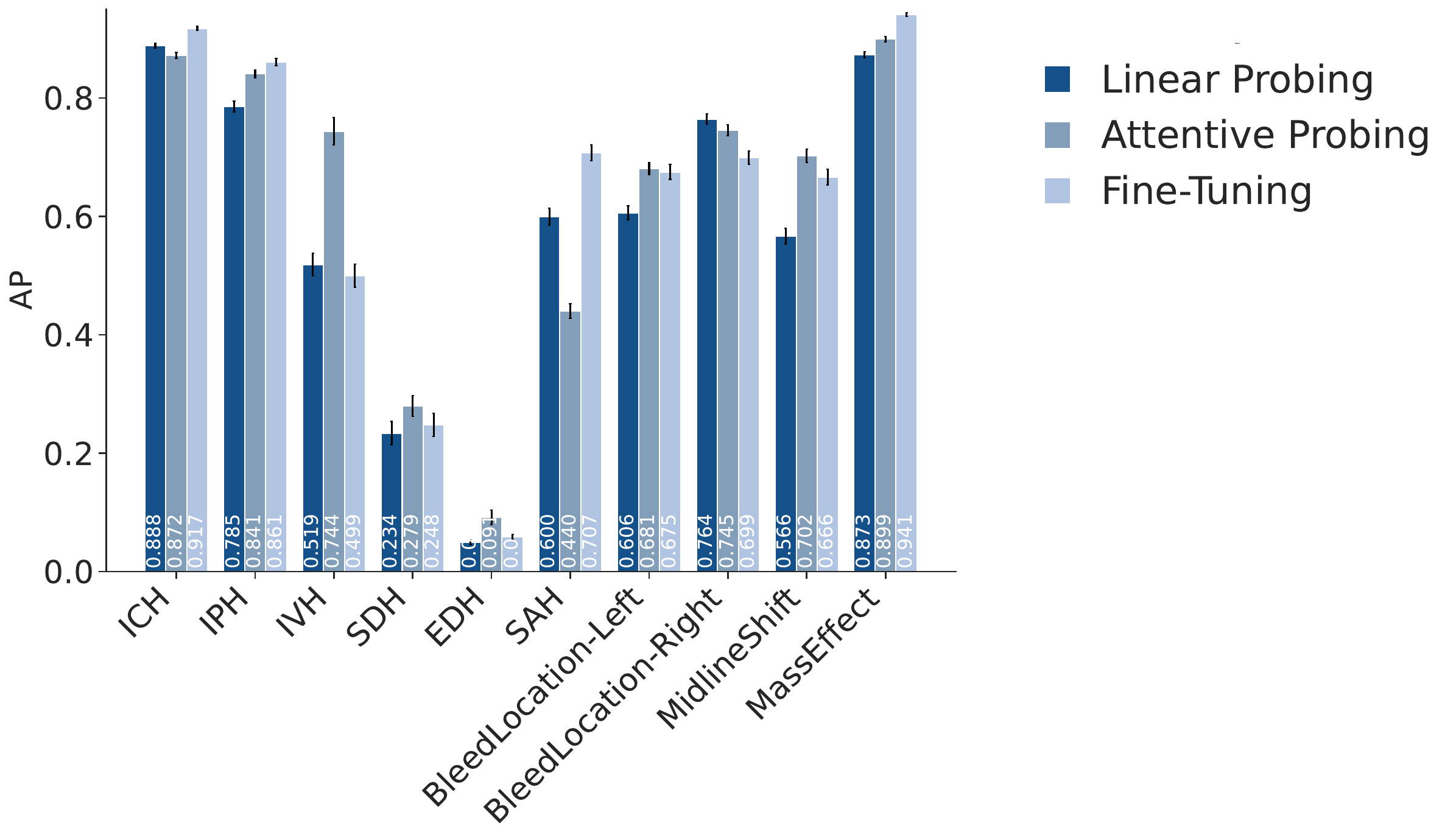} 
    \caption{\textbf{Performance comparison of supervised finetuning methods per pathology on the foundation model trained with DINO.} This plot breaks down the average performance across all diseases shown in Supplementary \Cref{fig:probing_comparison}. The results show that fine-tuning the entire network achieves the best performance in most scenarios. However, linear probing closely approaches finetuning performance for many diseases especially on small or imbalanced dataset, underscoring the capability of our pre-trained models to generate representations that adapt effectively to diverse disease detection tasks.}
    \label{fig:probing-comparison-perpath-dino}
\end{figure}

\begin{figure}
    \centering
    \makebox[\textwidth][l]{%
        \hspace{0.35\textwidth}\textbf{NYU Langone}
    } \\[0.2cm]
    \includegraphics[trim={0 0 0 0},clip,height=0.24\textwidth, width=0.3\textwidth]{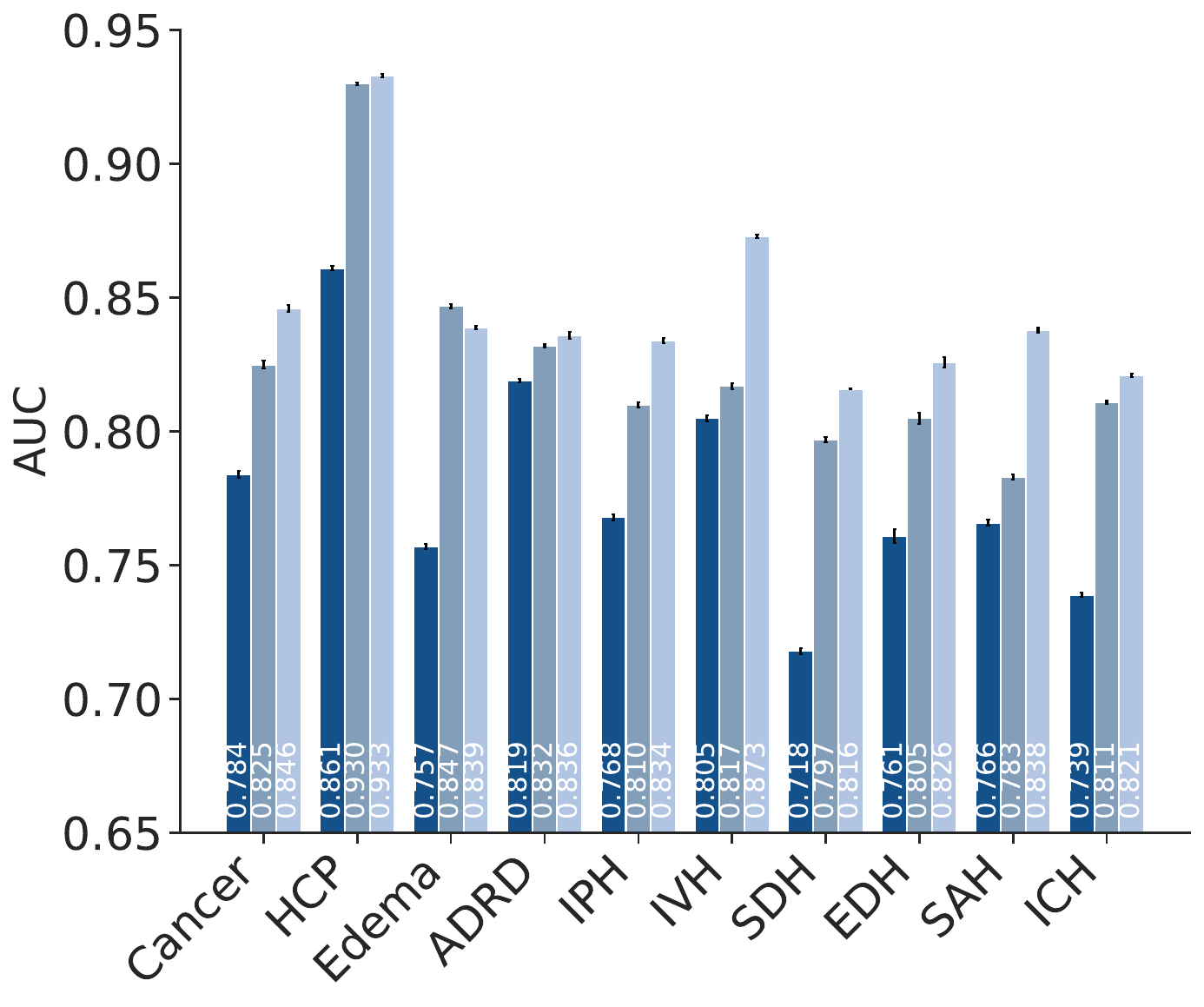}
    \includegraphics[trim={0 0 0 0},clip,height=0.24\textwidth, width=0.45\textwidth]{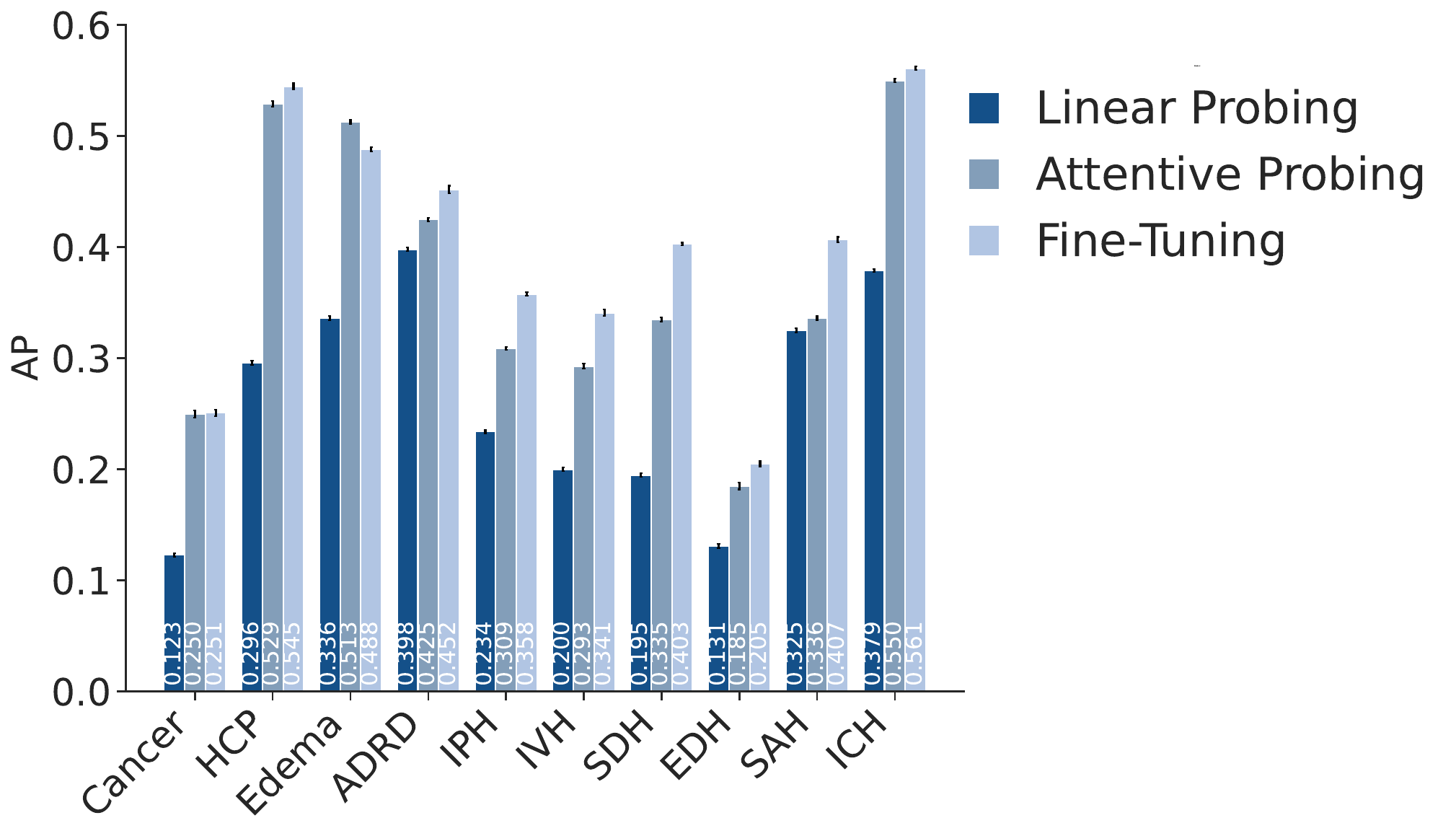}\\
    \makebox[\textwidth][l]{
        \hspace{0.35\textwidth}\textbf{NYU Long Island}
    } \\[0.2cm]
    \includegraphics[trim={0 0 0 0},clip,height=0.24\textwidth, width=0.3\textwidth]{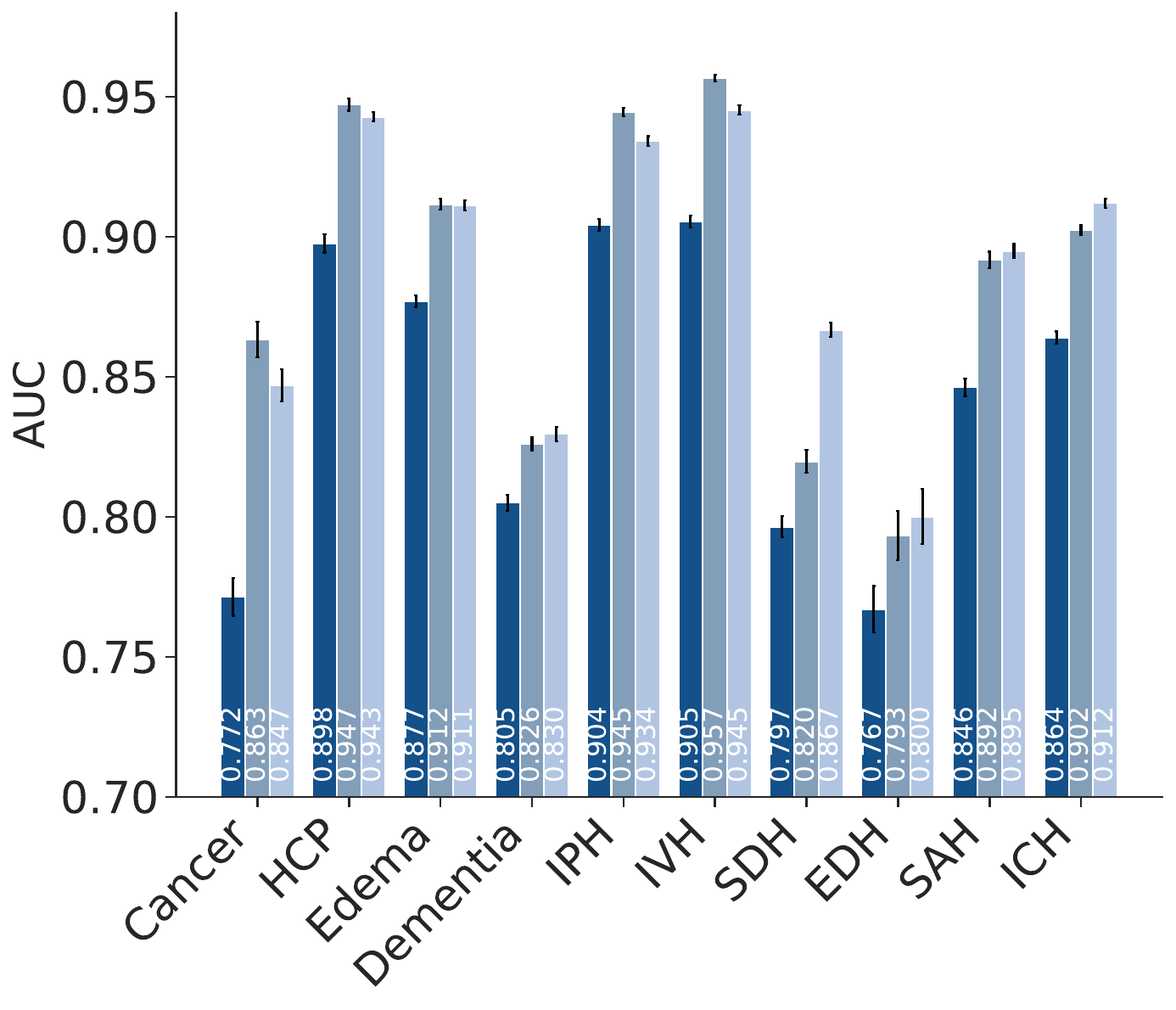}
    \includegraphics[trim={0 0 0 0},clip,height=0.24\textwidth, width=0.45\textwidth]{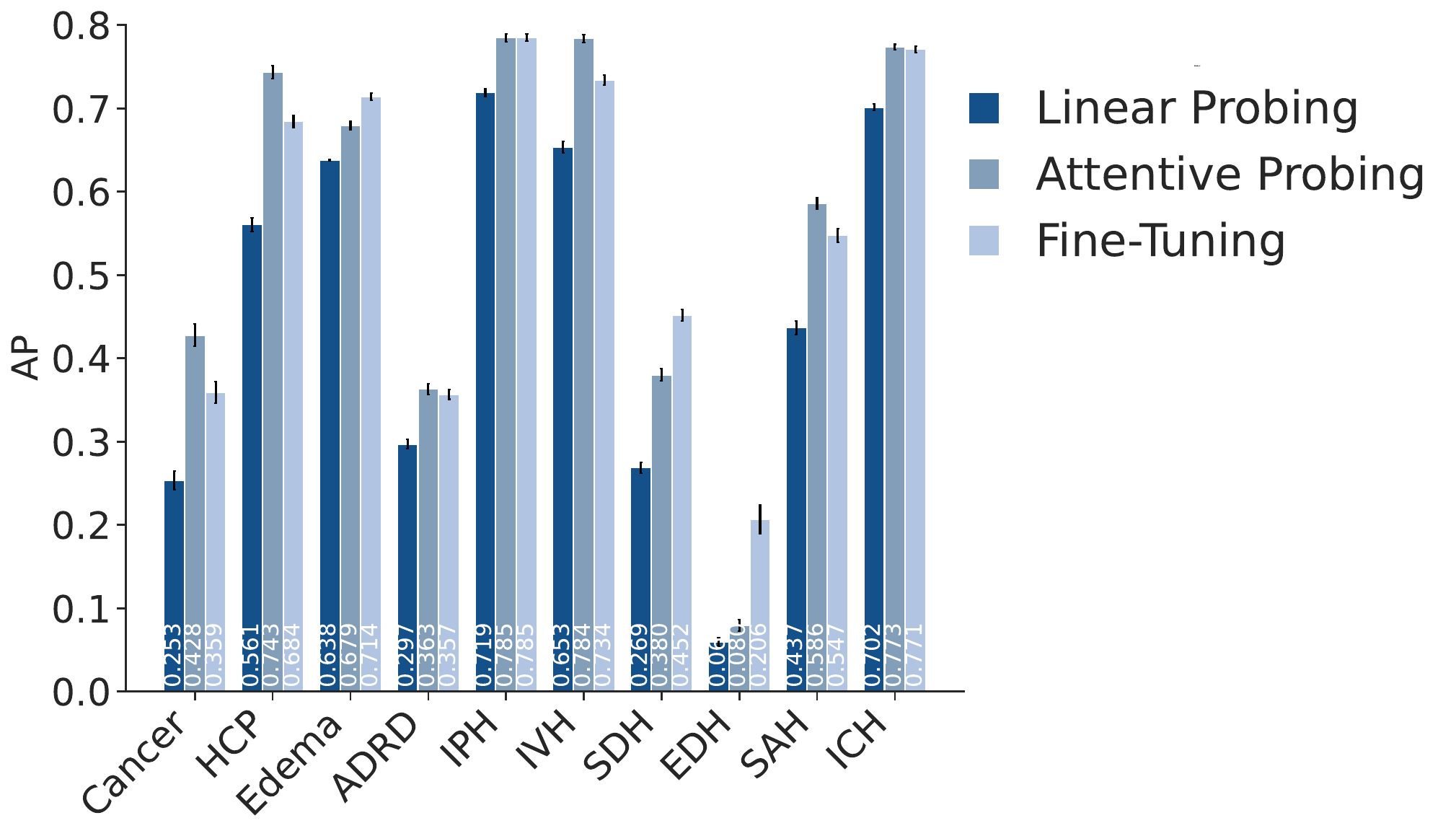} 
    \makebox[\textwidth][l]{
        \hspace{0.4\textwidth}\textbf{RSNA}
    } \\[0.2cm]
    \includegraphics[trim={0 0 0 0},clip,height=0.24\textwidth, width=0.3\textwidth]{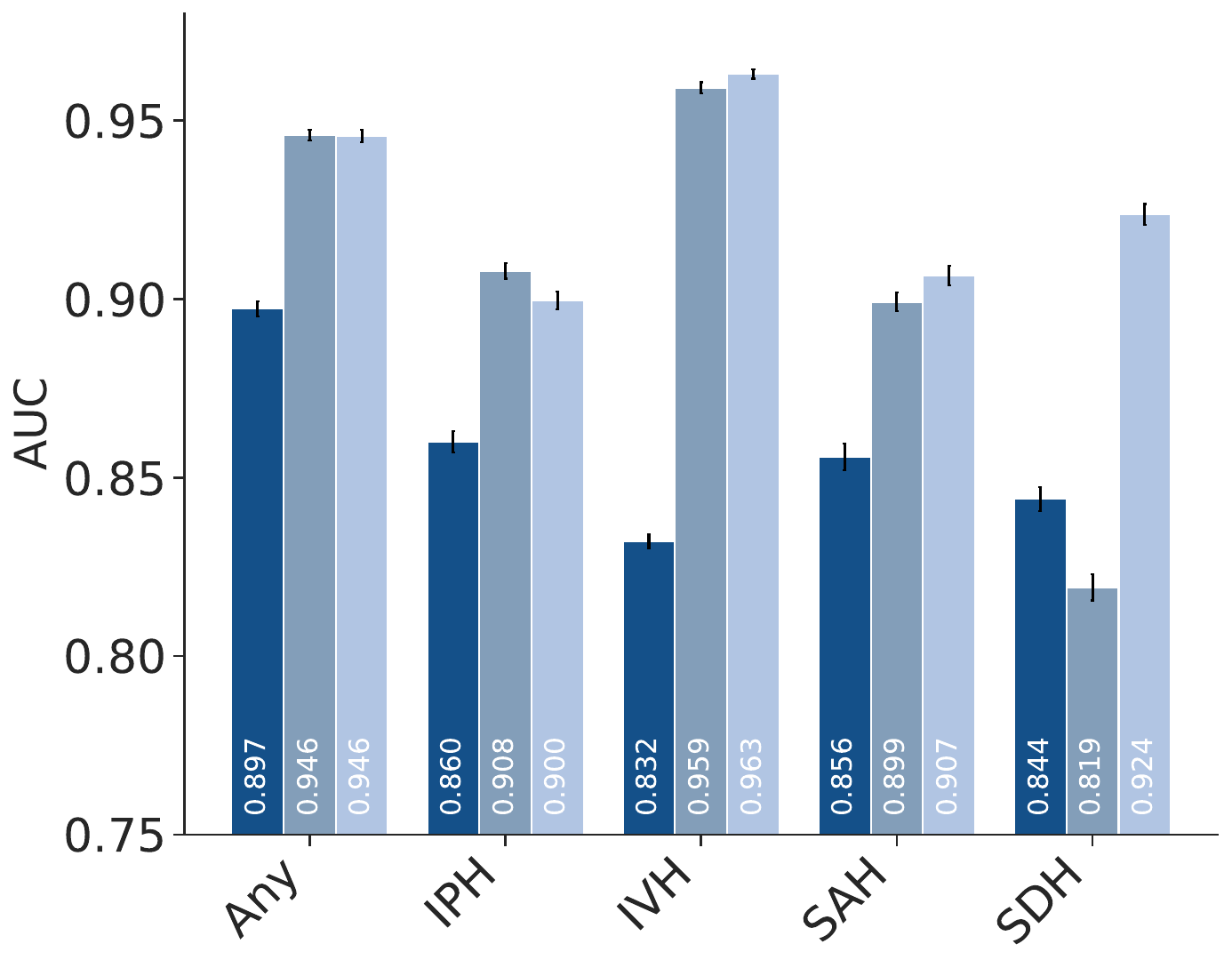}
    \includegraphics[height=0.24\textwidth, width=0.45\textwidth]{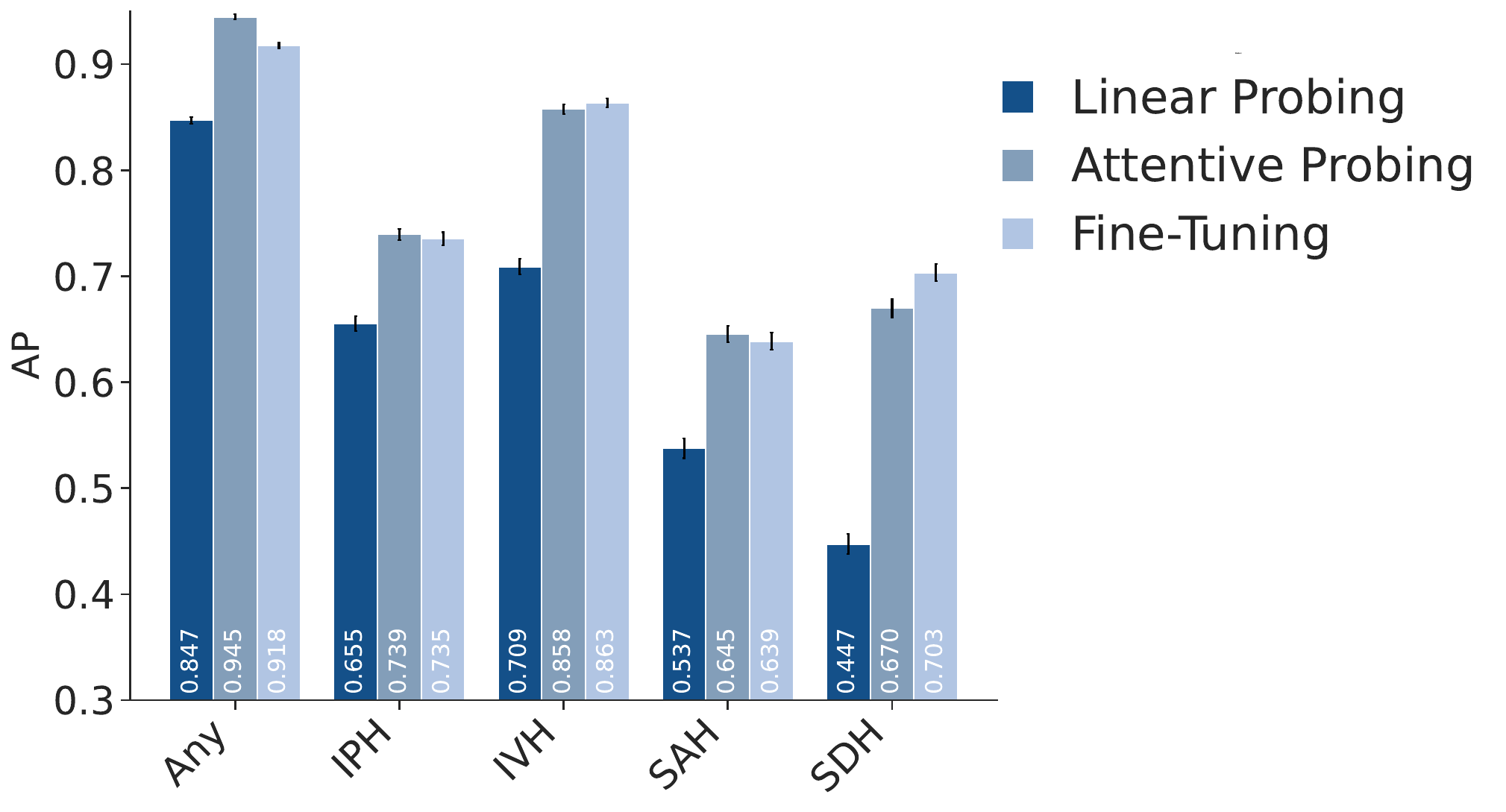} 
    \makebox[\textwidth][l]{
        \hspace{0.4\textwidth}\textbf{CQ500}
    } \\[0.2cm]
    \includegraphics[trim={0 0 120mm 0},clip,height=0.24\textwidth]{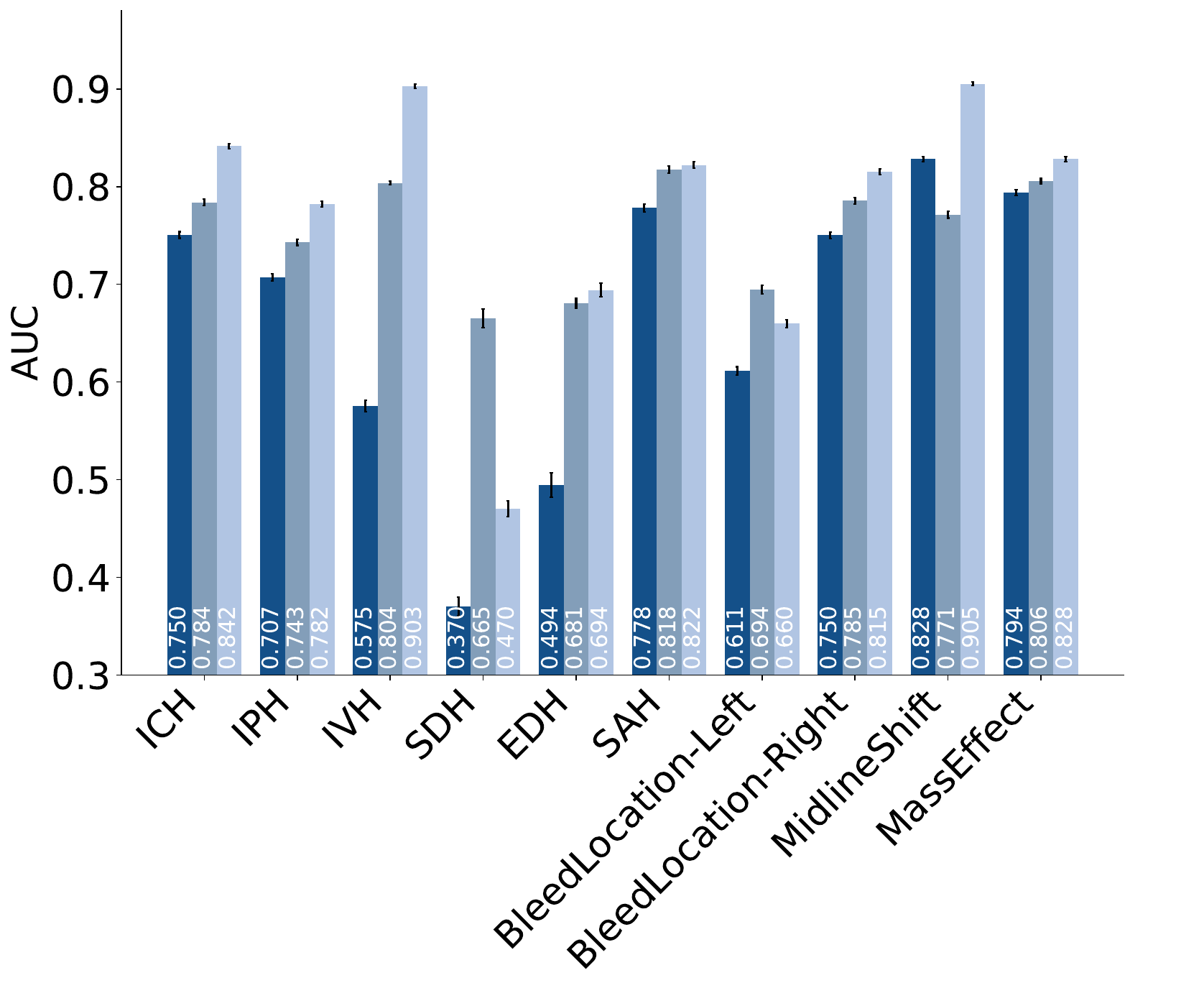}
    \includegraphics[trim={0 0 0 0},clip,height=0.24\textwidth]{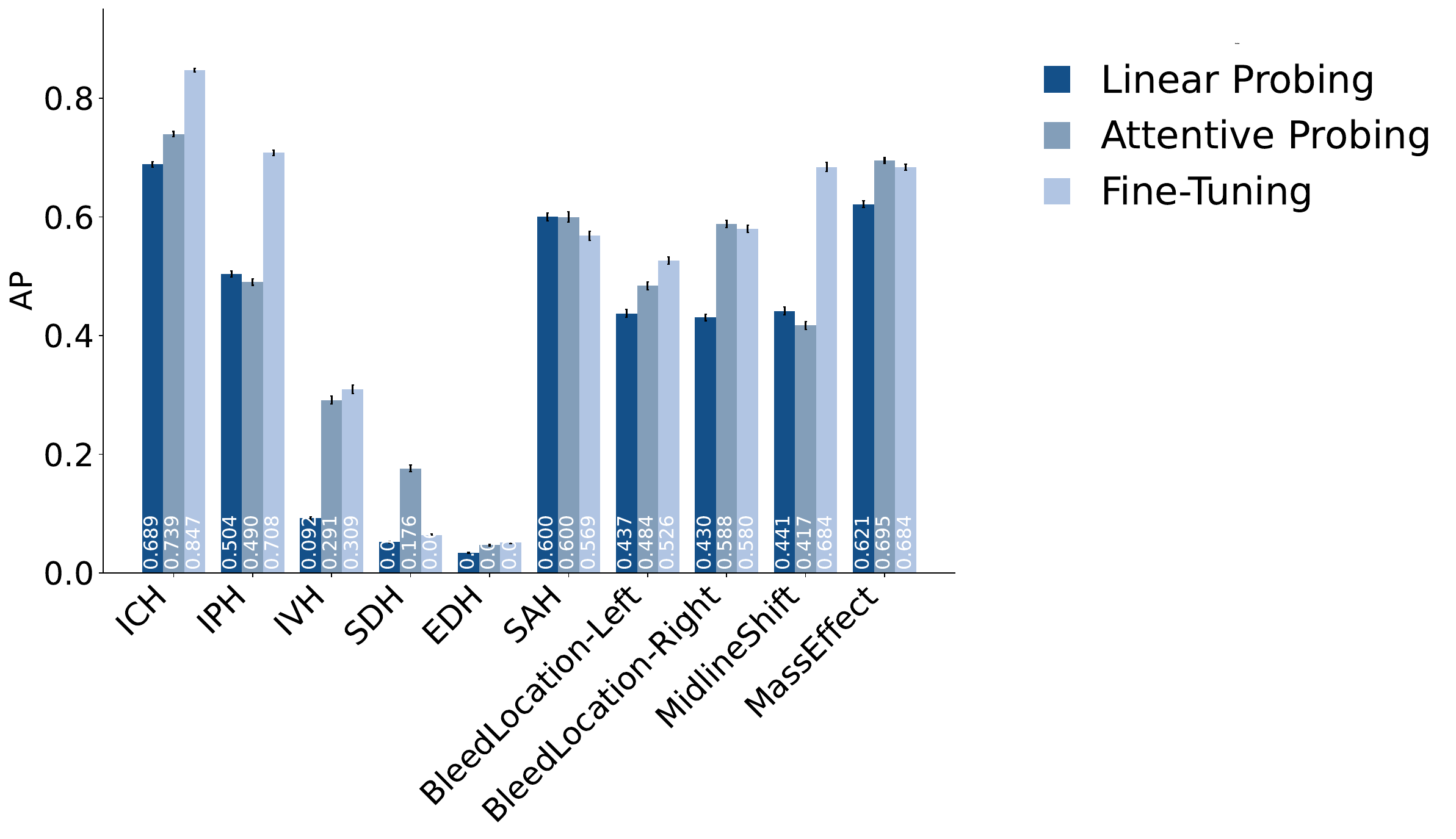} 
    \caption{\textbf{Performance comparison of supervised finetuning methods per pathology on the foundation model trained with MAE.} The results reveal that attentive probing is significantly more effective than linear probing, consistent with findings from~\cite{Chen2024}. Furthermore, for many diseases, the performance of probing models approaches that of fine-tuning, demonstrating that our pre-trained models produce adaptable representations capable of detecting diverse diseases.}
    \label{fig:probing-comparison-perpath}
\end{figure}

\newpage

\section{Time complexity increase with reduced patch size}
\label{apd:self_attention_rate}
Assume we have 3D image input of shape $H\times W\times D$, patch size $P$ and reducing factor $s$. By time complexity of self-attention $O(n^2 d)$ for sequence length $n$ and embedding dimension $d$, the new time complexity after reducing patch size can be derived as
\begin{align*}
    O(n^2d)&=O((\frac{H\times W\times D}{(\frac{P}{s})^3})^2d) \\
           &=O((\frac{H\times W\times D}{P^3})^2 s^6d)  \\
           &=O(s^6)O(n_{ori}^2d)
\end{align*}
where $n_{ori}=\frac{H\times W\times D}{P^3}$ is the original sequence length before reducing patch size.

\begin{figure*}[ht]
    \centering
    \begin{subfigure}[b]{\textwidth}
        \centering
        \includegraphics[width=\textwidth]{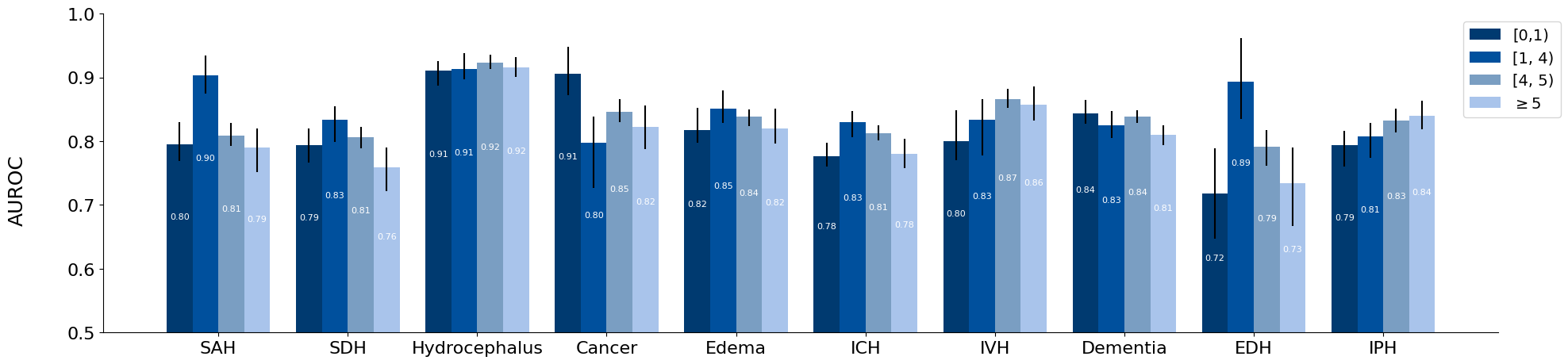}
        \caption{AUROC Performance}
    \end{subfigure}
    \hfill
    \begin{subfigure}[b]{\textwidth}
        \centering
        \includegraphics[width=\textwidth]{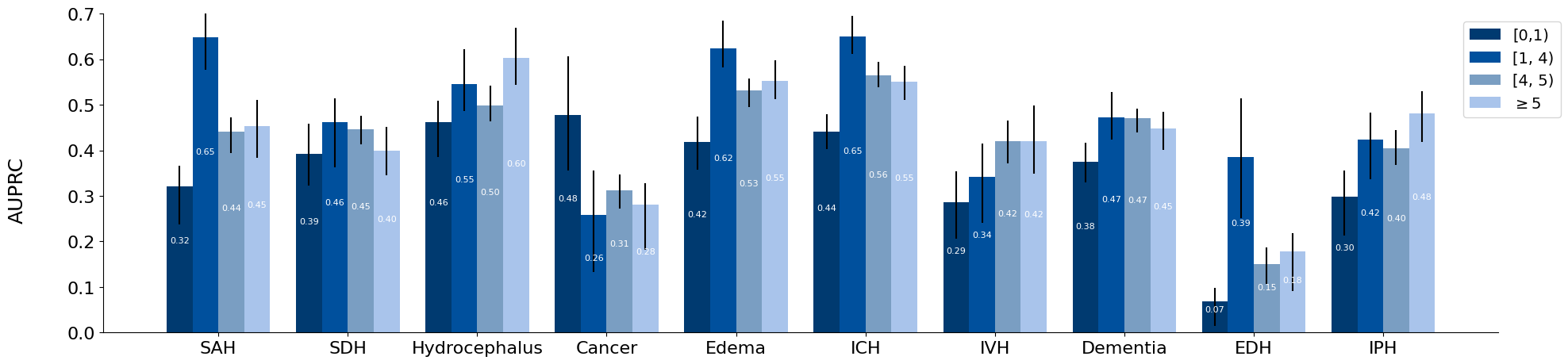}
        \caption{AUPRC Performance}
    \end{subfigure}
    \hfill
    \begin{subfigure}[b]{\textwidth}
        \centering
        \includegraphics[width=\textwidth]{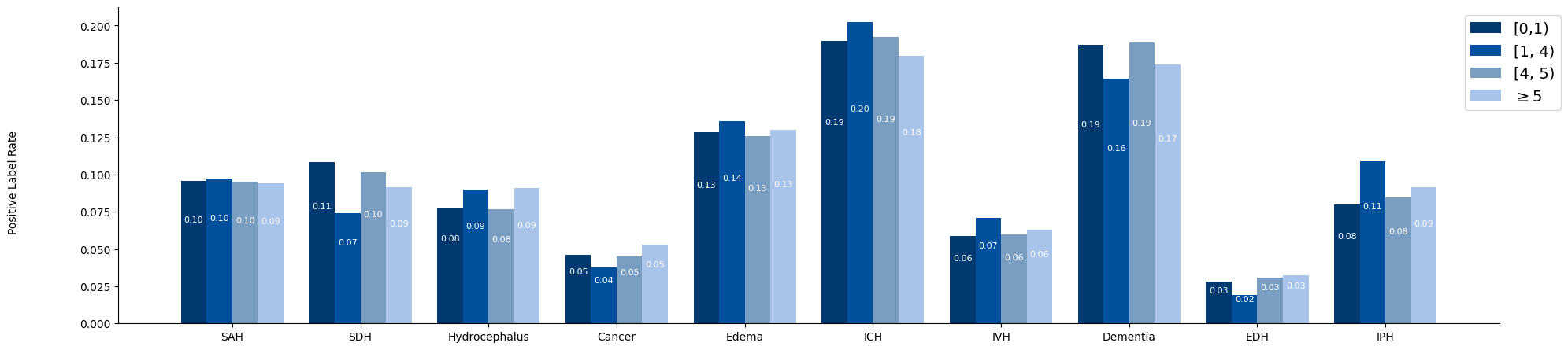}
        \caption{Ratio of Positive Labels}
    \end{subfigure}
    \caption{Performance for Each Slice Thickness Bin (Per Pathology).}
    \label{fig:slice_thickness_per_pathology}
\end{figure*}

\begin{figure*}[ht]
    \centering
    \begin{subfigure}[b]{0.3\textwidth}
        \centering
        \includegraphics[width=\textwidth]{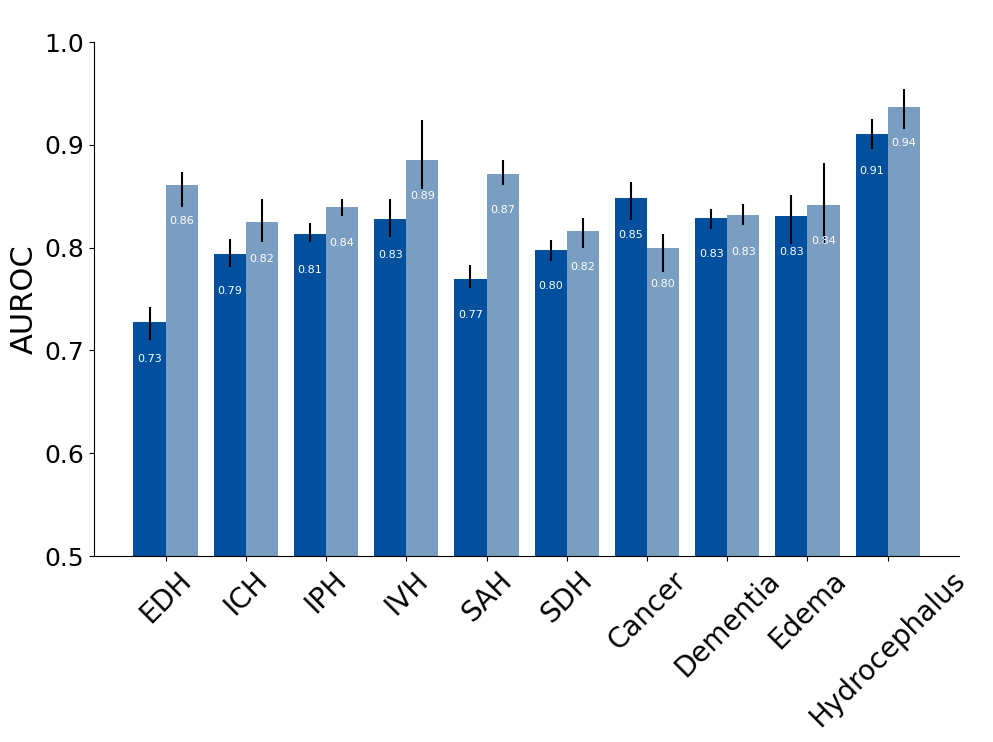}
        \caption{AUROC Performance}
    \end{subfigure}
    \hfill
    \begin{subfigure}[b]{0.3\textwidth}
        \centering
        \includegraphics[width=\textwidth]{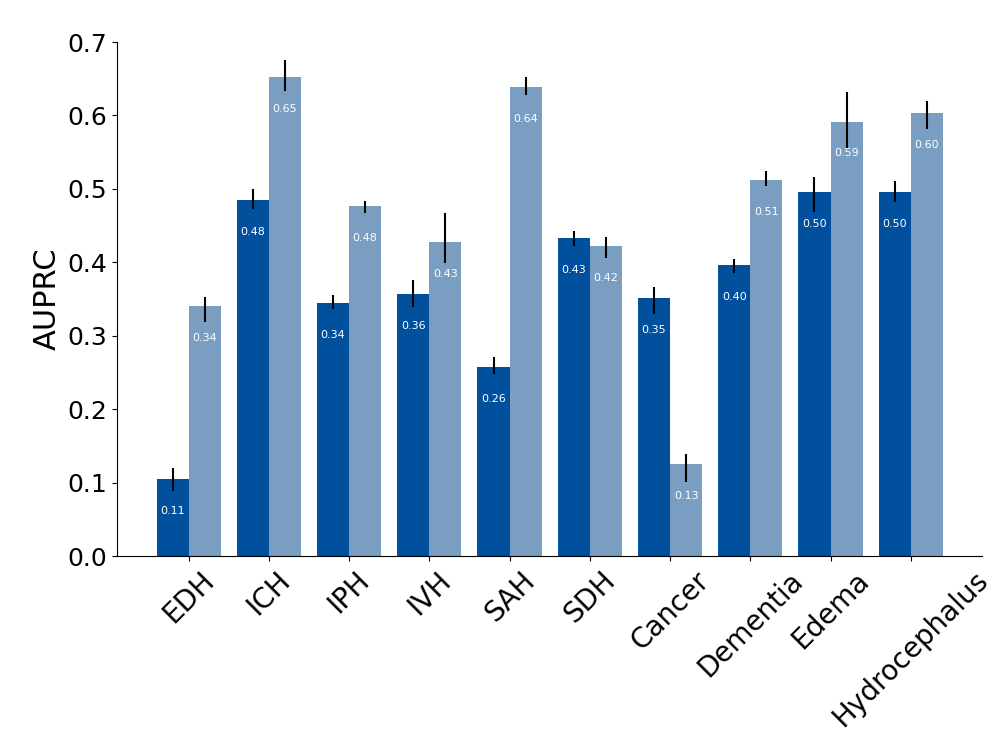}
        \caption{AUPRC Performance}
    \end{subfigure}
    \hfill
    \begin{subfigure}[b]{0.39\textwidth}
        \centering
        \includegraphics[width=\textwidth]{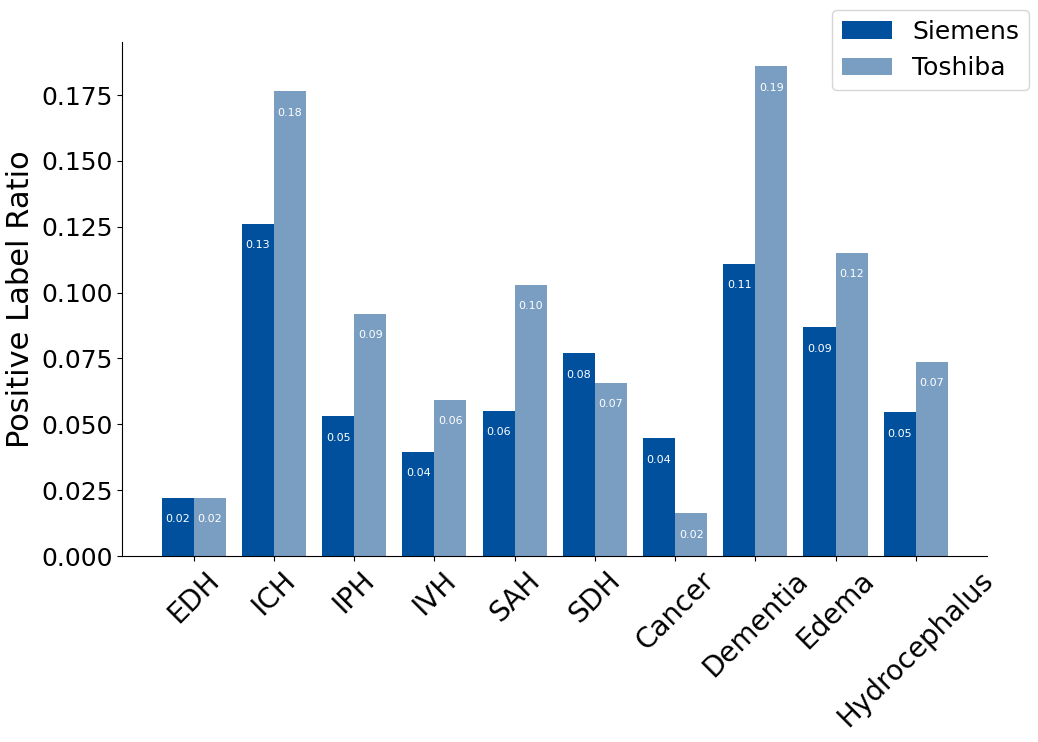}
        \caption{Distribution of Scans from Each Manufacturer}
    \end{subfigure}
    \caption{Performance for Each Manufacturer (Per Pathology).}
    \label{fig:manufacturer_per_pathology}
\end{figure*}


\begin{thebibliography}{10}
\urlstyle{rm}
\expandafter\ifx\csname url\endcsname\relax
  \def\url#1{\texttt{#1}}\fi
\expandafter\ifx\csname urlprefix\endcsname\relax\def\urlprefix{URL }\fi
\expandafter\ifx\csname doiprefix\endcsname\relax\def\doiprefix{DOI: }\fi
\providecommand{\bibinfo}[2]{#2}
\providecommand{\eprint}[2][]{\url{#2}}

\bibitem{flanders_construction_2020}
\bibinfo{author}{Flanders, A.~E.} \emph{et~al.}
\newblock \bibinfo{journal}{\bibinfo{title}{Construction of a {Machine} {Learning} {Dataset} through {Collaboration}: {The} {RSNA} 2019 {Brain} {CT} {Hemorrhage} {Challenge}}}.
\newblock {\emph{\JournalTitle{Radiology: Artificial Intelligence}}} \doiprefix\url{10.1148/ryai.2020190211} (\bibinfo{year}{2020}).
\newblock \bibinfo{note}{Publisher: Radiological Society of North America}.

\bibitem{CQ500}
\bibinfo{author}{Chilamkurthy, S.} \emph{et~al.}
\newblock \bibinfo{journal}{\bibinfo{title}{Deep learning algorithms for detection of critical findings in head ct scans: a retrospective study}}.
\newblock {\emph{\JournalTitle{The Lancet}}} \textbf{\bibinfo{volume}{392}}, \bibinfo{pages}{2388--2396}, \doiprefix\url{https://doi.org/10.1016/S0140-6736(18)31645-3} (\bibinfo{year}{2018}).

\bibitem{wang_deep_2021}
\bibinfo{author}{Wang, X.} \emph{et~al.}
\newblock \bibinfo{journal}{\bibinfo{title}{A deep learning algorithm for automatic detection and classification of acute intracranial hemorrhages in head {CT} scans}}.
\newblock {\emph{\JournalTitle{NeuroImage: Clinical}}} \textbf{\bibinfo{volume}{32}}, \bibinfo{pages}{102785}, \doiprefix\url{10.1016/j.nicl.2021.102785} (\bibinfo{year}{2021}).

\bibitem{yun_deep_2023}
\bibinfo{author}{Yun, T.~J.} \emph{et~al.}
\newblock \bibinfo{journal}{\bibinfo{title}{Deep learning based automatic detection algorithm for acute intracranial haemorrhage: a pivotal randomized clinical trial}}.
\newblock {\emph{\JournalTitle{npj Digital Medicine}}} \textbf{\bibinfo{volume}{6}}, \bibinfo{pages}{1--10}, \doiprefix\url{10.1038/s41746-023-00798-8} (\bibinfo{year}{2023}).
\newblock \bibinfo{note}{Publisher: Nature Publishing Group}.

\bibitem{pmlr-v139-radford21a}
\bibinfo{author}{Radford, A.} \emph{et~al.}
\newblock \bibinfo{title}{Learning transferable visual models from natural language supervision}.
\newblock In \bibinfo{editor}{Meila, M.} \& \bibinfo{editor}{Zhang, T.} (eds.) \emph{\bibinfo{booktitle}{Proceedings of the 38th International Conference on Machine Learning}}, vol. \bibinfo{volume}{139} of \emph{\bibinfo{series}{Proceedings of Machine Learning Research}}, \bibinfo{pages}{8748--8763} (\bibinfo{publisher}{PMLR}, \bibinfo{year}{2021}).

\bibitem{zhou2021ibot}
\bibinfo{author}{Zhou, J.} \emph{et~al.}
\newblock \bibinfo{journal}{\bibinfo{title}{ibot: Image bert pre-training with online tokenizer}}.
\newblock {\emph{\JournalTitle{International Conference on Learning Representations (ICLR)}}}  (\bibinfo{year}{2022}).

\bibitem{oquab2024dinov}
\bibinfo{author}{Oquab, M.} \emph{et~al.}
\newblock \bibinfo{journal}{\bibinfo{title}{{DINO}v2: Learning robust visual features without supervision}}.
\newblock {\emph{\JournalTitle{Transactions on Machine Learning Research}}}  (\bibinfo{year}{2024}).
\newblock \bibinfo{note}{Featured Certification}.

\bibitem{rishi24foundation}
\bibinfo{author}{Bommasani, R.} \emph{et~al.}
\newblock \bibinfo{journal}{\bibinfo{title}{On the opportunities and risks of foundation models}}.
\newblock {\emph{\JournalTitle{CoRR}}} \textbf{\bibinfo{volume}{abs/2108.07258}} (\bibinfo{year}{2021}).
\newblock \eprint{2108.07258}.

\bibitem{yao2024evaxfoundationmodelgeneral}
\bibinfo{author}{Yao, J.} \emph{et~al.}
\newblock \bibinfo{title}{Eva-x: A foundation model for general chest x-ray analysis with self-supervised learning} (\bibinfo{year}{2024}).
\newblock \eprint{2405.05237}.

\bibitem{wang_pathology_2024}
\bibinfo{author}{Wang, X.} \emph{et~al.}
\newblock \bibinfo{journal}{\bibinfo{title}{A pathology foundation model for cancer diagnosis and prognosis prediction}}.
\newblock {\emph{\JournalTitle{Nature}}} \textbf{\bibinfo{volume}{634}}, \bibinfo{pages}{970--978}, \doiprefix\url{10.1038/s41586-024-07894-z} (\bibinfo{year}{2024}).
\newblock \bibinfo{note}{Publisher: Nature Publishing Group}.

\bibitem{huang_visuallanguage_2023}
\bibinfo{author}{Huang, Z.}, \bibinfo{author}{Bianchi, F.}, \bibinfo{author}{Yuksekgonul, M.}, \bibinfo{author}{Montine, T.~J.} \& \bibinfo{author}{Zou, J.}
\newblock \bibinfo{journal}{\bibinfo{title}{A visual–language foundation model for pathology image analysis using medical {Twitter}}}.
\newblock {\emph{\JournalTitle{Nature Medicine}}} \textbf{\bibinfo{volume}{29}}, \bibinfo{pages}{2307--2316}, \doiprefix\url{10.1038/s41591-023-02504-3} (\bibinfo{year}{2023}).
\newblock \bibinfo{note}{Publisher: Nature Publishing Group}.

\bibitem{chen_towards_2024}
\bibinfo{author}{Chen, R.~J.} \emph{et~al.}
\newblock \bibinfo{journal}{\bibinfo{title}{Towards a general-purpose foundation model for computational pathology}}.
\newblock {\emph{\JournalTitle{Nature Medicine}}} \textbf{\bibinfo{volume}{30}}, \bibinfo{pages}{850--862}, \doiprefix\url{10.1038/s41591-024-02857-3} (\bibinfo{year}{2024}).
\newblock \bibinfo{note}{Publisher: Nature Publishing Group}.

\bibitem{Vorontsov2024}
\bibinfo{author}{Vorontsov, E.} \emph{et~al.}
\newblock \bibinfo{journal}{\bibinfo{title}{A foundation model for clinical-grade computational pathology and rare cancers detection}}.
\newblock {\emph{\JournalTitle{Nature Medicine}}} \textbf{\bibinfo{volume}{30}}, \bibinfo{pages}{2924--2935}, \doiprefix\url{10.1038/s41591-024-03141-0} (\bibinfo{year}{2024}).

\bibitem{zhou2023foundation}
\bibinfo{author}{Zhou, Y.} \emph{et~al.}
\newblock \bibinfo{journal}{\bibinfo{title}{A foundation model for generalizable disease detection from retinal images}}.
\newblock {\emph{\JournalTitle{Nature}}} \textbf{\bibinfo{volume}{622}}, \bibinfo{pages}{156--163} (\bibinfo{year}{2023}).

\bibitem{dong2024brainjepa}
\bibinfo{author}{Dong, Z.} \emph{et~al.}
\newblock \bibinfo{title}{Brain-{JEPA}: Brain dynamics foundation model with gradient positioning and spatiotemporal masking}.
\newblock In \emph{\bibinfo{booktitle}{The Thirty-eighth Annual Conference on Neural Information Processing Systems}} (\bibinfo{year}{2024}).

\bibitem{codella2024medimageinsight}
\bibinfo{author}{Codella, N. C.~F.} \emph{et~al.}
\newblock \bibinfo{title}{Medimageinsight: An open-source embedding model for general domain medical imaging} (\bibinfo{year}{2024}).
\newblock \eprint{2410.06542}.

\bibitem{yang2024advancingmultimodalmedicalcapabilities}
\bibinfo{author}{Yang, L.} \emph{et~al.}
\newblock \bibinfo{title}{Advancing multimodal medical capabilities of gemini} (\bibinfo{year}{2024}).
\newblock \eprint{2405.03162}.

\bibitem{zhang2024generalist}
\bibinfo{author}{Zhang, K.} \emph{et~al.}
\newblock \bibinfo{journal}{\bibinfo{title}{A generalist vision--language foundation model for diverse biomedical tasks}}.
\newblock {\emph{\JournalTitle{Nature Medicine}}} \bibinfo{pages}{1--13} (\bibinfo{year}{2024}).

\bibitem{Tang_2022_CVPR}
\bibinfo{author}{Tang, Y.} \emph{et~al.}
\newblock \bibinfo{title}{Self-supervised pre-training of swin transformers for 3d medical image analysis}.
\newblock In \emph{\bibinfo{booktitle}{Proceedings of the IEEE/CVF Conference on Computer Vision and Pattern Recognition (CVPR)}}, \bibinfo{pages}{20730--20740} (\bibinfo{year}{2022}).

\bibitem{blankemeier2024merlinvisionlanguagefoundation}
\bibinfo{author}{Blankemeier, L.} \emph{et~al.}
\newblock \bibinfo{title}{Merlin: A vision language foundation model for 3d computed tomography} (\bibinfo{year}{2024}).
\newblock \eprint{2406.06512}.

\bibitem{chen2020simple}
\bibinfo{author}{Chen, T.}, \bibinfo{author}{Kornblith, S.}, \bibinfo{author}{Norouzi, M.} \& \bibinfo{author}{Hinton, G.~E.}
\newblock \bibinfo{journal}{\bibinfo{title}{A simple framework for contrastive learning of visual representations}}.
\newblock {\emph{\JournalTitle{ArXiv}}} \textbf{\bibinfo{volume}{abs/2002.05709}} (\bibinfo{year}{2020}).

\bibitem{he2020momentum}
\bibinfo{author}{He, K.}, \bibinfo{author}{Fan, H.}, \bibinfo{author}{Wu, Y.}, \bibinfo{author}{Xie, S.} \& \bibinfo{author}{Girshick, R.}
\newblock \bibinfo{title}{Momentum contrast for unsupervised visual representation learning}.
\newblock In \emph{\bibinfo{booktitle}{Proceedings of the IEEE/CVF Conference on Computer Vision and Pattern Recognition}}, \bibinfo{pages}{9729--9738} (\bibinfo{year}{2020}).

\bibitem{caron2020unsupervised}
\bibinfo{author}{Caron, M.} \emph{et~al.}
\newblock \bibinfo{journal}{\bibinfo{title}{Unsupervised learning of visual features by contrasting cluster assignments}}.
\newblock {\emph{\JournalTitle{arXiv preprint arXiv:2006.09882}}}  (\bibinfo{year}{2020}).

\bibitem{caron2021emerging}
\bibinfo{author}{Caron, M.} \emph{et~al.}
\newblock \bibinfo{journal}{\bibinfo{title}{Emerging properties in self-supervised vision transformers}}.
\newblock {\emph{\JournalTitle{arXiv preprint arXiv:2104.14294}}}  (\bibinfo{year}{2021}).

\bibitem{bao2022beit}
\bibinfo{author}{Bao, H.}, \bibinfo{author}{Dong, L.}, \bibinfo{author}{Piao, S.} \& \bibinfo{author}{Wei, F.}
\newblock \bibinfo{title}{{BE}it: {BERT} pre-training of image transformers}.
\newblock In \emph{\bibinfo{booktitle}{International Conference on Learning Representations}} (\bibinfo{year}{2022}).

\bibitem{He2021MaskedAA}
\bibinfo{author}{He, K.} \emph{et~al.}
\newblock \bibinfo{journal}{\bibinfo{title}{Masked autoencoders are scalable vision learners}}.
\newblock {\emph{\JournalTitle{2022 IEEE/CVF Conference on Computer Vision and Pattern Recognition (CVPR)}}} \bibinfo{pages}{15979--15988} (\bibinfo{year}{2021}).

\bibitem{zbontar2021barlow}
\bibinfo{author}{Zbontar, J.}, \bibinfo{author}{Jing, L.}, \bibinfo{author}{Misra, I.}, \bibinfo{author}{LeCun, Y.} \& \bibinfo{author}{Deny, S.}
\newblock \bibinfo{journal}{\bibinfo{title}{Barlow twins: Self-supervised learning via redundancy reduction}}.
\newblock {\emph{\JournalTitle{arXiv preprint arXiv:2103.03230}}}  (\bibinfo{year}{2021}).

\bibitem{bardes2022vicreg}
\bibinfo{author}{Bardes, A.}, \bibinfo{author}{Ponce, J.} \& \bibinfo{author}{LeCun, Y.}
\newblock \bibinfo{title}{{VICR}eg: Variance-invariance-covariance regularization for self-supervised learning}.
\newblock In \emph{\bibinfo{booktitle}{International Conference on Learning Representations}} (\bibinfo{year}{2022}).

\bibitem{Liu_2023_CVPR}
\bibinfo{author}{Liu, K.} \emph{et~al.}
\newblock \bibinfo{title}{Multiple instance learning via iterative self-paced supervised contrastive learning}.
\newblock In \emph{\bibinfo{booktitle}{Proceedings of the IEEE/CVF Conference on Computer Vision and Pattern Recognition (CVPR)}}, \bibinfo{pages}{3355--3365} (\bibinfo{year}{2023}).

\bibitem{zhu2022interpretablepredictionlungsquamous}
\bibinfo{author}{Zhu, W.}, \bibinfo{author}{Fernandez{-}Granda, C.} \& \bibinfo{author}{Razavian, N.}
\newblock \bibinfo{title}{Interpretable prediction of lung squamous cell carcinoma recurrence with self-supervised learning}.
\newblock In \emph{\bibinfo{booktitle}{Proceedings of The 5th International Conference on Medical Imaging with Deep Learning}}, vol. \bibinfo{volume}{172} of \emph{\bibinfo{series}{Proceedings of Machine Learning Research}}, \bibinfo{pages}{1504--1522} (\bibinfo{publisher}{PMLR}, \bibinfo{year}{2022}).

\bibitem{Huang2023}
\bibinfo{author}{Huang, S.-C.} \emph{et~al.}
\newblock \bibinfo{journal}{\bibinfo{title}{Self-supervised learning for medical image classification: a systematic review and implementation guidelines}}.
\newblock {\emph{\JournalTitle{npj Digital Medicine}}} \textbf{\bibinfo{volume}{6}}, \bibinfo{pages}{74}, \doiprefix\url{10.1038/s41746-023-00811-0} (\bibinfo{year}{2023}).

\bibitem{azizi21big}
\bibinfo{author}{Azizi, S.} \emph{et~al.}
\newblock \bibinfo{title}{Big self-supervised models advance medical image classification}.
\newblock In \emph{\bibinfo{booktitle}{2021 IEEE/CVF International Conference on Computer Vision (ICCV)}}, \bibinfo{pages}{3458--3468}, \doiprefix\url{10.1109/ICCV48922.2021.00346} (\bibinfo{year}{2021}).

\bibitem{huang2023radiology}
\bibinfo{author}{Huang, H.}, \bibinfo{author}{Rawlekar, S.}, \bibinfo{author}{Chopra, S.} \& \bibinfo{author}{Deniz, C.~M.}
\newblock \bibinfo{title}{Radiology reports improve visual representations learned from radiographs}.
\newblock In \emph{\bibinfo{booktitle}{Medical Imaging with Deep Learning}} (\bibinfo{year}{2023}).

\bibitem{huang21GLoRIA}
\bibinfo{author}{Huang, S.-C.}, \bibinfo{author}{Shen, L.}, \bibinfo{author}{Lungren, M.~P.} \& \bibinfo{author}{Yeung, S.}
\newblock \bibinfo{title}{Gloria: A multimodal global-local representation learning framework for label-efficient medical image recognition}.
\newblock In \emph{\bibinfo{booktitle}{2021 IEEE/CVF International Conference on Computer Vision (ICCV)}}, \bibinfo{pages}{3922--3931}, \doiprefix\url{10.1109/ICCV48922.2021.00391} (\bibinfo{year}{2021}).

\bibitem{chen23masked}
\bibinfo{author}{Chen, Z.} \emph{et~al.}
\newblock \bibinfo{title}{Masked image modeling advances 3d medical image analysis}.
\newblock In \emph{\bibinfo{booktitle}{2023 IEEE/CVF Winter Conference on Applications of Computer Vision (WACV)}}, \bibinfo{pages}{1969--1979}, \doiprefix\url{10.1109/WACV56688.2023.00201} (\bibinfo{year}{2023}).

\bibitem{Azizi2023}
\bibinfo{author}{Azizi, S.} \emph{et~al.}
\newblock \bibinfo{journal}{\bibinfo{title}{Robust and data-efficient generalization of self-supervised machine learning for diagnostic imaging}}.
\newblock {\emph{\JournalTitle{Nature Biomedical Engineering}}} \textbf{\bibinfo{volume}{7}}, \bibinfo{pages}{756--779}, \doiprefix\url{10.1038/s41551-023-01049-7} (\bibinfo{year}{2023}).

\bibitem{dosovitskiy2020vit}
\bibinfo{author}{Dosovitskiy, A.} \emph{et~al.}
\newblock \bibinfo{journal}{\bibinfo{title}{An image is worth 16x16 words: Transformers for image recognition at scale}}.
\newblock {\emph{\JournalTitle{ICLR}}}  (\bibinfo{year}{2021}).

\bibitem{pai2025visionfoundationmodelscomputed}
\bibinfo{author}{Pai, S.} \emph{et~al.}
\newblock \bibinfo{title}{Vision foundation models for computed tomography} (\bibinfo{year}{2025}).
\newblock \eprint{2501.09001}.

\bibitem{pmlr-v80-ilse18a}
\bibinfo{author}{Ilse, M.}, \bibinfo{author}{Tomczak, J.} \& \bibinfo{author}{Welling, M.}
\newblock \bibinfo{title}{Attention-based deep multiple instance learning}.
\newblock In \bibinfo{editor}{Dy, J.} \& \bibinfo{editor}{Krause, A.} (eds.) \emph{\bibinfo{booktitle}{Proceedings of the 35th International Conference on Machine Learning}}, vol.~\bibinfo{volume}{80} of \emph{\bibinfo{series}{Proceedings of Machine Learning Research}}, \bibinfo{pages}{2127--2136} (\bibinfo{publisher}{PMLR}, \bibinfo{year}{2018}).

\bibitem{siméoni2025dinov3}
\bibinfo{author}{Siméoni, O.} \emph{et~al.}
\newblock \bibinfo{title}{Dinov3} (\bibinfo{year}{2025}).
\newblock \eprint{2508.10104}.

\bibitem{assran2025vjepa2selfsupervisedvideo}
\bibinfo{author}{Assran, M.} \emph{et~al.}
\newblock \bibinfo{title}{V-jepa 2: Self-supervised video models enable understanding, prediction and planning} (\bibinfo{year}{2025}).
\newblock \eprint{2506.09985}.

\bibitem{kaplan2020scalinglawsneurallanguage}
\bibinfo{author}{Kaplan, J.} \emph{et~al.}
\newblock \bibinfo{title}{Scaling laws for neural language models} (\bibinfo{year}{2020}).
\newblock \eprint{2001.08361}.

\bibitem{zhai22scalingvit}
\bibinfo{author}{Zhai, X.}, \bibinfo{author}{Kolesnikov, A.}, \bibinfo{author}{Houlsby, N.} \& \bibinfo{author}{Beyer, L.}
\newblock \bibinfo{title}{Scaling vision transformers}.
\newblock In \emph{\bibinfo{booktitle}{2022 IEEE/CVF Conference on Computer Vision and Pattern Recognition (CVPR)}}, \bibinfo{pages}{1204--1213}, \doiprefix\url{10.1109/CVPR52688.2022.01179} (\bibinfo{year}{2022}).

\bibitem{pmlr-v202-dehghani23a}
\bibinfo{author}{Dehghani, M.} \emph{et~al.}
\newblock \bibinfo{title}{Scaling vision transformers to 22 billion parameters}.
\newblock In \bibinfo{editor}{Krause, A.} \emph{et~al.} (eds.) \emph{\bibinfo{booktitle}{Proceedings of the 40th International Conference on Machine Learning}}, vol. \bibinfo{volume}{202} of \emph{\bibinfo{series}{Proceedings of Machine Learning Research}}, \bibinfo{pages}{7480--7512} (\bibinfo{publisher}{PMLR}, \bibinfo{year}{2023}).

\bibitem{li2024well}
\bibinfo{author}{Li, W.}, \bibinfo{author}{Yuille, A.} \& \bibinfo{author}{Zhou, Z.}
\newblock \bibinfo{title}{How well do supervised models transfer to 3d image segmentation}.
\newblock In \emph{\bibinfo{booktitle}{The Twelfth International Conference on Learning Representations}}, vol.~\bibinfo{volume}{1} (\bibinfo{year}{2024}).

\bibitem{Hemphill2015-yw}
\bibinfo{author}{Hemphill, J.~C., 3rd} \emph{et~al.}
\newblock \bibinfo{journal}{\bibinfo{title}{Guidelines for the management of spontaneous intracerebral hemorrhage: A guideline for healthcare professionals from the american heart {Association/American} stroke association}}.
\newblock {\emph{\JournalTitle{Stroke}}} \textbf{\bibinfo{volume}{46}}, \bibinfo{pages}{2032--2060} (\bibinfo{year}{2015}).

\bibitem{Qureshi2009-ve}
\bibinfo{author}{Qureshi, A.~I.}, \bibinfo{author}{Mendelow, A.~D.} \& \bibinfo{author}{Hanley, D.~F.}
\newblock \bibinfo{journal}{\bibinfo{title}{Intracerebral haemorrhage}}.
\newblock {\emph{\JournalTitle{Lancet}}} \textbf{\bibinfo{volume}{373}}, \bibinfo{pages}{1632--1644} (\bibinfo{year}{2009}).

\bibitem{Macellari2014-zj}
\bibinfo{author}{Macellari, F.}, \bibinfo{author}{Paciaroni, M.}, \bibinfo{author}{Agnelli, G.} \& \bibinfo{author}{Caso, V.}
\newblock \bibinfo{journal}{\bibinfo{title}{Neuroimaging in intracerebral hemorrhage}}.
\newblock {\emph{\JournalTitle{Stroke}}} \textbf{\bibinfo{volume}{45}}, \bibinfo{pages}{903--908} (\bibinfo{year}{2014}).

\bibitem{Morotti2022-lc}
\bibinfo{author}{Morotti, A.} \emph{et~al.}
\newblock \bibinfo{journal}{\bibinfo{title}{Intracerebral haemorrhage expansion: definitions, predictors, and prevention}}.
\newblock {\emph{\JournalTitle{Lancet Neurol}}} \textbf{\bibinfo{volume}{22}}, \bibinfo{pages}{159--171} (\bibinfo{year}{2022}).

\bibitem{Li2019-jx}
\bibinfo{author}{Li, H.}, \bibinfo{author}{Habes, M.}, \bibinfo{author}{Wolk, D.~A.}, \bibinfo{author}{Fan, Y.} \& \bibinfo{author}{{Alzheimer's Disease Neuroimaging Initiative and the Australian Imaging Biomarkers and Lifestyle Study of Aging}}.
\newblock \bibinfo{journal}{\bibinfo{title}{A deep learning model for early prediction of alzheimer's disease dementia based on hippocampal magnetic resonance imaging data}}.
\newblock {\emph{\JournalTitle{Alzheimers. Dement.}}} \textbf{\bibinfo{volume}{15}}, \bibinfo{pages}{1059--1070} (\bibinfo{year}{2019}).

\bibitem{pmlr-v116-liu20a}
\bibinfo{author}{Liu, S.}, \bibinfo{author}{Yadav, C.}, \bibinfo{author}{Fernandez-Granda, C.} \& \bibinfo{author}{Razavian, N.}
\newblock \bibinfo{title}{{On the design of convolutional neural networks for automatic detection of Alzheimer’s disease}}.
\newblock In \bibinfo{editor}{Dalca, A.~V.} \emph{et~al.} (eds.) \emph{\bibinfo{booktitle}{Proceedings of the Machine Learning for Health NeurIPS Workshop}}, vol. \bibinfo{volume}{116} of \emph{\bibinfo{series}{Proceedings of Machine Learning Research}}, \bibinfo{pages}{184--201} (\bibinfo{publisher}{PMLR}, \bibinfo{year}{2020}).

\bibitem{Xue2024}
\bibinfo{author}{Xue, C.} \emph{et~al.}
\newblock \bibinfo{journal}{\bibinfo{title}{Ai-based differential diagnosis of dementia etiologies on multimodal data}}.
\newblock {\emph{\JournalTitle{Nature Medicine}}} \textbf{\bibinfo{volume}{30}}, \bibinfo{pages}{2977--2989}, \doiprefix\url{10.1038/s41591-024-03118-z} (\bibinfo{year}{2024}).

\bibitem{https://doi.org/10.1002/neo2.10}
\bibinfo{author}{Agarwal, R.} \emph{et~al.}
\newblock \bibinfo{journal}{\bibinfo{title}{Effects of financial toxicity and socioeconomic status on mri follow-up time in multiple sclerosis}}.
\newblock {\emph{\JournalTitle{Clinical Neuroimaging}}} \textbf{\bibinfo{volume}{1}}, \bibinfo{pages}{e10}, \doiprefix\url{https://doi.org/10.1002/neo2.10} (\bibinfo{year}{2024}).
\newblock \eprint{https://onlinelibrary.wiley.com/doi/pdf/10.1002/neo2.10}.

\bibitem{lin_dementia_2020}
\bibinfo{author}{Lin, P.-J.} \emph{et~al.}
\newblock \bibinfo{journal}{\bibinfo{title}{Dementia diagnosis disparities by race and ethnicity}}.
\newblock {\emph{\JournalTitle{Alzheimer's \& Dementia}}} \textbf{\bibinfo{volume}{16}}, \bibinfo{pages}{e043183}, \doiprefix\url{10.1002/alz.043183} (\bibinfo{year}{2020}).

\bibitem{kim2021racial}
\bibinfo{author}{Kim, N.}
\newblock \bibinfo{journal}{\bibinfo{title}{Racial disparities in neurological care in the united states: An internal mechanism}}.
\newblock {\emph{\JournalTitle{HPHR}}} \textbf{\bibinfo{volume}{32}}, \doiprefix\url{10.54111/0001/FF11} (\bibinfo{year}{2021}).

\bibitem{hematoma_expansion}
\bibinfo{author}{Yu, B.} \emph{et~al.}
\newblock \bibinfo{journal}{\bibinfo{title}{Predicting hematoma expansion after ich: A comparison of clinician prediction with deep learning radiomics models}}.
\newblock {\emph{\JournalTitle{Neurocrit. Care}}}  (\bibinfo{year}{2025}).

\bibitem{Zhu2024-zd}
\bibinfo{author}{Zhu, W.} \emph{et~al.}
\newblock \bibinfo{journal}{\bibinfo{title}{Predicting risk of alzheimer's diseases and related dementias with {AI} foundation model on electronic health records}}.
\newblock {\emph{\JournalTitle{medRxiv}}}  (\bibinfo{year}{2024}).

\bibitem{li_first_2016}
\bibinfo{author}{Li, X.}, \bibinfo{author}{Morgan, P.~S.}, \bibinfo{author}{Ashburner, J.}, \bibinfo{author}{Smith, J.} \& \bibinfo{author}{Rorden, C.}
\newblock \bibinfo{journal}{\bibinfo{title}{The first step for neuroimaging data analysis: {DICOM} to {NIfTI} conversion}}.
\newblock {\emph{\JournalTitle{Journal of Neuroscience Methods}}} \textbf{\bibinfo{volume}{264}}, \bibinfo{pages}{47--56}, \doiprefix\url{10.1016/j.jneumeth.2016.03.001} (\bibinfo{year}{2016}).

\bibitem{MedSAM}
\bibinfo{author}{Ma, J.} \emph{et~al.}
\newblock \bibinfo{journal}{\bibinfo{title}{Segment anything in medical images}}.
\newblock {\emph{\JournalTitle{Nature Communications}}} \textbf{\bibinfo{volume}{15}}, \bibinfo{pages}{654} (\bibinfo{year}{2024}).

\bibitem{NIPS2017_3f5ee243}
\bibinfo{author}{Vaswani, A.} \emph{et~al.}
\newblock \bibinfo{title}{Attention is all you need}.
\newblock In \bibinfo{editor}{Guyon, I.} \emph{et~al.} (eds.) \emph{\bibinfo{booktitle}{Advances in Neural Information Processing Systems}}, vol.~\bibinfo{volume}{30} (\bibinfo{publisher}{Curran Associates, Inc.}, \bibinfo{year}{2017}).

\bibitem{loshchilov2018decoupled}
\bibinfo{author}{Loshchilov, I.} \& \bibinfo{author}{Hutter, F.}
\newblock \bibinfo{title}{Decoupled weight decay regularization}.
\newblock In \emph{\bibinfo{booktitle}{International Conference on Learning Representations}} (\bibinfo{year}{2019}).

\bibitem{ravi2024sam2}
\bibinfo{author}{Ravi, N.} \emph{et~al.}
\newblock \bibinfo{journal}{\bibinfo{title}{Sam 2: Segment anything in images and videos}}.
\newblock {\emph{\JournalTitle{arXiv preprint arXiv:2408.00714}}}  (\bibinfo{year}{2024}).

\bibitem{tong2022videomae}
\bibinfo{author}{Tong, Z.}, \bibinfo{author}{Song, Y.}, \bibinfo{author}{Wang, J.} \& \bibinfo{author}{Wang, L.}
\newblock \bibinfo{title}{Video{MAE}: Masked autoencoders are data-efficient learners for self-supervised video pre-training}.
\newblock In \emph{\bibinfo{booktitle}{Advances in Neural Information Processing Systems}} (\bibinfo{year}{2022}).

\bibitem{gupta2023siamese}
\bibinfo{author}{Gupta, A.}, \bibinfo{author}{Wu, J.}, \bibinfo{author}{Deng, J.} \& \bibinfo{author}{Fei-Fei, L.}
\newblock \bibinfo{title}{Siamese masked autoencoders}.
\newblock In \emph{\bibinfo{booktitle}{Thirty-seventh Conference on Neural Information Processing Systems}} (\bibinfo{year}{2023}).

\bibitem{zhou23self}
\bibinfo{author}{Zhou, L.} \emph{et~al.}
\newblock \bibinfo{title}{Self pre-training with masked autoencoders for medical image classification and segmentation}.
\newblock In \emph{\bibinfo{booktitle}{2023 IEEE 20th International Symposium on Biomedical Imaging (ISBI)}}, \bibinfo{pages}{1--6}, \doiprefix\url{10.1109/ISBI53787.2023.10230477} (\bibinfo{year}{2023}).

\bibitem{huang2022masked}
\bibinfo{author}{Huang, P.-Y.} \emph{et~al.}
\newblock \bibinfo{title}{Masked autoencoders that listen}.
\newblock In \bibinfo{editor}{Oh, A.~H.}, \bibinfo{editor}{Agarwal, A.}, \bibinfo{editor}{Belgrave, D.} \& \bibinfo{editor}{Cho, K.} (eds.) \emph{\bibinfo{booktitle}{Advances in Neural Information Processing Systems}} (\bibinfo{year}{2022}).

\bibitem{cong2022satmae}
\bibinfo{author}{Cong, Y.} \emph{et~al.}
\newblock \bibinfo{title}{Sat{MAE}: Pre-training transformers for temporal and multi-spectral satellite imagery}.
\newblock In \bibinfo{editor}{Oh, A.~H.}, \bibinfo{editor}{Agarwal, A.}, \bibinfo{editor}{Belgrave, D.} \& \bibinfo{editor}{Cho, K.} (eds.) \emph{\bibinfo{booktitle}{Advances in Neural Information Processing Systems}} (\bibinfo{year}{2022}).

\bibitem{yu2022coca}
\bibinfo{author}{Yu, J.} \emph{et~al.}
\newblock \bibinfo{journal}{\bibinfo{title}{Coca: Contrastive captioners are image-text foundation models}}.
\newblock {\emph{\JournalTitle{Transactions on Machine Learning Research}}}  (\bibinfo{year}{2022}).

\bibitem{yan2023videococavideotextmodelingzeroshot}
\bibinfo{author}{Yan, S.} \emph{et~al.}
\newblock \bibinfo{title}{Videococa: Video-text modeling with zero-shot transfer from contrastive captioners} (\bibinfo{year}{2023}).
\newblock \eprint{2212.04979}.

\bibitem{Chen2024}
\bibinfo{author}{Chen, X.} \emph{et~al.}
\newblock \bibinfo{journal}{\bibinfo{title}{Context autoencoder for self-supervised representation learning}}.
\newblock {\emph{\JournalTitle{International Journal of Computer Vision}}} \textbf{\bibinfo{volume}{132}}, \bibinfo{pages}{208--223}, \doiprefix\url{10.1007/s11263-023-01852-4} (\bibinfo{year}{2024}).

\bibitem{want19simpleshot}
\bibinfo{author}{Wang, Y.}, \bibinfo{author}{Chao, W.-L.}, \bibinfo{author}{Weinberger, K.~Q.} \& \bibinfo{author}{van~der Maaten, L.}
\newblock \bibinfo{journal}{\bibinfo{title}{Simpleshot: Revisiting nearest-neighbor classification for few-shot learning}}.
\newblock {\emph{\JournalTitle{arXiv preprint arXiv:1911.04623}}}  (\bibinfo{year}{2019}).

\bibitem{jake17proto}
\bibinfo{author}{Snell, J.}, \bibinfo{author}{Swersky, K.} \& \bibinfo{author}{Zemel, R.~S.}
\newblock \bibinfo{journal}{\bibinfo{title}{Prototypical networks for few-shot learning}}.
\newblock {\emph{\JournalTitle{CoRR}}} \textbf{\bibinfo{volume}{abs/1703.05175}} (\bibinfo{year}{2017}).
\newblock \eprint{1703.05175}.

\bibitem{hara3dcnns}
\bibinfo{author}{Hara, K.}, \bibinfo{author}{Kataoka, H.} \& \bibinfo{author}{Satoh, Y.}
\newblock \bibinfo{title}{Can spatiotemporal 3d cnns retrace the history of 2d cnns and imagenet?}
\newblock In \emph{\bibinfo{booktitle}{Proceedings of the IEEE Conference on Computer Vision and Pattern Recognition (CVPR)}}, \bibinfo{pages}{6546--6555} (\bibinfo{year}{2018}).

\bibitem{Chilamkurthy2018}
\bibinfo{author}{Chilamkurthy, S.} \emph{et~al.}
\newblock \bibinfo{journal}{\bibinfo{title}{Deep learning algorithms for detection of critical findings in head ct scans: a retrospective study}}.
\newblock {\emph{\JournalTitle{The Lancet}}} \textbf{\bibinfo{volume}{392}}, \bibinfo{pages}{2388--2396}, \doiprefix\url{10.1016/S0140-6736(18)31645-3} (\bibinfo{year}{2018}).

\bibitem{WANG2021102785}
\bibinfo{author}{Wang, X.} \emph{et~al.}
\newblock \bibinfo{journal}{\bibinfo{title}{A deep learning algorithm for automatic detection and classification of acute intracranial hemorrhages in head ct scans}}.
\newblock {\emph{\JournalTitle{NeuroImage: Clinical}}} \textbf{\bibinfo{volume}{32}}, \bibinfo{pages}{102785}, \doiprefix\url{https://doi.org/10.1016/j.nicl.2021.102785} (\bibinfo{year}{2021}).

\end{thebibliography}


\clearpage
\end{document}